\documentclass[10pt,twocolumn,letterpaper]{article}
\usepackage[protrusion=true,expansion=true,stretch=30,shrink=30]{microtype}
\usepackage{cvpr}

\usepackage{cuted} 

\usepackage{currfile} 

\usepackage{caption} 

\newcommand{\TODO}[1]{\textbf{\color{red}[TODO: #1]}}
\usepackage{booktabs}
\usepackage{array}
\usepackage{amsmath}
\usepackage{xcolor}
\usepackage{graphicx}
\renewcommand{\TODO}[1]{\textcolor{orange}{[TODO: #1]}}
\makeatletter
\newcommand{\defV}[2]{\expandafter\def\csname tav@#1\endcsname{#2}}
\newcommand{\V}[1]{\ifcsname tav@#1\endcsname\csname tav@#1\endcsname\else\underline{\hspace{1.5em}}\fi}
\makeatother
\defV{final/answers}{60{,}480}
\defV{final/answers/per_model}{6{,}048}
\defV{final/bound/100}{5.16}
\defV{final/bound/70}{4.91}
\defV{final/bound/70/diff}{9.82}
\defV{final/conditional/max}{0.80}
\defV{final/conditional/max/name}{Molmo2-O 7B}
\defV{final/conditional/min}{0.24}
\defV{final/conditional/min/name}{Qwen3.8 27B}
\defV{final/dec/crop/decidable}{8}
\defV{final/dec/crop/k}{5}
\defV{final/dec/crop/tested}{5}
\defV{final/dec/direct/decidable}{1}
\defV{final/dec/direct/k}{1}
\defV{final/dec/direct/tested}{5}
\defV{final/dec/gate/k}{6}
\defV{final/dec/gsi_placement/decidable}{6}
\defV{final/dec/gsi_placement/k}{5}
\defV{final/dec/gsi_placement/tested}{5}
\defV{final/dec/isobar/decidable}{7}
\defV{final/dec/isobar/k}{4}
\defV{final/dec/isobar/tested}{5}
\defV{final/dec/matched/decidable}{6}
\defV{final/dec/matched/k}{5}
\defV{final/dec/matched/tested}{7}
\defV{final/dec/reversal/decidable}{8}
\defV{final/dec/reversal/k}{6}
\defV{final/dec/reversal/n}{10}
\defV{final/dec/scatter/decidable}{10}
\defV{final/dec/scatter/k}{7}
\defV{final/dec/scatter/tested}{7}
\defV{final/dec/separation/decidable}{7}
\defV{final/dec/separation/k}{5}
\defV{final/dec/separation/tested}{6}
\defV{final/dec/white/decidable}{8}
\defV{final/dec/white/k}{7}
\defV{final/dec/white/tested}{10}
\defV{final/failed}{0}
\defV{final/frontier/gpt6_luna/crop/b}{77}
\defV{final/frontier/gpt6_luna/crop/c}{14}
\defV{final/frontier/gpt6_luna/crop/conflict/acc}{0.36}
\defV{final/frontier/gpt6_luna/crop/conflict/d}{$-$0.76}
\defV{final/frontier/gpt6_luna/crop/control/acc}{0.81}
\defV{final/frontier/gpt6_luna/crop/control/d}{+1.70}
\defV{final/frontier/gpt6_luna/crop/drop}{0.45}
\defV{final/frontier/gpt6_luna/crop/pholm}{$2\times10^{-11}$}
\defV{final/frontier/gpt6_luna/direct/b}{82}
\defV{final/frontier/gpt6_luna/direct/c}{8}
\defV{final/frontier/gpt6_luna/direct/conflict/acc}{0.21}
\defV{final/frontier/gpt6_luna/direct/conflict/d}{$-$1.72}
\defV{final/frontier/gpt6_luna/direct/control/acc}{0.74}
\defV{final/frontier/gpt6_luna/direct/control/d}{+1.35}
\defV{final/frontier/gpt6_luna/direct/drop}{0.53}
\defV{final/frontier/gpt6_luna/direct/pholm}{$3\times10^{-16}$}
\defV{final/frontier/gpt6_luna/matched/b}{70}
\defV{final/frontier/gpt6_luna/matched/c}{12}
\defV{final/frontier/gpt6_luna/matched/conflict/acc}{0.31}
\defV{final/frontier/gpt6_luna/matched/conflict/d}{$-$1.02}
\defV{final/frontier/gpt6_luna/matched/control/acc}{0.73}
\defV{final/frontier/gpt6_luna/matched/control/d}{+1.27}
\defV{final/frontier/gpt6_luna/matched/drop}{0.41}
\defV{final/frontier/gpt6_luna/matched/pholm}{$4\times10^{-11}$}
\defV{final/frontier/gpt6_luna/matched/status}{yes}
\defV{final/frontier/gpt6_luna/name}{GPT-6 Luna}
\defV{final/frontier/gpt6_luna/white/b}{90}
\defV{final/frontier/gpt6_luna/white/c}{4}
\defV{final/frontier/gpt6_luna/white/conflict/acc}{0.21}
\defV{final/frontier/gpt6_luna/white/conflict/d}{$-$2.08}
\defV{final/frontier/gpt6_luna/white/control/acc}{0.82}
\defV{final/frontier/gpt6_luna/white/control/d}{+1.86}
\defV{final/frontier/gpt6_luna/white/drop}{0.61}
\defV{final/frontier/gpt6_luna/white/pholm}{$6\times10^{-22}$}
\defV{final/frontier/gpt6_luna/white/status}{yes}
\defV{final/frontier/gpt6_sol/crop/b}{1}
\defV{final/frontier/gpt6_sol/crop/c}{0}
\defV{final/frontier/gpt6_sol/crop/conflict/acc}{0.99}
\defV{final/frontier/gpt6_sol/crop/conflict/d}{+4.49}
\defV{final/frontier/gpt6_sol/crop/control/acc}{1.00}
\defV{final/frontier/gpt6_sol/crop/control/d}{+4.91}
\defV{final/frontier/gpt6_sol/crop/drop}{0.01}
\defV{final/frontier/gpt6_sol/crop/pholm}{1.00}
\defV{final/frontier/gpt6_sol/direct/b}{56}
\defV{final/frontier/gpt6_sol/direct/c}{3}
\defV{final/frontier/gpt6_sol/direct/conflict/acc}{0.59}
\defV{final/frontier/gpt6_sol/direct/conflict/d}{+0.52}
\defV{final/frontier/gpt6_sol/direct/control/acc}{0.97}
\defV{final/frontier/gpt6_sol/direct/control/d}{+3.62}
\defV{final/frontier/gpt6_sol/direct/drop}{0.38}
\defV{final/frontier/gpt6_sol/direct/pholm}{$1\times10^{-13}$}
\defV{final/frontier/gpt6_sol/matched/b}{60}
\defV{final/frontier/gpt6_sol/matched/c}{1}
\defV{final/frontier/gpt6_sol/matched/conflict/acc}{0.55}
\defV{final/frontier/gpt6_sol/matched/conflict/d}{+0.29}
\defV{final/frontier/gpt6_sol/matched/control/acc}{0.97}
\defV{final/frontier/gpt6_sol/matched/control/d}{+3.68}
\defV{final/frontier/gpt6_sol/matched/drop}{0.42}
\defV{final/frontier/gpt6_sol/matched/pholm}{$1\times10^{-16}$}
\defV{final/frontier/gpt6_sol/matched/status}{yes}
\defV{final/frontier/gpt6_sol/name}{GPT-6.1 Sol}
\defV{final/frontier/gpt6_sol/white/b}{57}
\defV{final/frontier/gpt6_sol/white/c}{1}
\defV{final/frontier/gpt6_sol/white/conflict/acc}{0.59}
\defV{final/frontier/gpt6_sol/white/conflict/d}{+0.45}
\defV{final/frontier/gpt6_sol/white/control/acc}{0.99}
\defV{final/frontier/gpt6_sol/white/control/d}{+4.06}
\defV{final/frontier/gpt6_sol/white/drop}{0.40}
\defV{final/frontier/gpt6_sol/white/pholm}{$4\times10^{-16}$}
\defV{final/frontier/gpt6_sol/white/status}{yes}
\defV{final/frontier/items}{1120}
\defV{final/frontier/pairs}{140}
\defV{final/gemma4_12/answers}{6048}
\defV{final/gemma4_12/crop/b}{63}
\defV{final/gemma4_12/crop/c}{1}
\defV{final/gemma4_12/crop/ci}{[$-$4.75, $-$2.93]}
\defV{final/gemma4_12/crop/conflict/acc}{0.36}
\defV{final/gemma4_12/crop/conflict/c}{$-$1.29}
\defV{final/gemma4_12/crop/conflict/ci}{[$-$2.13, $-$0.92]}
\defV{final/gemma4_12/crop/conflict/d}{$-$1.47}
\defV{final/gemma4_12/crop/conflict/n}{140}
\defV{final/gemma4_12/crop/conflict/unparsed}{0}
\defV{final/gemma4_12/crop/control/acc}{0.81}
\defV{final/gemma4_12/crop/control/c}{$-$0.72}
\defV{final/gemma4_12/crop/control/d}{2.17}
\defV{final/gemma4_12/crop/control/n}{140}
\defV{final/gemma4_12/crop/control/unparsed}{0}
\defV{final/gemma4_12/crop/diff}{$-$3.64}
\defV{final/gemma4_12/crop/p}{$7\times10^{-18}$}
\defV{final/gemma4_12/crop/pairs}{70}
\defV{final/gemma4_12/crop/pholm}{--}
\defV{final/gemma4_12/crop/status}{--}
\defV{final/gemma4_12/cues_leader/b}{44}
\defV{final/gemma4_12/cues_leader/c}{1}
\defV{final/gemma4_12/cues_leader/ci}{[$-$3.84, $-$1.97]}
\defV{final/gemma4_12/cues_leader/conflict/acc}{0.65}
\defV{final/gemma4_12/cues_leader/conflict/c}{$-$0.04}
\defV{final/gemma4_12/cues_leader/conflict/ci}{[0.22, 1.33]}
\defV{final/gemma4_12/cues_leader/conflict/d}{0.74}
\defV{final/gemma4_12/cues_leader/conflict/n}{136}
\defV{final/gemma4_12/cues_leader/conflict/unparsed}{0}
\defV{final/gemma4_12/cues_leader/control/acc}{0.96}
\defV{final/gemma4_12/cues_leader/control/c}{0.25}
\defV{final/gemma4_12/cues_leader/control/d}{3.53}
\defV{final/gemma4_12/cues_leader/control/n}{136}
\defV{final/gemma4_12/cues_leader/control/unparsed}{0}
\defV{final/gemma4_12/cues_leader/diff}{$-$2.79}
\defV{final/gemma4_12/cues_leader/p}{$3\times10^{-12}$}
\defV{final/gemma4_12/cues_leader/pairs}{68}
\defV{final/gemma4_12/cues_leader/pholm}{--}
\defV{final/gemma4_12/cues_leader/status}{--}
\defV{final/gemma4_12/cues_region/b}{56}
\defV{final/gemma4_12/cues_region/c}{0}
\defV{final/gemma4_12/cues_region/ci}{[$-$4.06, $-$2.35]}
\defV{final/gemma4_12/cues_region/conflict/acc}{0.56}
\defV{final/gemma4_12/cues_region/conflict/c}{0.04}
\defV{final/gemma4_12/cues_region/conflict/ci}{[$-$0.25, 0.84]}
\defV{final/gemma4_12/cues_region/conflict/d}{0.28}
\defV{final/gemma4_12/cues_region/conflict/n}{140}
\defV{final/gemma4_12/cues_region/conflict/unparsed}{0}
\defV{final/gemma4_12/cues_region/control/acc}{0.96}
\defV{final/gemma4_12/cues_region/control/c}{0.00}
\defV{final/gemma4_12/cues_region/control/d}{3.30}
\defV{final/gemma4_12/cues_region/control/n}{140}
\defV{final/gemma4_12/cues_region/control/unparsed}{0}
\defV{final/gemma4_12/cues_region/diff}{$-$3.02}
\defV{final/gemma4_12/cues_region/p}{$3\times10^{-17}$}
\defV{final/gemma4_12/cues_region/pairs}{70}
\defV{final/gemma4_12/cues_region/pholm}{--}
\defV{final/gemma4_12/cues_region/status}{--}
\defV{final/gemma4_12/direct/b}{28}
\defV{final/gemma4_12/direct/c}{18}
\defV{final/gemma4_12/direct/ci}{[$-$0.92, 0.03]}
\defV{final/gemma4_12/direct/conflict/acc}{0.49}
\defV{final/gemma4_12/direct/conflict/c}{$-$0.50}
\defV{final/gemma4_12/direct/conflict/ci}{[$-$0.36, 0.28]}
\defV{final/gemma4_12/direct/conflict/d}{$-$0.04}
\defV{final/gemma4_12/direct/conflict/n}{140}
\defV{final/gemma4_12/direct/conflict/unparsed}{0}
\defV{final/gemma4_12/direct/control/acc}{0.56}
\defV{final/gemma4_12/direct/control/c}{$-$0.68}
\defV{final/gemma4_12/direct/control/d}{0.40}
\defV{final/gemma4_12/direct/control/n}{140}
\defV{final/gemma4_12/direct/control/unparsed}{0}
\defV{final/gemma4_12/direct/diff}{$-$0.44}
\defV{final/gemma4_12/direct/p}{0.18}
\defV{final/gemma4_12/direct/pairs}{70}
\defV{final/gemma4_12/direct/pholm}{--}
\defV{final/gemma4_12/direct/status}{n.d.}
\defV{final/gemma4_12/failed}{0}
\defV{final/gemma4_12/gate}{$-$0.23}
\defV{final/gemma4_12/gsi_placement/b}{30}
\defV{final/gemma4_12/gsi_placement/c}{28}
\defV{final/gemma4_12/gsi_placement/ci}{[$-$0.71, 0.54]}
\defV{final/gemma4_12/gsi_placement/conflict/acc}{0.51}
\defV{final/gemma4_12/gsi_placement/conflict/c}{$-$0.29}
\defV{final/gemma4_12/gsi_placement/conflict/ci}{[$-$0.34, 0.43]}
\defV{final/gemma4_12/gsi_placement/conflict/d}{0.04}
\defV{final/gemma4_12/gsi_placement/conflict/n}{128}
\defV{final/gemma4_12/gsi_placement/conflict/unparsed}{0}
\defV{final/gemma4_12/gsi_placement/control/acc}{0.52}
\defV{final/gemma4_12/gsi_placement/control/c}{$-$0.21}
\defV{final/gemma4_12/gsi_placement/control/d}{0.12}
\defV{final/gemma4_12/gsi_placement/control/n}{128}
\defV{final/gemma4_12/gsi_placement/control/unparsed}{0}
\defV{final/gemma4_12/gsi_placement/diff}{$-$0.08}
\defV{final/gemma4_12/gsi_placement/p}{0.90}
\defV{final/gemma4_12/gsi_placement/pairs}{64}
\defV{final/gemma4_12/gsi_placement/pholm}{--}
\defV{final/gemma4_12/gsi_placement/status}{n.d.}
\defV{final/gemma4_12/gsi_placement_k1/b}{11}
\defV{final/gemma4_12/gsi_placement_k1/c}{6}
\defV{final/gemma4_12/gsi_placement_k1/ci}{[$-$1.59, 0.30]}
\defV{final/gemma4_12/gsi_placement_k1/conflict/acc}{0.48}
\defV{final/gemma4_12/gsi_placement_k1/conflict/c}{$-$0.00}
\defV{final/gemma4_12/gsi_placement_k1/conflict/ci}{[$-$0.65, 0.43]}
\defV{final/gemma4_12/gsi_placement_k1/conflict/d}{$-$0.10}
\defV{final/gemma4_12/gsi_placement_k1/conflict/n}{46}
\defV{final/gemma4_12/gsi_placement_k1/conflict/unparsed}{0}
\defV{final/gemma4_12/gsi_placement_k1/control/acc}{0.59}
\defV{final/gemma4_12/gsi_placement_k1/control/c}{$-$0.50}
\defV{final/gemma4_12/gsi_placement_k1/control/d}{0.48}
\defV{final/gemma4_12/gsi_placement_k1/control/n}{46}
\defV{final/gemma4_12/gsi_placement_k1/control/unparsed}{0}
\defV{final/gemma4_12/gsi_placement_k1/diff}{$-$0.58}
\defV{final/gemma4_12/gsi_placement_k1/p}{0.33}
\defV{final/gemma4_12/gsi_placement_k1/pairs}{23}
\defV{final/gemma4_12/gsi_placement_k1/pholm}{--}
\defV{final/gemma4_12/gsi_placement_k1/status}{n.d.}
\defV{final/gemma4_12/isobar/b}{59}
\defV{final/gemma4_12/isobar/c}{11}
\defV{final/gemma4_12/isobar/ci}{[$-$2.43, $-$1.07]}
\defV{final/gemma4_12/isobar/conflict/acc}{0.66}
\defV{final/gemma4_12/isobar/conflict/c}{$-$0.30}
\defV{final/gemma4_12/isobar/conflict/ci}{[0.46, 1.30]}
\defV{final/gemma4_12/isobar/conflict/d}{0.85}
\defV{final/gemma4_12/isobar/conflict/n}{200}
\defV{final/gemma4_12/isobar/conflict/unparsed}{0}
\defV{final/gemma4_12/isobar/control/acc}{0.90}
\defV{final/gemma4_12/isobar/control/c}{$-$0.17}
\defV{final/gemma4_12/isobar/control/d}{2.55}
\defV{final/gemma4_12/isobar/control/n}{200}
\defV{final/gemma4_12/isobar/control/unparsed}{0}
\defV{final/gemma4_12/isobar/diff}{$-$1.70}
\defV{final/gemma4_12/isobar/p}{$4\times10^{-9}$}
\defV{final/gemma4_12/isobar/pairs}{100}
\defV{final/gemma4_12/isobar/pholm}{--}
\defV{final/gemma4_12/isobar/status}{--}
\defV{final/gemma4_12/isobar_text/b}{1}
\defV{final/gemma4_12/isobar_text/c}{5}
\defV{final/gemma4_12/isobar_text/ci}{[0.00, 0.49]}
\defV{final/gemma4_12/isobar_text/conflict/acc}{0.95}
\defV{final/gemma4_12/isobar_text/conflict/c}{0.63}
\defV{final/gemma4_12/isobar_text/conflict/ci}{[3.60, 4.28]}
\defV{final/gemma4_12/isobar_text/conflict/d}{3.90}
\defV{final/gemma4_12/isobar_text/conflict/n}{200}
\defV{final/gemma4_12/isobar_text/conflict/unparsed}{0}
\defV{final/gemma4_12/isobar_text/control/acc}{0.94}
\defV{final/gemma4_12/isobar_text/control/c}{0.74}
\defV{final/gemma4_12/isobar_text/control/d}{3.69}
\defV{final/gemma4_12/isobar_text/control/n}{200}
\defV{final/gemma4_12/isobar_text/control/unparsed}{0}
\defV{final/gemma4_12/isobar_text/diff}{+0.21}
\defV{final/gemma4_12/isobar_text/p}{0.22}
\defV{final/gemma4_12/isobar_text/pairs}{100}
\defV{final/gemma4_12/isobar_text/pholm}{--}
\defV{final/gemma4_12/isobar_text/status}{n.d.}
\defV{final/gemma4_12/linechart/b}{26}
\defV{final/gemma4_12/linechart/c}{12}
\defV{final/gemma4_12/linechart/ci}{[$-$1.51, 0.12]}
\defV{final/gemma4_12/linechart/conflict/acc}{0.74}
\defV{final/gemma4_12/linechart/conflict/c}{$-$0.09}
\defV{final/gemma4_12/linechart/conflict/ci}{[0.76, 1.87]}
\defV{final/gemma4_12/linechart/conflict/d}{1.29}
\defV{final/gemma4_12/linechart/conflict/n}{140}
\defV{final/gemma4_12/linechart/conflict/unparsed}{0}
\defV{final/gemma4_12/linechart/control/acc}{0.84}
\defV{final/gemma4_12/linechart/control/c}{$-$0.00}
\defV{final/gemma4_12/linechart/control/d}{1.97}
\defV{final/gemma4_12/linechart/control/n}{140}
\defV{final/gemma4_12/linechart/control/unparsed}{0}
\defV{final/gemma4_12/linechart/diff}{$-$0.68}
\defV{final/gemma4_12/linechart/p}{0.03}
\defV{final/gemma4_12/linechart/pairs}{70}
\defV{final/gemma4_12/linechart/pholm}{--}
\defV{final/gemma4_12/linechart/status}{--}
\defV{final/gemma4_12/linechart_colour/b}{0}
\defV{final/gemma4_12/linechart_colour/c}{0}
\defV{final/gemma4_12/linechart_colour/ci}{[0.00, 0.00]}
\defV{final/gemma4_12/linechart_colour/conflict/acc}{1.00}
\defV{final/gemma4_12/linechart_colour/conflict/c}{$-$0.00}
\defV{final/gemma4_12/linechart_colour/conflict/ci}{[4.91, 4.91]}
\defV{final/gemma4_12/linechart_colour/conflict/d}{4.91}
\defV{final/gemma4_12/linechart_colour/conflict/n}{140}
\defV{final/gemma4_12/linechart_colour/conflict/unparsed}{0}
\defV{final/gemma4_12/linechart_colour/control/acc}{1.00}
\defV{final/gemma4_12/linechart_colour/control/c}{$-$0.00}
\defV{final/gemma4_12/linechart_colour/control/d}{4.91}
\defV{final/gemma4_12/linechart_colour/control/n}{140}
\defV{final/gemma4_12/linechart_colour/control/unparsed}{0}
\defV{final/gemma4_12/linechart_colour/diff}{+0.00}
\defV{final/gemma4_12/linechart_colour/p}{1.00}
\defV{final/gemma4_12/linechart_colour/pairs}{70}
\defV{final/gemma4_12/linechart_colour/pholm}{--}
\defV{final/gemma4_12/linechart_colour/status}{n.d.}
\defV{final/gemma4_12/linechart_text/b}{0}
\defV{final/gemma4_12/linechart_text/c}{0}
\defV{final/gemma4_12/linechart_text/ci}{[0.00, 0.00]}
\defV{final/gemma4_12/linechart_text/conflict/acc}{1.00}
\defV{final/gemma4_12/linechart_text/conflict/c}{$-$0.00}
\defV{final/gemma4_12/linechart_text/conflict/ci}{[4.91, 4.91]}
\defV{final/gemma4_12/linechart_text/conflict/d}{4.91}
\defV{final/gemma4_12/linechart_text/conflict/n}{140}
\defV{final/gemma4_12/linechart_text/conflict/unparsed}{0}
\defV{final/gemma4_12/linechart_text/control/acc}{1.00}
\defV{final/gemma4_12/linechart_text/control/c}{$-$0.00}
\defV{final/gemma4_12/linechart_text/control/d}{4.91}
\defV{final/gemma4_12/linechart_text/control/n}{140}
\defV{final/gemma4_12/linechart_text/control/unparsed}{0}
\defV{final/gemma4_12/linechart_text/diff}{+0.00}
\defV{final/gemma4_12/linechart_text/p}{1.00}
\defV{final/gemma4_12/linechart_text/pairs}{70}
\defV{final/gemma4_12/linechart_text/pholm}{--}
\defV{final/gemma4_12/linechart_text/status}{n.d.}
\defV{final/gemma4_12/map/index_labels/acc}{0.62}
\defV{final/gemma4_12/map/index_labels/d}{0.74}
\defV{final/gemma4_12/map/index_labels/n}{120}
\defV{final/gemma4_12/map/structure_only/acc}{0.57}
\defV{final/gemma4_12/map/structure_only/d}{0.79}
\defV{final/gemma4_12/map/structure_only/n}{120}
\defV{final/gemma4_12/map/values_only/acc}{0.65}
\defV{final/gemma4_12/map/values_only/d}{0.82}
\defV{final/gemma4_12/map/values_only/n}{120}
\defV{final/gemma4_12/map/values_structure/acc}{0.90}
\defV{final/gemma4_12/map/values_structure/d}{2.85}
\defV{final/gemma4_12/map/values_structure/n}{120}
\defV{final/gemma4_12/matched/b}{14}
\defV{final/gemma4_12/matched/c}{7}
\defV{final/gemma4_12/matched/ci}{[$-$1.62, 0.22]}
\defV{final/gemma4_12/matched/conflict/acc}{0.44}
\defV{final/gemma4_12/matched/conflict/c}{$-$1.43}
\defV{final/gemma4_12/matched/conflict/ci}{[$-$1.58, $-$0.16]}
\defV{final/gemma4_12/matched/conflict/d}{$-$0.76}
\defV{final/gemma4_12/matched/conflict/n}{140}
\defV{final/gemma4_12/matched/conflict/unparsed}{0}
\defV{final/gemma4_12/matched/control/acc}{0.49}
\defV{final/gemma4_12/matched/control/c}{$-$1.47}
\defV{final/gemma4_12/matched/control/d}{$-$0.10}
\defV{final/gemma4_12/matched/control/n}{140}
\defV{final/gemma4_12/matched/control/unparsed}{0}
\defV{final/gemma4_12/matched/diff}{$-$0.66}
\defV{final/gemma4_12/matched/p}{0.19}
\defV{final/gemma4_12/matched/pairs}{70}
\defV{final/gemma4_12/matched/pholm}{--}
\defV{final/gemma4_12/matched/status}{n.d.}
\defV{final/gemma4_12/name}{Gemma-4 12B}
\defV{final/gemma4_12/pointer_linechart/b}{12}
\defV{final/gemma4_12/pointer_linechart/c}{6}
\defV{final/gemma4_12/pointer_linechart/ci}{[$-$1.35, 0.34]}
\defV{final/gemma4_12/pointer_linechart/conflict/acc}{0.75}
\defV{final/gemma4_12/pointer_linechart/conflict/c}{$-$0.05}
\defV{final/gemma4_12/pointer_linechart/conflict/ci}{[0.89, 1.99]}
\defV{final/gemma4_12/pointer_linechart/conflict/d}{1.37}
\defV{final/gemma4_12/pointer_linechart/conflict/n}{80}
\defV{final/gemma4_12/pointer_linechart/conflict/unparsed}{1}
\defV{final/gemma4_12/pointer_linechart/control/acc}{0.82}
\defV{final/gemma4_12/pointer_linechart/control/c}{$-$0.09}
\defV{final/gemma4_12/pointer_linechart/control/d}{1.82}
\defV{final/gemma4_12/pointer_linechart/control/n}{80}
\defV{final/gemma4_12/pointer_linechart/control/unparsed}{0}
\defV{final/gemma4_12/pointer_linechart/diff}{$-$0.45}
\defV{final/gemma4_12/pointer_linechart/p}{0.24}
\defV{final/gemma4_12/pointer_linechart/pairs}{40}
\defV{final/gemma4_12/pointer_linechart/pholm}{--}
\defV{final/gemma4_12/pointer_linechart/status}{--}
\defV{final/gemma4_12/pointer_scatter/b}{5}
\defV{final/gemma4_12/pointer_scatter/c}{0}
\defV{final/gemma4_12/pointer_scatter/ci}{[$-$2.29, $-$0.46]}
\defV{final/gemma4_12/pointer_scatter/conflict/acc}{0.94}
\defV{final/gemma4_12/pointer_scatter/conflict/c}{0.09}
\defV{final/gemma4_12/pointer_scatter/conflict/ci}{[2.21, 4.04]}
\defV{final/gemma4_12/pointer_scatter/conflict/d}{2.92}
\defV{final/gemma4_12/pointer_scatter/conflict/n}{80}
\defV{final/gemma4_12/pointer_scatter/conflict/unparsed}{0}
\defV{final/gemma4_12/pointer_scatter/control/acc}{1.00}
\defV{final/gemma4_12/pointer_scatter/control/c}{$-$0.00}
\defV{final/gemma4_12/pointer_scatter/control/d}{4.50}
\defV{final/gemma4_12/pointer_scatter/control/n}{80}
\defV{final/gemma4_12/pointer_scatter/control/unparsed}{0}
\defV{final/gemma4_12/pointer_scatter/diff}{$-$1.59}
\defV{final/gemma4_12/pointer_scatter/p}{0.06}
\defV{final/gemma4_12/pointer_scatter/pairs}{40}
\defV{final/gemma4_12/pointer_scatter/pholm}{--}
\defV{final/gemma4_12/pointer_scatter/status}{n.d.}
\defV{final/gemma4_12/public_data/b}{27}
\defV{final/gemma4_12/public_data/c}{7}
\defV{final/gemma4_12/public_data/ci}{[$-$2.75, $-$0.54]}
\defV{final/gemma4_12/public_data/conflict/acc}{0.61}
\defV{final/gemma4_12/public_data/conflict/c}{$-$0.36}
\defV{final/gemma4_12/public_data/conflict/ci}{[$-$0.07, 1.35]}
\defV{final/gemma4_12/public_data/conflict/d}{0.59}
\defV{final/gemma4_12/public_data/conflict/n}{80}
\defV{final/gemma4_12/public_data/conflict/unparsed}{0}
\defV{final/gemma4_12/public_data/control/acc}{0.86}
\defV{final/gemma4_12/public_data/control/c}{$-$0.28}
\defV{final/gemma4_12/public_data/control/d}{2.19}
\defV{final/gemma4_12/public_data/control/n}{80}
\defV{final/gemma4_12/public_data/control/unparsed}{0}
\defV{final/gemma4_12/public_data/diff}{$-$1.59}
\defV{final/gemma4_12/public_data/p}{$8\times10^{-4}$}
\defV{final/gemma4_12/public_data/pairs}{40}
\defV{final/gemma4_12/public_data/pholm}{--}
\defV{final/gemma4_12/public_data/status}{--}
\defV{final/gemma4_12/repo}{google/gemma-4-12B-it}
\defV{final/gemma4_12/revision}{707f0a3}
\defV{final/gemma4_12/scatter/b}{60}
\defV{final/gemma4_12/scatter/c}{3}
\defV{final/gemma4_12/scatter/ci}{[$-$3.31, $-$1.86]}
\defV{final/gemma4_12/scatter/conflict/acc}{0.67}
\defV{final/gemma4_12/scatter/conflict/c}{$-$0.05}
\defV{final/gemma4_12/scatter/conflict/ci}{[0.45, 1.34]}
\defV{final/gemma4_12/scatter/conflict/d}{0.87}
\defV{final/gemma4_12/scatter/conflict/n}{200}
\defV{final/gemma4_12/scatter/conflict/unparsed}{0}
\defV{final/gemma4_12/scatter/control/acc}{0.95}
\defV{final/gemma4_12/scatter/control/c}{0.05}
\defV{final/gemma4_12/scatter/control/d}{3.30}
\defV{final/gemma4_12/scatter/control/n}{200}
\defV{final/gemma4_12/scatter/control/unparsed}{0}
\defV{final/gemma4_12/scatter/diff}{$-$2.43}
\defV{final/gemma4_12/scatter/p}{$9\times10^{-15}$}
\defV{final/gemma4_12/scatter/pairs}{100}
\defV{final/gemma4_12/scatter/pholm}{--}
\defV{final/gemma4_12/scatter/status}{--}
\defV{final/gemma4_12/scatter_text/b}{0}
\defV{final/gemma4_12/scatter_text/c}{0}
\defV{final/gemma4_12/scatter_text/ci}{[0.00, 0.00]}
\defV{final/gemma4_12/scatter_text/conflict/acc}{1.00}
\defV{final/gemma4_12/scatter_text/conflict/c}{0.00}
\defV{final/gemma4_12/scatter_text/conflict/ci}{[5.16, 5.16]}
\defV{final/gemma4_12/scatter_text/conflict/d}{5.16}
\defV{final/gemma4_12/scatter_text/conflict/n}{200}
\defV{final/gemma4_12/scatter_text/conflict/unparsed}{0}
\defV{final/gemma4_12/scatter_text/control/acc}{1.00}
\defV{final/gemma4_12/scatter_text/control/c}{0.00}
\defV{final/gemma4_12/scatter_text/control/d}{5.16}
\defV{final/gemma4_12/scatter_text/control/n}{200}
\defV{final/gemma4_12/scatter_text/control/unparsed}{0}
\defV{final/gemma4_12/scatter_text/diff}{+0.00}
\defV{final/gemma4_12/scatter_text/p}{1.00}
\defV{final/gemma4_12/scatter_text/pairs}{100}
\defV{final/gemma4_12/scatter_text/pholm}{--}
\defV{final/gemma4_12/scatter_text/status}{n.d.}
\defV{final/gemma4_12/separation/b}{59}
\defV{final/gemma4_12/separation/c}{5}
\defV{final/gemma4_12/separation/ci}{[$-$2.76, $-$1.64]}
\defV{final/gemma4_12/separation/conflict/acc}{0.45}
\defV{final/gemma4_12/separation/conflict/c}{$-$0.28}
\defV{final/gemma4_12/separation/conflict/ci}{[$-$0.66, 0.18]}
\defV{final/gemma4_12/separation/conflict/d}{$-$0.24}
\defV{final/gemma4_12/separation/conflict/n}{140}
\defV{final/gemma4_12/separation/conflict/unparsed}{1}
\defV{final/gemma4_12/separation/control/acc}{0.84}
\defV{final/gemma4_12/separation/control/c}{$-$0.08}
\defV{final/gemma4_12/separation/control/d}{1.92}
\defV{final/gemma4_12/separation/control/n}{140}
\defV{final/gemma4_12/separation/control/unparsed}{0}
\defV{final/gemma4_12/separation/diff}{$-$2.16}
\defV{final/gemma4_12/separation/p}{$9\times10^{-13}$}
\defV{final/gemma4_12/separation/pairs}{70}
\defV{final/gemma4_12/separation/pholm}{--}
\defV{final/gemma4_12/separation/status}{--}
\defV{final/gemma4_12/separation_text/b}{0}
\defV{final/gemma4_12/separation_text/c}{0}
\defV{final/gemma4_12/separation_text/ci}{[0.00, 0.00]}
\defV{final/gemma4_12/separation_text/conflict/acc}{1.00}
\defV{final/gemma4_12/separation_text/conflict/c}{$-$0.00}
\defV{final/gemma4_12/separation_text/conflict/ci}{[4.91, 4.91]}
\defV{final/gemma4_12/separation_text/conflict/d}{4.91}
\defV{final/gemma4_12/separation_text/conflict/n}{140}
\defV{final/gemma4_12/separation_text/conflict/unparsed}{0}
\defV{final/gemma4_12/separation_text/control/acc}{1.00}
\defV{final/gemma4_12/separation_text/control/c}{$-$0.00}
\defV{final/gemma4_12/separation_text/control/d}{4.91}
\defV{final/gemma4_12/separation_text/control/n}{140}
\defV{final/gemma4_12/separation_text/control/unparsed}{0}
\defV{final/gemma4_12/separation_text/diff}{+0.00}
\defV{final/gemma4_12/separation_text/p}{1.00}
\defV{final/gemma4_12/separation_text/pairs}{70}
\defV{final/gemma4_12/separation_text/pholm}{--}
\defV{final/gemma4_12/separation_text/status}{n.d.}
\defV{final/gemma4_12/truncated}{2}
\defV{final/gemma4_12/truncated/pointer_control_linechart}{1}
\defV{final/gemma4_12/truncated/separation_shortcut_free}{1}
\defV{final/gemma4_12/unparsed}{2}
\defV{final/gemma4_12/unparsed/pct}{0.0}
\defV{final/gemma4_12/white/b}{12}
\defV{final/gemma4_12/white/c}{8}
\defV{final/gemma4_12/white/ci}{[$-$1.17, 0.45]}
\defV{final/gemma4_12/white/conditional}{0.83}
\defV{final/gemma4_12/white/conflict/acc}{0.49}
\defV{final/gemma4_12/white/conflict/c}{$-$1.34}
\defV{final/gemma4_12/white/conflict/ci}{[$-$0.85, 0.48]}
\defV{final/gemma4_12/white/conflict/d}{$-$0.17}
\defV{final/gemma4_12/white/conflict/n}{140}
\defV{final/gemma4_12/white/conflict/unparsed}{0}
\defV{final/gemma4_12/white/control/acc}{0.51}
\defV{final/gemma4_12/white/control/c}{$-$1.25}
\defV{final/gemma4_12/white/control/d}{0.15}
\defV{final/gemma4_12/white/control/n}{140}
\defV{final/gemma4_12/white/control/unparsed}{0}
\defV{final/gemma4_12/white/diff}{$-$0.33}
\defV{final/gemma4_12/white/p}{0.50}
\defV{final/gemma4_12/white/pairs}{70}
\defV{final/gemma4_12/white/pholm}{0.50}
\defV{final/gemma4_12/white/reversal}{no}
\defV{final/gemma4_12/white/status}{n.d.}
\defV{final/gemma4_31/answers}{6048}
\defV{final/gemma4_31/crop/b}{84}
\defV{final/gemma4_31/crop/c}{0}
\defV{final/gemma4_31/crop/ci}{[$-$5.64, $-$3.81]}
\defV{final/gemma4_31/crop/conflict/acc}{0.28}
\defV{final/gemma4_31/crop/conflict/c}{$-$0.94}
\defV{final/gemma4_31/crop/conflict/ci}{[$-$2.53, $-$1.21]}
\defV{final/gemma4_31/crop/conflict/d}{$-$1.74}
\defV{final/gemma4_31/crop/conflict/n}{140}
\defV{final/gemma4_31/crop/conflict/unparsed}{0}
\defV{final/gemma4_31/crop/control/acc}{0.88}
\defV{final/gemma4_31/crop/control/c}{$-$0.65}
\defV{final/gemma4_31/crop/control/d}{2.76}
\defV{final/gemma4_31/crop/control/n}{140}
\defV{final/gemma4_31/crop/control/unparsed}{0}
\defV{final/gemma4_31/crop/diff}{$-$4.50}
\defV{final/gemma4_31/crop/p}{$1\times10^{-25}$}
\defV{final/gemma4_31/crop/pairs}{70}
\defV{final/gemma4_31/crop/pholm}{$4\times10^{-25}$}
\defV{final/gemma4_31/crop/status}{yes}
\defV{final/gemma4_31/cues_leader/b}{30}
\defV{final/gemma4_31/cues_leader/c}{2}
\defV{final/gemma4_31/cues_leader/ci}{[$-$3.65, $-$1.21]}
\defV{final/gemma4_31/cues_leader/conflict/acc}{0.76}
\defV{final/gemma4_31/cues_leader/conflict/c}{$-$0.05}
\defV{final/gemma4_31/cues_leader/conflict/ci}{[0.86, 2.09]}
\defV{final/gemma4_31/cues_leader/conflict/d}{1.42}
\defV{final/gemma4_31/cues_leader/conflict/n}{136}
\defV{final/gemma4_31/cues_leader/conflict/unparsed}{0}
\defV{final/gemma4_31/cues_leader/control/acc}{0.97}
\defV{final/gemma4_31/cues_leader/control/c}{0.00}
\defV{final/gemma4_31/cues_leader/control/d}{3.59}
\defV{final/gemma4_31/cues_leader/control/n}{136}
\defV{final/gemma4_31/cues_leader/control/unparsed}{0}
\defV{final/gemma4_31/cues_leader/diff}{$-$2.17}
\defV{final/gemma4_31/cues_leader/p}{$2\times10^{-7}$}
\defV{final/gemma4_31/cues_leader/pairs}{68}
\defV{final/gemma4_31/cues_leader/pholm}{--}
\defV{final/gemma4_31/cues_leader/status}{--}
\defV{final/gemma4_31/cues_region/b}{31}
\defV{final/gemma4_31/cues_region/c}{2}
\defV{final/gemma4_31/cues_region/ci}{[$-$3.40, $-$1.31]}
\defV{final/gemma4_31/cues_region/conflict/acc}{0.76}
\defV{final/gemma4_31/cues_region/conflict/c}{$-$0.07}
\defV{final/gemma4_31/cues_region/conflict/ci}{[0.93, 2.02]}
\defV{final/gemma4_31/cues_region/conflict/d}{1.42}
\defV{final/gemma4_31/cues_region/conflict/n}{140}
\defV{final/gemma4_31/cues_region/conflict/unparsed}{0}
\defV{final/gemma4_31/cues_region/control/acc}{0.97}
\defV{final/gemma4_31/cues_region/control/c}{$-$0.00}
\defV{final/gemma4_31/cues_region/control/d}{3.62}
\defV{final/gemma4_31/cues_region/control/n}{140}
\defV{final/gemma4_31/cues_region/control/unparsed}{0}
\defV{final/gemma4_31/cues_region/diff}{$-$2.20}
\defV{final/gemma4_31/cues_region/p}{$1\times10^{-7}$}
\defV{final/gemma4_31/cues_region/pairs}{70}
\defV{final/gemma4_31/cues_region/pholm}{--}
\defV{final/gemma4_31/cues_region/status}{--}
\defV{final/gemma4_31/direct/b}{67}
\defV{final/gemma4_31/direct/c}{0}
\defV{final/gemma4_31/direct/ci}{[$-$4.87, $-$3.21]}
\defV{final/gemma4_31/direct/conflict/acc}{0.31}
\defV{final/gemma4_31/direct/conflict/c}{$-$1.01}
\defV{final/gemma4_31/direct/conflict/ci}{[$-$2.31, $-$1.09]}
\defV{final/gemma4_31/direct/conflict/d}{$-$1.60}
\defV{final/gemma4_31/direct/conflict/n}{140}
\defV{final/gemma4_31/direct/conflict/unparsed}{0}
\defV{final/gemma4_31/direct/control/acc}{0.79}
\defV{final/gemma4_31/direct/control/c}{$-$0.91}
\defV{final/gemma4_31/direct/control/d}{2.24}
\defV{final/gemma4_31/direct/control/n}{140}
\defV{final/gemma4_31/direct/control/unparsed}{0}
\defV{final/gemma4_31/direct/diff}{$-$3.84}
\defV{final/gemma4_31/direct/p}{$1\times10^{-20}$}
\defV{final/gemma4_31/direct/pairs}{70}
\defV{final/gemma4_31/direct/pholm}{$7\times10^{-20}$}
\defV{final/gemma4_31/direct/status}{yes}
\defV{final/gemma4_31/failed}{0}
\defV{final/gemma4_31/gate}{2.44}
\defV{final/gemma4_31/gsi_placement/b}{38}
\defV{final/gemma4_31/gsi_placement/c}{2}
\defV{final/gemma4_31/gsi_placement/ci}{[$-$3.12, $-$1.50]}
\defV{final/gemma4_31/gsi_placement/conflict/acc}{0.40}
\defV{final/gemma4_31/gsi_placement/conflict/c}{$-$1.05}
\defV{final/gemma4_31/gsi_placement/conflict/ci}{[$-$1.47, $-$0.44]}
\defV{final/gemma4_31/gsi_placement/conflict/d}{$-$0.87}
\defV{final/gemma4_31/gsi_placement/conflict/n}{128}
\defV{final/gemma4_31/gsi_placement/conflict/unparsed}{0}
\defV{final/gemma4_31/gsi_placement/control/acc}{0.68}
\defV{final/gemma4_31/gsi_placement/control/c}{$-$0.84}
\defV{final/gemma4_31/gsi_placement/control/d}{1.29}
\defV{final/gemma4_31/gsi_placement/control/n}{128}
\defV{final/gemma4_31/gsi_placement/control/unparsed}{0}
\defV{final/gemma4_31/gsi_placement/diff}{$-$2.15}
\defV{final/gemma4_31/gsi_placement/p}{$1\times10^{-9}$}
\defV{final/gemma4_31/gsi_placement/pairs}{64}
\defV{final/gemma4_31/gsi_placement/pholm}{$2\times10^{-8}$}
\defV{final/gemma4_31/gsi_placement/status}{yes}
\defV{final/gemma4_31/gsi_placement_k1/b}{7}
\defV{final/gemma4_31/gsi_placement_k1/c}{4}
\defV{final/gemma4_31/gsi_placement_k1/ci}{[$-$1.77, 0.52]}
\defV{final/gemma4_31/gsi_placement_k1/conflict/acc}{0.54}
\defV{final/gemma4_31/gsi_placement_k1/conflict/c}{$-$1.07}
\defV{final/gemma4_31/gsi_placement_k1/conflict/ci}{[$-$0.65, 1.43]}
\defV{final/gemma4_31/gsi_placement_k1/conflict/d}{0.37}
\defV{final/gemma4_31/gsi_placement_k1/conflict/n}{46}
\defV{final/gemma4_31/gsi_placement_k1/conflict/unparsed}{0}
\defV{final/gemma4_31/gsi_placement_k1/control/acc}{0.61}
\defV{final/gemma4_31/gsi_placement_k1/control/c}{$-$1.07}
\defV{final/gemma4_31/gsi_placement_k1/control/d}{0.92}
\defV{final/gemma4_31/gsi_placement_k1/control/n}{46}
\defV{final/gemma4_31/gsi_placement_k1/control/unparsed}{0}
\defV{final/gemma4_31/gsi_placement_k1/diff}{$-$0.55}
\defV{final/gemma4_31/gsi_placement_k1/p}{0.55}
\defV{final/gemma4_31/gsi_placement_k1/pairs}{23}
\defV{final/gemma4_31/gsi_placement_k1/pholm}{--}
\defV{final/gemma4_31/gsi_placement_k1/status}{--}
\defV{final/gemma4_31/isobar/b}{79}
\defV{final/gemma4_31/isobar/c}{7}
\defV{final/gemma4_31/isobar/ci}{[$-$3.17, $-$1.84]}
\defV{final/gemma4_31/isobar/conflict/acc}{0.55}
\defV{final/gemma4_31/isobar/conflict/c}{$-$0.37}
\defV{final/gemma4_31/isobar/conflict/ci}{[$-$0.16, 0.66]}
\defV{final/gemma4_31/isobar/conflict/d}{0.24}
\defV{final/gemma4_31/isobar/conflict/n}{200}
\defV{final/gemma4_31/isobar/conflict/unparsed}{0}
\defV{final/gemma4_31/isobar/control/acc}{0.91}
\defV{final/gemma4_31/isobar/control/c}{$-$0.27}
\defV{final/gemma4_31/isobar/control/d}{2.67}
\defV{final/gemma4_31/isobar/control/n}{200}
\defV{final/gemma4_31/isobar/control/unparsed}{0}
\defV{final/gemma4_31/isobar/diff}{$-$2.43}
\defV{final/gemma4_31/isobar/p}{$2\times10^{-16}$}
\defV{final/gemma4_31/isobar/pairs}{100}
\defV{final/gemma4_31/isobar/pholm}{$3\times10^{-15}$}
\defV{final/gemma4_31/isobar/status}{yes}
\defV{final/gemma4_31/isobar_text/b}{1}
\defV{final/gemma4_31/isobar_text/c}{0}
\defV{final/gemma4_31/isobar_text/ci}{[$-$0.76, 0.00]}
\defV{final/gemma4_31/isobar_text/conflict/acc}{0.99}
\defV{final/gemma4_31/isobar_text/conflict/c}{$-$0.20}
\defV{final/gemma4_31/isobar_text/conflict/ci}{[4.40, 5.16]}
\defV{final/gemma4_31/isobar_text/conflict/d}{4.75}
\defV{final/gemma4_31/isobar_text/conflict/n}{200}
\defV{final/gemma4_31/isobar_text/conflict/unparsed}{0}
\defV{final/gemma4_31/isobar_text/control/acc}{1.00}
\defV{final/gemma4_31/isobar_text/control/c}{0.00}
\defV{final/gemma4_31/isobar_text/control/d}{5.16}
\defV{final/gemma4_31/isobar_text/control/n}{200}
\defV{final/gemma4_31/isobar_text/control/unparsed}{0}
\defV{final/gemma4_31/isobar_text/diff}{$-$0.41}
\defV{final/gemma4_31/isobar_text/p}{1.00}
\defV{final/gemma4_31/isobar_text/pairs}{100}
\defV{final/gemma4_31/isobar_text/pholm}{--}
\defV{final/gemma4_31/isobar_text/status}{n.d.}
\defV{final/gemma4_31/linechart/b}{30}
\defV{final/gemma4_31/linechart/c}{10}
\defV{final/gemma4_31/linechart/ci}{[$-$1.76, $-$0.24]}
\defV{final/gemma4_31/linechart/conflict/acc}{0.70}
\defV{final/gemma4_31/linechart/conflict/c}{$-$0.04}
\defV{final/gemma4_31/linechart/conflict/ci}{[0.46, 1.61]}
\defV{final/gemma4_31/linechart/conflict/d}{1.03}
\defV{final/gemma4_31/linechart/conflict/n}{140}
\defV{final/gemma4_31/linechart/conflict/unparsed}{0}
\defV{final/gemma4_31/linechart/control/acc}{0.84}
\defV{final/gemma4_31/linechart/control/c}{0.12}
\defV{final/gemma4_31/linechart/control/d}{1.99}
\defV{final/gemma4_31/linechart/control/n}{140}
\defV{final/gemma4_31/linechart/control/unparsed}{0}
\defV{final/gemma4_31/linechart/diff}{$-$0.95}
\defV{final/gemma4_31/linechart/p}{0.002}
\defV{final/gemma4_31/linechart/pairs}{70}
\defV{final/gemma4_31/linechart/pholm}{--}
\defV{final/gemma4_31/linechart/status}{--}
\defV{final/gemma4_31/linechart_colour/b}{0}
\defV{final/gemma4_31/linechart_colour/c}{0}
\defV{final/gemma4_31/linechart_colour/ci}{[0.00, 0.00]}
\defV{final/gemma4_31/linechart_colour/conflict/acc}{1.00}
\defV{final/gemma4_31/linechart_colour/conflict/c}{$-$0.00}
\defV{final/gemma4_31/linechart_colour/conflict/ci}{[4.91, 4.91]}
\defV{final/gemma4_31/linechart_colour/conflict/d}{4.91}
\defV{final/gemma4_31/linechart_colour/conflict/n}{140}
\defV{final/gemma4_31/linechart_colour/conflict/unparsed}{0}
\defV{final/gemma4_31/linechart_colour/control/acc}{1.00}
\defV{final/gemma4_31/linechart_colour/control/c}{$-$0.00}
\defV{final/gemma4_31/linechart_colour/control/d}{4.91}
\defV{final/gemma4_31/linechart_colour/control/n}{140}
\defV{final/gemma4_31/linechart_colour/control/unparsed}{0}
\defV{final/gemma4_31/linechart_colour/diff}{+0.00}
\defV{final/gemma4_31/linechart_colour/p}{1.00}
\defV{final/gemma4_31/linechart_colour/pairs}{70}
\defV{final/gemma4_31/linechart_colour/pholm}{--}
\defV{final/gemma4_31/linechart_colour/status}{n.d.}
\defV{final/gemma4_31/linechart_text/b}{0}
\defV{final/gemma4_31/linechart_text/c}{0}
\defV{final/gemma4_31/linechart_text/ci}{[0.00, 0.00]}
\defV{final/gemma4_31/linechart_text/conflict/acc}{1.00}
\defV{final/gemma4_31/linechart_text/conflict/c}{$-$0.00}
\defV{final/gemma4_31/linechart_text/conflict/ci}{[4.91, 4.91]}
\defV{final/gemma4_31/linechart_text/conflict/d}{4.91}
\defV{final/gemma4_31/linechart_text/conflict/n}{140}
\defV{final/gemma4_31/linechart_text/conflict/unparsed}{0}
\defV{final/gemma4_31/linechart_text/control/acc}{1.00}
\defV{final/gemma4_31/linechart_text/control/c}{$-$0.00}
\defV{final/gemma4_31/linechart_text/control/d}{4.91}
\defV{final/gemma4_31/linechart_text/control/n}{140}
\defV{final/gemma4_31/linechart_text/control/unparsed}{0}
\defV{final/gemma4_31/linechart_text/diff}{+0.00}
\defV{final/gemma4_31/linechart_text/p}{1.00}
\defV{final/gemma4_31/linechart_text/pairs}{70}
\defV{final/gemma4_31/linechart_text/pholm}{--}
\defV{final/gemma4_31/linechart_text/status}{n.d.}
\defV{final/gemma4_31/map/index_labels/acc}{0.84}
\defV{final/gemma4_31/map/index_labels/d}{2.19}
\defV{final/gemma4_31/map/index_labels/n}{120}
\defV{final/gemma4_31/map/structure_only/acc}{0.78}
\defV{final/gemma4_31/map/structure_only/d}{1.53}
\defV{final/gemma4_31/map/structure_only/n}{120}
\defV{final/gemma4_31/map/values_only/acc}{0.78}
\defV{final/gemma4_31/map/values_only/d}{1.78}
\defV{final/gemma4_31/map/values_only/n}{120}
\defV{final/gemma4_31/map/values_structure/acc}{1.00}
\defV{final/gemma4_31/map/values_structure/d}{4.80}
\defV{final/gemma4_31/map/values_structure/n}{120}
\defV{final/gemma4_31/matched/b}{61}
\defV{final/gemma4_31/matched/c}{2}
\defV{final/gemma4_31/matched/ci}{[$-$4.41, $-$2.77]}
\defV{final/gemma4_31/matched/conflict/acc}{0.31}
\defV{final/gemma4_31/matched/conflict/c}{$-$0.91}
\defV{final/gemma4_31/matched/conflict/ci}{[$-$2.21, $-$1.02]}
\defV{final/gemma4_31/matched/conflict/d}{$-$1.47}
\defV{final/gemma4_31/matched/conflict/n}{140}
\defV{final/gemma4_31/matched/conflict/unparsed}{0}
\defV{final/gemma4_31/matched/control/acc}{0.73}
\defV{final/gemma4_31/matched/control/c}{$-$1.05}
\defV{final/gemma4_31/matched/control/d}{1.96}
\defV{final/gemma4_31/matched/control/n}{140}
\defV{final/gemma4_31/matched/control/unparsed}{0}
\defV{final/gemma4_31/matched/diff}{$-$3.43}
\defV{final/gemma4_31/matched/p}{$4\times10^{-16}$}
\defV{final/gemma4_31/matched/pairs}{70}
\defV{final/gemma4_31/matched/pholm}{$3\times10^{-15}$}
\defV{final/gemma4_31/matched/status}{yes}
\defV{final/gemma4_31/name}{Gemma-4 31B}
\defV{final/gemma4_31/pointer_linechart/b}{3}
\defV{final/gemma4_31/pointer_linechart/c}{0}
\defV{final/gemma4_31/pointer_linechart/ci}{[$-$1.13, 0.00]}
\defV{final/gemma4_31/pointer_linechart/conflict/acc}{0.89}
\defV{final/gemma4_31/pointer_linechart/conflict/c}{0.06}
\defV{final/gemma4_31/pointer_linechart/conflict/ci}{[1.81, 3.16]}
\defV{final/gemma4_31/pointer_linechart/conflict/d}{2.33}
\defV{final/gemma4_31/pointer_linechart/conflict/n}{80}
\defV{final/gemma4_31/pointer_linechart/conflict/unparsed}{0}
\defV{final/gemma4_31/pointer_linechart/control/acc}{0.93}
\defV{final/gemma4_31/pointer_linechart/control/c}{0.16}
\defV{final/gemma4_31/pointer_linechart/control/d}{2.77}
\defV{final/gemma4_31/pointer_linechart/control/n}{80}
\defV{final/gemma4_31/pointer_linechart/control/unparsed}{0}
\defV{final/gemma4_31/pointer_linechart/diff}{$-$0.44}
\defV{final/gemma4_31/pointer_linechart/p}{0.25}
\defV{final/gemma4_31/pointer_linechart/pairs}{40}
\defV{final/gemma4_31/pointer_linechart/pholm}{--}
\defV{final/gemma4_31/pointer_linechart/status}{n.d.}
\defV{final/gemma4_31/pointer_scatter/b}{5}
\defV{final/gemma4_31/pointer_scatter/c}{0}
\defV{final/gemma4_31/pointer_scatter/ci}{[$-$2.17, $-$0.46]}
\defV{final/gemma4_31/pointer_scatter/conflict/acc}{0.94}
\defV{final/gemma4_31/pointer_scatter/conflict/c}{$-$0.09}
\defV{final/gemma4_31/pointer_scatter/conflict/ci}{[2.33, 4.04]}
\defV{final/gemma4_31/pointer_scatter/conflict/d}{2.92}
\defV{final/gemma4_31/pointer_scatter/conflict/n}{80}
\defV{final/gemma4_31/pointer_scatter/conflict/unparsed}{0}
\defV{final/gemma4_31/pointer_scatter/control/acc}{1.00}
\defV{final/gemma4_31/pointer_scatter/control/c}{$-$0.00}
\defV{final/gemma4_31/pointer_scatter/control/d}{4.50}
\defV{final/gemma4_31/pointer_scatter/control/n}{80}
\defV{final/gemma4_31/pointer_scatter/control/unparsed}{0}
\defV{final/gemma4_31/pointer_scatter/diff}{$-$1.59}
\defV{final/gemma4_31/pointer_scatter/p}{0.06}
\defV{final/gemma4_31/pointer_scatter/pairs}{40}
\defV{final/gemma4_31/pointer_scatter/pholm}{--}
\defV{final/gemma4_31/pointer_scatter/status}{n.d.}
\defV{final/gemma4_31/public_data/b}{20}
\defV{final/gemma4_31/public_data/c}{2}
\defV{final/gemma4_31/public_data/ci}{[$-$3.54, $-$1.20]}
\defV{final/gemma4_31/public_data/conflict/acc}{0.75}
\defV{final/gemma4_31/public_data/conflict/c}{$-$0.23}
\defV{final/gemma4_31/public_data/conflict/ci}{[0.66, 2.13]}
\defV{final/gemma4_31/public_data/conflict/d}{1.35}
\defV{final/gemma4_31/public_data/conflict/n}{80}
\defV{final/gemma4_31/public_data/conflict/unparsed}{0}
\defV{final/gemma4_31/public_data/control/acc}{0.97}
\defV{final/gemma4_31/public_data/control/c}{0.00}
\defV{final/gemma4_31/public_data/control/d}{3.58}
\defV{final/gemma4_31/public_data/control/n}{80}
\defV{final/gemma4_31/public_data/control/unparsed}{0}
\defV{final/gemma4_31/public_data/diff}{$-$2.24}
\defV{final/gemma4_31/public_data/p}{$1\times10^{-4}$}
\defV{final/gemma4_31/public_data/pairs}{40}
\defV{final/gemma4_31/public_data/pholm}{--}
\defV{final/gemma4_31/public_data/status}{--}
\defV{final/gemma4_31/repo}{google/gemma-4-31B-it}
\defV{final/gemma4_31/revision}{fb9ae26}
\defV{final/gemma4_31/scatter/b}{41}
\defV{final/gemma4_31/scatter/c}{3}
\defV{final/gemma4_31/scatter/ci}{[$-$2.95, $-$1.27]}
\defV{final/gemma4_31/scatter/conflict/acc}{0.77}
\defV{final/gemma4_31/scatter/conflict/c}{$-$0.00}
\defV{final/gemma4_31/scatter/conflict/ci}{[0.98, 2.00]}
\defV{final/gemma4_31/scatter/conflict/d}{1.46}
\defV{final/gemma4_31/scatter/conflict/n}{200}
\defV{final/gemma4_31/scatter/conflict/unparsed}{0}
\defV{final/gemma4_31/scatter/control/acc}{0.96}
\defV{final/gemma4_31/scatter/control/c}{0.00}
\defV{final/gemma4_31/scatter/control/d}{3.40}
\defV{final/gemma4_31/scatter/control/n}{200}
\defV{final/gemma4_31/scatter/control/unparsed}{0}
\defV{final/gemma4_31/scatter/diff}{$-$1.94}
\defV{final/gemma4_31/scatter/p}{$2\times10^{-9}$}
\defV{final/gemma4_31/scatter/pairs}{100}
\defV{final/gemma4_31/scatter/pholm}{$2\times10^{-8}$}
\defV{final/gemma4_31/scatter/status}{yes}
\defV{final/gemma4_31/scatter_text/b}{0}
\defV{final/gemma4_31/scatter_text/c}{0}
\defV{final/gemma4_31/scatter_text/ci}{[0.00, 0.00]}
\defV{final/gemma4_31/scatter_text/conflict/acc}{1.00}
\defV{final/gemma4_31/scatter_text/conflict/c}{0.00}
\defV{final/gemma4_31/scatter_text/conflict/ci}{[5.16, 5.16]}
\defV{final/gemma4_31/scatter_text/conflict/d}{5.16}
\defV{final/gemma4_31/scatter_text/conflict/n}{200}
\defV{final/gemma4_31/scatter_text/conflict/unparsed}{0}
\defV{final/gemma4_31/scatter_text/control/acc}{1.00}
\defV{final/gemma4_31/scatter_text/control/c}{0.00}
\defV{final/gemma4_31/scatter_text/control/d}{5.16}
\defV{final/gemma4_31/scatter_text/control/n}{200}
\defV{final/gemma4_31/scatter_text/control/unparsed}{0}
\defV{final/gemma4_31/scatter_text/diff}{+0.00}
\defV{final/gemma4_31/scatter_text/p}{1.00}
\defV{final/gemma4_31/scatter_text/pairs}{100}
\defV{final/gemma4_31/scatter_text/pholm}{--}
\defV{final/gemma4_31/scatter_text/status}{n.d.}
\defV{final/gemma4_31/separation/b}{57}
\defV{final/gemma4_31/separation/c}{6}
\defV{final/gemma4_31/separation/ci}{[$-$2.59, $-$1.50]}
\defV{final/gemma4_31/separation/conflict/acc}{0.41}
\defV{final/gemma4_31/separation/conflict/c}{$-$0.41}
\defV{final/gemma4_31/separation/conflict/ci}{[$-$0.83, $-$0.12]}
\defV{final/gemma4_31/separation/conflict/d}{$-$0.46}
\defV{final/gemma4_31/separation/conflict/n}{140}
\defV{final/gemma4_31/separation/conflict/unparsed}{0}
\defV{final/gemma4_31/separation/control/acc}{0.78}
\defV{final/gemma4_31/separation/control/c}{$-$0.21}
\defV{final/gemma4_31/separation/control/d}{1.54}
\defV{final/gemma4_31/separation/control/n}{140}
\defV{final/gemma4_31/separation/control/unparsed}{0}
\defV{final/gemma4_31/separation/diff}{$-$2.01}
\defV{final/gemma4_31/separation/p}{$2\times10^{-11}$}
\defV{final/gemma4_31/separation/pairs}{70}
\defV{final/gemma4_31/separation/pholm}{$3\times10^{-10}$}
\defV{final/gemma4_31/separation/status}{yes}
\defV{final/gemma4_31/separation_text/b}{0}
\defV{final/gemma4_31/separation_text/c}{0}
\defV{final/gemma4_31/separation_text/ci}{[0.00, 0.00]}
\defV{final/gemma4_31/separation_text/conflict/acc}{1.00}
\defV{final/gemma4_31/separation_text/conflict/c}{$-$0.00}
\defV{final/gemma4_31/separation_text/conflict/ci}{[4.91, 4.91]}
\defV{final/gemma4_31/separation_text/conflict/d}{4.91}
\defV{final/gemma4_31/separation_text/conflict/n}{140}
\defV{final/gemma4_31/separation_text/conflict/unparsed}{0}
\defV{final/gemma4_31/separation_text/control/acc}{1.00}
\defV{final/gemma4_31/separation_text/control/c}{$-$0.00}
\defV{final/gemma4_31/separation_text/control/d}{4.91}
\defV{final/gemma4_31/separation_text/control/n}{140}
\defV{final/gemma4_31/separation_text/control/unparsed}{0}
\defV{final/gemma4_31/separation_text/diff}{+0.00}
\defV{final/gemma4_31/separation_text/p}{1.00}
\defV{final/gemma4_31/separation_text/pairs}{70}
\defV{final/gemma4_31/separation_text/pholm}{--}
\defV{final/gemma4_31/separation_text/status}{n.d.}
\defV{final/gemma4_31/truncated}{0}
\defV{final/gemma4_31/unparsed}{0}
\defV{final/gemma4_31/unparsed/pct}{0.0}
\defV{final/gemma4_31/white/b}{62}
\defV{final/gemma4_31/white/c}{1}
\defV{final/gemma4_31/white/ci}{[$-$4.95, $-$3.61]}
\defV{final/gemma4_31/white/conditional}{0.39}
\defV{final/gemma4_31/white/conflict/acc}{0.29}
\defV{final/gemma4_31/white/conflict/c}{$-$1.10}
\defV{final/gemma4_31/white/conflict/ci}{[$-$2.49, $-$1.37]}
\defV{final/gemma4_31/white/conflict/d}{$-$1.85}
\defV{final/gemma4_31/white/conflict/n}{140}
\defV{final/gemma4_31/white/conflict/unparsed}{0}
\defV{final/gemma4_31/white/control/acc}{0.73}
\defV{final/gemma4_31/white/control/c}{$-$1.28}
\defV{final/gemma4_31/white/control/d}{2.35}
\defV{final/gemma4_31/white/control/n}{140}
\defV{final/gemma4_31/white/control/unparsed}{0}
\defV{final/gemma4_31/white/diff}{$-$4.20}
\defV{final/gemma4_31/white/p}{$1\times10^{-17}$}
\defV{final/gemma4_31/white/pairs}{70}
\defV{final/gemma4_31/white/pholm}{$1\times10^{-16}$}
\defV{final/gemma4_31/white/reversal}{yes}
\defV{final/gemma4_31/white/status}{yes}
\defV{final/geom/placement/far/lo}{122}
\defV{final/geom/placement/near/hi}{55}
\defV{final/geom/placement/near/lo}{46}
\defV{final/geom/placement/near/median}{53}
\defV{final/geom/placement/neighbour/near}{53}
\defV{final/geom/placement/own/near}{53}
\defV{final/geom/separation/conflict/competing/median}{56}
\defV{final/geom/separation/conflict/true/median}{110}
\defV{final/geom/separation/control/competing/median}{110}
\defV{final/geom/separation/control/true/median}{50}
\defV{final/geom/transfer/near/hi}{118}
\defV{final/geom/transfer/near/lo}{42}
\defV{final/geom/transfer_isobar/bands}{57--73, 77--93 and 102--118}
\defV{final/geom/transfer_linechart/bands}{42--58, 72--88 and 102--118}
\defV{final/geom/transfer_scatter/bands}{42--58, 72--88 and 102--118}
\defV{final/glm46v_9/answers}{6048}
\defV{final/glm46v_9/crop/b}{5}
\defV{final/glm46v_9/crop/c}{2}
\defV{final/glm46v_9/crop/ci}{[$-$1.46, 0.00]}
\defV{final/glm46v_9/crop/conflict/acc}{0.50}
\defV{final/glm46v_9/crop/conflict/c}{$-$1.81}
\defV{final/glm46v_9/crop/conflict/ci}{[$-$0.65, 0.65]}
\defV{final/glm46v_9/crop/conflict/d}{0.00}
\defV{final/glm46v_9/crop/conflict/n}{140}
\defV{final/glm46v_9/crop/conflict/unparsed}{0}
\defV{final/glm46v_9/crop/control/acc}{0.52}
\defV{final/glm46v_9/crop/control/c}{$-$2.05}
\defV{final/glm46v_9/crop/control/d}{0.80}
\defV{final/glm46v_9/crop/control/n}{140}
\defV{final/glm46v_9/crop/control/unparsed}{0}
\defV{final/glm46v_9/crop/diff}{$-$0.80}
\defV{final/glm46v_9/crop/p}{0.45}
\defV{final/glm46v_9/crop/pairs}{70}
\defV{final/glm46v_9/crop/pholm}{--}
\defV{final/glm46v_9/crop/status}{n.d.}
\defV{final/glm46v_9/cues_leader/b}{48}
\defV{final/glm46v_9/cues_leader/c}{2}
\defV{final/glm46v_9/cues_leader/ci}{[$-$3.86, $-$2.09]}
\defV{final/glm46v_9/cues_leader/conflict/acc}{0.62}
\defV{final/glm46v_9/cues_leader/conflict/c}{0.04}
\defV{final/glm46v_9/cues_leader/conflict/ci}{[0.11, 1.11]}
\defV{final/glm46v_9/cues_leader/conflict/d}{0.59}
\defV{final/glm46v_9/cues_leader/conflict/n}{136}
\defV{final/glm46v_9/cues_leader/conflict/unparsed}{0}
\defV{final/glm46v_9/cues_leader/control/acc}{0.96}
\defV{final/glm46v_9/cues_leader/control/c}{$-$0.31}
\defV{final/glm46v_9/cues_leader/control/d}{3.43}
\defV{final/glm46v_9/cues_leader/control/n}{136}
\defV{final/glm46v_9/cues_leader/control/unparsed}{0}
\defV{final/glm46v_9/cues_leader/diff}{$-$2.84}
\defV{final/glm46v_9/cues_leader/p}{$2\times10^{-12}$}
\defV{final/glm46v_9/cues_leader/pairs}{68}
\defV{final/glm46v_9/cues_leader/pholm}{--}
\defV{final/glm46v_9/cues_leader/status}{--}
\defV{final/glm46v_9/cues_region/b}{63}
\defV{final/glm46v_9/cues_region/c}{2}
\defV{final/glm46v_9/cues_region/ci}{[$-$4.95, $-$3.30]}
\defV{final/glm46v_9/cues_region/conflict/acc}{0.55}
\defV{final/glm46v_9/cues_region/conflict/c}{$-$0.05}
\defV{final/glm46v_9/cues_region/conflict/ci}{[$-$0.28, 0.76]}
\defV{final/glm46v_9/cues_region/conflict/d}{0.25}
\defV{final/glm46v_9/cues_region/conflict/n}{140}
\defV{final/glm46v_9/cues_region/conflict/unparsed}{0}
\defV{final/glm46v_9/cues_region/control/acc}{0.99}
\defV{final/glm46v_9/cues_region/control/c}{$-$0.32}
\defV{final/glm46v_9/cues_region/control/d}{4.26}
\defV{final/glm46v_9/cues_region/control/n}{140}
\defV{final/glm46v_9/cues_region/control/unparsed}{0}
\defV{final/glm46v_9/cues_region/diff}{$-$4.02}
\defV{final/glm46v_9/cues_region/p}{$1\times10^{-16}$}
\defV{final/glm46v_9/cues_region/pairs}{70}
\defV{final/glm46v_9/cues_region/pholm}{--}
\defV{final/glm46v_9/cues_region/status}{--}
\defV{final/glm46v_9/direct/b}{9}
\defV{final/glm46v_9/direct/c}{10}
\defV{final/glm46v_9/direct/ci}{[$-$0.34, 0.87]}
\defV{final/glm46v_9/direct/conflict/acc}{0.52}
\defV{final/glm46v_9/direct/conflict/c}{$-$1.49}
\defV{final/glm46v_9/direct/conflict/ci}{[$-$0.12, 0.93]}
\defV{final/glm46v_9/direct/conflict/d}{0.32}
\defV{final/glm46v_9/direct/conflict/n}{140}
\defV{final/glm46v_9/direct/conflict/unparsed}{0}
\defV{final/glm46v_9/direct/control/acc}{0.51}
\defV{final/glm46v_9/direct/control/c}{$-$1.11}
\defV{final/glm46v_9/direct/control/d}{0.13}
\defV{final/glm46v_9/direct/control/n}{140}
\defV{final/glm46v_9/direct/control/unparsed}{0}
\defV{final/glm46v_9/direct/diff}{+0.19}
\defV{final/glm46v_9/direct/p}{1.00}
\defV{final/glm46v_9/direct/pairs}{70}
\defV{final/glm46v_9/direct/pholm}{--}
\defV{final/glm46v_9/direct/status}{n.d.}
\defV{final/glm46v_9/failed}{0}
\defV{final/glm46v_9/gate}{1.31}
\defV{final/glm46v_9/gsi_placement/b}{27}
\defV{final/glm46v_9/gsi_placement/c}{1}
\defV{final/glm46v_9/gsi_placement/ci}{[$-$2.39, $-$1.12]}
\defV{final/glm46v_9/gsi_placement/conflict/acc}{0.43}
\defV{final/glm46v_9/gsi_placement/conflict/c}{$-$1.07}
\defV{final/glm46v_9/gsi_placement/conflict/ci}{[$-$1.17, $-$0.16]}
\defV{final/glm46v_9/gsi_placement/conflict/d}{$-$0.61}
\defV{final/glm46v_9/gsi_placement/conflict/n}{128}
\defV{final/glm46v_9/gsi_placement/conflict/unparsed}{0}
\defV{final/glm46v_9/gsi_placement/control/acc}{0.63}
\defV{final/glm46v_9/gsi_placement/control/c}{$-$0.96}
\defV{final/glm46v_9/gsi_placement/control/d}{1.04}
\defV{final/glm46v_9/gsi_placement/control/n}{128}
\defV{final/glm46v_9/gsi_placement/control/unparsed}{0}
\defV{final/glm46v_9/gsi_placement/diff}{$-$1.66}
\defV{final/glm46v_9/gsi_placement/p}{$2\times10^{-7}$}
\defV{final/glm46v_9/gsi_placement/pairs}{64}
\defV{final/glm46v_9/gsi_placement/pholm}{--}
\defV{final/glm46v_9/gsi_placement/status}{--}
\defV{final/glm46v_9/gsi_placement_k1/b}{9}
\defV{final/glm46v_9/gsi_placement_k1/c}{1}
\defV{final/glm46v_9/gsi_placement_k1/ci}{[$-$2.89, $-$0.64]}
\defV{final/glm46v_9/gsi_placement_k1/conflict/acc}{0.48}
\defV{final/glm46v_9/gsi_placement_k1/conflict/c}{$-$1.14}
\defV{final/glm46v_9/gsi_placement_k1/conflict/ci}{[$-$1.24, 0.78]}
\defV{final/glm46v_9/gsi_placement_k1/conflict/d}{$-$0.18}
\defV{final/glm46v_9/gsi_placement_k1/conflict/n}{46}
\defV{final/glm46v_9/gsi_placement_k1/conflict/unparsed}{1}
\defV{final/glm46v_9/gsi_placement_k1/control/acc}{0.65}
\defV{final/glm46v_9/gsi_placement_k1/control/c}{$-$1.26}
\defV{final/glm46v_9/gsi_placement_k1/control/d}{1.55}
\defV{final/glm46v_9/gsi_placement_k1/control/n}{46}
\defV{final/glm46v_9/gsi_placement_k1/control/unparsed}{0}
\defV{final/glm46v_9/gsi_placement_k1/diff}{$-$1.73}
\defV{final/glm46v_9/gsi_placement_k1/p}{0.02}
\defV{final/glm46v_9/gsi_placement_k1/pairs}{23}
\defV{final/glm46v_9/gsi_placement_k1/pholm}{--}
\defV{final/glm46v_9/gsi_placement_k1/status}{--}
\defV{final/glm46v_9/isobar/b}{49}
\defV{final/glm46v_9/isobar/c}{28}
\defV{final/glm46v_9/isobar/ci}{[$-$1.04, $-$0.06]}
\defV{final/glm46v_9/isobar/conflict/acc}{0.61}
\defV{final/glm46v_9/isobar/conflict/c}{$-$0.23}
\defV{final/glm46v_9/isobar/conflict/ci}{[0.29, 0.95]}
\defV{final/glm46v_9/isobar/conflict/d}{0.61}
\defV{final/glm46v_9/isobar/conflict/n}{200}
\defV{final/glm46v_9/isobar/conflict/unparsed}{1}
\defV{final/glm46v_9/isobar/control/acc}{0.72}
\defV{final/glm46v_9/isobar/control/c}{$-$0.06}
\defV{final/glm46v_9/isobar/control/d}{1.15}
\defV{final/glm46v_9/isobar/control/n}{200}
\defV{final/glm46v_9/isobar/control/unparsed}{0}
\defV{final/glm46v_9/isobar/diff}{$-$0.54}
\defV{final/glm46v_9/isobar/p}{0.02}
\defV{final/glm46v_9/isobar/pairs}{100}
\defV{final/glm46v_9/isobar/pholm}{--}
\defV{final/glm46v_9/isobar/status}{--}
\defV{final/glm46v_9/isobar_text/b}{8}
\defV{final/glm46v_9/isobar_text/c}{13}
\defV{final/glm46v_9/isobar_text/ci}{[$-$0.20, 0.71]}
\defV{final/glm46v_9/isobar_text/conflict/acc}{0.50}
\defV{final/glm46v_9/isobar_text/conflict/c}{1.11}
\defV{final/glm46v_9/isobar_text/conflict/ci}{[$-$0.48, 0.46]}
\defV{final/glm46v_9/isobar_text/conflict/d}{0.00}
\defV{final/glm46v_9/isobar_text/conflict/n}{200}
\defV{final/glm46v_9/isobar_text/conflict/unparsed}{0}
\defV{final/glm46v_9/isobar_text/control/acc}{0.47}
\defV{final/glm46v_9/isobar_text/control/c}{1.24}
\defV{final/glm46v_9/isobar_text/control/d}{$-$0.27}
\defV{final/glm46v_9/isobar_text/control/n}{200}
\defV{final/glm46v_9/isobar_text/control/unparsed}{0}
\defV{final/glm46v_9/isobar_text/diff}{+0.27}
\defV{final/glm46v_9/isobar_text/p}{0.38}
\defV{final/glm46v_9/isobar_text/pairs}{100}
\defV{final/glm46v_9/isobar_text/pholm}{--}
\defV{final/glm46v_9/isobar_text/status}{n.d.}
\defV{final/glm46v_9/linechart/b}{47}
\defV{final/glm46v_9/linechart/c}{7}
\defV{final/glm46v_9/linechart/ci}{[$-$2.52, $-$1.05]}
\defV{final/glm46v_9/linechart/conflict/acc}{0.56}
\defV{final/glm46v_9/linechart/conflict/c}{$-$0.05}
\defV{final/glm46v_9/linechart/conflict/ci}{[$-$0.21, 0.84]}
\defV{final/glm46v_9/linechart/conflict/d}{0.32}
\defV{final/glm46v_9/linechart/conflict/n}{140}
\defV{final/glm46v_9/linechart/conflict/unparsed}{0}
\defV{final/glm46v_9/linechart/control/acc}{0.85}
\defV{final/glm46v_9/linechart/control/c}{$-$0.09}
\defV{final/glm46v_9/linechart/control/d}{2.04}
\defV{final/glm46v_9/linechart/control/n}{140}
\defV{final/glm46v_9/linechart/control/unparsed}{0}
\defV{final/glm46v_9/linechart/diff}{$-$1.72}
\defV{final/glm46v_9/linechart/p}{$2\times10^{-8}$}
\defV{final/glm46v_9/linechart/pairs}{70}
\defV{final/glm46v_9/linechart/pholm}{--}
\defV{final/glm46v_9/linechart/status}{--}
\defV{final/glm46v_9/linechart_colour/b}{6}
\defV{final/glm46v_9/linechart_colour/c}{3}
\defV{final/glm46v_9/linechart_colour/ci}{[$-$0.81, 0.32]}
\defV{final/glm46v_9/linechart_colour/conflict/acc}{0.91}
\defV{final/glm46v_9/linechart_colour/conflict/c}{0.21}
\defV{final/glm46v_9/linechart_colour/conflict/ci}{[2.06, 3.46]}
\defV{final/glm46v_9/linechart_colour/conflict/d}{2.64}
\defV{final/glm46v_9/linechart_colour/conflict/n}{140}
\defV{final/glm46v_9/linechart_colour/conflict/unparsed}{0}
\defV{final/glm46v_9/linechart_colour/control/acc}{0.93}
\defV{final/glm46v_9/linechart_colour/control/c}{0.10}
\defV{final/glm46v_9/linechart_colour/control/d}{2.86}
\defV{final/glm46v_9/linechart_colour/control/n}{140}
\defV{final/glm46v_9/linechart_colour/control/unparsed}{0}
\defV{final/glm46v_9/linechart_colour/diff}{$-$0.22}
\defV{final/glm46v_9/linechart_colour/p}{0.51}
\defV{final/glm46v_9/linechart_colour/pairs}{70}
\defV{final/glm46v_9/linechart_colour/pholm}{--}
\defV{final/glm46v_9/linechart_colour/status}{n.d.}
\defV{final/glm46v_9/linechart_text/b}{0}
\defV{final/glm46v_9/linechart_text/c}{0}
\defV{final/glm46v_9/linechart_text/ci}{[0.00, 0.00]}
\defV{final/glm46v_9/linechart_text/conflict/acc}{1.00}
\defV{final/glm46v_9/linechart_text/conflict/c}{$-$0.00}
\defV{final/glm46v_9/linechart_text/conflict/ci}{[4.91, 4.91]}
\defV{final/glm46v_9/linechart_text/conflict/d}{4.91}
\defV{final/glm46v_9/linechart_text/conflict/n}{140}
\defV{final/glm46v_9/linechart_text/conflict/unparsed}{0}
\defV{final/glm46v_9/linechart_text/control/acc}{1.00}
\defV{final/glm46v_9/linechart_text/control/c}{$-$0.00}
\defV{final/glm46v_9/linechart_text/control/d}{4.91}
\defV{final/glm46v_9/linechart_text/control/n}{140}
\defV{final/glm46v_9/linechart_text/control/unparsed}{0}
\defV{final/glm46v_9/linechart_text/diff}{+0.00}
\defV{final/glm46v_9/linechart_text/p}{1.00}
\defV{final/glm46v_9/linechart_text/pairs}{70}
\defV{final/glm46v_9/linechart_text/pholm}{--}
\defV{final/glm46v_9/linechart_text/status}{n.d.}
\defV{final/glm46v_9/map/index_labels/acc}{0.62}
\defV{final/glm46v_9/map/index_labels/d}{0.71}
\defV{final/glm46v_9/map/index_labels/n}{120}
\defV{final/glm46v_9/map/structure_only/acc}{0.72}
\defV{final/glm46v_9/map/structure_only/d}{1.28}
\defV{final/glm46v_9/map/structure_only/n}{120}
\defV{final/glm46v_9/map/values_only/acc}{0.61}
\defV{final/glm46v_9/map/values_only/d}{0.82}
\defV{final/glm46v_9/map/values_only/n}{120}
\defV{final/glm46v_9/map/values_structure/acc}{0.88}
\defV{final/glm46v_9/map/values_structure/d}{2.73}
\defV{final/glm46v_9/map/values_structure/n}{120}
\defV{final/glm46v_9/matched/b}{46}
\defV{final/glm46v_9/matched/c}{15}
\defV{final/glm46v_9/matched/ci}{[$-$2.22, $-$0.73]}
\defV{final/glm46v_9/matched/conflict/acc}{0.36}
\defV{final/glm46v_9/matched/conflict/c}{$-$0.79}
\defV{final/glm46v_9/matched/conflict/ci}{[$-$1.49, $-$0.42]}
\defV{final/glm46v_9/matched/conflict/d}{$-$0.93}
\defV{final/glm46v_9/matched/conflict/n}{140}
\defV{final/glm46v_9/matched/conflict/unparsed}{0}
\defV{final/glm46v_9/matched/control/acc}{0.59}
\defV{final/glm46v_9/matched/control/c}{$-$0.52}
\defV{final/glm46v_9/matched/control/d}{0.51}
\defV{final/glm46v_9/matched/control/n}{140}
\defV{final/glm46v_9/matched/control/unparsed}{1}
\defV{final/glm46v_9/matched/diff}{$-$1.43}
\defV{final/glm46v_9/matched/p}{$9\times10^{-5}$}
\defV{final/glm46v_9/matched/pairs}{70}
\defV{final/glm46v_9/matched/pholm}{--}
\defV{final/glm46v_9/matched/status}{--}
\defV{final/glm46v_9/name}{GLM-4.6V-Flash}
\defV{final/glm46v_9/pointer_linechart/b}{5}
\defV{final/glm46v_9/pointer_linechart/c}{3}
\defV{final/glm46v_9/pointer_linechart/ci}{[$-$1.23, 0.45]}
\defV{final/glm46v_9/pointer_linechart/conflict/acc}{0.88}
\defV{final/glm46v_9/pointer_linechart/conflict/c}{$-$0.11}
\defV{final/glm46v_9/pointer_linechart/conflict/ci}{[1.69, 3.02]}
\defV{final/glm46v_9/pointer_linechart/conflict/d}{2.23}
\defV{final/glm46v_9/pointer_linechart/conflict/n}{80}
\defV{final/glm46v_9/pointer_linechart/conflict/unparsed}{0}
\defV{final/glm46v_9/pointer_linechart/control/acc}{0.90}
\defV{final/glm46v_9/pointer_linechart/control/c}{$-$0.27}
\defV{final/glm46v_9/pointer_linechart/control/d}{2.55}
\defV{final/glm46v_9/pointer_linechart/control/n}{80}
\defV{final/glm46v_9/pointer_linechart/control/unparsed}{0}
\defV{final/glm46v_9/pointer_linechart/diff}{$-$0.32}
\defV{final/glm46v_9/pointer_linechart/p}{0.73}
\defV{final/glm46v_9/pointer_linechart/pairs}{40}
\defV{final/glm46v_9/pointer_linechart/pholm}{--}
\defV{final/glm46v_9/pointer_linechart/status}{n.d.}
\defV{final/glm46v_9/pointer_scatter/b}{2}
\defV{final/glm46v_9/pointer_scatter/c}{0}
\defV{final/glm46v_9/pointer_scatter/ci}{[$-$1.14, 0.00]}
\defV{final/glm46v_9/pointer_scatter/conflict/acc}{0.97}
\defV{final/glm46v_9/pointer_scatter/conflict/c}{0.35}
\defV{final/glm46v_9/pointer_scatter/conflict/ci}{[3.36, 4.50]}
\defV{final/glm46v_9/pointer_scatter/conflict/d}{3.80}
\defV{final/glm46v_9/pointer_scatter/conflict/n}{80}
\defV{final/glm46v_9/pointer_scatter/conflict/unparsed}{0}
\defV{final/glm46v_9/pointer_scatter/control/acc}{1.00}
\defV{final/glm46v_9/pointer_scatter/control/c}{$-$0.00}
\defV{final/glm46v_9/pointer_scatter/control/d}{4.50}
\defV{final/glm46v_9/pointer_scatter/control/n}{80}
\defV{final/glm46v_9/pointer_scatter/control/unparsed}{0}
\defV{final/glm46v_9/pointer_scatter/diff}{$-$0.70}
\defV{final/glm46v_9/pointer_scatter/p}{0.50}
\defV{final/glm46v_9/pointer_scatter/pairs}{40}
\defV{final/glm46v_9/pointer_scatter/pholm}{--}
\defV{final/glm46v_9/pointer_scatter/status}{n.d.}
\defV{final/glm46v_9/public_data/b}{28}
\defV{final/glm46v_9/public_data/c}{7}
\defV{final/glm46v_9/public_data/ci}{[$-$2.43, $-$0.79]}
\defV{final/glm46v_9/public_data/conflict/acc}{0.54}
\defV{final/glm46v_9/public_data/conflict/c}{$-$0.09}
\defV{final/glm46v_9/public_data/conflict/ci}{[$-$0.37, 0.76]}
\defV{final/glm46v_9/public_data/conflict/d}{0.18}
\defV{final/glm46v_9/public_data/conflict/n}{80}
\defV{final/glm46v_9/public_data/conflict/unparsed}{0}
\defV{final/glm46v_9/public_data/control/acc}{0.80}
\defV{final/glm46v_9/public_data/control/c}{$-$0.36}
\defV{final/glm46v_9/public_data/control/d}{1.74}
\defV{final/glm46v_9/public_data/control/n}{80}
\defV{final/glm46v_9/public_data/control/unparsed}{0}
\defV{final/glm46v_9/public_data/diff}{$-$1.55}
\defV{final/glm46v_9/public_data/p}{$5\times10^{-4}$}
\defV{final/glm46v_9/public_data/pairs}{40}
\defV{final/glm46v_9/public_data/pholm}{--}
\defV{final/glm46v_9/public_data/status}{--}
\defV{final/glm46v_9/repo}{zai-org/GLM-4.6V-Flash}
\defV{final/glm46v_9/revision}{411bb4d}
\defV{final/glm46v_9/scatter/b}{69}
\defV{final/glm46v_9/scatter/c}{1}
\defV{final/glm46v_9/scatter/ci}{[$-$4.60, $-$3.20]}
\defV{final/glm46v_9/scatter/conflict/acc}{0.65}
\defV{final/glm46v_9/scatter/conflict/c}{$-$0.00}
\defV{final/glm46v_9/scatter/conflict/ci}{[0.30, 1.25]}
\defV{final/glm46v_9/scatter/conflict/d}{0.76}
\defV{final/glm46v_9/scatter/conflict/n}{200}
\defV{final/glm46v_9/scatter/conflict/unparsed}{0}
\defV{final/glm46v_9/scatter/control/acc}{0.99}
\defV{final/glm46v_9/scatter/control/c}{$-$0.31}
\defV{final/glm46v_9/scatter/control/d}{4.54}
\defV{final/glm46v_9/scatter/control/n}{200}
\defV{final/glm46v_9/scatter/control/unparsed}{0}
\defV{final/glm46v_9/scatter/diff}{$-$3.78}
\defV{final/glm46v_9/scatter/p}{$1\times10^{-19}$}
\defV{final/glm46v_9/scatter/pairs}{100}
\defV{final/glm46v_9/scatter/pholm}{--}
\defV{final/glm46v_9/scatter/status}{--}
\defV{final/glm46v_9/scatter_text/b}{4}
\defV{final/glm46v_9/scatter_text/c}{3}
\defV{final/glm46v_9/scatter_text/ci}{[$-$0.35, 0.34]}
\defV{final/glm46v_9/scatter_text/conflict/acc}{0.91}
\defV{final/glm46v_9/scatter_text/conflict/c}{0.00}
\defV{final/glm46v_9/scatter_text/conflict/ci}{[2.21, 3.20]}
\defV{final/glm46v_9/scatter_text/conflict/d}{2.63}
\defV{final/glm46v_9/scatter_text/conflict/n}{200}
\defV{final/glm46v_9/scatter_text/conflict/unparsed}{1}
\defV{final/glm46v_9/scatter_text/control/acc}{0.91}
\defV{final/glm46v_9/scatter_text/control/c}{0.06}
\defV{final/glm46v_9/scatter_text/control/d}{2.64}
\defV{final/glm46v_9/scatter_text/control/n}{200}
\defV{final/glm46v_9/scatter_text/control/unparsed}{0}
\defV{final/glm46v_9/scatter_text/diff}{$-$0.01}
\defV{final/glm46v_9/scatter_text/p}{1.00}
\defV{final/glm46v_9/scatter_text/pairs}{100}
\defV{final/glm46v_9/scatter_text/pholm}{--}
\defV{final/glm46v_9/scatter_text/status}{n.d.}
\defV{final/glm46v_9/separation/b}{30}
\defV{final/glm46v_9/separation/c}{15}
\defV{final/glm46v_9/separation/ci}{[$-$1.39, $-$0.15]}
\defV{final/glm46v_9/separation/conflict/acc}{0.45}
\defV{final/glm46v_9/separation/conflict/c}{$-$0.83}
\defV{final/glm46v_9/separation/conflict/ci}{[$-$0.82, 0.15]}
\defV{final/glm46v_9/separation/conflict/d}{$-$0.31}
\defV{final/glm46v_9/separation/conflict/n}{140}
\defV{final/glm46v_9/separation/conflict/unparsed}{1}
\defV{final/glm46v_9/separation/control/acc}{0.56}
\defV{final/glm46v_9/separation/control/c}{$-$0.90}
\defV{final/glm46v_9/separation/control/d}{0.42}
\defV{final/glm46v_9/separation/control/n}{140}
\defV{final/glm46v_9/separation/control/unparsed}{0}
\defV{final/glm46v_9/separation/diff}{$-$0.73}
\defV{final/glm46v_9/separation/p}{0.04}
\defV{final/glm46v_9/separation/pairs}{70}
\defV{final/glm46v_9/separation/pholm}{--}
\defV{final/glm46v_9/separation/status}{n.d.}
\defV{final/glm46v_9/separation_text/b}{5}
\defV{final/glm46v_9/separation_text/c}{8}
\defV{final/glm46v_9/separation_text/ci}{[$-$1.03, 0.61]}
\defV{final/glm46v_9/separation_text/conflict/acc}{0.96}
\defV{final/glm46v_9/separation_text/conflict/c}{$-$0.14}
\defV{final/glm46v_9/separation_text/conflict/ci}{[2.78, 4.26]}
\defV{final/glm46v_9/separation_text/conflict/d}{3.34}
\defV{final/glm46v_9/separation_text/conflict/n}{140}
\defV{final/glm46v_9/separation_text/conflict/unparsed}{0}
\defV{final/glm46v_9/separation_text/control/acc}{0.94}
\defV{final/glm46v_9/separation_text/control/c}{$-$0.60}
\defV{final/glm46v_9/separation_text/control/d}{3.69}
\defV{final/glm46v_9/separation_text/control/n}{140}
\defV{final/glm46v_9/separation_text/control/unparsed}{2}
\defV{final/glm46v_9/separation_text/diff}{$-$0.36}
\defV{final/glm46v_9/separation_text/p}{0.58}
\defV{final/glm46v_9/separation_text/pairs}{70}
\defV{final/glm46v_9/separation_text/pholm}{--}
\defV{final/glm46v_9/separation_text/status}{--}
\defV{final/glm46v_9/truncated}{11}
\defV{final/glm46v_9/truncated/gsi_placement}{1}
\defV{final/glm46v_9/truncated/gsi_placement_development}{1}
\defV{final/glm46v_9/truncated/label_placement_boxes}{2}
\defV{final/glm46v_9/truncated/map_pairs}{2}
\defV{final/glm46v_9/truncated/separation_shortcut_free}{1}
\defV{final/glm46v_9/truncated/separation_shortcut_free_text}{2}
\defV{final/glm46v_9/truncated/transfer_isobar}{1}
\defV{final/glm46v_9/truncated/transfer_scatter_text}{1}
\defV{final/glm46v_9/unparsed}{11}
\defV{final/glm46v_9/unparsed/pct}{0.2}
\defV{final/glm46v_9/white/b}{46}
\defV{final/glm46v_9/white/c}{14}
\defV{final/glm46v_9/white/ci}{[$-$2.29, $-$0.92]}
\defV{final/glm46v_9/white/conditional}{0.42}
\defV{final/glm46v_9/white/conflict/acc}{0.34}
\defV{final/glm46v_9/white/conflict/c}{$-$0.77}
\defV{final/glm46v_9/white/conflict/ci}{[$-$1.74, $-$0.66]}
\defV{final/glm46v_9/white/conflict/d}{$-$1.12}
\defV{final/glm46v_9/white/conflict/n}{140}
\defV{final/glm46v_9/white/conflict/unparsed}{0}
\defV{final/glm46v_9/white/control/acc}{0.56}
\defV{final/glm46v_9/white/control/c}{$-$0.61}
\defV{final/glm46v_9/white/control/d}{0.42}
\defV{final/glm46v_9/white/control/n}{140}
\defV{final/glm46v_9/white/control/unparsed}{1}
\defV{final/glm46v_9/white/diff}{$-$1.54}
\defV{final/glm46v_9/white/p}{$4\times10^{-5}$}
\defV{final/glm46v_9/white/pairs}{70}
\defV{final/glm46v_9/white/pholm}{$3\times10^{-4}$}
\defV{final/glm46v_9/white/reversal}{no}
\defV{final/glm46v_9/white/status}{n.d.}
\defV{final/gpu}{NVIDIA A100-SXM4-80GB}
\defV{final/headline/control/max}{0.73}
\defV{final/headline/control/min}{0.44}
\defV{final/headline/maxabslambda}{1.56}
\defV{final/internvl35_38/answers}{6048}
\defV{final/internvl35_38/crop/b}{43}
\defV{final/internvl35_38/crop/c}{3}
\defV{final/internvl35_38/crop/ci}{[$-$3.30, $-$1.63]}
\defV{final/internvl35_38/crop/conflict/acc}{0.39}
\defV{final/internvl35_38/crop/conflict/c}{$-$0.92}
\defV{final/internvl35_38/crop/conflict/ci}{[$-$1.41, $-$0.30]}
\defV{final/internvl35_38/crop/conflict/d}{$-$0.81}
\defV{final/internvl35_38/crop/conflict/n}{140}
\defV{final/internvl35_38/crop/conflict/unparsed}{0}
\defV{final/internvl35_38/crop/control/acc}{0.68}
\defV{final/internvl35_38/crop/control/c}{$-$1.05}
\defV{final/internvl35_38/crop/control/d}{1.52}
\defV{final/internvl35_38/crop/control/n}{140}
\defV{final/internvl35_38/crop/control/unparsed}{0}
\defV{final/internvl35_38/crop/diff}{$-$2.34}
\defV{final/internvl35_38/crop/p}{$5\times10^{-10}$}
\defV{final/internvl35_38/crop/pairs}{70}
\defV{final/internvl35_38/crop/pholm}{$9\times10^{-10}$}
\defV{final/internvl35_38/crop/status}{yes}
\defV{final/internvl35_38/cues_leader/b}{71}
\defV{final/internvl35_38/cues_leader/c}{3}
\defV{final/internvl35_38/cues_leader/ci}{[$-$3.86, $-$2.28]}
\defV{final/internvl35_38/cues_leader/conflict/acc}{0.39}
\defV{final/internvl35_38/cues_leader/conflict/c}{0.17}
\defV{final/internvl35_38/cues_leader/conflict/ci}{[$-$1.11, $-$0.07]}
\defV{final/internvl35_38/cues_leader/conflict/d}{$-$0.56}
\defV{final/internvl35_38/cues_leader/conflict/n}{136}
\defV{final/internvl35_38/cues_leader/conflict/unparsed}{0}
\defV{final/internvl35_38/cues_leader/control/acc}{0.89}
\defV{final/internvl35_38/cues_leader/control/c}{$-$0.04}
\defV{final/internvl35_38/cues_leader/control/d}{2.39}
\defV{final/internvl35_38/cues_leader/control/n}{136}
\defV{final/internvl35_38/cues_leader/control/unparsed}{0}
\defV{final/internvl35_38/cues_leader/diff}{$-$2.95}
\defV{final/internvl35_38/cues_leader/p}{$7\times10^{-18}$}
\defV{final/internvl35_38/cues_leader/pairs}{68}
\defV{final/internvl35_38/cues_leader/pholm}{--}
\defV{final/internvl35_38/cues_leader/status}{--}
\defV{final/internvl35_38/cues_region/b}{79}
\defV{final/internvl35_38/cues_region/c}{1}
\defV{final/internvl35_38/cues_region/ci}{[$-$4.13, $-$2.63]}
\defV{final/internvl35_38/cues_region/conflict/acc}{0.34}
\defV{final/internvl35_38/cues_region/conflict/c}{$-$0.10}
\defV{final/internvl35_38/cues_region/conflict/ci}{[$-$1.38, $-$0.32]}
\defV{final/internvl35_38/cues_region/conflict/d}{$-$0.84}
\defV{final/internvl35_38/cues_region/conflict/n}{140}
\defV{final/internvl35_38/cues_region/conflict/unparsed}{0}
\defV{final/internvl35_38/cues_region/control/acc}{0.89}
\defV{final/internvl35_38/cues_region/control/c}{$-$0.11}
\defV{final/internvl35_38/cues_region/control/d}{2.44}
\defV{final/internvl35_38/cues_region/control/n}{140}
\defV{final/internvl35_38/cues_region/control/unparsed}{0}
\defV{final/internvl35_38/cues_region/diff}{$-$3.28}
\defV{final/internvl35_38/cues_region/p}{$1\times10^{-22}$}
\defV{final/internvl35_38/cues_region/pairs}{70}
\defV{final/internvl35_38/cues_region/pholm}{--}
\defV{final/internvl35_38/cues_region/status}{--}
\defV{final/internvl35_38/direct/b}{25}
\defV{final/internvl35_38/direct/c}{13}
\defV{final/internvl35_38/direct/ci}{[$-$1.49, 0.01]}
\defV{final/internvl35_38/direct/conflict/acc}{0.46}
\defV{final/internvl35_38/direct/conflict/c}{$-$1.09}
\defV{final/internvl35_38/direct/conflict/ci}{[$-$0.87, 0.20]}
\defV{final/internvl35_38/direct/conflict/d}{$-$0.32}
\defV{final/internvl35_38/direct/conflict/n}{140}
\defV{final/internvl35_38/direct/conflict/unparsed}{0}
\defV{final/internvl35_38/direct/control/acc}{0.55}
\defV{final/internvl35_38/direct/control/c}{$-$0.98}
\defV{final/internvl35_38/direct/control/d}{0.40}
\defV{final/internvl35_38/direct/control/n}{140}
\defV{final/internvl35_38/direct/control/unparsed}{0}
\defV{final/internvl35_38/direct/diff}{$-$0.72}
\defV{final/internvl35_38/direct/p}{0.07}
\defV{final/internvl35_38/direct/pairs}{70}
\defV{final/internvl35_38/direct/pholm}{0.29}
\defV{final/internvl35_38/direct/status}{n.d.}
\defV{final/internvl35_38/failed}{0}
\defV{final/internvl35_38/gate}{0.62}
\defV{final/internvl35_38/gsi_placement/b}{25}
\defV{final/internvl35_38/gsi_placement/c}{6}
\defV{final/internvl35_38/gsi_placement/ci}{[$-$2.34, $-$0.82]}
\defV{final/internvl35_38/gsi_placement/conflict/acc}{0.46}
\defV{final/internvl35_38/gsi_placement/conflict/c}{$-$1.10}
\defV{final/internvl35_38/gsi_placement/conflict/ci}{[$-$0.95, 0.16]}
\defV{final/internvl35_38/gsi_placement/conflict/d}{$-$0.35}
\defV{final/internvl35_38/gsi_placement/conflict/n}{128}
\defV{final/internvl35_38/gsi_placement/conflict/unparsed}{0}
\defV{final/internvl35_38/gsi_placement/control/acc}{0.61}
\defV{final/internvl35_38/gsi_placement/control/c}{$-$1.22}
\defV{final/internvl35_38/gsi_placement/control/d}{1.11}
\defV{final/internvl35_38/gsi_placement/control/n}{128}
\defV{final/internvl35_38/gsi_placement/control/unparsed}{0}
\defV{final/internvl35_38/gsi_placement/diff}{$-$1.46}
\defV{final/internvl35_38/gsi_placement/p}{$9\times10^{-4}$}
\defV{final/internvl35_38/gsi_placement/pairs}{64}
\defV{final/internvl35_38/gsi_placement/pholm}{0.005}
\defV{final/internvl35_38/gsi_placement/status}{yes}
\defV{final/internvl35_38/gsi_placement_k1/b}{12}
\defV{final/internvl35_38/gsi_placement_k1/c}{8}
\defV{final/internvl35_38/gsi_placement_k1/ci}{[$-$1.83, 0.75]}
\defV{final/internvl35_38/gsi_placement_k1/conflict/acc}{0.43}
\defV{final/internvl35_38/gsi_placement_k1/conflict/c}{$-$0.56}
\defV{final/internvl35_38/gsi_placement_k1/conflict/ci}{[$-$1.27, 0.44]}
\defV{final/internvl35_38/gsi_placement_k1/conflict/d}{$-$0.37}
\defV{final/internvl35_38/gsi_placement_k1/conflict/n}{46}
\defV{final/internvl35_38/gsi_placement_k1/conflict/unparsed}{0}
\defV{final/internvl35_38/gsi_placement_k1/control/acc}{0.52}
\defV{final/internvl35_38/gsi_placement_k1/control/c}{$-$0.68}
\defV{final/internvl35_38/gsi_placement_k1/control/d}{0.13}
\defV{final/internvl35_38/gsi_placement_k1/control/n}{46}
\defV{final/internvl35_38/gsi_placement_k1/control/unparsed}{0}
\defV{final/internvl35_38/gsi_placement_k1/diff}{$-$0.50}
\defV{final/internvl35_38/gsi_placement_k1/p}{0.50}
\defV{final/internvl35_38/gsi_placement_k1/pairs}{23}
\defV{final/internvl35_38/gsi_placement_k1/pholm}{--}
\defV{final/internvl35_38/gsi_placement_k1/status}{n.d.}
\defV{final/internvl35_38/isobar/b}{65}
\defV{final/internvl35_38/isobar/c}{28}
\defV{final/internvl35_38/isobar/ci}{[$-$1.61, $-$0.53]}
\defV{final/internvl35_38/isobar/conflict/acc}{0.52}
\defV{final/internvl35_38/isobar/conflict/c}{0.02}
\defV{final/internvl35_38/isobar/conflict/ci}{[$-$0.23, 0.43]}
\defV{final/internvl35_38/isobar/conflict/d}{0.10}
\defV{final/internvl35_38/isobar/conflict/n}{200}
\defV{final/internvl35_38/isobar/conflict/unparsed}{0}
\defV{final/internvl35_38/isobar/control/acc}{0.70}
\defV{final/internvl35_38/isobar/control/c}{0.37}
\defV{final/internvl35_38/isobar/control/d}{1.14}
\defV{final/internvl35_38/isobar/control/n}{200}
\defV{final/internvl35_38/isobar/control/unparsed}{0}
\defV{final/internvl35_38/isobar/diff}{$-$1.04}
\defV{final/internvl35_38/isobar/p}{$2\times10^{-4}$}
\defV{final/internvl35_38/isobar/pairs}{100}
\defV{final/internvl35_38/isobar/pholm}{0.001}
\defV{final/internvl35_38/isobar/status}{yes}
\defV{final/internvl35_38/isobar_text/b}{15}
\defV{final/internvl35_38/isobar_text/c}{14}
\defV{final/internvl35_38/isobar_text/ci}{[$-$0.54, 0.48]}
\defV{final/internvl35_38/isobar_text/conflict/acc}{0.86}
\defV{final/internvl35_38/isobar_text/conflict/c}{$-$0.07}
\defV{final/internvl35_38/isobar_text/conflict/ci}{[1.81, 2.67]}
\defV{final/internvl35_38/isobar_text/conflict/d}{2.18}
\defV{final/internvl35_38/isobar_text/conflict/n}{200}
\defV{final/internvl35_38/isobar_text/conflict/unparsed}{0}
\defV{final/internvl35_38/isobar_text/control/acc}{0.87}
\defV{final/internvl35_38/isobar_text/control/c}{0.05}
\defV{final/internvl35_38/isobar_text/control/d}{2.22}
\defV{final/internvl35_38/isobar_text/control/n}{200}
\defV{final/internvl35_38/isobar_text/control/unparsed}{0}
\defV{final/internvl35_38/isobar_text/diff}{$-$0.04}
\defV{final/internvl35_38/isobar_text/p}{1.00}
\defV{final/internvl35_38/isobar_text/pairs}{100}
\defV{final/internvl35_38/isobar_text/pholm}{--}
\defV{final/internvl35_38/isobar_text/status}{--}
\defV{final/internvl35_38/linechart/b}{34}
\defV{final/internvl35_38/linechart/c}{10}
\defV{final/internvl35_38/linechart/ci}{[$-$1.65, $-$0.37]}
\defV{final/internvl35_38/linechart/conflict/acc}{0.61}
\defV{final/internvl35_38/linechart/conflict/c}{0.05}
\defV{final/internvl35_38/linechart/conflict/ci}{[0.00, 1.05]}
\defV{final/internvl35_38/linechart/conflict/d}{0.54}
\defV{final/internvl35_38/linechart/conflict/n}{140}
\defV{final/internvl35_38/linechart/conflict/unparsed}{0}
\defV{final/internvl35_38/linechart/control/acc}{0.78}
\defV{final/internvl35_38/linechart/control/c}{$-$0.12}
\defV{final/internvl35_38/linechart/control/d}{1.52}
\defV{final/internvl35_38/linechart/control/n}{140}
\defV{final/internvl35_38/linechart/control/unparsed}{0}
\defV{final/internvl35_38/linechart/diff}{$-$0.98}
\defV{final/internvl35_38/linechart/p}{$4\times10^{-4}$}
\defV{final/internvl35_38/linechart/pairs}{70}
\defV{final/internvl35_38/linechart/pholm}{--}
\defV{final/internvl35_38/linechart/status}{--}
\defV{final/internvl35_38/linechart_colour/b}{1}
\defV{final/internvl35_38/linechart_colour/c}{1}
\defV{final/internvl35_38/linechart_colour/ci}{[$-$0.80, 0.02]}
\defV{final/internvl35_38/linechart_colour/conflict/acc}{0.99}
\defV{final/internvl35_38/linechart_colour/conflict/c}{$-$0.21}
\defV{final/internvl35_38/linechart_colour/conflict/ci}{[4.11, 4.91]}
\defV{final/internvl35_38/linechart_colour/conflict/d}{4.49}
\defV{final/internvl35_38/linechart_colour/conflict/n}{140}
\defV{final/internvl35_38/linechart_colour/conflict/unparsed}{0}
\defV{final/internvl35_38/linechart_colour/control/acc}{0.99}
\defV{final/internvl35_38/linechart_colour/control/c}{0.00}
\defV{final/internvl35_38/linechart_colour/control/d}{4.91}
\defV{final/internvl35_38/linechart_colour/control/n}{140}
\defV{final/internvl35_38/linechart_colour/control/unparsed}{1}
\defV{final/internvl35_38/linechart_colour/diff}{$-$0.42}
\defV{final/internvl35_38/linechart_colour/p}{1.00}
\defV{final/internvl35_38/linechart_colour/pairs}{70}
\defV{final/internvl35_38/linechart_colour/pholm}{--}
\defV{final/internvl35_38/linechart_colour/status}{n.d.}
\defV{final/internvl35_38/linechart_text/b}{0}
\defV{final/internvl35_38/linechart_text/c}{0}
\defV{final/internvl35_38/linechart_text/ci}{[0.00, 0.00]}
\defV{final/internvl35_38/linechart_text/conflict/acc}{1.00}
\defV{final/internvl35_38/linechart_text/conflict/c}{$-$0.00}
\defV{final/internvl35_38/linechart_text/conflict/ci}{[4.91, 4.91]}
\defV{final/internvl35_38/linechart_text/conflict/d}{4.91}
\defV{final/internvl35_38/linechart_text/conflict/n}{140}
\defV{final/internvl35_38/linechart_text/conflict/unparsed}{0}
\defV{final/internvl35_38/linechart_text/control/acc}{1.00}
\defV{final/internvl35_38/linechart_text/control/c}{$-$0.00}
\defV{final/internvl35_38/linechart_text/control/d}{4.91}
\defV{final/internvl35_38/linechart_text/control/n}{140}
\defV{final/internvl35_38/linechart_text/control/unparsed}{0}
\defV{final/internvl35_38/linechart_text/diff}{+0.00}
\defV{final/internvl35_38/linechart_text/p}{1.00}
\defV{final/internvl35_38/linechart_text/pairs}{70}
\defV{final/internvl35_38/linechart_text/pholm}{--}
\defV{final/internvl35_38/linechart_text/status}{n.d.}
\defV{final/internvl35_38/map/index_labels/acc}{0.60}
\defV{final/internvl35_38/map/index_labels/d}{0.84}
\defV{final/internvl35_38/map/index_labels/n}{120}
\defV{final/internvl35_38/map/structure_only/acc}{0.58}
\defV{final/internvl35_38/map/structure_only/d}{0.52}
\defV{final/internvl35_38/map/structure_only/n}{120}
\defV{final/internvl35_38/map/values_only/acc}{0.73}
\defV{final/internvl35_38/map/values_only/d}{1.41}
\defV{final/internvl35_38/map/values_only/n}{120}
\defV{final/internvl35_38/map/values_structure/acc}{0.82}
\defV{final/internvl35_38/map/values_structure/d}{1.90}
\defV{final/internvl35_38/map/values_structure/n}{120}
\defV{final/internvl35_38/matched/b}{45}
\defV{final/internvl35_38/matched/c}{14}
\defV{final/internvl35_38/matched/ci}{[$-$2.22, $-$0.80]}
\defV{final/internvl35_38/matched/conflict/acc}{0.39}
\defV{final/internvl35_38/matched/conflict/c}{$-$0.59}
\defV{final/internvl35_38/matched/conflict/ci}{[$-$1.21, $-$0.21]}
\defV{final/internvl35_38/matched/conflict/d}{$-$0.68}
\defV{final/internvl35_38/matched/conflict/n}{140}
\defV{final/internvl35_38/matched/conflict/unparsed}{0}
\defV{final/internvl35_38/matched/control/acc}{0.61}
\defV{final/internvl35_38/matched/control/c}{$-$0.86}
\defV{final/internvl35_38/matched/control/d}{0.77}
\defV{final/internvl35_38/matched/control/n}{140}
\defV{final/internvl35_38/matched/control/unparsed}{0}
\defV{final/internvl35_38/matched/diff}{$-$1.45}
\defV{final/internvl35_38/matched/p}{$7\times10^{-5}$}
\defV{final/internvl35_38/matched/pairs}{70}
\defV{final/internvl35_38/matched/pholm}{$2\times10^{-4}$}
\defV{final/internvl35_38/matched/status}{yes}
\defV{final/internvl35_38/name}{InternVL3.5 38B}
\defV{final/internvl35_38/pointer_linechart/b}{11}
\defV{final/internvl35_38/pointer_linechart/c}{6}
\defV{final/internvl35_38/pointer_linechart/ci}{[$-$0.99, 0.22]}
\defV{final/internvl35_38/pointer_linechart/conflict/acc}{0.62}
\defV{final/internvl35_38/pointer_linechart/conflict/c}{$-$0.00}
\defV{final/internvl35_38/pointer_linechart/conflict/ci}{[0.00, 1.31]}
\defV{final/internvl35_38/pointer_linechart/conflict/d}{0.62}
\defV{final/internvl35_38/pointer_linechart/conflict/n}{80}
\defV{final/internvl35_38/pointer_linechart/conflict/unparsed}{0}
\defV{final/internvl35_38/pointer_linechart/control/acc}{0.69}
\defV{final/internvl35_38/pointer_linechart/control/c}{0.08}
\defV{final/internvl35_38/pointer_linechart/control/d}{1.00}
\defV{final/internvl35_38/pointer_linechart/control/n}{80}
\defV{final/internvl35_38/pointer_linechart/control/unparsed}{1}
\defV{final/internvl35_38/pointer_linechart/diff}{$-$0.38}
\defV{final/internvl35_38/pointer_linechart/p}{0.33}
\defV{final/internvl35_38/pointer_linechart/pairs}{40}
\defV{final/internvl35_38/pointer_linechart/pholm}{--}
\defV{final/internvl35_38/pointer_linechart/status}{--}
\defV{final/internvl35_38/pointer_scatter/b}{6}
\defV{final/internvl35_38/pointer_scatter/c}{0}
\defV{final/internvl35_38/pointer_scatter/ci}{[$-$1.87, $-$0.21]}
\defV{final/internvl35_38/pointer_scatter/conflict/acc}{0.91}
\defV{final/internvl35_38/pointer_scatter/conflict/c}{0.28}
\defV{final/internvl35_38/pointer_scatter/conflict/ci}{[2.37, 4.01]}
\defV{final/internvl35_38/pointer_scatter/conflict/d}{2.99}
\defV{final/internvl35_38/pointer_scatter/conflict/n}{80}
\defV{final/internvl35_38/pointer_scatter/conflict/unparsed}{2}
\defV{final/internvl35_38/pointer_scatter/control/acc}{0.99}
\defV{final/internvl35_38/pointer_scatter/control/c}{0.23}
\defV{final/internvl35_38/pointer_scatter/control/d}{4.04}
\defV{final/internvl35_38/pointer_scatter/control/n}{80}
\defV{final/internvl35_38/pointer_scatter/control/unparsed}{0}
\defV{final/internvl35_38/pointer_scatter/diff}{$-$1.05}
\defV{final/internvl35_38/pointer_scatter/p}{0.03}
\defV{final/internvl35_38/pointer_scatter/pairs}{40}
\defV{final/internvl35_38/pointer_scatter/pholm}{--}
\defV{final/internvl35_38/pointer_scatter/status}{n.d.}
\defV{final/internvl35_38/public_data/b}{25}
\defV{final/internvl35_38/public_data/c}{13}
\defV{final/internvl35_38/public_data/ci}{[$-$1.59, 0.00]}
\defV{final/internvl35_38/public_data/conflict/acc}{0.56}
\defV{final/internvl35_38/public_data/conflict/c}{$-$0.09}
\defV{final/internvl35_38/public_data/conflict/ci}{[$-$0.26, 0.95]}
\defV{final/internvl35_38/public_data/conflict/d}{0.31}
\defV{final/internvl35_38/public_data/conflict/n}{80}
\defV{final/internvl35_38/public_data/conflict/unparsed}{0}
\defV{final/internvl35_38/public_data/control/acc}{0.71}
\defV{final/internvl35_38/public_data/control/c}{0.04}
\defV{final/internvl35_38/public_data/control/d}{1.09}
\defV{final/internvl35_38/public_data/control/n}{80}
\defV{final/internvl35_38/public_data/control/unparsed}{0}
\defV{final/internvl35_38/public_data/diff}{$-$0.78}
\defV{final/internvl35_38/public_data/p}{0.07}
\defV{final/internvl35_38/public_data/pairs}{40}
\defV{final/internvl35_38/public_data/pholm}{--}
\defV{final/internvl35_38/public_data/status}{--}
\defV{final/internvl35_38/repo}{OpenGVLab/InternVL3\_5-38B}
\defV{final/internvl35_38/revision}{de99855}
\defV{final/internvl35_38/scatter/b}{95}
\defV{final/internvl35_38/scatter/c}{5}
\defV{final/internvl35_38/scatter/ci}{[$-$2.93, $-$1.92]}
\defV{final/internvl35_38/scatter/conflict/acc}{0.34}
\defV{final/internvl35_38/scatter/conflict/c}{0.07}
\defV{final/internvl35_38/scatter/conflict/ci}{[$-$1.30, $-$0.45]}
\defV{final/internvl35_38/scatter/conflict/d}{$-$0.85}
\defV{final/internvl35_38/scatter/conflict/n}{200}
\defV{final/internvl35_38/scatter/conflict/unparsed}{0}
\defV{final/internvl35_38/scatter/control/acc}{0.79}
\defV{final/internvl35_38/scatter/control/c}{0.02}
\defV{final/internvl35_38/scatter/control/d}{1.56}
\defV{final/internvl35_38/scatter/control/n}{200}
\defV{final/internvl35_38/scatter/control/unparsed}{0}
\defV{final/internvl35_38/scatter/diff}{$-$2.40}
\defV{final/internvl35_38/scatter/p}{$1\times10^{-22}$}
\defV{final/internvl35_38/scatter/pairs}{100}
\defV{final/internvl35_38/scatter/pholm}{$3\times10^{-21}$}
\defV{final/internvl35_38/scatter/status}{yes}
\defV{final/internvl35_38/scatter_text/b}{2}
\defV{final/internvl35_38/scatter_text/c}{2}
\defV{final/internvl35_38/scatter_text/ci}{[$-$0.70, 0.73]}
\defV{final/internvl35_38/scatter_text/conflict/acc}{0.97}
\defV{final/internvl35_38/scatter_text/conflict/c}{0.07}
\defV{final/internvl35_38/scatter_text/conflict/ci}{[3.26, 4.75]}
\defV{final/internvl35_38/scatter_text/conflict/d}{3.78}
\defV{final/internvl35_38/scatter_text/conflict/n}{200}
\defV{final/internvl35_38/scatter_text/conflict/unparsed}{0}
\defV{final/internvl35_38/scatter_text/control/acc}{0.97}
\defV{final/internvl35_38/scatter_text/control/c}{$-$0.07}
\defV{final/internvl35_38/scatter_text/control/d}{3.78}
\defV{final/internvl35_38/scatter_text/control/n}{200}
\defV{final/internvl35_38/scatter_text/control/unparsed}{0}
\defV{final/internvl35_38/scatter_text/diff}{+0.00}
\defV{final/internvl35_38/scatter_text/p}{1.00}
\defV{final/internvl35_38/scatter_text/pairs}{100}
\defV{final/internvl35_38/scatter_text/pholm}{--}
\defV{final/internvl35_38/scatter_text/status}{n.d.}
\defV{final/internvl35_38/separation/b}{42}
\defV{final/internvl35_38/separation/c}{8}
\defV{final/internvl35_38/separation/ci}{[$-$2.24, $-$1.03]}
\defV{final/internvl35_38/separation/conflict/acc}{0.32}
\defV{final/internvl35_38/separation/conflict/c}{$-$0.74}
\defV{final/internvl35_38/separation/conflict/ci}{[$-$1.74, $-$0.74]}
\defV{final/internvl35_38/separation/conflict/d}{$-$1.19}
\defV{final/internvl35_38/separation/conflict/n}{140}
\defV{final/internvl35_38/separation/conflict/unparsed}{0}
\defV{final/internvl35_38/separation/control/acc}{0.56}
\defV{final/internvl35_38/separation/control/c}{$-$0.63}
\defV{final/internvl35_38/separation/control/d}{0.39}
\defV{final/internvl35_38/separation/control/n}{140}
\defV{final/internvl35_38/separation/control/unparsed}{0}
\defV{final/internvl35_38/separation/diff}{$-$1.58}
\defV{final/internvl35_38/separation/p}{$1\times10^{-6}$}
\defV{final/internvl35_38/separation/pairs}{70}
\defV{final/internvl35_38/separation/pholm}{--}
\defV{final/internvl35_38/separation/status}{n.d.}
\defV{final/internvl35_38/separation_text/b}{17}
\defV{final/internvl35_38/separation_text/c}{9}
\defV{final/internvl35_38/separation_text/ci}{[$-$0.89, 0.71]}
\defV{final/internvl35_38/separation_text/conflict/acc}{0.77}
\defV{final/internvl35_38/separation_text/conflict/c}{$-$0.45}
\defV{final/internvl35_38/separation_text/conflict/ci}{[1.81, 3.06]}
\defV{final/internvl35_38/separation_text/conflict/d}{2.27}
\defV{final/internvl35_38/separation_text/conflict/n}{140}
\defV{final/internvl35_38/separation_text/conflict/unparsed}{13}
\defV{final/internvl35_38/separation_text/control/acc}{0.83}
\defV{final/internvl35_38/separation_text/control/c}{$-$0.30}
\defV{final/internvl35_38/separation_text/control/d}{2.39}
\defV{final/internvl35_38/separation_text/control/n}{140}
\defV{final/internvl35_38/separation_text/control/unparsed}{8}
\defV{final/internvl35_38/separation_text/diff}{$-$0.12}
\defV{final/internvl35_38/separation_text/p}{0.17}
\defV{final/internvl35_38/separation_text/pairs}{70}
\defV{final/internvl35_38/separation_text/pholm}{--}
\defV{final/internvl35_38/separation_text/status}{--}
\defV{final/internvl35_38/truncated}{0}
\defV{final/internvl35_38/unparsed}{25}
\defV{final/internvl35_38/unparsed/pct}{0.4}
\defV{final/internvl35_38/white/b}{41}
\defV{final/internvl35_38/white/c}{18}
\defV{final/internvl35_38/white/ci}{[$-$1.85, $-$0.38]}
\defV{final/internvl35_38/white/conditional}{0.50}
\defV{final/internvl35_38/white/conflict/acc}{0.42}
\defV{final/internvl35_38/white/conflict/c}{$-$0.68}
\defV{final/internvl35_38/white/conflict/ci}{[$-$1.02, 0.00]}
\defV{final/internvl35_38/white/conflict/d}{$-$0.49}
\defV{final/internvl35_38/white/conflict/n}{140}
\defV{final/internvl35_38/white/conflict/unparsed}{0}
\defV{final/internvl35_38/white/control/acc}{0.59}
\defV{final/internvl35_38/white/control/c}{$-$0.76}
\defV{final/internvl35_38/white/control/d}{0.57}
\defV{final/internvl35_38/white/control/n}{140}
\defV{final/internvl35_38/white/control/unparsed}{0}
\defV{final/internvl35_38/white/diff}{$-$1.06}
\defV{final/internvl35_38/white/p}{0.004}
\defV{final/internvl35_38/white/pairs}{70}
\defV{final/internvl35_38/white/pholm}{0.01}
\defV{final/internvl35_38/white/reversal}{no}
\defV{final/internvl35_38/white/status}{yes}
\defV{final/llavaov2_8/answers}{6048}
\defV{final/llavaov2_8/crop/b}{63}
\defV{final/llavaov2_8/crop/c}{16}
\defV{final/llavaov2_8/crop/ci}{[$-$2.57, $-$1.22]}
\defV{final/llavaov2_8/crop/conflict/acc}{0.34}
\defV{final/llavaov2_8/crop/conflict/c}{$-$0.42}
\defV{final/llavaov2_8/crop/conflict/ci}{[$-$1.37, $-$0.43]}
\defV{final/llavaov2_8/crop/conflict/d}{$-$0.89}
\defV{final/llavaov2_8/crop/conflict/n}{140}
\defV{final/llavaov2_8/crop/conflict/unparsed}{2}
\defV{final/llavaov2_8/crop/control/acc}{0.67}
\defV{final/llavaov2_8/crop/control/c}{$-$0.43}
\defV{final/llavaov2_8/crop/control/d}{0.99}
\defV{final/llavaov2_8/crop/control/n}{140}
\defV{final/llavaov2_8/crop/control/unparsed}{1}
\defV{final/llavaov2_8/crop/diff}{$-$1.88}
\defV{final/llavaov2_8/crop/p}{$9\times10^{-8}$}
\defV{final/llavaov2_8/crop/pairs}{70}
\defV{final/llavaov2_8/crop/pholm}{$9\times10^{-8}$}
\defV{final/llavaov2_8/crop/status}{yes}
\defV{final/llavaov2_8/cues_leader/b}{57}
\defV{final/llavaov2_8/cues_leader/c}{2}
\defV{final/llavaov2_8/cues_leader/ci}{[$-$2.69, $-$1.63]}
\defV{final/llavaov2_8/cues_leader/conflict/acc}{0.38}
\defV{final/llavaov2_8/cues_leader/conflict/c}{$-$0.15}
\defV{final/llavaov2_8/cues_leader/conflict/ci}{[$-$0.99, $-$0.24]}
\defV{final/llavaov2_8/cues_leader/conflict/d}{$-$0.60}
\defV{final/llavaov2_8/cues_leader/conflict/n}{136}
\defV{final/llavaov2_8/cues_leader/conflict/unparsed}{2}
\defV{final/llavaov2_8/cues_leader/control/acc}{0.78}
\defV{final/llavaov2_8/cues_leader/control/c}{$-$0.05}
\defV{final/llavaov2_8/cues_leader/control/d}{1.52}
\defV{final/llavaov2_8/cues_leader/control/n}{136}
\defV{final/llavaov2_8/cues_leader/control/unparsed}{0}
\defV{final/llavaov2_8/cues_leader/diff}{$-$2.12}
\defV{final/llavaov2_8/cues_leader/p}{$6\times10^{-15}$}
\defV{final/llavaov2_8/cues_leader/pairs}{68}
\defV{final/llavaov2_8/cues_leader/pholm}{--}
\defV{final/llavaov2_8/cues_leader/status}{--}
\defV{final/llavaov2_8/cues_region/b}{58}
\defV{final/llavaov2_8/cues_region/c}{6}
\defV{final/llavaov2_8/cues_region/ci}{[$-$2.64, $-$1.41]}
\defV{final/llavaov2_8/cues_region/conflict/acc}{0.39}
\defV{final/llavaov2_8/cues_region/conflict/c}{$-$0.14}
\defV{final/llavaov2_8/cues_region/conflict/ci}{[$-$0.99, $-$0.20]}
\defV{final/llavaov2_8/cues_region/conflict/d}{$-$0.57}
\defV{final/llavaov2_8/cues_region/conflict/n}{140}
\defV{final/llavaov2_8/cues_region/conflict/unparsed}{1}
\defV{final/llavaov2_8/cues_region/control/acc}{0.76}
\defV{final/llavaov2_8/cues_region/control/c}{$-$0.11}
\defV{final/llavaov2_8/cues_region/control/d}{1.42}
\defV{final/llavaov2_8/cues_region/control/n}{140}
\defV{final/llavaov2_8/cues_region/control/unparsed}{1}
\defV{final/llavaov2_8/cues_region/diff}{$-$1.98}
\defV{final/llavaov2_8/cues_region/p}{$9\times10^{-12}$}
\defV{final/llavaov2_8/cues_region/pairs}{70}
\defV{final/llavaov2_8/cues_region/pholm}{--}
\defV{final/llavaov2_8/cues_region/status}{--}
\defV{final/llavaov2_8/direct/b}{5}
\defV{final/llavaov2_8/direct/c}{5}
\defV{final/llavaov2_8/direct/ci}{[$-$0.22, 0.26]}
\defV{final/llavaov2_8/direct/conflict/acc}{0.49}
\defV{final/llavaov2_8/direct/conflict/c}{0.85}
\defV{final/llavaov2_8/direct/conflict/ci}{[$-$0.23, 0.12]}
\defV{final/llavaov2_8/direct/conflict/d}{$-$0.05}
\defV{final/llavaov2_8/direct/conflict/n}{140}
\defV{final/llavaov2_8/direct/conflict/unparsed}{0}
\defV{final/llavaov2_8/direct/control/acc}{0.49}
\defV{final/llavaov2_8/direct/control/c}{1.08}
\defV{final/llavaov2_8/direct/control/d}{$-$0.06}
\defV{final/llavaov2_8/direct/control/n}{140}
\defV{final/llavaov2_8/direct/control/unparsed}{0}
\defV{final/llavaov2_8/direct/diff}{+0.01}
\defV{final/llavaov2_8/direct/p}{1.00}
\defV{final/llavaov2_8/direct/pairs}{70}
\defV{final/llavaov2_8/direct/pholm}{1.00}
\defV{final/llavaov2_8/direct/status}{n.d.}
\defV{final/llavaov2_8/failed}{0}
\defV{final/llavaov2_8/gate}{$-$0.05}
\defV{final/llavaov2_8/gsi_placement/b}{5}
\defV{final/llavaov2_8/gsi_placement/c}{5}
\defV{final/llavaov2_8/gsi_placement/ci}{[$-$0.96, 0.98]}
\defV{final/llavaov2_8/gsi_placement/conflict/acc}{0.47}
\defV{final/llavaov2_8/gsi_placement/conflict/c}{$-$1.76}
\defV{final/llavaov2_8/gsi_placement/conflict/ci}{[$-$0.44, 1.13]}
\defV{final/llavaov2_8/gsi_placement/conflict/d}{0.38}
\defV{final/llavaov2_8/gsi_placement/conflict/n}{128}
\defV{final/llavaov2_8/gsi_placement/conflict/unparsed}{11}
\defV{final/llavaov2_8/gsi_placement/control/acc}{0.47}
\defV{final/llavaov2_8/gsi_placement/control/c}{$-$2.18}
\defV{final/llavaov2_8/gsi_placement/control/d}{0.43}
\defV{final/llavaov2_8/gsi_placement/control/n}{128}
\defV{final/llavaov2_8/gsi_placement/control/unparsed}{9}
\defV{final/llavaov2_8/gsi_placement/diff}{$-$0.04}
\defV{final/llavaov2_8/gsi_placement/p}{1.00}
\defV{final/llavaov2_8/gsi_placement/pairs}{64}
\defV{final/llavaov2_8/gsi_placement/pholm}{--}
\defV{final/llavaov2_8/gsi_placement/status}{n.d.}
\defV{final/llavaov2_8/gsi_placement_k1/b}{4}
\defV{final/llavaov2_8/gsi_placement_k1/c}{0}
\defV{final/llavaov2_8/gsi_placement_k1/ci}{[$-$2.44, 0.00]}
\defV{final/llavaov2_8/gsi_placement_k1/conflict/acc}{0.46}
\defV{final/llavaov2_8/gsi_placement_k1/conflict/c}{$-$1.65}
\defV{final/llavaov2_8/gsi_placement_k1/conflict/ci}{[$-$1.30, 0.00]}
\defV{final/llavaov2_8/gsi_placement_k1/conflict/d}{$-$0.78}
\defV{final/llavaov2_8/gsi_placement_k1/conflict/n}{46}
\defV{final/llavaov2_8/gsi_placement_k1/conflict/unparsed}{0}
\defV{final/llavaov2_8/gsi_placement_k1/control/acc}{0.54}
\defV{final/llavaov2_8/gsi_placement_k1/control/c}{$-$1.65}
\defV{final/llavaov2_8/gsi_placement_k1/control/d}{0.78}
\defV{final/llavaov2_8/gsi_placement_k1/control/n}{46}
\defV{final/llavaov2_8/gsi_placement_k1/control/unparsed}{0}
\defV{final/llavaov2_8/gsi_placement_k1/diff}{$-$1.56}
\defV{final/llavaov2_8/gsi_placement_k1/p}{0.12}
\defV{final/llavaov2_8/gsi_placement_k1/pairs}{23}
\defV{final/llavaov2_8/gsi_placement_k1/pholm}{--}
\defV{final/llavaov2_8/gsi_placement_k1/status}{n.d.}
\defV{final/llavaov2_8/isobar/b}{43}
\defV{final/llavaov2_8/isobar/c}{26}
\defV{final/llavaov2_8/isobar/ci}{[$-$0.96, $-$0.03]}
\defV{final/llavaov2_8/isobar/conflict/acc}{0.45}
\defV{final/llavaov2_8/isobar/conflict/c}{0.50}
\defV{final/llavaov2_8/isobar/conflict/ci}{[$-$0.68, 0.12]}
\defV{final/llavaov2_8/isobar/conflict/d}{$-$0.28}
\defV{final/llavaov2_8/isobar/conflict/n}{200}
\defV{final/llavaov2_8/isobar/conflict/unparsed}{0}
\defV{final/llavaov2_8/isobar/control/acc}{0.54}
\defV{final/llavaov2_8/isobar/control/c}{0.45}
\defV{final/llavaov2_8/isobar/control/d}{0.19}
\defV{final/llavaov2_8/isobar/control/n}{200}
\defV{final/llavaov2_8/isobar/control/unparsed}{0}
\defV{final/llavaov2_8/isobar/diff}{$-$0.47}
\defV{final/llavaov2_8/isobar/p}{0.05}
\defV{final/llavaov2_8/isobar/pairs}{100}
\defV{final/llavaov2_8/isobar/pholm}{--}
\defV{final/llavaov2_8/isobar/status}{n.d.}
\defV{final/llavaov2_8/isobar_text/b}{17}
\defV{final/llavaov2_8/isobar_text/c}{17}
\defV{final/llavaov2_8/isobar_text/ci}{[$-$0.32, 0.33]}
\defV{final/llavaov2_8/isobar_text/conflict/acc}{0.46}
\defV{final/llavaov2_8/isobar_text/conflict/c}{0.21}
\defV{final/llavaov2_8/isobar_text/conflict/ci}{[$-$0.59, 0.13]}
\defV{final/llavaov2_8/isobar_text/conflict/d}{$-$0.23}
\defV{final/llavaov2_8/isobar_text/conflict/n}{200}
\defV{final/llavaov2_8/isobar_text/conflict/unparsed}{0}
\defV{final/llavaov2_8/isobar_text/control/acc}{0.46}
\defV{final/llavaov2_8/isobar_text/control/c}{0.21}
\defV{final/llavaov2_8/isobar_text/control/d}{$-$0.23}
\defV{final/llavaov2_8/isobar_text/control/n}{200}
\defV{final/llavaov2_8/isobar_text/control/unparsed}{0}
\defV{final/llavaov2_8/isobar_text/diff}{+0.00}
\defV{final/llavaov2_8/isobar_text/p}{1.00}
\defV{final/llavaov2_8/isobar_text/pairs}{100}
\defV{final/llavaov2_8/isobar_text/pholm}{--}
\defV{final/llavaov2_8/isobar_text/status}{n.d.}
\defV{final/llavaov2_8/linechart/b}{30}
\defV{final/llavaov2_8/linechart/c}{12}
\defV{final/llavaov2_8/linechart/ci}{[$-$1.21, $-$0.14]}
\defV{final/llavaov2_8/linechart/conflict/acc}{0.56}
\defV{final/llavaov2_8/linechart/conflict/c}{$-$0.11}
\defV{final/llavaov2_8/linechart/conflict/ci}{[$-$0.25, 0.80]}
\defV{final/llavaov2_8/linechart/conflict/d}{0.29}
\defV{final/llavaov2_8/linechart/conflict/n}{140}
\defV{final/llavaov2_8/linechart/conflict/unparsed}{0}
\defV{final/llavaov2_8/linechart/control/acc}{0.69}
\defV{final/llavaov2_8/linechart/control/c}{$-$0.04}
\defV{final/llavaov2_8/linechart/control/d}{0.95}
\defV{final/llavaov2_8/linechart/control/n}{140}
\defV{final/llavaov2_8/linechart/control/unparsed}{0}
\defV{final/llavaov2_8/linechart/diff}{$-$0.67}
\defV{final/llavaov2_8/linechart/p}{0.008}
\defV{final/llavaov2_8/linechart/pairs}{70}
\defV{final/llavaov2_8/linechart/pholm}{--}
\defV{final/llavaov2_8/linechart/status}{--}
\defV{final/llavaov2_8/linechart_colour/b}{16}
\defV{final/llavaov2_8/linechart_colour/c}{5}
\defV{final/llavaov2_8/linechart_colour/ci}{[$-$1.01, $-$0.06]}
\defV{final/llavaov2_8/linechart_colour/conflict/acc}{0.76}
\defV{final/llavaov2_8/linechart_colour/conflict/c}{$-$0.11}
\defV{final/llavaov2_8/linechart_colour/conflict/ci}{[0.96, 1.98]}
\defV{final/llavaov2_8/linechart_colour/conflict/d}{1.42}
\defV{final/llavaov2_8/linechart_colour/conflict/n}{140}
\defV{final/llavaov2_8/linechart_colour/conflict/unparsed}{1}
\defV{final/llavaov2_8/linechart_colour/control/acc}{0.84}
\defV{final/llavaov2_8/linechart_colour/control/c}{$-$0.08}
\defV{final/llavaov2_8/linechart_colour/control/d}{1.92}
\defV{final/llavaov2_8/linechart_colour/control/n}{140}
\defV{final/llavaov2_8/linechart_colour/control/unparsed}{0}
\defV{final/llavaov2_8/linechart_colour/diff}{$-$0.51}
\defV{final/llavaov2_8/linechart_colour/p}{0.03}
\defV{final/llavaov2_8/linechart_colour/pairs}{70}
\defV{final/llavaov2_8/linechart_colour/pholm}{--}
\defV{final/llavaov2_8/linechart_colour/status}{--}
\defV{final/llavaov2_8/linechart_text/b}{0}
\defV{final/llavaov2_8/linechart_text/c}{0}
\defV{final/llavaov2_8/linechart_text/ci}{[0.00, 0.00]}
\defV{final/llavaov2_8/linechart_text/conflict/acc}{1.00}
\defV{final/llavaov2_8/linechart_text/conflict/c}{$-$0.00}
\defV{final/llavaov2_8/linechart_text/conflict/ci}{[4.91, 4.91]}
\defV{final/llavaov2_8/linechart_text/conflict/d}{4.91}
\defV{final/llavaov2_8/linechart_text/conflict/n}{140}
\defV{final/llavaov2_8/linechart_text/conflict/unparsed}{0}
\defV{final/llavaov2_8/linechart_text/control/acc}{1.00}
\defV{final/llavaov2_8/linechart_text/control/c}{$-$0.00}
\defV{final/llavaov2_8/linechart_text/control/d}{4.91}
\defV{final/llavaov2_8/linechart_text/control/n}{140}
\defV{final/llavaov2_8/linechart_text/control/unparsed}{0}
\defV{final/llavaov2_8/linechart_text/diff}{+0.00}
\defV{final/llavaov2_8/linechart_text/p}{1.00}
\defV{final/llavaov2_8/linechart_text/pairs}{70}
\defV{final/llavaov2_8/linechart_text/pholm}{--}
\defV{final/llavaov2_8/linechart_text/status}{n.d.}
\defV{final/llavaov2_8/map/index_labels/acc}{0.41}
\defV{final/llavaov2_8/map/index_labels/d}{0.12}
\defV{final/llavaov2_8/map/index_labels/n}{120}
\defV{final/llavaov2_8/map/structure_only/acc}{0.54}
\defV{final/llavaov2_8/map/structure_only/d}{0.27}
\defV{final/llavaov2_8/map/structure_only/n}{120}
\defV{final/llavaov2_8/map/values_only/acc}{0.13}
\defV{final/llavaov2_8/map/values_only/d}{0.31}
\defV{final/llavaov2_8/map/values_only/n}{120}
\defV{final/llavaov2_8/map/values_structure/acc}{0.62}
\defV{final/llavaov2_8/map/values_structure/d}{1.05}
\defV{final/llavaov2_8/map/values_structure/n}{120}
\defV{final/llavaov2_8/matched/b}{46}
\defV{final/llavaov2_8/matched/c}{18}
\defV{final/llavaov2_8/matched/ci}{[$-$2.53, $-$1.14]}
\defV{final/llavaov2_8/matched/conflict/acc}{0.26}
\defV{final/llavaov2_8/matched/conflict/c}{0.12}
\defV{final/llavaov2_8/matched/conflict/ci}{[$-$1.27, $-$0.46]}
\defV{final/llavaov2_8/matched/conflict/d}{$-$0.82}
\defV{final/llavaov2_8/matched/conflict/n}{140}
\defV{final/llavaov2_8/matched/conflict/unparsed}{30}
\defV{final/llavaov2_8/matched/control/acc}{0.46}
\defV{final/llavaov2_8/matched/control/c}{$-$0.03}
\defV{final/llavaov2_8/matched/control/d}{0.98}
\defV{final/llavaov2_8/matched/control/n}{140}
\defV{final/llavaov2_8/matched/control/unparsed}{46}
\defV{final/llavaov2_8/matched/diff}{$-$1.80}
\defV{final/llavaov2_8/matched/p}{$6\times10^{-4}$}
\defV{final/llavaov2_8/matched/pairs}{70}
\defV{final/llavaov2_8/matched/pholm}{0.001}
\defV{final/llavaov2_8/matched/status}{yes}
\defV{final/llavaov2_8/name}{LLaVA-OneVision-2 8B}
\defV{final/llavaov2_8/pointer_linechart/b}{11}
\defV{final/llavaov2_8/pointer_linechart/c}{12}
\defV{final/llavaov2_8/pointer_linechart/ci}{[$-$0.63, 0.72]}
\defV{final/llavaov2_8/pointer_linechart/conflict/acc}{0.62}
\defV{final/llavaov2_8/pointer_linechart/conflict/c}{0.33}
\defV{final/llavaov2_8/pointer_linechart/conflict/ci}{[0.25, 1.39]}
\defV{final/llavaov2_8/pointer_linechart/conflict/d}{0.78}
\defV{final/llavaov2_8/pointer_linechart/conflict/n}{80}
\defV{final/llavaov2_8/pointer_linechart/conflict/unparsed}{3}
\defV{final/llavaov2_8/pointer_linechart/control/acc}{0.61}
\defV{final/llavaov2_8/pointer_linechart/control/c}{0.13}
\defV{final/llavaov2_8/pointer_linechart/control/d}{0.72}
\defV{final/llavaov2_8/pointer_linechart/control/n}{80}
\defV{final/llavaov2_8/pointer_linechart/control/unparsed}{4}
\defV{final/llavaov2_8/pointer_linechart/diff}{+0.06}
\defV{final/llavaov2_8/pointer_linechart/p}{1.00}
\defV{final/llavaov2_8/pointer_linechart/pairs}{40}
\defV{final/llavaov2_8/pointer_linechart/pholm}{--}
\defV{final/llavaov2_8/pointer_linechart/status}{--}
\defV{final/llavaov2_8/pointer_scatter/b}{15}
\defV{final/llavaov2_8/pointer_scatter/c}{5}
\defV{final/llavaov2_8/pointer_scatter/ci}{[$-$1.60, 0.55]}
\defV{final/llavaov2_8/pointer_scatter/conflict/acc}{0.74}
\defV{final/llavaov2_8/pointer_scatter/conflict/c}{0.01}
\defV{final/llavaov2_8/pointer_scatter/conflict/ci}{[1.66, 3.17]}
\defV{final/llavaov2_8/pointer_scatter/conflict/d}{2.25}
\defV{final/llavaov2_8/pointer_scatter/conflict/n}{80}
\defV{final/llavaov2_8/pointer_scatter/conflict/unparsed}{13}
\defV{final/llavaov2_8/pointer_scatter/control/acc}{0.86}
\defV{final/llavaov2_8/pointer_scatter/control/c}{0.18}
\defV{final/llavaov2_8/pointer_scatter/control/d}{2.70}
\defV{final/llavaov2_8/pointer_scatter/control/n}{80}
\defV{final/llavaov2_8/pointer_scatter/control/unparsed}{5}
\defV{final/llavaov2_8/pointer_scatter/diff}{$-$0.45}
\defV{final/llavaov2_8/pointer_scatter/p}{0.04}
\defV{final/llavaov2_8/pointer_scatter/pairs}{40}
\defV{final/llavaov2_8/pointer_scatter/pholm}{--}
\defV{final/llavaov2_8/pointer_scatter/status}{--}
\defV{final/llavaov2_8/public_data/b}{25}
\defV{final/llavaov2_8/public_data/c}{8}
\defV{final/llavaov2_8/public_data/ci}{[$-$1.82, $-$0.39]}
\defV{final/llavaov2_8/public_data/conflict/acc}{0.41}
\defV{final/llavaov2_8/public_data/conflict/c}{0.22}
\defV{final/llavaov2_8/public_data/conflict/ci}{[$-$1.00, 0.07]}
\defV{final/llavaov2_8/public_data/conflict/d}{$-$0.44}
\defV{final/llavaov2_8/public_data/conflict/n}{80}
\defV{final/llavaov2_8/public_data/conflict/unparsed}{0}
\defV{final/llavaov2_8/public_data/control/acc}{0.62}
\defV{final/llavaov2_8/public_data/control/c}{0.19}
\defV{final/llavaov2_8/public_data/control/d}{0.63}
\defV{final/llavaov2_8/public_data/control/n}{80}
\defV{final/llavaov2_8/public_data/control/unparsed}{0}
\defV{final/llavaov2_8/public_data/diff}{$-$1.07}
\defV{final/llavaov2_8/public_data/p}{0.005}
\defV{final/llavaov2_8/public_data/pairs}{40}
\defV{final/llavaov2_8/public_data/pholm}{--}
\defV{final/llavaov2_8/public_data/status}{--}
\defV{final/llavaov2_8/repo}{lmms-lab-encoder/LLaVA-OneVision-2-8B-Instruct}
\defV{final/llavaov2_8/revision}{878cde7}
\defV{final/llavaov2_8/scatter/b}{58}
\defV{final/llavaov2_8/scatter/c}{6}
\defV{final/llavaov2_8/scatter/ci}{[$-$1.79, $-$0.94]}
\defV{final/llavaov2_8/scatter/conflict/acc}{0.43}
\defV{final/llavaov2_8/scatter/conflict/c}{0.07}
\defV{final/llavaov2_8/scatter/conflict/ci}{[$-$0.68, 0.05]}
\defV{final/llavaov2_8/scatter/conflict/d}{$-$0.31}
\defV{final/llavaov2_8/scatter/conflict/n}{200}
\defV{final/llavaov2_8/scatter/conflict/unparsed}{1}
\defV{final/llavaov2_8/scatter/control/acc}{0.69}
\defV{final/llavaov2_8/scatter/control/c}{0.06}
\defV{final/llavaov2_8/scatter/control/d}{1.03}
\defV{final/llavaov2_8/scatter/control/n}{200}
\defV{final/llavaov2_8/scatter/control/unparsed}{1}
\defV{final/llavaov2_8/scatter/diff}{$-$1.34}
\defV{final/llavaov2_8/scatter/p}{$9\times10^{-12}$}
\defV{final/llavaov2_8/scatter/pairs}{100}
\defV{final/llavaov2_8/scatter/pholm}{$2\times10^{-10}$}
\defV{final/llavaov2_8/scatter/status}{yes}
\defV{final/llavaov2_8/scatter_text/b}{8}
\defV{final/llavaov2_8/scatter_text/c}{6}
\defV{final/llavaov2_8/scatter_text/ci}{[$-$0.63, 0.48]}
\defV{final/llavaov2_8/scatter_text/conflict/acc}{0.88}
\defV{final/llavaov2_8/scatter_text/conflict/c}{$-$0.04}
\defV{final/llavaov2_8/scatter_text/conflict/ci}{[2.46, 3.61]}
\defV{final/llavaov2_8/scatter_text/conflict/d}{2.91}
\defV{final/llavaov2_8/scatter_text/conflict/n}{200}
\defV{final/llavaov2_8/scatter_text/conflict/unparsed}{11}
\defV{final/llavaov2_8/scatter_text/control/acc}{0.89}
\defV{final/llavaov2_8/scatter_text/control/c}{$-$0.00}
\defV{final/llavaov2_8/scatter_text/control/d}{2.99}
\defV{final/llavaov2_8/scatter_text/control/n}{200}
\defV{final/llavaov2_8/scatter_text/control/unparsed}{10}
\defV{final/llavaov2_8/scatter_text/diff}{$-$0.08}
\defV{final/llavaov2_8/scatter_text/p}{0.79}
\defV{final/llavaov2_8/scatter_text/pairs}{100}
\defV{final/llavaov2_8/scatter_text/pholm}{--}
\defV{final/llavaov2_8/scatter_text/status}{--}
\defV{final/llavaov2_8/separation/b}{46}
\defV{final/llavaov2_8/separation/c}{11}
\defV{final/llavaov2_8/separation/ci}{[$-$1.93, $-$0.82]}
\defV{final/llavaov2_8/separation/conflict/acc}{0.43}
\defV{final/llavaov2_8/separation/conflict/c}{$-$0.30}
\defV{final/llavaov2_8/separation/conflict/ci}{[$-$0.72, 0.00]}
\defV{final/llavaov2_8/separation/conflict/d}{$-$0.35}
\defV{final/llavaov2_8/separation/conflict/n}{140}
\defV{final/llavaov2_8/separation/conflict/unparsed}{1}
\defV{final/llavaov2_8/separation/control/acc}{0.68}
\defV{final/llavaov2_8/separation/control/c}{$-$0.29}
\defV{final/llavaov2_8/separation/control/d}{0.98}
\defV{final/llavaov2_8/separation/control/n}{140}
\defV{final/llavaov2_8/separation/control/unparsed}{1}
\defV{final/llavaov2_8/separation/diff}{$-$1.33}
\defV{final/llavaov2_8/separation/p}{$3\times10^{-6}$}
\defV{final/llavaov2_8/separation/pairs}{70}
\defV{final/llavaov2_8/separation/pholm}{$4\times10^{-5}$}
\defV{final/llavaov2_8/separation/status}{yes}
\defV{final/llavaov2_8/separation_text/b}{10}
\defV{final/llavaov2_8/separation_text/c}{17}
\defV{final/llavaov2_8/separation_text/ci}{[$-$0.06, 0.66]}
\defV{final/llavaov2_8/separation_text/conflict/acc}{0.57}
\defV{final/llavaov2_8/separation_text/conflict/c}{$-$0.37}
\defV{final/llavaov2_8/separation_text/conflict/ci}{[0.04, 0.92]}
\defV{final/llavaov2_8/separation_text/conflict/d}{0.48}
\defV{final/llavaov2_8/separation_text/conflict/n}{140}
\defV{final/llavaov2_8/separation_text/conflict/unparsed}{4}
\defV{final/llavaov2_8/separation_text/control/acc}{0.52}
\defV{final/llavaov2_8/separation_text/control/c}{$-$0.36}
\defV{final/llavaov2_8/separation_text/control/d}{0.18}
\defV{final/llavaov2_8/separation_text/control/n}{140}
\defV{final/llavaov2_8/separation_text/control/unparsed}{3}
\defV{final/llavaov2_8/separation_text/diff}{+0.30}
\defV{final/llavaov2_8/separation_text/p}{0.25}
\defV{final/llavaov2_8/separation_text/pairs}{70}
\defV{final/llavaov2_8/separation_text/pholm}{--}
\defV{final/llavaov2_8/separation_text/status}{n.d.}
\defV{final/llavaov2_8/truncated}{316}
\defV{final/llavaov2_8/truncated/gsi_placement}{20}
\defV{final/llavaov2_8/truncated/gsi_placement_development}{5}
\defV{final/llavaov2_8/truncated/label_placement_boxes}{164}
\defV{final/llavaov2_8/truncated/map_pairs}{124}
\defV{final/llavaov2_8/truncated/separation_shortcut_free}{2}
\defV{final/llavaov2_8/truncated/separation_shortcut_free_text}{1}
\defV{final/llavaov2_8/unparsed}{376}
\defV{final/llavaov2_8/unparsed/pct}{6.2}
\defV{final/llavaov2_8/white/b}{44}
\defV{final/llavaov2_8/white/c}{16}
\defV{final/llavaov2_8/white/ci}{[$-$2.51, $-$0.95]}
\defV{final/llavaov2_8/white/conditional}{0.29}
\defV{final/llavaov2_8/white/conflict/acc}{0.24}
\defV{final/llavaov2_8/white/conflict/c}{0.30}
\defV{final/llavaov2_8/white/conflict/ci}{[$-$1.35, $-$0.37]}
\defV{final/llavaov2_8/white/conflict/d}{$-$0.85}
\defV{final/llavaov2_8/white/conflict/n}{140}
\defV{final/llavaov2_8/white/conflict/unparsed}{39}
\defV{final/llavaov2_8/white/control/acc}{0.44}
\defV{final/llavaov2_8/white/control/c}{$-$0.12}
\defV{final/llavaov2_8/white/control/d}{0.83}
\defV{final/llavaov2_8/white/control/n}{140}
\defV{final/llavaov2_8/white/control/unparsed}{47}
\defV{final/llavaov2_8/white/diff}{$-$1.67}
\defV{final/llavaov2_8/white/p}{$4\times10^{-4}$}
\defV{final/llavaov2_8/white/pairs}{70}
\defV{final/llavaov2_8/white/pholm}{0.002}
\defV{final/llavaov2_8/white/reversal}{yes}
\defV{final/llavaov2_8/white/status}{yes}
\defV{final/map/index_labels/max}{2.19}
\defV{final/map/index_labels/min}{$-$0.16}
\defV{final/map/n}{120}
\defV{final/map/pairs}{60}
\defV{final/map/structure_only/max}{1.85}
\defV{final/map/structure_only/min}{$-$0.09}
\defV{final/map/values_only/max}{1.97}
\defV{final/map/values_only/min}{0.31}
\defV{final/map/values_structure/max}{4.80}
\defV{final/map/values_structure/min}{1.05}
\defV{final/mistral32_24/answers}{6048}
\defV{final/mistral32_24/crop/b}{67}
\defV{final/mistral32_24/crop/c}{10}
\defV{final/mistral32_24/crop/ci}{[$-$3.17, $-$1.75]}
\defV{final/mistral32_24/crop/conflict/acc}{0.34}
\defV{final/mistral32_24/crop/conflict/c}{$-$0.47}
\defV{final/mistral32_24/crop/conflict/ci}{[$-$1.39, $-$0.54]}
\defV{final/mistral32_24/crop/conflict/d}{$-$0.93}
\defV{final/mistral32_24/crop/conflict/n}{140}
\defV{final/mistral32_24/crop/conflict/unparsed}{0}
\defV{final/mistral32_24/crop/control/acc}{0.74}
\defV{final/mistral32_24/crop/control/c}{$-$0.52}
\defV{final/mistral32_24/crop/control/d}{1.46}
\defV{final/mistral32_24/crop/control/n}{140}
\defV{final/mistral32_24/crop/control/unparsed}{0}
\defV{final/mistral32_24/crop/diff}{$-$2.39}
\defV{final/mistral32_24/crop/p}{$2\times10^{-11}$}
\defV{final/mistral32_24/crop/pairs}{70}
\defV{final/mistral32_24/crop/pholm}{--}
\defV{final/mistral32_24/crop/status}{--}
\defV{final/mistral32_24/cues_leader/b}{59}
\defV{final/mistral32_24/cues_leader/c}{2}
\defV{final/mistral32_24/cues_leader/ci}{[$-$3.43, $-$1.93]}
\defV{final/mistral32_24/cues_leader/conflict/acc}{0.48}
\defV{final/mistral32_24/cues_leader/conflict/c}{0.02}
\defV{final/mistral32_24/cues_leader/conflict/ci}{[$-$0.59, 0.40]}
\defV{final/mistral32_24/cues_leader/conflict/d}{$-$0.11}
\defV{final/mistral32_24/cues_leader/conflict/n}{136}
\defV{final/mistral32_24/cues_leader/conflict/unparsed}{0}
\defV{final/mistral32_24/cues_leader/control/acc}{0.90}
\defV{final/mistral32_24/cues_leader/control/c}{0.00}
\defV{final/mistral32_24/cues_leader/control/d}{2.47}
\defV{final/mistral32_24/cues_leader/control/n}{136}
\defV{final/mistral32_24/cues_leader/control/unparsed}{0}
\defV{final/mistral32_24/cues_leader/diff}{$-$2.58}
\defV{final/mistral32_24/cues_leader/p}{$2\times10^{-15}$}
\defV{final/mistral32_24/cues_leader/pairs}{68}
\defV{final/mistral32_24/cues_leader/pholm}{--}
\defV{final/mistral32_24/cues_leader/status}{--}
\defV{final/mistral32_24/cues_region/b}{67}
\defV{final/mistral32_24/cues_region/c}{0}
\defV{final/mistral32_24/cues_region/ci}{[$-$3.93, $-$2.52]}
\defV{final/mistral32_24/cues_region/conflict/acc}{0.45}
\defV{final/mistral32_24/cues_region/conflict/c}{$-$0.05}
\defV{final/mistral32_24/cues_region/conflict/ci}{[$-$0.76, 0.25]}
\defV{final/mistral32_24/cues_region/conflict/d}{$-$0.25}
\defV{final/mistral32_24/cues_region/conflict/n}{140}
\defV{final/mistral32_24/cues_region/conflict/unparsed}{0}
\defV{final/mistral32_24/cues_region/control/acc}{0.93}
\defV{final/mistral32_24/cues_region/control/c}{$-$0.10}
\defV{final/mistral32_24/cues_region/control/d}{2.86}
\defV{final/mistral32_24/cues_region/control/n}{140}
\defV{final/mistral32_24/cues_region/control/unparsed}{0}
\defV{final/mistral32_24/cues_region/diff}{$-$3.11}
\defV{final/mistral32_24/cues_region/p}{$1\times10^{-20}$}
\defV{final/mistral32_24/cues_region/pairs}{70}
\defV{final/mistral32_24/cues_region/pholm}{--}
\defV{final/mistral32_24/cues_region/status}{--}
\defV{final/mistral32_24/direct/b}{0}
\defV{final/mistral32_24/direct/c}{0}
\defV{final/mistral32_24/direct/ci}{[0.00, 0.00]}
\defV{final/mistral32_24/direct/conflict/acc}{0.50}
\defV{final/mistral32_24/direct/conflict/c}{$-$2.46}
\defV{final/mistral32_24/direct/conflict/ci}{[0.00, 0.00]}
\defV{final/mistral32_24/direct/conflict/d}{0.00}
\defV{final/mistral32_24/direct/conflict/n}{140}
\defV{final/mistral32_24/direct/conflict/unparsed}{0}
\defV{final/mistral32_24/direct/control/acc}{0.50}
\defV{final/mistral32_24/direct/control/c}{$-$2.46}
\defV{final/mistral32_24/direct/control/d}{0.00}
\defV{final/mistral32_24/direct/control/n}{140}
\defV{final/mistral32_24/direct/control/unparsed}{0}
\defV{final/mistral32_24/direct/diff}{+0.00}
\defV{final/mistral32_24/direct/p}{1.00}
\defV{final/mistral32_24/direct/pairs}{70}
\defV{final/mistral32_24/direct/pholm}{--}
\defV{final/mistral32_24/direct/status}{n.d.}
\defV{final/mistral32_24/failed}{0}
\defV{final/mistral32_24/gate}{$-$1.10}
\defV{final/mistral32_24/gsi_placement/b}{38}
\defV{final/mistral32_24/gsi_placement/c}{12}
\defV{final/mistral32_24/gsi_placement/ci}{[$-$1.56, $-$0.51]}
\defV{final/mistral32_24/gsi_placement/conflict/acc}{0.41}
\defV{final/mistral32_24/gsi_placement/conflict/c}{0.04}
\defV{final/mistral32_24/gsi_placement/conflict/ci}{[$-$0.88, $-$0.08]}
\defV{final/mistral32_24/gsi_placement/conflict/d}{$-$0.47}
\defV{final/mistral32_24/gsi_placement/conflict/n}{128}
\defV{final/mistral32_24/gsi_placement/conflict/unparsed}{0}
\defV{final/mistral32_24/gsi_placement/control/acc}{0.61}
\defV{final/mistral32_24/gsi_placement/control/c}{$-$0.20}
\defV{final/mistral32_24/gsi_placement/control/d}{0.56}
\defV{final/mistral32_24/gsi_placement/control/n}{128}
\defV{final/mistral32_24/gsi_placement/control/unparsed}{0}
\defV{final/mistral32_24/gsi_placement/diff}{$-$1.03}
\defV{final/mistral32_24/gsi_placement/p}{$3\times10^{-4}$}
\defV{final/mistral32_24/gsi_placement/pairs}{64}
\defV{final/mistral32_24/gsi_placement/pholm}{--}
\defV{final/mistral32_24/gsi_placement/status}{n.d.}
\defV{final/mistral32_24/gsi_placement_k1/b}{12}
\defV{final/mistral32_24/gsi_placement_k1/c}{9}
\defV{final/mistral32_24/gsi_placement_k1/ci}{[$-$1.34, 0.57]}
\defV{final/mistral32_24/gsi_placement_k1/conflict/acc}{0.52}
\defV{final/mistral32_24/gsi_placement_k1/conflict/c}{$-$0.43}
\defV{final/mistral32_24/gsi_placement_k1/conflict/ci}{[$-$0.48, 0.77]}
\defV{final/mistral32_24/gsi_placement_k1/conflict/d}{0.11}
\defV{final/mistral32_24/gsi_placement_k1/conflict/n}{46}
\defV{final/mistral32_24/gsi_placement_k1/conflict/unparsed}{0}
\defV{final/mistral32_24/gsi_placement_k1/control/acc}{0.59}
\defV{final/mistral32_24/gsi_placement_k1/control/c}{$-$0.63}
\defV{final/mistral32_24/gsi_placement_k1/control/d}{0.51}
\defV{final/mistral32_24/gsi_placement_k1/control/n}{46}
\defV{final/mistral32_24/gsi_placement_k1/control/unparsed}{0}
\defV{final/mistral32_24/gsi_placement_k1/diff}{$-$0.40}
\defV{final/mistral32_24/gsi_placement_k1/p}{0.66}
\defV{final/mistral32_24/gsi_placement_k1/pairs}{23}
\defV{final/mistral32_24/gsi_placement_k1/pholm}{--}
\defV{final/mistral32_24/gsi_placement_k1/status}{--}
\defV{final/mistral32_24/isobar/b}{79}
\defV{final/mistral32_24/isobar/c}{16}
\defV{final/mistral32_24/isobar/ci}{[$-$2.20, $-$1.23]}
\defV{final/mistral32_24/isobar/conflict/acc}{0.45}
\defV{final/mistral32_24/isobar/conflict/c}{$-$0.05}
\defV{final/mistral32_24/isobar/conflict/ci}{[$-$0.61, 0.10]}
\defV{final/mistral32_24/isobar/conflict/d}{$-$0.25}
\defV{final/mistral32_24/isobar/conflict/n}{200}
\defV{final/mistral32_24/isobar/conflict/unparsed}{0}
\defV{final/mistral32_24/isobar/control/acc}{0.77}
\defV{final/mistral32_24/isobar/control/c}{0.02}
\defV{final/mistral32_24/isobar/control/d}{1.43}
\defV{final/mistral32_24/isobar/control/n}{200}
\defV{final/mistral32_24/isobar/control/unparsed}{0}
\defV{final/mistral32_24/isobar/diff}{$-$1.68}
\defV{final/mistral32_24/isobar/p}{$3\times10^{-11}$}
\defV{final/mistral32_24/isobar/pairs}{100}
\defV{final/mistral32_24/isobar/pholm}{$5\times10^{-10}$}
\defV{final/mistral32_24/isobar/status}{yes}
\defV{final/mistral32_24/isobar_text/b}{3}
\defV{final/mistral32_24/isobar_text/c}{6}
\defV{final/mistral32_24/isobar_text/ci}{[$-$0.10, 0.29]}
\defV{final/mistral32_24/isobar_text/conflict/acc}{0.61}
\defV{final/mistral32_24/isobar_text/conflict/c}{$-$0.56}
\defV{final/mistral32_24/isobar_text/conflict/ci}{[0.43, 0.99]}
\defV{final/mistral32_24/isobar_text/conflict/d}{0.68}
\defV{final/mistral32_24/isobar_text/conflict/n}{200}
\defV{final/mistral32_24/isobar_text/conflict/unparsed}{0}
\defV{final/mistral32_24/isobar_text/control/acc}{0.60}
\defV{final/mistral32_24/isobar_text/control/c}{$-$0.60}
\defV{final/mistral32_24/isobar_text/control/d}{0.60}
\defV{final/mistral32_24/isobar_text/control/n}{200}
\defV{final/mistral32_24/isobar_text/control/unparsed}{0}
\defV{final/mistral32_24/isobar_text/diff}{+0.08}
\defV{final/mistral32_24/isobar_text/p}{0.51}
\defV{final/mistral32_24/isobar_text/pairs}{100}
\defV{final/mistral32_24/isobar_text/pholm}{--}
\defV{final/mistral32_24/isobar_text/status}{n.d.}
\defV{final/mistral32_24/linechart/b}{17}
\defV{final/mistral32_24/linechart/c}{4}
\defV{final/mistral32_24/linechart/ci}{[$-$0.80, $-$0.15]}
\defV{final/mistral32_24/linechart/conflict/acc}{0.51}
\defV{final/mistral32_24/linechart/conflict/c}{0.07}
\defV{final/mistral32_24/linechart/conflict/ci}{[$-$0.21, 0.36]}
\defV{final/mistral32_24/linechart/conflict/d}{0.07}
\defV{final/mistral32_24/linechart/conflict/n}{140}
\defV{final/mistral32_24/linechart/conflict/unparsed}{0}
\defV{final/mistral32_24/linechart/control/acc}{0.61}
\defV{final/mistral32_24/linechart/control/c}{0.05}
\defV{final/mistral32_24/linechart/control/d}{0.54}
\defV{final/mistral32_24/linechart/control/n}{140}
\defV{final/mistral32_24/linechart/control/unparsed}{0}
\defV{final/mistral32_24/linechart/diff}{$-$0.47}
\defV{final/mistral32_24/linechart/p}{0.007}
\defV{final/mistral32_24/linechart/pairs}{70}
\defV{final/mistral32_24/linechart/pholm}{--}
\defV{final/mistral32_24/linechart/status}{--}
\defV{final/mistral32_24/linechart_colour/b}{13}
\defV{final/mistral32_24/linechart_colour/c}{0}
\defV{final/mistral32_24/linechart_colour/ci}{[$-$0.78, $-$0.27]}
\defV{final/mistral32_24/linechart_colour/conflict/acc}{0.63}
\defV{final/mistral32_24/linechart_colour/conflict/c}{$-$0.00}
\defV{final/mistral32_24/linechart_colour/conflict/ci}{[0.32, 1.03]}
\defV{final/mistral32_24/linechart_colour/conflict/d}{0.65}
\defV{final/mistral32_24/linechart_colour/conflict/n}{140}
\defV{final/mistral32_24/linechart_colour/conflict/unparsed}{0}
\defV{final/mistral32_24/linechart_colour/control/acc}{0.72}
\defV{final/mistral32_24/linechart_colour/control/c}{$-$0.06}
\defV{final/mistral32_24/linechart_colour/control/d}{1.16}
\defV{final/mistral32_24/linechart_colour/control/n}{140}
\defV{final/mistral32_24/linechart_colour/control/unparsed}{0}
\defV{final/mistral32_24/linechart_colour/diff}{$-$0.51}
\defV{final/mistral32_24/linechart_colour/p}{$2\times10^{-4}$}
\defV{final/mistral32_24/linechart_colour/pairs}{70}
\defV{final/mistral32_24/linechart_colour/pholm}{--}
\defV{final/mistral32_24/linechart_colour/status}{--}
\defV{final/mistral32_24/linechart_text/b}{0}
\defV{final/mistral32_24/linechart_text/c}{0}
\defV{final/mistral32_24/linechart_text/ci}{[0.00, 0.00]}
\defV{final/mistral32_24/linechart_text/conflict/acc}{1.00}
\defV{final/mistral32_24/linechart_text/conflict/c}{$-$0.00}
\defV{final/mistral32_24/linechart_text/conflict/ci}{[4.91, 4.91]}
\defV{final/mistral32_24/linechart_text/conflict/d}{4.91}
\defV{final/mistral32_24/linechart_text/conflict/n}{140}
\defV{final/mistral32_24/linechart_text/conflict/unparsed}{0}
\defV{final/mistral32_24/linechart_text/control/acc}{1.00}
\defV{final/mistral32_24/linechart_text/control/c}{$-$0.00}
\defV{final/mistral32_24/linechart_text/control/d}{4.91}
\defV{final/mistral32_24/linechart_text/control/n}{140}
\defV{final/mistral32_24/linechart_text/control/unparsed}{0}
\defV{final/mistral32_24/linechart_text/diff}{+0.00}
\defV{final/mistral32_24/linechart_text/p}{1.00}
\defV{final/mistral32_24/linechart_text/pairs}{70}
\defV{final/mistral32_24/linechart_text/pholm}{--}
\defV{final/mistral32_24/linechart_text/status}{n.d.}
\defV{final/mistral32_24/map/index_labels/acc}{0.66}
\defV{final/mistral32_24/map/index_labels/d}{0.83}
\defV{final/mistral32_24/map/index_labels/n}{120}
\defV{final/mistral32_24/map/structure_only/acc}{0.52}
\defV{final/mistral32_24/map/structure_only/d}{0.15}
\defV{final/mistral32_24/map/structure_only/n}{120}
\defV{final/mistral32_24/map/values_only/acc}{0.72}
\defV{final/mistral32_24/map/values_only/d}{1.32}
\defV{final/mistral32_24/map/values_only/n}{120}
\defV{final/mistral32_24/map/values_structure/acc}{0.93}
\defV{final/mistral32_24/map/values_structure/d}{3.13}
\defV{final/mistral32_24/map/values_structure/n}{120}
\defV{final/mistral32_24/matched/b}{54}
\defV{final/mistral32_24/matched/c}{19}
\defV{final/mistral32_24/matched/ci}{[$-$2.09, $-$0.74]}
\defV{final/mistral32_24/matched/conflict/acc}{0.33}
\defV{final/mistral32_24/matched/conflict/c}{$-$0.45}
\defV{final/mistral32_24/matched/conflict/ci}{[$-$1.46, $-$0.53]}
\defV{final/mistral32_24/matched/conflict/d}{$-$0.97}
\defV{final/mistral32_24/matched/conflict/n}{140}
\defV{final/mistral32_24/matched/conflict/unparsed}{0}
\defV{final/mistral32_24/matched/control/acc}{0.58}
\defV{final/mistral32_24/matched/control/c}{$-$0.47}
\defV{final/mistral32_24/matched/control/d}{0.44}
\defV{final/mistral32_24/matched/control/n}{140}
\defV{final/mistral32_24/matched/control/unparsed}{0}
\defV{final/mistral32_24/matched/diff}{$-$1.40}
\defV{final/mistral32_24/matched/p}{$5\times10^{-5}$}
\defV{final/mistral32_24/matched/pairs}{70}
\defV{final/mistral32_24/matched/pholm}{$2\times10^{-4}$}
\defV{final/mistral32_24/matched/status}{n.d.}
\defV{final/mistral32_24/name}{Mistral-Small-3.2 24B}
\defV{final/mistral32_24/pointer_linechart/b}{11}
\defV{final/mistral32_24/pointer_linechart/c}{6}
\defV{final/mistral32_24/pointer_linechart/ci}{[$-$1.14, 0.37]}
\defV{final/mistral32_24/pointer_linechart/conflict/acc}{0.75}
\defV{final/mistral32_24/pointer_linechart/conflict/c}{0.14}
\defV{final/mistral32_24/pointer_linechart/conflict/ci}{[0.90, 2.16]}
\defV{final/mistral32_24/pointer_linechart/conflict/d}{1.45}
\defV{final/mistral32_24/pointer_linechart/conflict/n}{80}
\defV{final/mistral32_24/pointer_linechart/conflict/unparsed}{2}
\defV{final/mistral32_24/pointer_linechart/control/acc}{0.81}
\defV{final/mistral32_24/pointer_linechart/control/c}{0.18}
\defV{final/mistral32_24/pointer_linechart/control/d}{1.82}
\defV{final/mistral32_24/pointer_linechart/control/n}{80}
\defV{final/mistral32_24/pointer_linechart/control/unparsed}{1}
\defV{final/mistral32_24/pointer_linechart/diff}{$-$0.37}
\defV{final/mistral32_24/pointer_linechart/p}{0.33}
\defV{final/mistral32_24/pointer_linechart/pairs}{40}
\defV{final/mistral32_24/pointer_linechart/pholm}{--}
\defV{final/mistral32_24/pointer_linechart/status}{--}
\defV{final/mistral32_24/pointer_scatter/b}{6}
\defV{final/mistral32_24/pointer_scatter/c}{1}
\defV{final/mistral32_24/pointer_scatter/ci}{[$-$2.22, $-$0.14]}
\defV{final/mistral32_24/pointer_scatter/conflict/acc}{0.93}
\defV{final/mistral32_24/pointer_scatter/conflict/c}{$-$0.16}
\defV{final/mistral32_24/pointer_scatter/conflict/ci}{[2.20, 3.80]}
\defV{final/mistral32_24/pointer_scatter/conflict/d}{2.77}
\defV{final/mistral32_24/pointer_scatter/conflict/n}{80}
\defV{final/mistral32_24/pointer_scatter/conflict/unparsed}{0}
\defV{final/mistral32_24/pointer_scatter/control/acc}{0.99}
\defV{final/mistral32_24/pointer_scatter/control/c}{$-$0.23}
\defV{final/mistral32_24/pointer_scatter/control/d}{4.04}
\defV{final/mistral32_24/pointer_scatter/control/n}{80}
\defV{final/mistral32_24/pointer_scatter/control/unparsed}{0}
\defV{final/mistral32_24/pointer_scatter/diff}{$-$1.27}
\defV{final/mistral32_24/pointer_scatter/p}{0.12}
\defV{final/mistral32_24/pointer_scatter/pairs}{40}
\defV{final/mistral32_24/pointer_scatter/pholm}{--}
\defV{final/mistral32_24/pointer_scatter/status}{n.d.}
\defV{final/mistral32_24/public_data/b}{31}
\defV{final/mistral32_24/public_data/c}{9}
\defV{final/mistral32_24/public_data/ci}{[$-$2.43, $-$0.63]}
\defV{final/mistral32_24/public_data/conflict/acc}{0.47}
\defV{final/mistral32_24/public_data/conflict/c}{0.06}
\defV{final/mistral32_24/public_data/conflict/ci}{[$-$0.72, 0.43]}
\defV{final/mistral32_24/public_data/conflict/d}{$-$0.12}
\defV{final/mistral32_24/public_data/conflict/n}{80}
\defV{final/mistral32_24/public_data/conflict/unparsed}{0}
\defV{final/mistral32_24/public_data/control/acc}{0.75}
\defV{final/mistral32_24/public_data/control/c}{0.23}
\defV{final/mistral32_24/public_data/control/d}{1.35}
\defV{final/mistral32_24/public_data/control/n}{80}
\defV{final/mistral32_24/public_data/control/unparsed}{0}
\defV{final/mistral32_24/public_data/diff}{$-$1.47}
\defV{final/mistral32_24/public_data/p}{$7\times10^{-4}$}
\defV{final/mistral32_24/public_data/pairs}{40}
\defV{final/mistral32_24/public_data/pholm}{--}
\defV{final/mistral32_24/public_data/status}{--}
\defV{final/mistral32_24/repo}{mistralai/Mistral-Small-3.2-24B-Instruct-2506}
\defV{final/mistral32_24/revision}{95a6d26}
\defV{final/mistral32_24/scatter/b}{76}
\defV{final/mistral32_24/scatter/c}{4}
\defV{final/mistral32_24/scatter/ci}{[$-$2.78, $-$1.65]}
\defV{final/mistral32_24/scatter/conflict/acc}{0.52}
\defV{final/mistral32_24/scatter/conflict/c}{0.04}
\defV{final/mistral32_24/scatter/conflict/ci}{[$-$0.33, 0.51]}
\defV{final/mistral32_24/scatter/conflict/d}{0.09}
\defV{final/mistral32_24/scatter/conflict/n}{200}
\defV{final/mistral32_24/scatter/conflict/unparsed}{1}
\defV{final/mistral32_24/scatter/control/acc}{0.88}
\defV{final/mistral32_24/scatter/control/c}{$-$0.12}
\defV{final/mistral32_24/scatter/control/d}{2.28}
\defV{final/mistral32_24/scatter/control/n}{200}
\defV{final/mistral32_24/scatter/control/unparsed}{0}
\defV{final/mistral32_24/scatter/diff}{$-$2.19}
\defV{final/mistral32_24/scatter/p}{$3\times10^{-18}$}
\defV{final/mistral32_24/scatter/pairs}{100}
\defV{final/mistral32_24/scatter/pholm}{$6\times10^{-17}$}
\defV{final/mistral32_24/scatter/status}{yes}
\defV{final/mistral32_24/scatter_text/b}{2}
\defV{final/mistral32_24/scatter_text/c}{3}
\defV{final/mistral32_24/scatter_text/ci}{[$-$0.62, 0.83]}
\defV{final/mistral32_24/scatter_text/conflict/acc}{0.98}
\defV{final/mistral32_24/scatter_text/conflict/c}{$-$0.18}
\defV{final/mistral32_24/scatter_text/conflict/ci}{[3.48, 4.75]}
\defV{final/mistral32_24/scatter_text/conflict/d}{3.99}
\defV{final/mistral32_24/scatter_text/conflict/n}{200}
\defV{final/mistral32_24/scatter_text/conflict/unparsed}{0}
\defV{final/mistral32_24/scatter_text/control/acc}{0.97}
\defV{final/mistral32_24/scatter_text/control/c}{0.24}
\defV{final/mistral32_24/scatter_text/control/d}{3.87}
\defV{final/mistral32_24/scatter_text/control/n}{200}
\defV{final/mistral32_24/scatter_text/control/unparsed}{0}
\defV{final/mistral32_24/scatter_text/diff}{+0.12}
\defV{final/mistral32_24/scatter_text/p}{1.00}
\defV{final/mistral32_24/scatter_text/pairs}{100}
\defV{final/mistral32_24/scatter_text/pholm}{--}
\defV{final/mistral32_24/scatter_text/status}{n.d.}
\defV{final/mistral32_24/separation/b}{30}
\defV{final/mistral32_24/separation/c}{10}
\defV{final/mistral32_24/separation/ci}{[$-$1.65, $-$0.51]}
\defV{final/mistral32_24/separation/conflict/acc}{0.43}
\defV{final/mistral32_24/separation/conflict/c}{$-$0.85}
\defV{final/mistral32_24/separation/conflict/ci}{[$-$1.05, 0.00]}
\defV{final/mistral32_24/separation/conflict/d}{$-$0.51}
\defV{final/mistral32_24/separation/conflict/n}{140}
\defV{final/mistral32_24/separation/conflict/unparsed}{0}
\defV{final/mistral32_24/separation/control/acc}{0.57}
\defV{final/mistral32_24/separation/control/c}{$-$0.91}
\defV{final/mistral32_24/separation/control/d}{0.53}
\defV{final/mistral32_24/separation/control/n}{140}
\defV{final/mistral32_24/separation/control/unparsed}{0}
\defV{final/mistral32_24/separation/diff}{$-$1.04}
\defV{final/mistral32_24/separation/p}{0.002}
\defV{final/mistral32_24/separation/pairs}{70}
\defV{final/mistral32_24/separation/pholm}{0.009}
\defV{final/mistral32_24/separation/status}{yes}
\defV{final/mistral32_24/separation_text/b}{3}
\defV{final/mistral32_24/separation_text/c}{4}
\defV{final/mistral32_24/separation_text/ci}{[$-$0.42, 0.80]}
\defV{final/mistral32_24/separation_text/conflict/acc}{0.95}
\defV{final/mistral32_24/separation_text/conflict/c}{$-$0.14}
\defV{final/mistral32_24/separation_text/conflict/ci}{[2.78, 4.25]}
\defV{final/mistral32_24/separation_text/conflict/d}{3.33}
\defV{final/mistral32_24/separation_text/conflict/n}{140}
\defV{final/mistral32_24/separation_text/conflict/unparsed}{1}
\defV{final/mistral32_24/separation_text/control/acc}{0.94}
\defV{final/mistral32_24/separation_text/control/c}{$-$0.07}
\defV{final/mistral32_24/separation_text/control/d}{3.17}
\defV{final/mistral32_24/separation_text/control/n}{140}
\defV{final/mistral32_24/separation_text/control/unparsed}{1}
\defV{final/mistral32_24/separation_text/diff}{+0.16}
\defV{final/mistral32_24/separation_text/p}{1.00}
\defV{final/mistral32_24/separation_text/pairs}{70}
\defV{final/mistral32_24/separation_text/pholm}{--}
\defV{final/mistral32_24/separation_text/status}{n.d.}
\defV{final/mistral32_24/truncated}{2}
\defV{final/mistral32_24/truncated/separation_shortcut_free_text}{2}
\defV{final/mistral32_24/unparsed}{10}
\defV{final/mistral32_24/unparsed/pct}{0.2}
\defV{final/mistral32_24/white/b}{56}
\defV{final/mistral32_24/white/c}{13}
\defV{final/mistral32_24/white/ci}{[$-$2.56, $-$1.16]}
\defV{final/mistral32_24/white/conditional}{0.39}
\defV{final/mistral32_24/white/conflict/acc}{0.35}
\defV{final/mistral32_24/white/conflict/c}{$-$0.59}
\defV{final/mistral32_24/white/conflict/ci}{[$-$1.45, $-$0.41]}
\defV{final/mistral32_24/white/conflict/d}{$-$0.89}
\defV{final/mistral32_24/white/conflict/n}{140}
\defV{final/mistral32_24/white/conflict/unparsed}{1}
\defV{final/mistral32_24/white/control/acc}{0.66}
\defV{final/mistral32_24/white/control/c}{$-$0.39}
\defV{final/mistral32_24/white/control/d}{0.93}
\defV{final/mistral32_24/white/control/n}{140}
\defV{final/mistral32_24/white/control/unparsed}{2}
\defV{final/mistral32_24/white/diff}{$-$1.82}
\defV{final/mistral32_24/white/p}{$2\times10^{-7}$}
\defV{final/mistral32_24/white/pairs}{70}
\defV{final/mistral32_24/white/pholm}{$1\times10^{-6}$}
\defV{final/mistral32_24/white/reversal}{yes}
\defV{final/mistral32_24/white/status}{yes}
\defV{final/models/n}{10}
\defV{final/molmo2o_7/answers}{6048}
\defV{final/molmo2o_7/crop/b}{21}
\defV{final/molmo2o_7/crop/c}{11}
\defV{final/molmo2o_7/crop/ci}{[$-$1.36, $-$0.00]}
\defV{final/molmo2o_7/crop/conflict/acc}{0.44}
\defV{final/molmo2o_7/crop/conflict/c}{$-$1.07}
\defV{final/molmo2o_7/crop/conflict/ci}{[$-$1.06, 0.00]}
\defV{final/molmo2o_7/crop/conflict/d}{$-$0.50}
\defV{final/molmo2o_7/crop/conflict/n}{140}
\defV{final/molmo2o_7/crop/conflict/unparsed}{3}
\defV{final/molmo2o_7/crop/control/acc}{0.51}
\defV{final/molmo2o_7/crop/control/c}{$-$1.10}
\defV{final/molmo2o_7/crop/control/d}{0.14}
\defV{final/molmo2o_7/crop/control/n}{140}
\defV{final/molmo2o_7/crop/control/unparsed}{3}
\defV{final/molmo2o_7/crop/diff}{$-$0.64}
\defV{final/molmo2o_7/crop/p}{0.11}
\defV{final/molmo2o_7/crop/pairs}{70}
\defV{final/molmo2o_7/crop/pholm}{--}
\defV{final/molmo2o_7/crop/status}{n.d.}
\defV{final/molmo2o_7/cues_leader/b}{59}
\defV{final/molmo2o_7/cues_leader/c}{7}
\defV{final/molmo2o_7/cues_leader/ci}{[$-$2.80, $-$1.48]}
\defV{final/molmo2o_7/cues_leader/conflict/acc}{0.40}
\defV{final/molmo2o_7/cues_leader/conflict/c}{0.11}
\defV{final/molmo2o_7/cues_leader/conflict/ci}{[$-$0.93, $-$0.01]}
\defV{final/molmo2o_7/cues_leader/conflict/d}{$-$0.49}
\defV{final/molmo2o_7/cues_leader/conflict/n}{136}
\defV{final/molmo2o_7/cues_leader/conflict/unparsed}{2}
\defV{final/molmo2o_7/cues_leader/control/acc}{0.78}
\defV{final/molmo2o_7/cues_leader/control/c}{$-$0.15}
\defV{final/molmo2o_7/cues_leader/control/d}{1.61}
\defV{final/molmo2o_7/cues_leader/control/n}{136}
\defV{final/molmo2o_7/cues_leader/control/unparsed}{2}
\defV{final/molmo2o_7/cues_leader/diff}{$-$2.10}
\defV{final/molmo2o_7/cues_leader/p}{$2\times10^{-11}$}
\defV{final/molmo2o_7/cues_leader/pairs}{68}
\defV{final/molmo2o_7/cues_leader/pholm}{--}
\defV{final/molmo2o_7/cues_leader/status}{--}
\defV{final/molmo2o_7/cues_region/b}{62}
\defV{final/molmo2o_7/cues_region/c}{11}
\defV{final/molmo2o_7/cues_region/ci}{[$-$2.81, $-$1.31]}
\defV{final/molmo2o_7/cues_region/conflict/acc}{0.43}
\defV{final/molmo2o_7/cues_region/conflict/c}{$-$0.07}
\defV{final/molmo2o_7/cues_region/conflict/ci}{[$-$0.77, 0.19]}
\defV{final/molmo2o_7/cues_region/conflict/d}{$-$0.29}
\defV{final/molmo2o_7/cues_region/conflict/n}{140}
\defV{final/molmo2o_7/cues_region/conflict/unparsed}{4}
\defV{final/molmo2o_7/cues_region/control/acc}{0.79}
\defV{final/molmo2o_7/cues_region/control/c}{0.06}
\defV{final/molmo2o_7/cues_region/control/d}{1.73}
\defV{final/molmo2o_7/cues_region/control/n}{140}
\defV{final/molmo2o_7/cues_region/control/unparsed}{3}
\defV{final/molmo2o_7/cues_region/diff}{$-$2.02}
\defV{final/molmo2o_7/cues_region/p}{$9\times10^{-10}$}
\defV{final/molmo2o_7/cues_region/pairs}{70}
\defV{final/molmo2o_7/cues_region/pholm}{--}
\defV{final/molmo2o_7/cues_region/status}{--}
\defV{final/molmo2o_7/direct/b}{1}
\defV{final/molmo2o_7/direct/c}{0}
\defV{final/molmo2o_7/direct/ci}{[$-$0.80, 0.00]}
\defV{final/molmo2o_7/direct/conflict/acc}{0.49}
\defV{final/molmo2o_7/direct/conflict/c}{$-$2.24}
\defV{final/molmo2o_7/direct/conflict/ci}{[$-$0.80, 0.00]}
\defV{final/molmo2o_7/direct/conflict/d}{$-$0.42}
\defV{final/molmo2o_7/direct/conflict/n}{140}
\defV{final/molmo2o_7/direct/conflict/unparsed}{0}
\defV{final/molmo2o_7/direct/control/acc}{0.50}
\defV{final/molmo2o_7/direct/control/c}{$-$2.46}
\defV{final/molmo2o_7/direct/control/d}{0.00}
\defV{final/molmo2o_7/direct/control/n}{140}
\defV{final/molmo2o_7/direct/control/unparsed}{0}
\defV{final/molmo2o_7/direct/diff}{$-$0.42}
\defV{final/molmo2o_7/direct/p}{1.00}
\defV{final/molmo2o_7/direct/pairs}{70}
\defV{final/molmo2o_7/direct/pholm}{--}
\defV{final/molmo2o_7/direct/status}{n.d.}
\defV{final/molmo2o_7/failed}{0}
\defV{final/molmo2o_7/gate}{0.00}
\defV{final/molmo2o_7/gsi_placement/b}{3}
\defV{final/molmo2o_7/gsi_placement/c}{7}
\defV{final/molmo2o_7/gsi_placement/ci}{[$-$0.52, 1.57]}
\defV{final/molmo2o_7/gsi_placement/conflict/acc}{0.52}
\defV{final/molmo2o_7/gsi_placement/conflict/c}{$-$1.74}
\defV{final/molmo2o_7/gsi_placement/conflict/ci}{[$-$0.23, 1.22]}
\defV{final/molmo2o_7/gsi_placement/conflict/d}{0.51}
\defV{final/molmo2o_7/gsi_placement/conflict/n}{128}
\defV{final/molmo2o_7/gsi_placement/conflict/unparsed}{0}
\defV{final/molmo2o_7/gsi_placement/control/acc}{0.49}
\defV{final/molmo2o_7/gsi_placement/control/c}{$-$1.60}
\defV{final/molmo2o_7/gsi_placement/control/d}{$-$0.01}
\defV{final/molmo2o_7/gsi_placement/control/n}{128}
\defV{final/molmo2o_7/gsi_placement/control/unparsed}{1}
\defV{final/molmo2o_7/gsi_placement/diff}{+0.52}
\defV{final/molmo2o_7/gsi_placement/p}{0.34}
\defV{final/molmo2o_7/gsi_placement/pairs}{64}
\defV{final/molmo2o_7/gsi_placement/pholm}{--}
\defV{final/molmo2o_7/gsi_placement/status}{n.d.}
\defV{final/molmo2o_7/gsi_placement_k1/b}{1}
\defV{final/molmo2o_7/gsi_placement_k1/c}{2}
\defV{final/molmo2o_7/gsi_placement_k1/ci}{[$-$0.06, 0.06]}
\defV{final/molmo2o_7/gsi_placement_k1/conflict/acc}{0.48}
\defV{final/molmo2o_7/gsi_placement_k1/conflict/c}{$-$2.03}
\defV{final/molmo2o_7/gsi_placement_k1/conflict/ci}{[$-$0.06, 0.00]}
\defV{final/molmo2o_7/gsi_placement_k1/conflict/d}{$-$0.02}
\defV{final/molmo2o_7/gsi_placement_k1/conflict/n}{46}
\defV{final/molmo2o_7/gsi_placement_k1/conflict/unparsed}{1}
\defV{final/molmo2o_7/gsi_placement_k1/control/acc}{0.46}
\defV{final/molmo2o_7/gsi_placement_k1/control/c}{$-$2.01}
\defV{final/molmo2o_7/gsi_placement_k1/control/d}{$-$0.02}
\defV{final/molmo2o_7/gsi_placement_k1/control/n}{46}
\defV{final/molmo2o_7/gsi_placement_k1/control/unparsed}{3}
\defV{final/molmo2o_7/gsi_placement_k1/diff}{+0.00}
\defV{final/molmo2o_7/gsi_placement_k1/p}{1.00}
\defV{final/molmo2o_7/gsi_placement_k1/pairs}{23}
\defV{final/molmo2o_7/gsi_placement_k1/pholm}{--}
\defV{final/molmo2o_7/gsi_placement_k1/status}{n.d.}
\defV{final/molmo2o_7/isobar/b}{47}
\defV{final/molmo2o_7/isobar/c}{33}
\defV{final/molmo2o_7/isobar/ci}{[$-$0.90, 0.03]}
\defV{final/molmo2o_7/isobar/conflict/acc}{0.46}
\defV{final/molmo2o_7/isobar/conflict/c}{$-$0.52}
\defV{final/molmo2o_7/isobar/conflict/ci}{[$-$0.56, 0.09]}
\defV{final/molmo2o_7/isobar/conflict/d}{$-$0.23}
\defV{final/molmo2o_7/isobar/conflict/n}{200}
\defV{final/molmo2o_7/isobar/conflict/unparsed}{0}
\defV{final/molmo2o_7/isobar/control/acc}{0.53}
\defV{final/molmo2o_7/isobar/control/c}{$-$0.61}
\defV{final/molmo2o_7/isobar/control/d}{0.18}
\defV{final/molmo2o_7/isobar/control/n}{200}
\defV{final/molmo2o_7/isobar/control/unparsed}{0}
\defV{final/molmo2o_7/isobar/diff}{$-$0.41}
\defV{final/molmo2o_7/isobar/p}{0.15}
\defV{final/molmo2o_7/isobar/pairs}{100}
\defV{final/molmo2o_7/isobar/pholm}{--}
\defV{final/molmo2o_7/isobar/status}{n.d.}
\defV{final/molmo2o_7/isobar_text/b}{7}
\defV{final/molmo2o_7/isobar_text/c}{5}
\defV{final/molmo2o_7/isobar_text/ci}{[$-$0.23, 0.12]}
\defV{final/molmo2o_7/isobar_text/conflict/acc}{0.49}
\defV{final/molmo2o_7/isobar_text/conflict/c}{$-$0.20}
\defV{final/molmo2o_7/isobar_text/conflict/ci}{[$-$0.29, 0.18]}
\defV{final/molmo2o_7/isobar_text/conflict/d}{$-$0.05}
\defV{final/molmo2o_7/isobar_text/conflict/n}{200}
\defV{final/molmo2o_7/isobar_text/conflict/unparsed}{0}
\defV{final/molmo2o_7/isobar_text/control/acc}{0.50}
\defV{final/molmo2o_7/isobar_text/control/c}{$-$0.23}
\defV{final/molmo2o_7/isobar_text/control/d}{0.00}
\defV{final/molmo2o_7/isobar_text/control/n}{200}
\defV{final/molmo2o_7/isobar_text/control/unparsed}{0}
\defV{final/molmo2o_7/isobar_text/diff}{$-$0.05}
\defV{final/molmo2o_7/isobar_text/p}{0.77}
\defV{final/molmo2o_7/isobar_text/pairs}{100}
\defV{final/molmo2o_7/isobar_text/pholm}{--}
\defV{final/molmo2o_7/isobar_text/status}{n.d.}
\defV{final/molmo2o_7/linechart/b}{17}
\defV{final/molmo2o_7/linechart/c}{8}
\defV{final/molmo2o_7/linechart/ci}{[$-$0.78, 0.16]}
\defV{final/molmo2o_7/linechart/conflict/acc}{0.57}
\defV{final/molmo2o_7/linechart/conflict/c}{0.00}
\defV{final/molmo2o_7/linechart/conflict/ci}{[0.06, 1.13]}
\defV{final/molmo2o_7/linechart/conflict/d}{0.60}
\defV{final/molmo2o_7/linechart/conflict/n}{140}
\defV{final/molmo2o_7/linechart/conflict/unparsed}{11}
\defV{final/molmo2o_7/linechart/control/acc}{0.64}
\defV{final/molmo2o_7/linechart/control/c}{0.10}
\defV{final/molmo2o_7/linechart/control/d}{0.89}
\defV{final/molmo2o_7/linechart/control/n}{140}
\defV{final/molmo2o_7/linechart/control/unparsed}{8}
\defV{final/molmo2o_7/linechart/diff}{$-$0.29}
\defV{final/molmo2o_7/linechart/p}{0.11}
\defV{final/molmo2o_7/linechart/pairs}{70}
\defV{final/molmo2o_7/linechart/pholm}{--}
\defV{final/molmo2o_7/linechart/status}{--}
\defV{final/molmo2o_7/linechart_colour/b}{6}
\defV{final/molmo2o_7/linechart_colour/c}{5}
\defV{final/molmo2o_7/linechart_colour/ci}{[$-$1.31, 0.12]}
\defV{final/molmo2o_7/linechart_colour/conflict/acc}{0.88}
\defV{final/molmo2o_7/linechart_colour/conflict/c}{0.05}
\defV{final/molmo2o_7/linechart_colour/conflict/ci}{[1.94, 3.33]}
\defV{final/molmo2o_7/linechart_colour/conflict/d}{2.55}
\defV{final/molmo2o_7/linechart_colour/conflict/n}{140}
\defV{final/molmo2o_7/linechart_colour/conflict/unparsed}{4}
\defV{final/molmo2o_7/linechart_colour/control/acc}{0.89}
\defV{final/molmo2o_7/linechart_colour/control/c}{0.12}
\defV{final/molmo2o_7/linechart_colour/control/d}{3.01}
\defV{final/molmo2o_7/linechart_colour/control/n}{140}
\defV{final/molmo2o_7/linechart_colour/control/unparsed}{8}
\defV{final/molmo2o_7/linechart_colour/diff}{$-$0.47}
\defV{final/molmo2o_7/linechart_colour/p}{1.00}
\defV{final/molmo2o_7/linechart_colour/pairs}{70}
\defV{final/molmo2o_7/linechart_colour/pholm}{--}
\defV{final/molmo2o_7/linechart_colour/status}{--}
\defV{final/molmo2o_7/linechart_text/b}{0}
\defV{final/molmo2o_7/linechart_text/c}{0}
\defV{final/molmo2o_7/linechart_text/ci}{[0.00, 0.00]}
\defV{final/molmo2o_7/linechart_text/conflict/acc}{1.00}
\defV{final/molmo2o_7/linechart_text/conflict/c}{$-$0.00}
\defV{final/molmo2o_7/linechart_text/conflict/ci}{[4.91, 4.91]}
\defV{final/molmo2o_7/linechart_text/conflict/d}{4.91}
\defV{final/molmo2o_7/linechart_text/conflict/n}{140}
\defV{final/molmo2o_7/linechart_text/conflict/unparsed}{0}
\defV{final/molmo2o_7/linechart_text/control/acc}{1.00}
\defV{final/molmo2o_7/linechart_text/control/c}{$-$0.00}
\defV{final/molmo2o_7/linechart_text/control/d}{4.91}
\defV{final/molmo2o_7/linechart_text/control/n}{140}
\defV{final/molmo2o_7/linechart_text/control/unparsed}{0}
\defV{final/molmo2o_7/linechart_text/diff}{+0.00}
\defV{final/molmo2o_7/linechart_text/p}{1.00}
\defV{final/molmo2o_7/linechart_text/pairs}{70}
\defV{final/molmo2o_7/linechart_text/pholm}{--}
\defV{final/molmo2o_7/linechart_text/status}{n.d.}
\defV{final/molmo2o_7/map/index_labels/acc}{0.49}
\defV{final/molmo2o_7/map/index_labels/d}{$-$0.16}
\defV{final/molmo2o_7/map/index_labels/n}{120}
\defV{final/molmo2o_7/map/structure_only/acc}{0.49}
\defV{final/molmo2o_7/map/structure_only/d}{$-$0.09}
\defV{final/molmo2o_7/map/structure_only/n}{120}
\defV{final/molmo2o_7/map/values_only/acc}{0.59}
\defV{final/molmo2o_7/map/values_only/d}{0.80}
\defV{final/molmo2o_7/map/values_only/n}{120}
\defV{final/molmo2o_7/map/values_structure/acc}{0.84}
\defV{final/molmo2o_7/map/values_structure/d}{2.11}
\defV{final/molmo2o_7/map/values_structure/n}{120}
\defV{final/molmo2o_7/matched/b}{9}
\defV{final/molmo2o_7/matched/c}{7}
\defV{final/molmo2o_7/matched/ci}{[$-$0.91, 0.77]}
\defV{final/molmo2o_7/matched/conflict/acc}{0.49}
\defV{final/molmo2o_7/matched/conflict/c}{$-$1.57}
\defV{final/molmo2o_7/matched/conflict/ci}{[$-$0.37, 0.68]}
\defV{final/molmo2o_7/matched/conflict/d}{0.13}
\defV{final/molmo2o_7/matched/conflict/n}{140}
\defV{final/molmo2o_7/matched/conflict/unparsed}{6}
\defV{final/molmo2o_7/matched/control/acc}{0.50}
\defV{final/molmo2o_7/matched/control/c}{$-$1.53}
\defV{final/molmo2o_7/matched/control/d}{0.22}
\defV{final/molmo2o_7/matched/control/n}{140}
\defV{final/molmo2o_7/matched/control/unparsed}{2}
\defV{final/molmo2o_7/matched/diff}{$-$0.09}
\defV{final/molmo2o_7/matched/p}{0.80}
\defV{final/molmo2o_7/matched/pairs}{70}
\defV{final/molmo2o_7/matched/pholm}{--}
\defV{final/molmo2o_7/matched/status}{n.d.}
\defV{final/molmo2o_7/name}{Molmo2-O 7B}
\defV{final/molmo2o_7/pointer_linechart/b}{7}
\defV{final/molmo2o_7/pointer_linechart/c}{6}
\defV{final/molmo2o_7/pointer_linechart/ci}{[$-$0.60, 0.49]}
\defV{final/molmo2o_7/pointer_linechart/conflict/acc}{0.71}
\defV{final/molmo2o_7/pointer_linechart/conflict/c}{0.08}
\defV{final/molmo2o_7/pointer_linechart/conflict/ci}{[0.67, 1.79]}
\defV{final/molmo2o_7/pointer_linechart/conflict/d}{1.14}
\defV{final/molmo2o_7/pointer_linechart/conflict/n}{80}
\defV{final/molmo2o_7/pointer_linechart/conflict/unparsed}{1}
\defV{final/molmo2o_7/pointer_linechart/control/acc}{0.72}
\defV{final/molmo2o_7/pointer_linechart/control/c}{0.03}
\defV{final/molmo2o_7/pointer_linechart/control/d}{1.22}
\defV{final/molmo2o_7/pointer_linechart/control/n}{80}
\defV{final/molmo2o_7/pointer_linechart/control/unparsed}{1}
\defV{final/molmo2o_7/pointer_linechart/diff}{$-$0.07}
\defV{final/molmo2o_7/pointer_linechart/p}{1.00}
\defV{final/molmo2o_7/pointer_linechart/pairs}{40}
\defV{final/molmo2o_7/pointer_linechart/pholm}{--}
\defV{final/molmo2o_7/pointer_linechart/status}{--}
\defV{final/molmo2o_7/pointer_scatter/b}{19}
\defV{final/molmo2o_7/pointer_scatter/c}{8}
\defV{final/molmo2o_7/pointer_scatter/ci}{[$-$1.59, 0.06]}
\defV{final/molmo2o_7/pointer_scatter/conflict/acc}{0.69}
\defV{final/molmo2o_7/pointer_scatter/conflict/c}{$-$0.05}
\defV{final/molmo2o_7/pointer_scatter/conflict/ci}{[0.90, 2.33]}
\defV{final/molmo2o_7/pointer_scatter/conflict/d}{1.53}
\defV{final/molmo2o_7/pointer_scatter/conflict/n}{80}
\defV{final/molmo2o_7/pointer_scatter/conflict/unparsed}{10}
\defV{final/molmo2o_7/pointer_scatter/control/acc}{0.82}
\defV{final/molmo2o_7/pointer_scatter/control/c}{0.05}
\defV{final/molmo2o_7/pointer_scatter/control/d}{2.26}
\defV{final/molmo2o_7/pointer_scatter/control/n}{80}
\defV{final/molmo2o_7/pointer_scatter/control/unparsed}{5}
\defV{final/molmo2o_7/pointer_scatter/diff}{$-$0.73}
\defV{final/molmo2o_7/pointer_scatter/p}{0.05}
\defV{final/molmo2o_7/pointer_scatter/pairs}{40}
\defV{final/molmo2o_7/pointer_scatter/pholm}{--}
\defV{final/molmo2o_7/pointer_scatter/status}{--}
\defV{final/molmo2o_7/public_data/b}{17}
\defV{final/molmo2o_7/public_data/c}{19}
\defV{final/molmo2o_7/public_data/ci}{[$-$0.56, 0.76]}
\defV{final/molmo2o_7/public_data/conflict/acc}{0.50}
\defV{final/molmo2o_7/public_data/conflict/c}{$-$0.44}
\defV{final/molmo2o_7/public_data/conflict/ci}{[$-$0.46, 0.49]}
\defV{final/molmo2o_7/public_data/conflict/d}{0.00}
\defV{final/molmo2o_7/public_data/conflict/n}{80}
\defV{final/molmo2o_7/public_data/conflict/unparsed}{0}
\defV{final/molmo2o_7/public_data/control/acc}{0.47}
\defV{final/molmo2o_7/public_data/control/c}{$-$0.33}
\defV{final/molmo2o_7/public_data/control/d}{$-$0.09}
\defV{final/molmo2o_7/public_data/control/n}{80}
\defV{final/molmo2o_7/public_data/control/unparsed}{1}
\defV{final/molmo2o_7/public_data/diff}{+0.09}
\defV{final/molmo2o_7/public_data/p}{0.87}
\defV{final/molmo2o_7/public_data/pairs}{40}
\defV{final/molmo2o_7/public_data/pholm}{--}
\defV{final/molmo2o_7/public_data/status}{n.d.}
\defV{final/molmo2o_7/repo}{allenai/Molmo2-O-7B}
\defV{final/molmo2o_7/revision}{7844106}
\defV{final/molmo2o_7/scatter/b}{87}
\defV{final/molmo2o_7/scatter/c}{11}
\defV{final/molmo2o_7/scatter/ci}{[$-$2.55, $-$1.52]}
\defV{final/molmo2o_7/scatter/conflict/acc}{0.38}
\defV{final/molmo2o_7/scatter/conflict/c}{0.06}
\defV{final/molmo2o_7/scatter/conflict/ci}{[$-$0.96, $-$0.18]}
\defV{final/molmo2o_7/scatter/conflict/d}{$-$0.58}
\defV{final/molmo2o_7/scatter/conflict/n}{200}
\defV{final/molmo2o_7/scatter/conflict/unparsed}{3}
\defV{final/molmo2o_7/scatter/control/acc}{0.76}
\defV{final/molmo2o_7/scatter/control/c}{$-$0.18}
\defV{final/molmo2o_7/scatter/control/d}{1.44}
\defV{final/molmo2o_7/scatter/control/n}{200}
\defV{final/molmo2o_7/scatter/control/unparsed}{1}
\defV{final/molmo2o_7/scatter/diff}{$-$2.02}
\defV{final/molmo2o_7/scatter/p}{$8\times10^{-16}$}
\defV{final/molmo2o_7/scatter/pairs}{100}
\defV{final/molmo2o_7/scatter/pholm}{--}
\defV{final/molmo2o_7/scatter/status}{--}
\defV{final/molmo2o_7/scatter_text/b}{4}
\defV{final/molmo2o_7/scatter_text/c}{3}
\defV{final/molmo2o_7/scatter_text/ci}{[$-$0.18, 0.26]}
\defV{final/molmo2o_7/scatter_text/conflict/acc}{0.85}
\defV{final/molmo2o_7/scatter_text/conflict/c}{$-$0.04}
\defV{final/molmo2o_7/scatter_text/conflict/ci}{[1.91, 2.74]}
\defV{final/molmo2o_7/scatter_text/conflict/d}{2.28}
\defV{final/molmo2o_7/scatter_text/conflict/n}{200}
\defV{final/molmo2o_7/scatter_text/conflict/unparsed}{6}
\defV{final/molmo2o_7/scatter_text/control/acc}{0.85}
\defV{final/molmo2o_7/scatter_text/control/c}{$-$0.07}
\defV{final/molmo2o_7/scatter_text/control/d}{2.25}
\defV{final/molmo2o_7/scatter_text/control/n}{200}
\defV{final/molmo2o_7/scatter_text/control/unparsed}{4}
\defV{final/molmo2o_7/scatter_text/diff}{+0.03}
\defV{final/molmo2o_7/scatter_text/p}{1.00}
\defV{final/molmo2o_7/scatter_text/pairs}{100}
\defV{final/molmo2o_7/scatter_text/pholm}{--}
\defV{final/molmo2o_7/scatter_text/status}{n.d.}
\defV{final/molmo2o_7/separation/b}{17}
\defV{final/molmo2o_7/separation/c}{20}
\defV{final/molmo2o_7/separation/ci}{[$-$0.46, 0.74]}
\defV{final/molmo2o_7/separation/conflict/acc}{0.51}
\defV{final/molmo2o_7/separation/conflict/c}{$-$0.70}
\defV{final/molmo2o_7/separation/conflict/ci}{[$-$0.36, 0.52]}
\defV{final/molmo2o_7/separation/conflict/d}{0.06}
\defV{final/molmo2o_7/separation/conflict/n}{140}
\defV{final/molmo2o_7/separation/conflict/unparsed}{1}
\defV{final/molmo2o_7/separation/control/acc}{0.49}
\defV{final/molmo2o_7/separation/control/c}{$-$0.89}
\defV{final/molmo2o_7/separation/control/d}{$-$0.04}
\defV{final/molmo2o_7/separation/control/n}{140}
\defV{final/molmo2o_7/separation/control/unparsed}{3}
\defV{final/molmo2o_7/separation/diff}{+0.10}
\defV{final/molmo2o_7/separation/p}{0.74}
\defV{final/molmo2o_7/separation/pairs}{70}
\defV{final/molmo2o_7/separation/pholm}{--}
\defV{final/molmo2o_7/separation/status}{n.d.}
\defV{final/molmo2o_7/separation_text/b}{2}
\defV{final/molmo2o_7/separation_text/c}{1}
\defV{final/molmo2o_7/separation_text/ci}{[$-$0.87, 0.07]}
\defV{final/molmo2o_7/separation_text/conflict/acc}{0.76}
\defV{final/molmo2o_7/separation_text/conflict/c}{$-$0.96}
\defV{final/molmo2o_7/separation_text/conflict/ci}{[1.58, 2.82]}
\defV{final/molmo2o_7/separation_text/conflict/d}{2.14}
\defV{final/molmo2o_7/separation_text/conflict/n}{140}
\defV{final/molmo2o_7/separation_text/conflict/unparsed}{0}
\defV{final/molmo2o_7/separation_text/control/acc}{0.77}
\defV{final/molmo2o_7/separation_text/control/c}{$-$1.17}
\defV{final/molmo2o_7/separation_text/control/d}{2.56}
\defV{final/molmo2o_7/separation_text/control/n}{140}
\defV{final/molmo2o_7/separation_text/control/unparsed}{0}
\defV{final/molmo2o_7/separation_text/diff}{$-$0.42}
\defV{final/molmo2o_7/separation_text/p}{1.00}
\defV{final/molmo2o_7/separation_text/pairs}{70}
\defV{final/molmo2o_7/separation_text/pholm}{--}
\defV{final/molmo2o_7/separation_text/status}{n.d.}
\defV{final/molmo2o_7/truncated}{13}
\defV{final/molmo2o_7/truncated/label_placement_boxes}{1}
\defV{final/molmo2o_7/truncated/separation_shortcut_free}{1}
\defV{final/molmo2o_7/truncated/transfer_scatter}{1}
\defV{final/molmo2o_7/truncated/transfer_scatter_text}{10}
\defV{final/molmo2o_7/unparsed}{101}
\defV{final/molmo2o_7/unparsed/pct}{1.7}
\defV{final/molmo2o_7/white/b}{15}
\defV{final/molmo2o_7/white/c}{7}
\defV{final/molmo2o_7/white/ci}{[$-$2.06, $-$0.25]}
\defV{final/molmo2o_7/white/conditional}{0.80}
\defV{final/molmo2o_7/white/conflict/acc}{0.49}
\defV{final/molmo2o_7/white/conflict/c}{$-$1.33}
\defV{final/molmo2o_7/white/conflict/ci}{[$-$0.70, 0.34]}
\defV{final/molmo2o_7/white/conflict/d}{$-$0.16}
\defV{final/molmo2o_7/white/conflict/n}{140}
\defV{final/molmo2o_7/white/conflict/unparsed}{1}
\defV{final/molmo2o_7/white/control/acc}{0.54}
\defV{final/molmo2o_7/white/control/c}{$-$1.56}
\defV{final/molmo2o_7/white/control/d}{0.91}
\defV{final/molmo2o_7/white/control/n}{140}
\defV{final/molmo2o_7/white/control/unparsed}{2}
\defV{final/molmo2o_7/white/diff}{$-$1.08}
\defV{final/molmo2o_7/white/p}{0.13}
\defV{final/molmo2o_7/white/pairs}{70}
\defV{final/molmo2o_7/white/pholm}{0.27}
\defV{final/molmo2o_7/white/reversal}{no}
\defV{final/molmo2o_7/white/status}{no}
\defV{final/nat/acc/max}{0.62}
\defV{final/nat/acc/min}{0.25}
\defV{final/nat/harmful}{1}
\defV{final/nat/items}{120}
\defV{final/nat/lower}{9}
\defV{final/nat/models}{10}
\defV{final/package/torch}{2.13.0}
\defV{final/package/transformers}{5.17.0}
\defV{final/package/vllm}{0.31.0}
\defV{final/prev/neighbour}{0.056}
\defV{final/prev/neighbour/ci}{[0.049, 0.064]}
\defV{final/prev/other}{0.70}
\defV{final/prev/own}{6,277}
\defV{final/prev/points}{10,000}
\defV{final/prev/ratio}{2.1}
\defV{final/prev/share}{0.78}
\defV{final/prev/share/ci}{[0.76, 0.79]}
\defV{final/prev/windows}{400}
\defV{final/probe/crop/min}{0.94}
\defV{final/probe/far/hi}{2}
\defV{final/probe/far/max}{0.94}
\defV{final/probe/far/min}{0.76}
\defV{final/probe/gap}{$-$0.05}
\defV{final/probe/gemma4_12/drop}{0.03}
\defV{final/probe/gemma4_12/far}{0.81}
\defV{final/probe/gemma4_12/item/b}{7}
\defV{final/probe/gemma4_12/item/c}{1}
\defV{final/probe/gemma4_12/item/conflict}{0.48}
\defV{final/probe/gemma4_12/item/control}{0.60}
\defV{final/probe/gemma4_12/item/kept}{52}
\defV{final/probe/gemma4_12/near}{0.69}
\defV{final/probe/gemma4_31/drop}{0.44}
\defV{final/probe/gemma4_31/far}{0.94}
\defV{final/probe/gemma4_31/item/b}{53}
\defV{final/probe/gemma4_31/item/c}{0}
\defV{final/probe/gemma4_31/item/conflict}{0.30}
\defV{final/probe/gemma4_31/item/control}{0.74}
\defV{final/probe/gemma4_31/item/kept}{121}
\defV{final/probe/gemma4_31/near}{0.95}
\defV{final/probe/glm46v_9/drop}{0.23}
\defV{final/probe/glm46v_9/far}{0.94}
\defV{final/probe/glm46v_9/item/b}{37}
\defV{final/probe/glm46v_9/item/c}{8}
\defV{final/probe/glm46v_9/item/conflict}{0.30}
\defV{final/probe/glm46v_9/item/control}{0.59}
\defV{final/probe/glm46v_9/item/kept}{102}
\defV{final/probe/glm46v_9/near}{0.93}
\defV{final/probe/internvl35_38/drop}{0.16}
\defV{final/probe/internvl35_38/far}{0.84}
\defV{final/probe/internvl35_38/item/b}{21}
\defV{final/probe/internvl35_38/item/c}{8}
\defV{final/probe/internvl35_38/item/conflict}{0.40}
\defV{final/probe/internvl35_38/item/control}{0.58}
\defV{final/probe/internvl35_38/item/kept}{72}
\defV{final/probe/internvl35_38/near}{0.78}
\defV{final/probe/item/k}{7}
\defV{final/probe/item/kept/max}{121}
\defV{final/probe/item/kept/min}{52}
\defV{final/probe/item/kholm}{7}
\defV{final/probe/llavaov2_8/drop}{0.20}
\defV{final/probe/llavaov2_8/far}{0.76}
\defV{final/probe/llavaov2_8/item/b}{26}
\defV{final/probe/llavaov2_8/item/c}{7}
\defV{final/probe/llavaov2_8/item/conflict}{0.19}
\defV{final/probe/llavaov2_8/item/control}{0.48}
\defV{final/probe/llavaov2_8/item/kept}{67}
\defV{final/probe/llavaov2_8/near}{0.74}
\defV{final/probe/mistral32_24/drop}{0.31}
\defV{final/probe/mistral32_24/far}{0.81}
\defV{final/probe/mistral32_24/item/b}{31}
\defV{final/probe/mistral32_24/item/c}{2}
\defV{final/probe/mistral32_24/item/conflict}{0.30}
\defV{final/probe/mistral32_24/item/control}{0.77}
\defV{final/probe/mistral32_24/item/kept}{61}
\defV{final/probe/mistral32_24/near}{0.73}
\defV{final/probe/molmo2o_7/drop}{0.06}
\defV{final/probe/molmo2o_7/far}{0.88}
\defV{final/probe/molmo2o_7/item/b}{6}
\defV{final/probe/molmo2o_7/item/c}{3}
\defV{final/probe/molmo2o_7/item/conflict}{0.44}
\defV{final/probe/molmo2o_7/item/control}{0.48}
\defV{final/probe/molmo2o_7/item/kept}{77}
\defV{final/probe/molmo2o_7/near}{0.79}
\defV{final/probe/n}{140}
\defV{final/probe/qwen35_4/drop}{0.17}
\defV{final/probe/qwen35_4/far}{0.86}
\defV{final/probe/qwen35_4/item/b}{24}
\defV{final/probe/qwen35_4/item/c}{6}
\defV{final/probe/qwen35_4/item/conflict}{0.35}
\defV{final/probe/qwen35_4/item/control}{0.58}
\defV{final/probe/qwen35_4/item/kept}{77}
\defV{final/probe/qwen35_4/near}{0.82}
\defV{final/probe/qwen35_9/drop}{0.36}
\defV{final/probe/qwen35_9/far}{0.88}
\defV{final/probe/qwen35_9/item/b}{40}
\defV{final/probe/qwen35_9/item/c}{3}
\defV{final/probe/qwen35_9/item/conflict}{0.32}
\defV{final/probe/qwen35_9/item/control}{0.72}
\defV{final/probe/qwen35_9/item/kept}{92}
\defV{final/probe/qwen35_9/near}{0.84}
\defV{final/probe/qwen38_27/drop}{0.49}
\defV{final/probe/qwen38_27/far}{0.86}
\defV{final/probe/qwen38_27/item/b}{51}
\defV{final/probe/qwen38_27/item/c}{1}
\defV{final/probe/qwen38_27/item/conflict}{0.16}
\defV{final/probe/qwen38_27/item/control}{0.79}
\defV{final/probe/qwen38_27/item/kept}{80}
\defV{final/probe/qwen38_27/near}{0.79}
\defV{final/probe/vllm}{0.30.0}
\defV{final/probe/vllm/other}{Mistral-Small-3.2 24B}
\defV{final/python}{3.13.13}
\defV{final/qwen35_4/answers}{6048}
\defV{final/qwen35_4/crop/b}{74}
\defV{final/qwen35_4/crop/c}{2}
\defV{final/qwen35_4/crop/ci}{[$-$4.82, $-$3.21]}
\defV{final/qwen35_4/crop/conflict/acc}{0.27}
\defV{final/qwen35_4/crop/conflict/c}{$-$0.79}
\defV{final/qwen35_4/crop/conflict/ci}{[$-$2.20, $-$0.94]}
\defV{final/qwen35_4/crop/conflict/d}{$-$1.46}
\defV{final/qwen35_4/crop/conflict/n}{140}
\defV{final/qwen35_4/crop/conflict/unparsed}{6}
\defV{final/qwen35_4/crop/control/acc}{0.79}
\defV{final/qwen35_4/crop/control/c}{$-$0.82}
\defV{final/qwen35_4/crop/control/d}{2.41}
\defV{final/qwen35_4/crop/control/n}{140}
\defV{final/qwen35_4/crop/control/unparsed}{7}
\defV{final/qwen35_4/crop/diff}{$-$3.87}
\defV{final/qwen35_4/crop/p}{$8\times10^{-20}$}
\defV{final/qwen35_4/crop/pairs}{70}
\defV{final/qwen35_4/crop/pholm}{--}
\defV{final/qwen35_4/crop/status}{--}
\defV{final/qwen35_4/cues_leader/b}{63}
\defV{final/qwen35_4/cues_leader/c}{3}
\defV{final/qwen35_4/cues_leader/ci}{[$-$4.79, $-$2.53]}
\defV{final/qwen35_4/cues_leader/conflict/acc}{0.53}
\defV{final/qwen35_4/cues_leader/conflict/c}{$-$0.05}
\defV{final/qwen35_4/cues_leader/conflict/ci}{[$-$0.37, 0.74]}
\defV{final/qwen35_4/cues_leader/conflict/d}{0.17}
\defV{final/qwen35_4/cues_leader/conflict/n}{136}
\defV{final/qwen35_4/cues_leader/conflict/unparsed}{1}
\defV{final/qwen35_4/cues_leader/control/acc}{0.97}
\defV{final/qwen35_4/cues_leader/control/c}{0.19}
\defV{final/qwen35_4/cues_leader/control/d}{3.66}
\defV{final/qwen35_4/cues_leader/control/n}{136}
\defV{final/qwen35_4/cues_leader/control/unparsed}{0}
\defV{final/qwen35_4/cues_leader/diff}{$-$3.49}
\defV{final/qwen35_4/cues_leader/p}{$1\times10^{-15}$}
\defV{final/qwen35_4/cues_leader/pairs}{68}
\defV{final/qwen35_4/cues_leader/pholm}{--}
\defV{final/qwen35_4/cues_leader/status}{--}
\defV{final/qwen35_4/cues_region/b}{72}
\defV{final/qwen35_4/cues_region/c}{6}
\defV{final/qwen35_4/cues_region/ci}{[$-$4.30, $-$2.19]}
\defV{final/qwen35_4/cues_region/conflict/acc}{0.46}
\defV{final/qwen35_4/cues_region/conflict/c}{0.03}
\defV{final/qwen35_4/cues_region/conflict/ci}{[$-$0.69, 0.38]}
\defV{final/qwen35_4/cues_region/conflict/d}{$-$0.16}
\defV{final/qwen35_4/cues_region/conflict/n}{140}
\defV{final/qwen35_4/cues_region/conflict/unparsed}{1}
\defV{final/qwen35_4/cues_region/control/acc}{0.94}
\defV{final/qwen35_4/cues_region/control/c}{0.05}
\defV{final/qwen35_4/cues_region/control/d}{2.95}
\defV{final/qwen35_4/cues_region/control/n}{140}
\defV{final/qwen35_4/cues_region/control/unparsed}{0}
\defV{final/qwen35_4/cues_region/diff}{$-$3.11}
\defV{final/qwen35_4/cues_region/p}{$2\times10^{-15}$}
\defV{final/qwen35_4/cues_region/pairs}{70}
\defV{final/qwen35_4/cues_region/pholm}{--}
\defV{final/qwen35_4/cues_region/status}{--}
\defV{final/qwen35_4/direct/b}{10}
\defV{final/qwen35_4/direct/c}{7}
\defV{final/qwen35_4/direct/ci}{[$-$1.08, 0.13]}
\defV{final/qwen35_4/direct/conflict/acc}{0.47}
\defV{final/qwen35_4/direct/conflict/c}{$-$1.42}
\defV{final/qwen35_4/direct/conflict/ci}{[$-$0.70, 0.20]}
\defV{final/qwen35_4/direct/conflict/d}{$-$0.20}
\defV{final/qwen35_4/direct/conflict/n}{140}
\defV{final/qwen35_4/direct/conflict/unparsed}{3}
\defV{final/qwen35_4/direct/control/acc}{0.49}
\defV{final/qwen35_4/direct/control/c}{$-$1.31}
\defV{final/qwen35_4/direct/control/d}{0.17}
\defV{final/qwen35_4/direct/control/n}{140}
\defV{final/qwen35_4/direct/control/unparsed}{6}
\defV{final/qwen35_4/direct/diff}{$-$0.38}
\defV{final/qwen35_4/direct/p}{0.63}
\defV{final/qwen35_4/direct/pairs}{70}
\defV{final/qwen35_4/direct/pholm}{--}
\defV{final/qwen35_4/direct/status}{n.d.}
\defV{final/qwen35_4/failed}{0}
\defV{final/qwen35_4/gate}{0.67}
\defV{final/qwen35_4/gsi_placement/b}{46}
\defV{final/qwen35_4/gsi_placement/c}{17}
\defV{final/qwen35_4/gsi_placement/ci}{[$-$1.93, $-$0.55]}
\defV{final/qwen35_4/gsi_placement/conflict/acc}{0.37}
\defV{final/qwen35_4/gsi_placement/conflict/c}{$-$0.56}
\defV{final/qwen35_4/gsi_placement/conflict/ci}{[$-$0.95, 0.04]}
\defV{final/qwen35_4/gsi_placement/conflict/d}{$-$0.45}
\defV{final/qwen35_4/gsi_placement/conflict/n}{128}
\defV{final/qwen35_4/gsi_placement/conflict/unparsed}{16}
\defV{final/qwen35_4/gsi_placement/control/acc}{0.59}
\defV{final/qwen35_4/gsi_placement/control/c}{$-$0.34}
\defV{final/qwen35_4/gsi_placement/control/d}{0.77}
\defV{final/qwen35_4/gsi_placement/control/n}{128}
\defV{final/qwen35_4/gsi_placement/control/unparsed}{11}
\defV{final/qwen35_4/gsi_placement/diff}{$-$1.22}
\defV{final/qwen35_4/gsi_placement/p}{$3\times10^{-4}$}
\defV{final/qwen35_4/gsi_placement/pairs}{64}
\defV{final/qwen35_4/gsi_placement/pholm}{0.002}
\defV{final/qwen35_4/gsi_placement/status}{yes}
\defV{final/qwen35_4/gsi_placement_k1/b}{17}
\defV{final/qwen35_4/gsi_placement_k1/c}{2}
\defV{final/qwen35_4/gsi_placement_k1/ci}{[$-$3.05, $-$0.70]}
\defV{final/qwen35_4/gsi_placement_k1/conflict/acc}{0.24}
\defV{final/qwen35_4/gsi_placement_k1/conflict/c}{$-$0.66}
\defV{final/qwen35_4/gsi_placement_k1/conflict/ci}{[$-$2.15, $-$0.13]}
\defV{final/qwen35_4/gsi_placement_k1/conflict/d}{$-$1.03}
\defV{final/qwen35_4/gsi_placement_k1/conflict/n}{46}
\defV{final/qwen35_4/gsi_placement_k1/conflict/unparsed}{10}
\defV{final/qwen35_4/gsi_placement_k1/control/acc}{0.57}
\defV{final/qwen35_4/gsi_placement_k1/control/c}{$-$0.50}
\defV{final/qwen35_4/gsi_placement_k1/control/d}{0.65}
\defV{final/qwen35_4/gsi_placement_k1/control/n}{46}
\defV{final/qwen35_4/gsi_placement_k1/control/unparsed}{4}
\defV{final/qwen35_4/gsi_placement_k1/diff}{$-$1.69}
\defV{final/qwen35_4/gsi_placement_k1/p}{$7\times10^{-4}$}
\defV{final/qwen35_4/gsi_placement_k1/pairs}{23}
\defV{final/qwen35_4/gsi_placement_k1/pholm}{--}
\defV{final/qwen35_4/gsi_placement_k1/status}{--}
\defV{final/qwen35_4/isobar/b}{46}
\defV{final/qwen35_4/isobar/c}{32}
\defV{final/qwen35_4/isobar/ci}{[$-$0.86, 0.11]}
\defV{final/qwen35_4/isobar/conflict/acc}{0.54}
\defV{final/qwen35_4/isobar/conflict/c}{0.02}
\defV{final/qwen35_4/isobar/conflict/ci}{[$-$0.15, 0.58]}
\defV{final/qwen35_4/isobar/conflict/d}{0.20}
\defV{final/qwen35_4/isobar/conflict/n}{200}
\defV{final/qwen35_4/isobar/conflict/unparsed}{0}
\defV{final/qwen35_4/isobar/control/acc}{0.61}
\defV{final/qwen35_4/isobar/control/c}{$-$0.04}
\defV{final/qwen35_4/isobar/control/d}{0.57}
\defV{final/qwen35_4/isobar/control/n}{200}
\defV{final/qwen35_4/isobar/control/unparsed}{1}
\defV{final/qwen35_4/isobar/diff}{$-$0.37}
\defV{final/qwen35_4/isobar/p}{0.14}
\defV{final/qwen35_4/isobar/pairs}{100}
\defV{final/qwen35_4/isobar/pholm}{0.14}
\defV{final/qwen35_4/isobar/status}{no}
\defV{final/qwen35_4/isobar_text/b}{8}
\defV{final/qwen35_4/isobar_text/c}{13}
\defV{final/qwen35_4/isobar_text/ci}{[$-$0.19, 0.62]}
\defV{final/qwen35_4/isobar_text/conflict/acc}{0.86}
\defV{final/qwen35_4/isobar_text/conflict/c}{0.02}
\defV{final/qwen35_4/isobar_text/conflict/ci}{[1.81, 2.65]}
\defV{final/qwen35_4/isobar_text/conflict/d}{2.17}
\defV{final/qwen35_4/isobar_text/conflict/n}{200}
\defV{final/qwen35_4/isobar_text/conflict/unparsed}{0}
\defV{final/qwen35_4/isobar_text/control/acc}{0.84}
\defV{final/qwen35_4/isobar_text/control/c}{$-$0.00}
\defV{final/qwen35_4/isobar_text/control/d}{1.96}
\defV{final/qwen35_4/isobar_text/control/n}{200}
\defV{final/qwen35_4/isobar_text/control/unparsed}{0}
\defV{final/qwen35_4/isobar_text/diff}{+0.21}
\defV{final/qwen35_4/isobar_text/p}{0.38}
\defV{final/qwen35_4/isobar_text/pairs}{100}
\defV{final/qwen35_4/isobar_text/pholm}{--}
\defV{final/qwen35_4/isobar_text/status}{--}
\defV{final/qwen35_4/linechart/b}{18}
\defV{final/qwen35_4/linechart/c}{15}
\defV{final/qwen35_4/linechart/ci}{[$-$0.73, 0.58]}
\defV{final/qwen35_4/linechart/conflict/acc}{0.76}
\defV{final/qwen35_4/linechart/conflict/c}{$-$0.16}
\defV{final/qwen35_4/linechart/conflict/ci}{[0.88, 2.05]}
\defV{final/qwen35_4/linechart/conflict/d}{1.42}
\defV{final/qwen35_4/linechart/conflict/n}{140}
\defV{final/qwen35_4/linechart/conflict/unparsed}{1}
\defV{final/qwen35_4/linechart/control/acc}{0.78}
\defV{final/qwen35_4/linechart/control/c}{0.07}
\defV{final/qwen35_4/linechart/control/d}{1.51}
\defV{final/qwen35_4/linechart/control/n}{140}
\defV{final/qwen35_4/linechart/control/unparsed}{0}
\defV{final/qwen35_4/linechart/diff}{$-$0.09}
\defV{final/qwen35_4/linechart/p}{0.73}
\defV{final/qwen35_4/linechart/pairs}{70}
\defV{final/qwen35_4/linechart/pholm}{--}
\defV{final/qwen35_4/linechart/status}{--}
\defV{final/qwen35_4/linechart_colour/b}{6}
\defV{final/qwen35_4/linechart_colour/c}{3}
\defV{final/qwen35_4/linechart_colour/ci}{[$-$1.28, 0.96]}
\defV{final/qwen35_4/linechart_colour/conflict/acc}{0.95}
\defV{final/qwen35_4/linechart_colour/conflict/c}{0.60}
\defV{final/qwen35_4/linechart_colour/conflict/ci}{[3.39, 4.11]}
\defV{final/qwen35_4/linechart_colour/conflict/d}{3.71}
\defV{final/qwen35_4/linechart_colour/conflict/n}{140}
\defV{final/qwen35_4/linechart_colour/conflict/unparsed}{0}
\defV{final/qwen35_4/linechart_colour/control/acc}{0.97}
\defV{final/qwen35_4/linechart_colour/control/c}{$-$0.00}
\defV{final/qwen35_4/linechart_colour/control/d}{3.62}
\defV{final/qwen35_4/linechart_colour/control/n}{140}
\defV{final/qwen35_4/linechart_colour/control/unparsed}{0}
\defV{final/qwen35_4/linechart_colour/diff}{+0.09}
\defV{final/qwen35_4/linechart_colour/p}{0.51}
\defV{final/qwen35_4/linechart_colour/pairs}{70}
\defV{final/qwen35_4/linechart_colour/pholm}{--}
\defV{final/qwen35_4/linechart_colour/status}{n.d.}
\defV{final/qwen35_4/linechart_text/b}{0}
\defV{final/qwen35_4/linechart_text/c}{0}
\defV{final/qwen35_4/linechart_text/ci}{[0.00, 0.00]}
\defV{final/qwen35_4/linechart_text/conflict/acc}{1.00}
\defV{final/qwen35_4/linechart_text/conflict/c}{$-$0.00}
\defV{final/qwen35_4/linechart_text/conflict/ci}{[4.91, 4.91]}
\defV{final/qwen35_4/linechart_text/conflict/d}{4.91}
\defV{final/qwen35_4/linechart_text/conflict/n}{140}
\defV{final/qwen35_4/linechart_text/conflict/unparsed}{0}
\defV{final/qwen35_4/linechart_text/control/acc}{1.00}
\defV{final/qwen35_4/linechart_text/control/c}{$-$0.00}
\defV{final/qwen35_4/linechart_text/control/d}{4.91}
\defV{final/qwen35_4/linechart_text/control/n}{140}
\defV{final/qwen35_4/linechart_text/control/unparsed}{0}
\defV{final/qwen35_4/linechart_text/diff}{+0.00}
\defV{final/qwen35_4/linechart_text/p}{1.00}
\defV{final/qwen35_4/linechart_text/pairs}{70}
\defV{final/qwen35_4/linechart_text/pholm}{--}
\defV{final/qwen35_4/linechart_text/status}{n.d.}
\defV{final/qwen35_4/map/index_labels/acc}{0.64}
\defV{final/qwen35_4/map/index_labels/d}{0.95}
\defV{final/qwen35_4/map/index_labels/n}{120}
\defV{final/qwen35_4/map/structure_only/acc}{0.60}
\defV{final/qwen35_4/map/structure_only/d}{1.02}
\defV{final/qwen35_4/map/structure_only/n}{120}
\defV{final/qwen35_4/map/values_only/acc}{0.61}
\defV{final/qwen35_4/map/values_only/d}{0.82}
\defV{final/qwen35_4/map/values_only/n}{120}
\defV{final/qwen35_4/map/values_structure/acc}{0.88}
\defV{final/qwen35_4/map/values_structure/d}{2.39}
\defV{final/qwen35_4/map/values_structure/n}{120}
\defV{final/qwen35_4/matched/b}{34}
\defV{final/qwen35_4/matched/c}{13}
\defV{final/qwen35_4/matched/ci}{[$-$1.71, $-$0.22]}
\defV{final/qwen35_4/matched/conflict/acc}{0.35}
\defV{final/qwen35_4/matched/conflict/c}{$-$0.63}
\defV{final/qwen35_4/matched/conflict/ci}{[$-$1.14, $-$0.22]}
\defV{final/qwen35_4/matched/conflict/d}{$-$0.65}
\defV{final/qwen35_4/matched/conflict/n}{140}
\defV{final/qwen35_4/matched/conflict/unparsed}{12}
\defV{final/qwen35_4/matched/control/acc}{0.50}
\defV{final/qwen35_4/matched/control/c}{$-$0.99}
\defV{final/qwen35_4/matched/control/d}{0.31}
\defV{final/qwen35_4/matched/control/n}{140}
\defV{final/qwen35_4/matched/control/unparsed}{17}
\defV{final/qwen35_4/matched/diff}{$-$0.96}
\defV{final/qwen35_4/matched/p}{0.003}
\defV{final/qwen35_4/matched/pairs}{70}
\defV{final/qwen35_4/matched/pholm}{0.003}
\defV{final/qwen35_4/matched/status}{n.d.}
\defV{final/qwen35_4/name}{Qwen3.5 4B}
\defV{final/qwen35_4/pointer_linechart/b}{6}
\defV{final/qwen35_4/pointer_linechart/c}{5}
\defV{final/qwen35_4/pointer_linechart/ci}{[$-$1.56, 0.66]}
\defV{final/qwen35_4/pointer_linechart/conflict/acc}{0.88}
\defV{final/qwen35_4/pointer_linechart/conflict/c}{$-$0.07}
\defV{final/qwen35_4/pointer_linechart/conflict/ci}{[1.69, 3.34]}
\defV{final/qwen35_4/pointer_linechart/conflict/d}{2.32}
\defV{final/qwen35_4/pointer_linechart/conflict/n}{80}
\defV{final/qwen35_4/pointer_linechart/conflict/unparsed}{1}
\defV{final/qwen35_4/pointer_linechart/control/acc}{0.89}
\defV{final/qwen35_4/pointer_linechart/control/c}{$-$0.40}
\defV{final/qwen35_4/pointer_linechart/control/d}{2.76}
\defV{final/qwen35_4/pointer_linechart/control/n}{80}
\defV{final/qwen35_4/pointer_linechart/control/unparsed}{2}
\defV{final/qwen35_4/pointer_linechart/diff}{$-$0.45}
\defV{final/qwen35_4/pointer_linechart/p}{1.00}
\defV{final/qwen35_4/pointer_linechart/pairs}{40}
\defV{final/qwen35_4/pointer_linechart/pholm}{--}
\defV{final/qwen35_4/pointer_linechart/status}{--}
\defV{final/qwen35_4/pointer_scatter/b}{10}
\defV{final/qwen35_4/pointer_scatter/c}{1}
\defV{final/qwen35_4/pointer_scatter/ci}{[$-$2.78, $-$0.64]}
\defV{final/qwen35_4/pointer_scatter/conflict/acc}{0.88}
\defV{final/qwen35_4/pointer_scatter/conflict/c}{0.07}
\defV{final/qwen35_4/pointer_scatter/conflict/ci}{[1.59, 3.34]}
\defV{final/qwen35_4/pointer_scatter/conflict/d}{2.32}
\defV{final/qwen35_4/pointer_scatter/conflict/n}{80}
\defV{final/qwen35_4/pointer_scatter/conflict/unparsed}{1}
\defV{final/qwen35_4/pointer_scatter/control/acc}{0.99}
\defV{final/qwen35_4/pointer_scatter/control/c}{0.23}
\defV{final/qwen35_4/pointer_scatter/control/d}{4.04}
\defV{final/qwen35_4/pointer_scatter/control/n}{80}
\defV{final/qwen35_4/pointer_scatter/control/unparsed}{0}
\defV{final/qwen35_4/pointer_scatter/diff}{$-$1.72}
\defV{final/qwen35_4/pointer_scatter/p}{0.01}
\defV{final/qwen35_4/pointer_scatter/pairs}{40}
\defV{final/qwen35_4/pointer_scatter/pholm}{--}
\defV{final/qwen35_4/pointer_scatter/status}{--}
\defV{final/qwen35_4/public_data/b}{20}
\defV{final/qwen35_4/public_data/c}{11}
\defV{final/qwen35_4/public_data/ci}{[$-$1.39, 0.15]}
\defV{final/qwen35_4/public_data/conflict/acc}{0.54}
\defV{final/qwen35_4/public_data/conflict/c}{$-$0.15}
\defV{final/qwen35_4/public_data/conflict/ci}{[$-$0.33, 0.70]}
\defV{final/qwen35_4/public_data/conflict/d}{0.19}
\defV{final/qwen35_4/public_data/conflict/n}{80}
\defV{final/qwen35_4/public_data/conflict/unparsed}{0}
\defV{final/qwen35_4/public_data/control/acc}{0.65}
\defV{final/qwen35_4/public_data/control/c}{$-$0.05}
\defV{final/qwen35_4/public_data/control/d}{0.79}
\defV{final/qwen35_4/public_data/control/n}{80}
\defV{final/qwen35_4/public_data/control/unparsed}{1}
\defV{final/qwen35_4/public_data/diff}{$-$0.61}
\defV{final/qwen35_4/public_data/p}{0.15}
\defV{final/qwen35_4/public_data/pairs}{40}
\defV{final/qwen35_4/public_data/pholm}{--}
\defV{final/qwen35_4/public_data/status}{--}
\defV{final/qwen35_4/repo}{Qwen/Qwen3.5-4B}
\defV{final/qwen35_4/revision}{851bf6e}
\defV{final/qwen35_4/scatter/b}{73}
\defV{final/qwen35_4/scatter/c}{2}
\defV{final/qwen35_4/scatter/ci}{[$-$3.92, $-$2.21]}
\defV{final/qwen35_4/scatter/conflict/acc}{0.60}
\defV{final/qwen35_4/scatter/conflict/c}{$-$0.01}
\defV{final/qwen35_4/scatter/conflict/ci}{[0.07, 0.98]}
\defV{final/qwen35_4/scatter/conflict/d}{0.53}
\defV{final/qwen35_4/scatter/conflict/n}{200}
\defV{final/qwen35_4/scatter/conflict/unparsed}{0}
\defV{final/qwen35_4/scatter/control/acc}{0.96}
\defV{final/qwen35_4/scatter/control/c}{0.00}
\defV{final/qwen35_4/scatter/control/d}{3.40}
\defV{final/qwen35_4/scatter/control/n}{200}
\defV{final/qwen35_4/scatter/control/unparsed}{0}
\defV{final/qwen35_4/scatter/diff}{$-$2.87}
\defV{final/qwen35_4/scatter/p}{$2\times10^{-19}$}
\defV{final/qwen35_4/scatter/pairs}{100}
\defV{final/qwen35_4/scatter/pholm}{$3\times10^{-18}$}
\defV{final/qwen35_4/scatter/status}{yes}
\defV{final/qwen35_4/scatter_text/b}{0}
\defV{final/qwen35_4/scatter_text/c}{0}
\defV{final/qwen35_4/scatter_text/ci}{[0.00, 0.00]}
\defV{final/qwen35_4/scatter_text/conflict/acc}{1.00}
\defV{final/qwen35_4/scatter_text/conflict/c}{0.00}
\defV{final/qwen35_4/scatter_text/conflict/ci}{[5.16, 5.16]}
\defV{final/qwen35_4/scatter_text/conflict/d}{5.16}
\defV{final/qwen35_4/scatter_text/conflict/n}{200}
\defV{final/qwen35_4/scatter_text/conflict/unparsed}{0}
\defV{final/qwen35_4/scatter_text/control/acc}{1.00}
\defV{final/qwen35_4/scatter_text/control/c}{0.00}
\defV{final/qwen35_4/scatter_text/control/d}{5.16}
\defV{final/qwen35_4/scatter_text/control/n}{200}
\defV{final/qwen35_4/scatter_text/control/unparsed}{0}
\defV{final/qwen35_4/scatter_text/diff}{+0.00}
\defV{final/qwen35_4/scatter_text/p}{1.00}
\defV{final/qwen35_4/scatter_text/pairs}{100}
\defV{final/qwen35_4/scatter_text/pholm}{--}
\defV{final/qwen35_4/scatter_text/status}{n.d.}
\defV{final/qwen35_4/separation/b}{43}
\defV{final/qwen35_4/separation/c}{23}
\defV{final/qwen35_4/separation/ci}{[$-$1.54, $-$0.35]}
\defV{final/qwen35_4/separation/conflict/acc}{0.44}
\defV{final/qwen35_4/separation/conflict/c}{$-$0.22}
\defV{final/qwen35_4/separation/conflict/ci}{[$-$0.55, 0.24]}
\defV{final/qwen35_4/separation/conflict/d}{$-$0.16}
\defV{final/qwen35_4/separation/conflict/n}{140}
\defV{final/qwen35_4/separation/conflict/unparsed}{9}
\defV{final/qwen35_4/separation/control/acc}{0.58}
\defV{final/qwen35_4/separation/control/c}{$-$0.26}
\defV{final/qwen35_4/separation/control/d}{0.75}
\defV{final/qwen35_4/separation/control/n}{140}
\defV{final/qwen35_4/separation/control/unparsed}{14}
\defV{final/qwen35_4/separation/diff}{$-$0.91}
\defV{final/qwen35_4/separation/p}{0.02}
\defV{final/qwen35_4/separation/pairs}{70}
\defV{final/qwen35_4/separation/pholm}{0.04}
\defV{final/qwen35_4/separation/status}{no}
\defV{final/qwen35_4/separation_text/b}{9}
\defV{final/qwen35_4/separation_text/c}{10}
\defV{final/qwen35_4/separation_text/ci}{[$-$0.41, 1.14]}
\defV{final/qwen35_4/separation_text/conflict/acc}{0.92}
\defV{final/qwen35_4/separation_text/conflict/c}{0.11}
\defV{final/qwen35_4/separation_text/conflict/ci}{[3.26, 4.86]}
\defV{final/qwen35_4/separation_text/conflict/d}{3.79}
\defV{final/qwen35_4/separation_text/conflict/n}{140}
\defV{final/qwen35_4/separation_text/conflict/unparsed}{8}
\defV{final/qwen35_4/separation_text/control/acc}{0.91}
\defV{final/qwen35_4/separation_text/control/c}{0.28}
\defV{final/qwen35_4/separation_text/control/d}{3.50}
\defV{final/qwen35_4/separation_text/control/n}{140}
\defV{final/qwen35_4/separation_text/control/unparsed}{7}
\defV{final/qwen35_4/separation_text/diff}{+0.29}
\defV{final/qwen35_4/separation_text/p}{1.00}
\defV{final/qwen35_4/separation_text/pairs}{70}
\defV{final/qwen35_4/separation_text/pholm}{--}
\defV{final/qwen35_4/separation_text/status}{--}
\defV{final/qwen35_4/truncated}{213}
\defV{final/qwen35_4/truncated/grouping_cues}{2}
\defV{final/qwen35_4/truncated/gsi_placement}{41}
\defV{final/qwen35_4/truncated/gsi_placement_development}{10}
\defV{final/qwen35_4/truncated/label_placement_boxes}{82}
\defV{final/qwen35_4/truncated/map_pairs}{33}
\defV{final/qwen35_4/truncated/pointer_control_linechart}{3}
\defV{final/qwen35_4/truncated/pointer_control_scatter}{1}
\defV{final/qwen35_4/truncated/public_data}{1}
\defV{final/qwen35_4/truncated/separation_shortcut_free}{23}
\defV{final/qwen35_4/truncated/separation_shortcut_free_text}{15}
\defV{final/qwen35_4/truncated/transfer_isobar}{1}
\defV{final/qwen35_4/truncated/transfer_linechart}{1}
\defV{final/qwen35_4/unparsed}{213}
\defV{final/qwen35_4/unparsed/pct}{3.5}
\defV{final/qwen35_4/white/b}{38}
\defV{final/qwen35_4/white/c}{14}
\defV{final/qwen35_4/white/ci}{[$-$2.69, $-$0.97]}
\defV{final/qwen35_4/white/conditional}{0.51}
\defV{final/qwen35_4/white/conflict/acc}{0.38}
\defV{final/qwen35_4/white/conflict/c}{$-$0.88}
\defV{final/qwen35_4/white/conflict/ci}{[$-$1.26, $-$0.10]}
\defV{final/qwen35_4/white/conflict/d}{$-$0.63}
\defV{final/qwen35_4/white/conflict/n}{140}
\defV{final/qwen35_4/white/conflict/unparsed}{12}
\defV{final/qwen35_4/white/control/acc}{0.55}
\defV{final/qwen35_4/white/control/c}{$-$1.07}
\defV{final/qwen35_4/white/control/d}{1.07}
\defV{final/qwen35_4/white/control/n}{140}
\defV{final/qwen35_4/white/control/unparsed}{19}
\defV{final/qwen35_4/white/diff}{$-$1.70}
\defV{final/qwen35_4/white/p}{0.001}
\defV{final/qwen35_4/white/pairs}{70}
\defV{final/qwen35_4/white/pholm}{0.005}
\defV{final/qwen35_4/white/reversal}{yes}
\defV{final/qwen35_4/white/status}{yes}
\defV{final/qwen35_9/answers}{6048}
\defV{final/qwen35_9/crop/b}{92}
\defV{final/qwen35_9/crop/c}{0}
\defV{final/qwen35_9/crop/ci}{[$-$5.78, $-$4.53]}
\defV{final/qwen35_9/crop/conflict/acc}{0.23}
\defV{final/qwen35_9/crop/conflict/c}{$-$0.74}
\defV{final/qwen35_9/crop/conflict/ci}{[$-$2.55, $-$1.32]}
\defV{final/qwen35_9/crop/conflict/d}{$-$1.80}
\defV{final/qwen35_9/crop/conflict/n}{140}
\defV{final/qwen35_9/crop/conflict/unparsed}{5}
\defV{final/qwen35_9/crop/control/acc}{0.89}
\defV{final/qwen35_9/crop/control/c}{$-$0.84}
\defV{final/qwen35_9/crop/control/d}{3.22}
\defV{final/qwen35_9/crop/control/n}{140}
\defV{final/qwen35_9/crop/control/unparsed}{1}
\defV{final/qwen35_9/crop/diff}{$-$5.03}
\defV{final/qwen35_9/crop/p}{$4\times10^{-28}$}
\defV{final/qwen35_9/crop/pairs}{70}
\defV{final/qwen35_9/crop/pholm}{$2\times10^{-27}$}
\defV{final/qwen35_9/crop/status}{yes}
\defV{final/qwen35_9/cues_leader/b}{54}
\defV{final/qwen35_9/cues_leader/c}{5}
\defV{final/qwen35_9/cues_leader/ci}{[$-$3.52, $-$1.69]}
\defV{final/qwen35_9/cues_leader/conflict/acc}{0.57}
\defV{final/qwen35_9/cues_leader/conflict/c}{0.09}
\defV{final/qwen35_9/cues_leader/conflict/ci}{[$-$0.18, 0.86]}
\defV{final/qwen35_9/cues_leader/conflict/d}{0.33}
\defV{final/qwen35_9/cues_leader/conflict/n}{136}
\defV{final/qwen35_9/cues_leader/conflict/unparsed}{0}
\defV{final/qwen35_9/cues_leader/control/acc}{0.93}
\defV{final/qwen35_9/cues_leader/control/c}{0.10}
\defV{final/qwen35_9/cues_leader/control/d}{2.83}
\defV{final/qwen35_9/cues_leader/control/n}{136}
\defV{final/qwen35_9/cues_leader/control/unparsed}{0}
\defV{final/qwen35_9/cues_leader/diff}{$-$2.50}
\defV{final/qwen35_9/cues_leader/p}{$2\times10^{-11}$}
\defV{final/qwen35_9/cues_leader/pairs}{68}
\defV{final/qwen35_9/cues_leader/pholm}{--}
\defV{final/qwen35_9/cues_leader/status}{--}
\defV{final/qwen35_9/cues_region/b}{81}
\defV{final/qwen35_9/cues_region/c}{4}
\defV{final/qwen35_9/cues_region/ci}{[$-$4.60, $-$2.58]}
\defV{final/qwen35_9/cues_region/conflict/acc}{0.38}
\defV{final/qwen35_9/cues_region/conflict/c}{0.06}
\defV{final/qwen35_9/cues_region/conflict/ci}{[$-$1.13, $-$0.11]}
\defV{final/qwen35_9/cues_region/conflict/d}{$-$0.61}
\defV{final/qwen35_9/cues_region/conflict/n}{140}
\defV{final/qwen35_9/cues_region/conflict/unparsed}{0}
\defV{final/qwen35_9/cues_region/control/acc}{0.93}
\defV{final/qwen35_9/cues_region/control/c}{$-$0.10}
\defV{final/qwen35_9/cues_region/control/d}{2.86}
\defV{final/qwen35_9/cues_region/control/n}{140}
\defV{final/qwen35_9/cues_region/control/unparsed}{0}
\defV{final/qwen35_9/cues_region/diff}{$-$3.47}
\defV{final/qwen35_9/cues_region/p}{$1\times10^{-19}$}
\defV{final/qwen35_9/cues_region/pairs}{70}
\defV{final/qwen35_9/cues_region/pholm}{--}
\defV{final/qwen35_9/cues_region/status}{--}
\defV{final/qwen35_9/direct/b}{5}
\defV{final/qwen35_9/direct/c}{2}
\defV{final/qwen35_9/direct/ci}{[$-$1.57, 0.16]}
\defV{final/qwen35_9/direct/conflict/acc}{0.49}
\defV{final/qwen35_9/direct/conflict/c}{$-$1.73}
\defV{final/qwen35_9/direct/conflict/ci}{[$-$0.80, 0.42]}
\defV{final/qwen35_9/direct/conflict/d}{$-$0.16}
\defV{final/qwen35_9/direct/conflict/n}{140}
\defV{final/qwen35_9/direct/conflict/unparsed}{0}
\defV{final/qwen35_9/direct/control/acc}{0.51}
\defV{final/qwen35_9/direct/control/c}{$-$2.13}
\defV{final/qwen35_9/direct/control/d}{0.65}
\defV{final/qwen35_9/direct/control/n}{140}
\defV{final/qwen35_9/direct/control/unparsed}{0}
\defV{final/qwen35_9/direct/diff}{$-$0.80}
\defV{final/qwen35_9/direct/p}{0.45}
\defV{final/qwen35_9/direct/pairs}{70}
\defV{final/qwen35_9/direct/pholm}{0.91}
\defV{final/qwen35_9/direct/status}{n.d.}
\defV{final/qwen35_9/failed}{0}
\defV{final/qwen35_9/gate}{1.69}
\defV{final/qwen35_9/gsi_placement/b}{28}
\defV{final/qwen35_9/gsi_placement/c}{4}
\defV{final/qwen35_9/gsi_placement/ci}{[$-$3.05, $-$1.24]}
\defV{final/qwen35_9/gsi_placement/conflict/acc}{0.39}
\defV{final/qwen35_9/gsi_placement/conflict/c}{$-$1.19}
\defV{final/qwen35_9/gsi_placement/conflict/ci}{[$-$1.59, $-$0.20]}
\defV{final/qwen35_9/gsi_placement/conflict/d}{$-$0.76}
\defV{final/qwen35_9/gsi_placement/conflict/n}{128}
\defV{final/qwen35_9/gsi_placement/conflict/unparsed}{9}
\defV{final/qwen35_9/gsi_placement/control/acc}{0.58}
\defV{final/qwen35_9/gsi_placement/control/c}{$-$1.34}
\defV{final/qwen35_9/gsi_placement/control/d}{1.26}
\defV{final/qwen35_9/gsi_placement/control/n}{128}
\defV{final/qwen35_9/gsi_placement/control/unparsed}{7}
\defV{final/qwen35_9/gsi_placement/diff}{$-$2.03}
\defV{final/qwen35_9/gsi_placement/p}{$2\times10^{-5}$}
\defV{final/qwen35_9/gsi_placement/pairs}{64}
\defV{final/qwen35_9/gsi_placement/pholm}{$2\times10^{-4}$}
\defV{final/qwen35_9/gsi_placement/status}{yes}
\defV{final/qwen35_9/gsi_placement_k1/b}{6}
\defV{final/qwen35_9/gsi_placement_k1/c}{2}
\defV{final/qwen35_9/gsi_placement_k1/ci}{[$-$2.45, 0.51]}
\defV{final/qwen35_9/gsi_placement_k1/conflict/acc}{0.43}
\defV{final/qwen35_9/gsi_placement_k1/conflict/c}{$-$1.26}
\defV{final/qwen35_9/gsi_placement_k1/conflict/ci}{[$-$1.39, 0.53]}
\defV{final/qwen35_9/gsi_placement_k1/conflict/d}{$-$0.46}
\defV{final/qwen35_9/gsi_placement_k1/conflict/n}{46}
\defV{final/qwen35_9/gsi_placement_k1/conflict/unparsed}{3}
\defV{final/qwen35_9/gsi_placement_k1/control/acc}{0.52}
\defV{final/qwen35_9/gsi_placement_k1/control/c}{$-$1.27}
\defV{final/qwen35_9/gsi_placement_k1/control/d}{0.49}
\defV{final/qwen35_9/gsi_placement_k1/control/n}{46}
\defV{final/qwen35_9/gsi_placement_k1/control/unparsed}{2}
\defV{final/qwen35_9/gsi_placement_k1/diff}{$-$0.95}
\defV{final/qwen35_9/gsi_placement_k1/p}{0.29}
\defV{final/qwen35_9/gsi_placement_k1/pairs}{23}
\defV{final/qwen35_9/gsi_placement_k1/pholm}{--}
\defV{final/qwen35_9/gsi_placement_k1/status}{n.d.}
\defV{final/qwen35_9/isobar/b}{47}
\defV{final/qwen35_9/isobar/c}{31}
\defV{final/qwen35_9/isobar/ci}{[$-$0.87, 0.08]}
\defV{final/qwen35_9/isobar/conflict/acc}{0.52}
\defV{final/qwen35_9/isobar/conflict/c}{$-$0.23}
\defV{final/qwen35_9/isobar/conflict/ci}{[$-$0.22, 0.44]}
\defV{final/qwen35_9/isobar/conflict/d}{0.10}
\defV{final/qwen35_9/isobar/conflict/n}{200}
\defV{final/qwen35_9/isobar/conflict/unparsed}{2}
\defV{final/qwen35_9/isobar/control/acc}{0.59}
\defV{final/qwen35_9/isobar/control/c}{$-$0.10}
\defV{final/qwen35_9/isobar/control/d}{0.49}
\defV{final/qwen35_9/isobar/control/n}{200}
\defV{final/qwen35_9/isobar/control/unparsed}{1}
\defV{final/qwen35_9/isobar/diff}{$-$0.40}
\defV{final/qwen35_9/isobar/p}{0.09}
\defV{final/qwen35_9/isobar/pairs}{100}
\defV{final/qwen35_9/isobar/pholm}{--}
\defV{final/qwen35_9/isobar/status}{n.d.}
\defV{final/qwen35_9/isobar_text/b}{4}
\defV{final/qwen35_9/isobar_text/c}{3}
\defV{final/qwen35_9/isobar_text/ci}{[$-$0.41, 0.20]}
\defV{final/qwen35_9/isobar_text/conflict/acc}{0.90}
\defV{final/qwen35_9/isobar_text/conflict/c}{$-$0.05}
\defV{final/qwen35_9/isobar_text/conflict/ci}{[2.17, 3.05]}
\defV{final/qwen35_9/isobar_text/conflict/d}{2.52}
\defV{final/qwen35_9/isobar_text/conflict/n}{200}
\defV{final/qwen35_9/isobar_text/conflict/unparsed}{0}
\defV{final/qwen35_9/isobar_text/control/acc}{0.91}
\defV{final/qwen35_9/isobar_text/control/c}{$-$0.14}
\defV{final/qwen35_9/isobar_text/control/d}{2.60}
\defV{final/qwen35_9/isobar_text/control/n}{200}
\defV{final/qwen35_9/isobar_text/control/unparsed}{0}
\defV{final/qwen35_9/isobar_text/diff}{$-$0.08}
\defV{final/qwen35_9/isobar_text/p}{1.00}
\defV{final/qwen35_9/isobar_text/pairs}{100}
\defV{final/qwen35_9/isobar_text/pholm}{--}
\defV{final/qwen35_9/isobar_text/status}{n.d.}
\defV{final/qwen35_9/linechart/b}{24}
\defV{final/qwen35_9/linechart/c}{12}
\defV{final/qwen35_9/linechart/ci}{[$-$1.33, 0.10]}
\defV{final/qwen35_9/linechart/conflict/acc}{0.74}
\defV{final/qwen35_9/linechart/conflict/c}{$-$0.04}
\defV{final/qwen35_9/linechart/conflict/ci}{[0.76, 1.84]}
\defV{final/qwen35_9/linechart/conflict/d}{1.28}
\defV{final/qwen35_9/linechart/conflict/n}{140}
\defV{final/qwen35_9/linechart/conflict/unparsed}{0}
\defV{final/qwen35_9/linechart/control/acc}{0.83}
\defV{final/qwen35_9/linechart/control/c}{$-$0.05}
\defV{final/qwen35_9/linechart/control/d}{1.86}
\defV{final/qwen35_9/linechart/control/n}{140}
\defV{final/qwen35_9/linechart/control/unparsed}{0}
\defV{final/qwen35_9/linechart/diff}{$-$0.58}
\defV{final/qwen35_9/linechart/p}{0.07}
\defV{final/qwen35_9/linechart/pairs}{70}
\defV{final/qwen35_9/linechart/pholm}{--}
\defV{final/qwen35_9/linechart/status}{--}
\defV{final/qwen35_9/linechart_colour/b}{1}
\defV{final/qwen35_9/linechart_colour/c}{0}
\defV{final/qwen35_9/linechart_colour/ci}{[$-$0.80, 0.00]}
\defV{final/qwen35_9/linechart_colour/conflict/acc}{0.98}
\defV{final/qwen35_9/linechart_colour/conflict/c}{0.11}
\defV{final/qwen35_9/linechart_colour/conflict/ci}{[3.07, 4.91]}
\defV{final/qwen35_9/linechart_colour/conflict/d}{3.84}
\defV{final/qwen35_9/linechart_colour/conflict/n}{140}
\defV{final/qwen35_9/linechart_colour/conflict/unparsed}{0}
\defV{final/qwen35_9/linechart_colour/control/acc}{0.99}
\defV{final/qwen35_9/linechart_colour/control/c}{$-$0.00}
\defV{final/qwen35_9/linechart_colour/control/d}{4.06}
\defV{final/qwen35_9/linechart_colour/control/n}{140}
\defV{final/qwen35_9/linechart_colour/control/unparsed}{0}
\defV{final/qwen35_9/linechart_colour/diff}{$-$0.22}
\defV{final/qwen35_9/linechart_colour/p}{1.00}
\defV{final/qwen35_9/linechart_colour/pairs}{70}
\defV{final/qwen35_9/linechart_colour/pholm}{--}
\defV{final/qwen35_9/linechart_colour/status}{n.d.}
\defV{final/qwen35_9/linechart_text/b}{0}
\defV{final/qwen35_9/linechart_text/c}{0}
\defV{final/qwen35_9/linechart_text/ci}{[0.00, 0.00]}
\defV{final/qwen35_9/linechart_text/conflict/acc}{1.00}
\defV{final/qwen35_9/linechart_text/conflict/c}{$-$0.00}
\defV{final/qwen35_9/linechart_text/conflict/ci}{[4.91, 4.91]}
\defV{final/qwen35_9/linechart_text/conflict/d}{4.91}
\defV{final/qwen35_9/linechart_text/conflict/n}{140}
\defV{final/qwen35_9/linechart_text/conflict/unparsed}{0}
\defV{final/qwen35_9/linechart_text/control/acc}{1.00}
\defV{final/qwen35_9/linechart_text/control/c}{$-$0.00}
\defV{final/qwen35_9/linechart_text/control/d}{4.91}
\defV{final/qwen35_9/linechart_text/control/n}{140}
\defV{final/qwen35_9/linechart_text/control/unparsed}{0}
\defV{final/qwen35_9/linechart_text/diff}{+0.00}
\defV{final/qwen35_9/linechart_text/p}{1.00}
\defV{final/qwen35_9/linechart_text/pairs}{70}
\defV{final/qwen35_9/linechart_text/pholm}{--}
\defV{final/qwen35_9/linechart_text/status}{n.d.}
\defV{final/qwen35_9/map/index_labels/acc}{0.75}
\defV{final/qwen35_9/map/index_labels/d}{1.55}
\defV{final/qwen35_9/map/index_labels/n}{120}
\defV{final/qwen35_9/map/structure_only/acc}{0.64}
\defV{final/qwen35_9/map/structure_only/d}{1.15}
\defV{final/qwen35_9/map/structure_only/n}{120}
\defV{final/qwen35_9/map/values_only/acc}{0.62}
\defV{final/qwen35_9/map/values_only/d}{1.25}
\defV{final/qwen35_9/map/values_only/n}{120}
\defV{final/qwen35_9/map/values_structure/acc}{0.97}
\defV{final/qwen35_9/map/values_structure/d}{3.71}
\defV{final/qwen35_9/map/values_structure/n}{120}
\defV{final/qwen35_9/matched/b}{44}
\defV{final/qwen35_9/matched/c}{9}
\defV{final/qwen35_9/matched/ci}{[$-$2.62, $-$1.26]}
\defV{final/qwen35_9/matched/conflict/acc}{0.38}
\defV{final/qwen35_9/matched/conflict/c}{$-$0.72}
\defV{final/qwen35_9/matched/conflict/ci}{[$-$1.04, $-$0.10]}
\defV{final/qwen35_9/matched/conflict/d}{$-$0.57}
\defV{final/qwen35_9/matched/conflict/n}{140}
\defV{final/qwen35_9/matched/conflict/unparsed}{10}
\defV{final/qwen35_9/matched/control/acc}{0.63}
\defV{final/qwen35_9/matched/control/c}{$-$0.71}
\defV{final/qwen35_9/matched/control/d}{1.31}
\defV{final/qwen35_9/matched/control/n}{140}
\defV{final/qwen35_9/matched/control/unparsed}{14}
\defV{final/qwen35_9/matched/diff}{$-$1.88}
\defV{final/qwen35_9/matched/p}{$1\times10^{-6}$}
\defV{final/qwen35_9/matched/pairs}{70}
\defV{final/qwen35_9/matched/pholm}{$6\times10^{-6}$}
\defV{final/qwen35_9/matched/status}{yes}
\defV{final/qwen35_9/name}{Qwen3.5 9B}
\defV{final/qwen35_9/pointer_linechart/b}{18}
\defV{final/qwen35_9/pointer_linechart/c}{8}
\defV{final/qwen35_9/pointer_linechart/ci}{[$-$1.71, 0.03]}
\defV{final/qwen35_9/pointer_linechart/conflict/acc}{0.68}
\defV{final/qwen35_9/pointer_linechart/conflict/c}{$-$0.03}
\defV{final/qwen35_9/pointer_linechart/conflict/ci}{[0.47, 1.94]}
\defV{final/qwen35_9/pointer_linechart/conflict/d}{1.13}
\defV{final/qwen35_9/pointer_linechart/conflict/n}{80}
\defV{final/qwen35_9/pointer_linechart/conflict/unparsed}{5}
\defV{final/qwen35_9/pointer_linechart/control/acc}{0.80}
\defV{final/qwen35_9/pointer_linechart/control/c}{0.38}
\defV{final/qwen35_9/pointer_linechart/control/d}{1.92}
\defV{final/qwen35_9/pointer_linechart/control/n}{80}
\defV{final/qwen35_9/pointer_linechart/control/unparsed}{2}
\defV{final/qwen35_9/pointer_linechart/diff}{$-$0.79}
\defV{final/qwen35_9/pointer_linechart/p}{0.08}
\defV{final/qwen35_9/pointer_linechart/pairs}{40}
\defV{final/qwen35_9/pointer_linechart/pholm}{--}
\defV{final/qwen35_9/pointer_linechart/status}{--}
\defV{final/qwen35_9/pointer_scatter/b}{16}
\defV{final/qwen35_9/pointer_scatter/c}{1}
\defV{final/qwen35_9/pointer_scatter/ci}{[$-$2.82, $-$1.26]}
\defV{final/qwen35_9/pointer_scatter/conflict/acc}{0.78}
\defV{final/qwen35_9/pointer_scatter/conflict/c}{0.12}
\defV{final/qwen35_9/pointer_scatter/conflict/ci}{[0.85, 2.46]}
\defV{final/qwen35_9/pointer_scatter/conflict/d}{1.54}
\defV{final/qwen35_9/pointer_scatter/conflict/n}{80}
\defV{final/qwen35_9/pointer_scatter/conflict/unparsed}{1}
\defV{final/qwen35_9/pointer_scatter/control/acc}{0.96}
\defV{final/qwen35_9/pointer_scatter/control/c}{0.44}
\defV{final/qwen35_9/pointer_scatter/control/d}{3.62}
\defV{final/qwen35_9/pointer_scatter/control/n}{80}
\defV{final/qwen35_9/pointer_scatter/control/unparsed}{0}
\defV{final/qwen35_9/pointer_scatter/diff}{$-$2.08}
\defV{final/qwen35_9/pointer_scatter/p}{$3\times10^{-4}$}
\defV{final/qwen35_9/pointer_scatter/pairs}{40}
\defV{final/qwen35_9/pointer_scatter/pholm}{--}
\defV{final/qwen35_9/pointer_scatter/status}{--}
\defV{final/qwen35_9/public_data/b}{18}
\defV{final/qwen35_9/public_data/c}{7}
\defV{final/qwen35_9/public_data/ci}{[$-$1.44, $-$0.07]}
\defV{final/qwen35_9/public_data/conflict/acc}{0.60}
\defV{final/qwen35_9/public_data/conflict/c}{$-$0.04}
\defV{final/qwen35_9/public_data/conflict/ci}{[0.00, 1.07]}
\defV{final/qwen35_9/public_data/conflict/d}{0.53}
\defV{final/qwen35_9/public_data/conflict/n}{80}
\defV{final/qwen35_9/public_data/conflict/unparsed}{1}
\defV{final/qwen35_9/public_data/control/acc}{0.74}
\defV{final/qwen35_9/public_data/control/c}{0.04}
\defV{final/qwen35_9/public_data/control/d}{1.24}
\defV{final/qwen35_9/public_data/control/n}{80}
\defV{final/qwen35_9/public_data/control/unparsed}{0}
\defV{final/qwen35_9/public_data/diff}{$-$0.71}
\defV{final/qwen35_9/public_data/p}{0.04}
\defV{final/qwen35_9/public_data/pairs}{40}
\defV{final/qwen35_9/public_data/pholm}{--}
\defV{final/qwen35_9/public_data/status}{--}
\defV{final/qwen35_9/repo}{Qwen/Qwen3.5-9B}
\defV{final/qwen35_9/revision}{c202236}
\defV{final/qwen35_9/scatter/b}{82}
\defV{final/qwen35_9/scatter/c}{3}
\defV{final/qwen35_9/scatter/ci}{[$-$4.24, $-$2.39]}
\defV{final/qwen35_9/scatter/conflict/acc}{0.56}
\defV{final/qwen35_9/scatter/conflict/c}{0.01}
\defV{final/qwen35_9/scatter/conflict/ci}{[$-$0.12, 0.77]}
\defV{final/qwen35_9/scatter/conflict/d}{0.32}
\defV{final/qwen35_9/scatter/conflict/n}{200}
\defV{final/qwen35_9/scatter/conflict/unparsed}{0}
\defV{final/qwen35_9/scatter/control/acc}{0.96}
\defV{final/qwen35_9/scatter/control/c}{0.11}
\defV{final/qwen35_9/scatter/control/d}{3.42}
\defV{final/qwen35_9/scatter/control/n}{200}
\defV{final/qwen35_9/scatter/control/unparsed}{0}
\defV{final/qwen35_9/scatter/diff}{$-$3.10}
\defV{final/qwen35_9/scatter/p}{$5\times10^{-21}$}
\defV{final/qwen35_9/scatter/pairs}{100}
\defV{final/qwen35_9/scatter/pholm}{$1\times10^{-19}$}
\defV{final/qwen35_9/scatter/status}{yes}
\defV{final/qwen35_9/scatter_text/b}{1}
\defV{final/qwen35_9/scatter_text/c}{0}
\defV{final/qwen35_9/scatter_text/ci}{[$-$0.76, 0.00]}
\defV{final/qwen35_9/scatter_text/conflict/acc}{0.99}
\defV{final/qwen35_9/scatter_text/conflict/c}{$-$0.20}
\defV{final/qwen35_9/scatter_text/conflict/ci}{[4.39, 5.16]}
\defV{final/qwen35_9/scatter_text/conflict/d}{4.75}
\defV{final/qwen35_9/scatter_text/conflict/n}{200}
\defV{final/qwen35_9/scatter_text/conflict/unparsed}{1}
\defV{final/qwen35_9/scatter_text/control/acc}{0.99}
\defV{final/qwen35_9/scatter_text/control/c}{$-$0.00}
\defV{final/qwen35_9/scatter_text/control/d}{5.16}
\defV{final/qwen35_9/scatter_text/control/n}{200}
\defV{final/qwen35_9/scatter_text/control/unparsed}{1}
\defV{final/qwen35_9/scatter_text/diff}{$-$0.41}
\defV{final/qwen35_9/scatter_text/p}{1.00}
\defV{final/qwen35_9/scatter_text/pairs}{100}
\defV{final/qwen35_9/scatter_text/pholm}{--}
\defV{final/qwen35_9/scatter_text/status}{n.d.}
\defV{final/qwen35_9/separation/b}{44}
\defV{final/qwen35_9/separation/c}{10}
\defV{final/qwen35_9/separation/ci}{[$-$2.42, $-$1.02]}
\defV{final/qwen35_9/separation/conflict/acc}{0.27}
\defV{final/qwen35_9/separation/conflict/c}{$-$0.65}
\defV{final/qwen35_9/separation/conflict/ci}{[$-$1.64, $-$0.70]}
\defV{final/qwen35_9/separation/conflict/d}{$-$1.12}
\defV{final/qwen35_9/separation/conflict/n}{140}
\defV{final/qwen35_9/separation/conflict/unparsed}{17}
\defV{final/qwen35_9/separation/control/acc}{0.51}
\defV{final/qwen35_9/separation/control/c}{$-$0.45}
\defV{final/qwen35_9/separation/control/d}{0.53}
\defV{final/qwen35_9/separation/control/n}{140}
\defV{final/qwen35_9/separation/control/unparsed}{19}
\defV{final/qwen35_9/separation/diff}{$-$1.65}
\defV{final/qwen35_9/separation/p}{$3\times10^{-6}$}
\defV{final/qwen35_9/separation/pairs}{70}
\defV{final/qwen35_9/separation/pholm}{$4\times10^{-5}$}
\defV{final/qwen35_9/separation/status}{yes}
\defV{final/qwen35_9/separation_text/b}{10}
\defV{final/qwen35_9/separation_text/c}{16}
\defV{final/qwen35_9/separation_text/ci}{[$-$0.75, 0.83]}
\defV{final/qwen35_9/separation_text/conflict/acc}{0.92}
\defV{final/qwen35_9/separation_text/conflict/c}{0.22}
\defV{final/qwen35_9/separation_text/conflict/ci}{[4.02, 4.88]}
\defV{final/qwen35_9/separation_text/conflict/d}{4.43}
\defV{final/qwen35_9/separation_text/conflict/n}{140}
\defV{final/qwen35_9/separation_text/conflict/unparsed}{10}
\defV{final/qwen35_9/separation_text/control/acc}{0.88}
\defV{final/qwen35_9/separation_text/control/c}{$-$0.22}
\defV{final/qwen35_9/separation_text/control/d}{4.39}
\defV{final/qwen35_9/separation_text/control/n}{140}
\defV{final/qwen35_9/separation_text/control/unparsed}{16}
\defV{final/qwen35_9/separation_text/diff}{+0.04}
\defV{final/qwen35_9/separation_text/p}{0.33}
\defV{final/qwen35_9/separation_text/pairs}{70}
\defV{final/qwen35_9/separation_text/pholm}{--}
\defV{final/qwen35_9/separation_text/status}{--}
\defV{final/qwen35_9/truncated}{187}
\defV{final/qwen35_9/truncated/gsi_placement}{21}
\defV{final/qwen35_9/truncated/gsi_placement_development}{4}
\defV{final/qwen35_9/truncated/label_placement_boxes}{65}
\defV{final/qwen35_9/truncated/map_pairs}{21}
\defV{final/qwen35_9/truncated/pointer_control_linechart}{7}
\defV{final/qwen35_9/truncated/pointer_control_scatter}{1}
\defV{final/qwen35_9/truncated/public_data}{1}
\defV{final/qwen35_9/truncated/separation_shortcut_free}{36}
\defV{final/qwen35_9/truncated/separation_shortcut_free_text}{26}
\defV{final/qwen35_9/truncated/transfer_isobar}{3}
\defV{final/qwen35_9/truncated/transfer_scatter_text}{2}
\defV{final/qwen35_9/unparsed}{187}
\defV{final/qwen35_9/unparsed/pct}{3.1}
\defV{final/qwen35_9/white/b}{55}
\defV{final/qwen35_9/white/c}{4}
\defV{final/qwen35_9/white/ci}{[$-$3.53, $-$2.11]}
\defV{final/qwen35_9/white/conditional}{0.41}
\defV{final/qwen35_9/white/conflict/acc}{0.31}
\defV{final/qwen35_9/white/conflict/c}{$-$0.58}
\defV{final/qwen35_9/white/conflict/ci}{[$-$1.32, $-$0.36]}
\defV{final/qwen35_9/white/conflict/d}{$-$0.82}
\defV{final/qwen35_9/white/conflict/n}{140}
\defV{final/qwen35_9/white/conflict/unparsed}{18}
\defV{final/qwen35_9/white/control/acc}{0.67}
\defV{final/qwen35_9/white/control/c}{$-$0.83}
\defV{final/qwen35_9/white/control/d}{1.87}
\defV{final/qwen35_9/white/control/n}{140}
\defV{final/qwen35_9/white/control/unparsed}{17}
\defV{final/qwen35_9/white/diff}{$-$2.69}
\defV{final/qwen35_9/white/p}{$2\times10^{-12}$}
\defV{final/qwen35_9/white/pairs}{70}
\defV{final/qwen35_9/white/pholm}{$1\times10^{-11}$}
\defV{final/qwen35_9/white/reversal}{yes}
\defV{final/qwen35_9/white/status}{yes}
\defV{final/qwen38_27/answers}{6048}
\defV{final/qwen38_27/crop/b}{99}
\defV{final/qwen38_27/crop/c}{5}
\defV{final/qwen38_27/crop/ci}{[$-$5.64, $-$4.13]}
\defV{final/qwen38_27/crop/conflict/acc}{0.21}
\defV{final/qwen38_27/crop/conflict/c}{$-$0.40}
\defV{final/qwen38_27/crop/conflict/ci}{[$-$2.11, $-$0.97]}
\defV{final/qwen38_27/crop/conflict/d}{$-$1.49}
\defV{final/qwen38_27/crop/conflict/n}{140}
\defV{final/qwen38_27/crop/conflict/unparsed}{13}
\defV{final/qwen38_27/crop/control/acc}{0.89}
\defV{final/qwen38_27/crop/control/c}{$-$0.79}
\defV{final/qwen38_27/crop/control/d}{3.31}
\defV{final/qwen38_27/crop/control/n}{140}
\defV{final/qwen38_27/crop/control/unparsed}{3}
\defV{final/qwen38_27/crop/diff}{$-$4.80}
\defV{final/qwen38_27/crop/p}{$1\times10^{-23}$}
\defV{final/qwen38_27/crop/pairs}{70}
\defV{final/qwen38_27/crop/pholm}{$3\times10^{-23}$}
\defV{final/qwen38_27/crop/status}{yes}
\defV{final/qwen38_27/cues_leader/b}{42}
\defV{final/qwen38_27/cues_leader/c}{3}
\defV{final/qwen38_27/cues_leader/ci}{[$-$3.83, $-$2.36]}
\defV{final/qwen38_27/cues_leader/conflict/acc}{0.67}
\defV{final/qwen38_27/cues_leader/conflict/c}{0.09}
\defV{final/qwen38_27/cues_leader/conflict/ci}{[0.73, 1.87]}
\defV{final/qwen38_27/cues_leader/conflict/d}{1.23}
\defV{final/qwen38_27/cues_leader/conflict/n}{136}
\defV{final/qwen38_27/cues_leader/conflict/unparsed}{12}
\defV{final/qwen38_27/cues_leader/control/acc}{0.96}
\defV{final/qwen38_27/cues_leader/control/c}{$-$0.33}
\defV{final/qwen38_27/cues_leader/control/d}{4.22}
\defV{final/qwen38_27/cues_leader/control/n}{136}
\defV{final/qwen38_27/cues_leader/control/unparsed}{4}
\defV{final/qwen38_27/cues_leader/diff}{$-$2.99}
\defV{final/qwen38_27/cues_leader/p}{$9\times10^{-10}$}
\defV{final/qwen38_27/cues_leader/pairs}{68}
\defV{final/qwen38_27/cues_leader/pholm}{--}
\defV{final/qwen38_27/cues_leader/status}{--}
\defV{final/qwen38_27/cues_region/b}{69}
\defV{final/qwen38_27/cues_region/c}{2}
\defV{final/qwen38_27/cues_region/ci}{[$-$5.08, $-$2.83]}
\defV{final/qwen38_27/cues_region/conflict/acc}{0.50}
\defV{final/qwen38_27/cues_region/conflict/c}{$-$0.11}
\defV{final/qwen38_27/cues_region/conflict/ci}{[$-$0.36, 0.61]}
\defV{final/qwen38_27/cues_region/conflict/d}{0.11}
\defV{final/qwen38_27/cues_region/conflict/n}{140}
\defV{final/qwen38_27/cues_region/conflict/unparsed}{6}
\defV{final/qwen38_27/cues_region/control/acc}{0.98}
\defV{final/qwen38_27/cues_region/control/c}{$-$0.11}
\defV{final/qwen38_27/cues_region/control/d}{3.84}
\defV{final/qwen38_27/cues_region/control/n}{140}
\defV{final/qwen38_27/cues_region/control/unparsed}{0}
\defV{final/qwen38_27/cues_region/diff}{$-$3.73}
\defV{final/qwen38_27/cues_region/p}{$2\times10^{-18}$}
\defV{final/qwen38_27/cues_region/pairs}{70}
\defV{final/qwen38_27/cues_region/pholm}{--}
\defV{final/qwen38_27/cues_region/status}{--}
\defV{final/qwen38_27/direct/b}{9}
\defV{final/qwen38_27/direct/c}{4}
\defV{final/qwen38_27/direct/ci}{[$-$1.12, 0.00]}
\defV{final/qwen38_27/direct/conflict/acc}{0.46}
\defV{final/qwen38_27/direct/conflict/c}{$-$1.41}
\defV{final/qwen38_27/direct/conflict/ci}{[$-$1.03, $-$0.14]}
\defV{final/qwen38_27/direct/conflict/d}{$-$0.48}
\defV{final/qwen38_27/direct/conflict/n}{140}
\defV{final/qwen38_27/direct/conflict/unparsed}{0}
\defV{final/qwen38_27/direct/control/acc}{0.50}
\defV{final/qwen38_27/direct/control/c}{$-$1.42}
\defV{final/qwen38_27/direct/control/d}{0.00}
\defV{final/qwen38_27/direct/control/n}{140}
\defV{final/qwen38_27/direct/control/unparsed}{0}
\defV{final/qwen38_27/direct/diff}{$-$0.48}
\defV{final/qwen38_27/direct/p}{0.27}
\defV{final/qwen38_27/direct/pairs}{70}
\defV{final/qwen38_27/direct/pholm}{0.80}
\defV{final/qwen38_27/direct/status}{n.d.}
\defV{final/qwen38_27/failed}{0}
\defV{final/qwen38_27/gate}{0.86}
\defV{final/qwen38_27/gsi_placement/b}{25}
\defV{final/qwen38_27/gsi_placement/c}{6}
\defV{final/qwen38_27/gsi_placement/ci}{[$-$2.98, $-$0.71]}
\defV{final/qwen38_27/gsi_placement/conflict/acc}{0.11}
\defV{final/qwen38_27/gsi_placement/conflict/c}{$-$1.00}
\defV{final/qwen38_27/gsi_placement/conflict/ci}{[$-$1.48, 0.08]}
\defV{final/qwen38_27/gsi_placement/conflict/d}{$-$0.64}
\defV{final/qwen38_27/gsi_placement/conflict/n}{128}
\defV{final/qwen38_27/gsi_placement/conflict/unparsed}{96}
\defV{final/qwen38_27/gsi_placement/control/acc}{0.26}
\defV{final/qwen38_27/gsi_placement/control/c}{$-$0.84}
\defV{final/qwen38_27/gsi_placement/control/d}{1.05}
\defV{final/qwen38_27/gsi_placement/control/n}{128}
\defV{final/qwen38_27/gsi_placement/control/unparsed}{81}
\defV{final/qwen38_27/gsi_placement/diff}{$-$1.69}
\defV{final/qwen38_27/gsi_placement/p}{$9\times10^{-4}$}
\defV{final/qwen38_27/gsi_placement/pairs}{64}
\defV{final/qwen38_27/gsi_placement/pholm}{0.005}
\defV{final/qwen38_27/gsi_placement/status}{yes}
\defV{final/qwen38_27/gsi_placement_k1/b}{5}
\defV{final/qwen38_27/gsi_placement_k1/c}{2}
\defV{final/qwen38_27/gsi_placement_k1/ci}{[$-$1.17, 0.59]}
\defV{final/qwen38_27/gsi_placement_k1/conflict/acc}{0.09}
\defV{final/qwen38_27/gsi_placement_k1/conflict/c}{$-$1.12}
\defV{final/qwen38_27/gsi_placement_k1/conflict/ci}{[$-$0.92, 0.69]}
\defV{final/qwen38_27/gsi_placement_k1/conflict/d}{0.05}
\defV{final/qwen38_27/gsi_placement_k1/conflict/n}{46}
\defV{final/qwen38_27/gsi_placement_k1/conflict/unparsed}{33}
\defV{final/qwen38_27/gsi_placement_k1/control/acc}{0.15}
\defV{final/qwen38_27/gsi_placement_k1/control/c}{$-$1.46}
\defV{final/qwen38_27/gsi_placement_k1/control/d}{0.15}
\defV{final/qwen38_27/gsi_placement_k1/control/n}{46}
\defV{final/qwen38_27/gsi_placement_k1/control/unparsed}{34}
\defV{final/qwen38_27/gsi_placement_k1/diff}{$-$0.10}
\defV{final/qwen38_27/gsi_placement_k1/p}{0.45}
\defV{final/qwen38_27/gsi_placement_k1/pairs}{23}
\defV{final/qwen38_27/gsi_placement_k1/pholm}{--}
\defV{final/qwen38_27/gsi_placement_k1/status}{n.d.}
\defV{final/qwen38_27/isobar/b}{39}
\defV{final/qwen38_27/isobar/c}{17}
\defV{final/qwen38_27/isobar/ci}{[$-$1.52, $-$0.33]}
\defV{final/qwen38_27/isobar/conflict/acc}{0.73}
\defV{final/qwen38_27/isobar/conflict/c}{0.14}
\defV{final/qwen38_27/isobar/conflict/ci}{[0.87, 1.65]}
\defV{final/qwen38_27/isobar/conflict/d}{1.24}
\defV{final/qwen38_27/isobar/conflict/n}{200}
\defV{final/qwen38_27/isobar/conflict/unparsed}{1}
\defV{final/qwen38_27/isobar/control/acc}{0.84}
\defV{final/qwen38_27/isobar/control/c}{0.33}
\defV{final/qwen38_27/isobar/control/d}{2.10}
\defV{final/qwen38_27/isobar/control/n}{200}
\defV{final/qwen38_27/isobar/control/unparsed}{1}
\defV{final/qwen38_27/isobar/diff}{$-$0.86}
\defV{final/qwen38_27/isobar/p}{0.005}
\defV{final/qwen38_27/isobar/pairs}{100}
\defV{final/qwen38_27/isobar/pholm}{0.01}
\defV{final/qwen38_27/isobar/status}{yes}
\defV{final/qwen38_27/isobar_text/b}{0}
\defV{final/qwen38_27/isobar_text/c}{3}
\defV{final/qwen38_27/isobar_text/ci}{[0.00, 1.13]}
\defV{final/qwen38_27/isobar_text/conflict/acc}{1.00}
\defV{final/qwen38_27/isobar_text/conflict/c}{0.00}
\defV{final/qwen38_27/isobar_text/conflict/ci}{[5.16, 5.16]}
\defV{final/qwen38_27/isobar_text/conflict/d}{5.16}
\defV{final/qwen38_27/isobar_text/conflict/n}{200}
\defV{final/qwen38_27/isobar_text/conflict/unparsed}{0}
\defV{final/qwen38_27/isobar_text/control/acc}{0.98}
\defV{final/qwen38_27/isobar_text/control/c}{$-$0.38}
\defV{final/qwen38_27/isobar_text/control/d}{4.40}
\defV{final/qwen38_27/isobar_text/control/n}{200}
\defV{final/qwen38_27/isobar_text/control/unparsed}{0}
\defV{final/qwen38_27/isobar_text/diff}{+0.76}
\defV{final/qwen38_27/isobar_text/p}{0.25}
\defV{final/qwen38_27/isobar_text/pairs}{100}
\defV{final/qwen38_27/isobar_text/pholm}{--}
\defV{final/qwen38_27/isobar_text/status}{n.d.}
\defV{final/qwen38_27/linechart/b}{25}
\defV{final/qwen38_27/linechart/c}{11}
\defV{final/qwen38_27/linechart/ci}{[$-$1.51, 0.08]}
\defV{final/qwen38_27/linechart/conflict/acc}{0.75}
\defV{final/qwen38_27/linechart/conflict/c}{0.20}
\defV{final/qwen38_27/linechart/conflict/ci}{[0.78, 1.99]}
\defV{final/qwen38_27/linechart/conflict/d}{1.35}
\defV{final/qwen38_27/linechart/conflict/n}{140}
\defV{final/qwen38_27/linechart/conflict/unparsed}{0}
\defV{final/qwen38_27/linechart/control/acc}{0.85}
\defV{final/qwen38_27/linechart/control/c}{0.09}
\defV{final/qwen38_27/linechart/control/d}{2.04}
\defV{final/qwen38_27/linechart/control/n}{140}
\defV{final/qwen38_27/linechart/control/unparsed}{0}
\defV{final/qwen38_27/linechart/diff}{$-$0.69}
\defV{final/qwen38_27/linechart/p}{0.03}
\defV{final/qwen38_27/linechart/pairs}{70}
\defV{final/qwen38_27/linechart/pholm}{--}
\defV{final/qwen38_27/linechart/status}{--}
\defV{final/qwen38_27/linechart_colour/b}{1}
\defV{final/qwen38_27/linechart_colour/c}{1}
\defV{final/qwen38_27/linechart_colour/ci}{[$-$0.00, 0.00]}
\defV{final/qwen38_27/linechart_colour/conflict/acc}{0.99}
\defV{final/qwen38_27/linechart_colour/conflict/c}{$-$0.21}
\defV{final/qwen38_27/linechart_colour/conflict/ci}{[4.11, 4.91]}
\defV{final/qwen38_27/linechart_colour/conflict/d}{4.49}
\defV{final/qwen38_27/linechart_colour/conflict/n}{140}
\defV{final/qwen38_27/linechart_colour/conflict/unparsed}{0}
\defV{final/qwen38_27/linechart_colour/control/acc}{0.99}
\defV{final/qwen38_27/linechart_colour/control/c}{0.21}
\defV{final/qwen38_27/linechart_colour/control/d}{4.49}
\defV{final/qwen38_27/linechart_colour/control/n}{140}
\defV{final/qwen38_27/linechart_colour/control/unparsed}{0}
\defV{final/qwen38_27/linechart_colour/diff}{+0.00}
\defV{final/qwen38_27/linechart_colour/p}{1.00}
\defV{final/qwen38_27/linechart_colour/pairs}{70}
\defV{final/qwen38_27/linechart_colour/pholm}{--}
\defV{final/qwen38_27/linechart_colour/status}{n.d.}
\defV{final/qwen38_27/linechart_text/b}{0}
\defV{final/qwen38_27/linechart_text/c}{0}
\defV{final/qwen38_27/linechart_text/ci}{[0.00, 0.00]}
\defV{final/qwen38_27/linechart_text/conflict/acc}{1.00}
\defV{final/qwen38_27/linechart_text/conflict/c}{$-$0.00}
\defV{final/qwen38_27/linechart_text/conflict/ci}{[4.91, 4.91]}
\defV{final/qwen38_27/linechart_text/conflict/d}{4.91}
\defV{final/qwen38_27/linechart_text/conflict/n}{140}
\defV{final/qwen38_27/linechart_text/conflict/unparsed}{0}
\defV{final/qwen38_27/linechart_text/control/acc}{1.00}
\defV{final/qwen38_27/linechart_text/control/c}{$-$0.00}
\defV{final/qwen38_27/linechart_text/control/d}{4.91}
\defV{final/qwen38_27/linechart_text/control/n}{140}
\defV{final/qwen38_27/linechart_text/control/unparsed}{0}
\defV{final/qwen38_27/linechart_text/diff}{+0.00}
\defV{final/qwen38_27/linechart_text/p}{1.00}
\defV{final/qwen38_27/linechart_text/pairs}{70}
\defV{final/qwen38_27/linechart_text/pholm}{--}
\defV{final/qwen38_27/linechart_text/status}{n.d.}
\defV{final/qwen38_27/map/index_labels/acc}{0.61}
\defV{final/qwen38_27/map/index_labels/d}{1.25}
\defV{final/qwen38_27/map/index_labels/n}{120}
\defV{final/qwen38_27/map/structure_only/acc}{0.40}
\defV{final/qwen38_27/map/structure_only/d}{1.85}
\defV{final/qwen38_27/map/structure_only/n}{120}
\defV{final/qwen38_27/map/values_only/acc}{0.41}
\defV{final/qwen38_27/map/values_only/d}{1.97}
\defV{final/qwen38_27/map/values_only/n}{120}
\defV{final/qwen38_27/map/values_structure/acc}{0.99}
\defV{final/qwen38_27/map/values_structure/d}{4.79}
\defV{final/qwen38_27/map/values_structure/n}{120}
\defV{final/qwen38_27/matched/b}{81}
\defV{final/qwen38_27/matched/c}{8}
\defV{final/qwen38_27/matched/ci}{[$-$4.31, $-$2.59]}
\defV{final/qwen38_27/matched/conflict/acc}{0.20}
\defV{final/qwen38_27/matched/conflict/c}{$-$0.50}
\defV{final/qwen38_27/matched/conflict/ci}{[$-$1.80, $-$0.80]}
\defV{final/qwen38_27/matched/conflict/d}{$-$1.27}
\defV{final/qwen38_27/matched/conflict/n}{140}
\defV{final/qwen38_27/matched/conflict/unparsed}{35}
\defV{final/qwen38_27/matched/control/acc}{0.72}
\defV{final/qwen38_27/matched/control/c}{$-$0.33}
\defV{final/qwen38_27/matched/control/d}{2.05}
\defV{final/qwen38_27/matched/control/n}{140}
\defV{final/qwen38_27/matched/control/unparsed}{20}
\defV{final/qwen38_27/matched/diff}{$-$3.32}
\defV{final/qwen38_27/matched/p}{$3\times10^{-16}$}
\defV{final/qwen38_27/matched/pairs}{70}
\defV{final/qwen38_27/matched/pholm}{$2\times10^{-15}$}
\defV{final/qwen38_27/matched/status}{yes}
\defV{final/qwen38_27/name}{Qwen3.8 27B}
\defV{final/qwen38_27/pointer_linechart/b}{9}
\defV{final/qwen38_27/pointer_linechart/c}{4}
\defV{final/qwen38_27/pointer_linechart/ci}{[$-$1.43, 0.20]}
\defV{final/qwen38_27/pointer_linechart/conflict/acc}{0.81}
\defV{final/qwen38_27/pointer_linechart/conflict/c}{$-$0.04}
\defV{final/qwen38_27/pointer_linechart/conflict/ci}{[1.24, 2.48]}
\defV{final/qwen38_27/pointer_linechart/conflict/d}{1.72}
\defV{final/qwen38_27/pointer_linechart/conflict/n}{80}
\defV{final/qwen38_27/pointer_linechart/conflict/unparsed}{0}
\defV{final/qwen38_27/pointer_linechart/control/acc}{0.88}
\defV{final/qwen38_27/pointer_linechart/control/c}{0.23}
\defV{final/qwen38_27/pointer_linechart/control/d}{2.27}
\defV{final/qwen38_27/pointer_linechart/control/n}{80}
\defV{final/qwen38_27/pointer_linechart/control/unparsed}{0}
\defV{final/qwen38_27/pointer_linechart/diff}{$-$0.55}
\defV{final/qwen38_27/pointer_linechart/p}{0.27}
\defV{final/qwen38_27/pointer_linechart/pairs}{40}
\defV{final/qwen38_27/pointer_linechart/pholm}{--}
\defV{final/qwen38_27/pointer_linechart/status}{--}
\defV{final/qwen38_27/pointer_scatter/b}{17}
\defV{final/qwen38_27/pointer_scatter/c}{1}
\defV{final/qwen38_27/pointer_scatter/ci}{[$-$3.29, $-$1.12]}
\defV{final/qwen38_27/pointer_scatter/conflict/acc}{0.78}
\defV{final/qwen38_27/pointer_scatter/conflict/c}{0.16}
\defV{final/qwen38_27/pointer_scatter/conflict/ci}{[0.86, 2.28]}
\defV{final/qwen38_27/pointer_scatter/conflict/d}{1.49}
\defV{final/qwen38_27/pointer_scatter/conflict/n}{80}
\defV{final/qwen38_27/pointer_scatter/conflict/unparsed}{0}
\defV{final/qwen38_27/pointer_scatter/control/acc}{0.97}
\defV{final/qwen38_27/pointer_scatter/control/c}{0.00}
\defV{final/qwen38_27/pointer_scatter/control/d}{3.58}
\defV{final/qwen38_27/pointer_scatter/control/n}{80}
\defV{final/qwen38_27/pointer_scatter/control/unparsed}{0}
\defV{final/qwen38_27/pointer_scatter/diff}{$-$2.10}
\defV{final/qwen38_27/pointer_scatter/p}{$1\times10^{-4}$}
\defV{final/qwen38_27/pointer_scatter/pairs}{40}
\defV{final/qwen38_27/pointer_scatter/pholm}{--}
\defV{final/qwen38_27/pointer_scatter/status}{--}
\defV{final/qwen38_27/public_data/b}{17}
\defV{final/qwen38_27/public_data/c}{3}
\defV{final/qwen38_27/public_data/ci}{[$-$2.50, $-$0.64]}
\defV{final/qwen38_27/public_data/conflict/acc}{0.75}
\defV{final/qwen38_27/public_data/conflict/c}{0.23}
\defV{final/qwen38_27/public_data/conflict/ci}{[0.67, 2.13]}
\defV{final/qwen38_27/public_data/conflict/d}{1.35}
\defV{final/qwen38_27/public_data/conflict/n}{80}
\defV{final/qwen38_27/public_data/conflict/unparsed}{0}
\defV{final/qwen38_27/public_data/control/acc}{0.93}
\defV{final/qwen38_27/public_data/control/c}{0.16}
\defV{final/qwen38_27/public_data/control/d}{2.77}
\defV{final/qwen38_27/public_data/control/n}{80}
\defV{final/qwen38_27/public_data/control/unparsed}{0}
\defV{final/qwen38_27/public_data/diff}{$-$1.43}
\defV{final/qwen38_27/public_data/p}{0.003}
\defV{final/qwen38_27/public_data/pairs}{40}
\defV{final/qwen38_27/public_data/pholm}{--}
\defV{final/qwen38_27/public_data/status}{--}
\defV{final/qwen38_27/repo}{Qwen/Qwen3.8-27B}
\defV{final/qwen38_27/revision}{1d4bf0f}
\defV{final/qwen38_27/scatter/b}{47}
\defV{final/qwen38_27/scatter/c}{0}
\defV{final/qwen38_27/scatter/ci}{[$-$3.35, $-$1.62]}
\defV{final/qwen38_27/scatter/conflict/acc}{0.74}
\defV{final/qwen38_27/scatter/conflict/c}{0.03}
\defV{final/qwen38_27/scatter/conflict/ci}{[1.04, 2.08]}
\defV{final/qwen38_27/scatter/conflict/d}{1.52}
\defV{final/qwen38_27/scatter/conflict/n}{200}
\defV{final/qwen38_27/scatter/conflict/unparsed}{10}
\defV{final/qwen38_27/scatter/control/acc}{0.97}
\defV{final/qwen38_27/scatter/control/c}{0.07}
\defV{final/qwen38_27/scatter/control/d}{3.78}
\defV{final/qwen38_27/scatter/control/n}{200}
\defV{final/qwen38_27/scatter/control/unparsed}{0}
\defV{final/qwen38_27/scatter/diff}{$-$2.26}
\defV{final/qwen38_27/scatter/p}{$1\times10^{-14}$}
\defV{final/qwen38_27/scatter/pairs}{100}
\defV{final/qwen38_27/scatter/pholm}{$3\times10^{-13}$}
\defV{final/qwen38_27/scatter/status}{yes}
\defV{final/qwen38_27/scatter_text/b}{0}
\defV{final/qwen38_27/scatter_text/c}{0}
\defV{final/qwen38_27/scatter_text/ci}{[0.00, 0.00]}
\defV{final/qwen38_27/scatter_text/conflict/acc}{1.00}
\defV{final/qwen38_27/scatter_text/conflict/c}{0.00}
\defV{final/qwen38_27/scatter_text/conflict/ci}{[5.16, 5.16]}
\defV{final/qwen38_27/scatter_text/conflict/d}{5.16}
\defV{final/qwen38_27/scatter_text/conflict/n}{200}
\defV{final/qwen38_27/scatter_text/conflict/unparsed}{0}
\defV{final/qwen38_27/scatter_text/control/acc}{1.00}
\defV{final/qwen38_27/scatter_text/control/c}{0.00}
\defV{final/qwen38_27/scatter_text/control/d}{5.16}
\defV{final/qwen38_27/scatter_text/control/n}{200}
\defV{final/qwen38_27/scatter_text/control/unparsed}{0}
\defV{final/qwen38_27/scatter_text/diff}{+0.00}
\defV{final/qwen38_27/scatter_text/p}{1.00}
\defV{final/qwen38_27/scatter_text/pairs}{100}
\defV{final/qwen38_27/scatter_text/pholm}{--}
\defV{final/qwen38_27/scatter_text/status}{n.d.}
\defV{final/qwen38_27/separation/b}{41}
\defV{final/qwen38_27/separation/c}{9}
\defV{final/qwen38_27/separation/ci}{[$-$2.67, $-$1.24]}
\defV{final/qwen38_27/separation/conflict/acc}{0.21}
\defV{final/qwen38_27/separation/conflict/c}{$-$0.22}
\defV{final/qwen38_27/separation/conflict/ci}{[$-$1.58, $-$0.50]}
\defV{final/qwen38_27/separation/conflict/d}{$-$1.01}
\defV{final/qwen38_27/separation/conflict/n}{140}
\defV{final/qwen38_27/separation/conflict/unparsed}{45}
\defV{final/qwen38_27/separation/control/acc}{0.44}
\defV{final/qwen38_27/separation/control/c}{$-$0.07}
\defV{final/qwen38_27/separation/control/d}{0.87}
\defV{final/qwen38_27/separation/control/n}{140}
\defV{final/qwen38_27/separation/control/unparsed}{49}
\defV{final/qwen38_27/separation/diff}{$-$1.89}
\defV{final/qwen38_27/separation/p}{$6\times10^{-6}$}
\defV{final/qwen38_27/separation/pairs}{70}
\defV{final/qwen38_27/separation/pholm}{$6\times10^{-5}$}
\defV{final/qwen38_27/separation/status}{yes}
\defV{final/qwen38_27/separation_text/b}{0}
\defV{final/qwen38_27/separation_text/c}{0}
\defV{final/qwen38_27/separation_text/ci}{[0.00, 0.00]}
\defV{final/qwen38_27/separation_text/conflict/acc}{1.00}
\defV{final/qwen38_27/separation_text/conflict/c}{$-$0.00}
\defV{final/qwen38_27/separation_text/conflict/ci}{[4.91, 4.91]}
\defV{final/qwen38_27/separation_text/conflict/d}{4.91}
\defV{final/qwen38_27/separation_text/conflict/n}{140}
\defV{final/qwen38_27/separation_text/conflict/unparsed}{0}
\defV{final/qwen38_27/separation_text/control/acc}{1.00}
\defV{final/qwen38_27/separation_text/control/c}{$-$0.00}
\defV{final/qwen38_27/separation_text/control/d}{4.91}
\defV{final/qwen38_27/separation_text/control/n}{140}
\defV{final/qwen38_27/separation_text/control/unparsed}{0}
\defV{final/qwen38_27/separation_text/diff}{+0.00}
\defV{final/qwen38_27/separation_text/p}{1.00}
\defV{final/qwen38_27/separation_text/pairs}{70}
\defV{final/qwen38_27/separation_text/pholm}{--}
\defV{final/qwen38_27/separation_text/status}{n.d.}
\defV{final/qwen38_27/truncated}{689}
\defV{final/qwen38_27/truncated/grouping_cues}{22}
\defV{final/qwen38_27/truncated/gsi_placement}{244}
\defV{final/qwen38_27/truncated/gsi_placement_development}{48}
\defV{final/qwen38_27/truncated/label_placement_boxes}{130}
\defV{final/qwen38_27/truncated/map_pairs}{139}
\defV{final/qwen38_27/truncated/separation_shortcut_free}{94}
\defV{final/qwen38_27/truncated/transfer_isobar}{2}
\defV{final/qwen38_27/truncated/transfer_scatter}{10}
\defV{final/qwen38_27/unparsed}{689}
\defV{final/qwen38_27/unparsed/pct}{11.4}
\defV{final/qwen38_27/white/b}{74}
\defV{final/qwen38_27/white/c}{5}
\defV{final/qwen38_27/white/ci}{[$-$4.39, $-$2.70]}
\defV{final/qwen38_27/white/conditional}{0.24}
\defV{final/qwen38_27/white/conflict/acc}{0.21}
\defV{final/qwen38_27/white/conflict/c}{$-$0.45}
\defV{final/qwen38_27/white/conflict/ci}{[$-$1.90, $-$0.73]}
\defV{final/qwen38_27/white/conflict/d}{$-$1.25}
\defV{final/qwen38_27/white/conflict/n}{140}
\defV{final/qwen38_27/white/conflict/unparsed}{32}
\defV{final/qwen38_27/white/control/acc}{0.70}
\defV{final/qwen38_27/white/control/c}{$-$0.30}
\defV{final/qwen38_27/white/control/d}{2.17}
\defV{final/qwen38_27/white/control/n}{140}
\defV{final/qwen38_27/white/control/unparsed}{27}
\defV{final/qwen38_27/white/diff}{$-$3.42}
\defV{final/qwen38_27/white/p}{$8\times10^{-17}$}
\defV{final/qwen38_27/white/pairs}{70}
\defV{final/qwen38_27/white/pholm}{$7\times10^{-16}$}
\defV{final/qwen38_27/white/reversal}{yes}
\defV{final/qwen38_27/white/status}{yes}
\defV{final/rule/gate}{yes}
\defV{final/rule/gsi_placement}{yes}
\defV{final/rule/isobar}{yes}
\defV{final/rule/matched}{yes}
\defV{final/rule/reversal}{yes}
\defV{final/rule/reversal/all}{no}
\defV{final/rule/scatter}{yes}
\defV{final/rule/separation}{yes}
\defV{final/set/grouping_cues/q}{552}
\defV{final/set/gsi_placement/q}{348}
\defV{final/set/gsi_placement_development/q}{68}
\defV{final/set/label_placement_boxes/q}{1{,}120}
\defV{final/set/map_pairs/q}{480}
\defV{final/set/pointer_control_linechart/q}{160}
\defV{final/set/pointer_control_scatter/q}{160}
\defV{final/set/public_data/q}{160}
\defV{final/set/separation_shortcut_free/q}{280}
\defV{final/set/separation_shortcut_free_text/q}{280}
\defV{final/set/transfer_isobar/q}{400}
\defV{final/set/transfer_isobar_text/q}{400}
\defV{final/set/transfer_linechart/q}{560}
\defV{final/set/transfer_linechart_text/q}{280}
\defV{final/set/transfer_scatter/q}{400}
\defV{final/set/transfer_scatter_text/q}{400}
\defV{final/spot/decided}{8}
\defV{final/spot/gemma4_12/named/conflict}{0.46}
\defV{final/spot/gemma4_12/named/control}{0.54}
\defV{final/spot/gemma4_12/named/drop}{0.08}
\defV{final/spot/gemma4_12/named/p}{0.02}
\defV{final/spot/gemma4_12/named/unanswered}{0}
\defV{final/spot/gemma4_12/original/conflict}{0.49}
\defV{final/spot/gemma4_12/original/control}{0.51}
\defV{final/spot/gemma4_12/original/drop}{0.03}
\defV{final/spot/gemma4_12/original/p}{0.50}
\defV{final/spot/gemma4_12/original/unanswered}{0}
\defV{final/spot/gemma4_12/value/conflict}{0.46}
\defV{final/spot/gemma4_12/value/control}{0.56}
\defV{final/spot/gemma4_12/value/drop}{0.11}
\defV{final/spot/gemma4_12/value/near}{0.21}
\defV{final/spot/gemma4_12/value/p}{0.006}
\defV{final/spot/gemma4_12/value/unanswered}{9}
\defV{final/spot/gemma4_31/named/conflict}{0.27}
\defV{final/spot/gemma4_31/named/control}{0.79}
\defV{final/spot/gemma4_31/named/drop}{0.52}
\defV{final/spot/gemma4_31/named/p}{$3\times10^{-19}$}
\defV{final/spot/gemma4_31/named/unanswered}{0}
\defV{final/spot/gemma4_31/original/conflict}{0.29}
\defV{final/spot/gemma4_31/original/control}{0.73}
\defV{final/spot/gemma4_31/original/drop}{0.44}
\defV{final/spot/gemma4_31/original/p}{$1\times10^{-17}$}
\defV{final/spot/gemma4_31/original/unanswered}{0}
\defV{final/spot/gemma4_31/value/conflict}{0.32}
\defV{final/spot/gemma4_31/value/control}{0.79}
\defV{final/spot/gemma4_31/value/drop}{0.46}
\defV{final/spot/gemma4_31/value/near}{0.49}
\defV{final/spot/gemma4_31/value/p}{$5\times10^{-17}$}
\defV{final/spot/gemma4_31/value/unanswered}{0}
\defV{final/spot/glm46v_9/named/conflict}{0.40}
\defV{final/spot/glm46v_9/named/control}{0.66}
\defV{final/spot/glm46v_9/named/drop}{0.26}
\defV{final/spot/glm46v_9/named/p}{$4\times10^{-7}$}
\defV{final/spot/glm46v_9/named/unanswered}{0}
\defV{final/spot/glm46v_9/original/conflict}{0.34}
\defV{final/spot/glm46v_9/original/control}{0.56}
\defV{final/spot/glm46v_9/original/drop}{0.23}
\defV{final/spot/glm46v_9/original/p}{$4\times10^{-5}$}
\defV{final/spot/glm46v_9/original/unanswered}{0}
\defV{final/spot/glm46v_9/value/conflict}{0.49}
\defV{final/spot/glm46v_9/value/control}{0.52}
\defV{final/spot/glm46v_9/value/drop}{0.03}
\defV{final/spot/glm46v_9/value/near}{0.64}
\defV{final/spot/glm46v_9/value/p}{0.58}
\defV{final/spot/glm46v_9/value/unanswered}{0}
\defV{final/spot/internvl35_38/named/conflict}{0.47}
\defV{final/spot/internvl35_38/named/control}{0.61}
\defV{final/spot/internvl35_38/named/drop}{0.14}
\defV{final/spot/internvl35_38/named/p}{0.001}
\defV{final/spot/internvl35_38/named/unanswered}{0}
\defV{final/spot/internvl35_38/original/conflict}{0.42}
\defV{final/spot/internvl35_38/original/control}{0.59}
\defV{final/spot/internvl35_38/original/drop}{0.16}
\defV{final/spot/internvl35_38/original/p}{0.004}
\defV{final/spot/internvl35_38/original/unanswered}{0}
\defV{final/spot/internvl35_38/value/conflict}{0.44}
\defV{final/spot/internvl35_38/value/control}{0.51}
\defV{final/spot/internvl35_38/value/drop}{0.06}
\defV{final/spot/internvl35_38/value/near}{0.71}
\defV{final/spot/internvl35_38/value/p}{0.12}
\defV{final/spot/internvl35_38/value/unanswered}{0}
\defV{final/spot/items}{280}
\defV{final/spot/llavaov2_8/named/conflict}{0.35}
\defV{final/spot/llavaov2_8/named/control}{0.63}
\defV{final/spot/llavaov2_8/named/drop}{0.28}
\defV{final/spot/llavaov2_8/named/p}{$5\times10^{-7}$}
\defV{final/spot/llavaov2_8/named/unanswered}{0}
\defV{final/spot/llavaov2_8/original/conflict}{0.24}
\defV{final/spot/llavaov2_8/original/control}{0.44}
\defV{final/spot/llavaov2_8/original/drop}{0.20}
\defV{final/spot/llavaov2_8/original/p}{$4\times10^{-4}$}
\defV{final/spot/llavaov2_8/original/unanswered}{0}
\defV{final/spot/llavaov2_8/value/conflict}{0.53}
\defV{final/spot/llavaov2_8/value/control}{0.54}
\defV{final/spot/llavaov2_8/value/drop}{0.01}
\defV{final/spot/llavaov2_8/value/near}{0.55}
\defV{final/spot/llavaov2_8/value/p}{1.00}
\defV{final/spot/llavaov2_8/value/unanswered}{1}
\defV{final/spot/mistral32_24/named/conflict}{0.44}
\defV{final/spot/mistral32_24/named/control}{0.55}
\defV{final/spot/mistral32_24/named/drop}{0.11}
\defV{final/spot/mistral32_24/named/p}{0.009}
\defV{final/spot/mistral32_24/named/unanswered}{0}
\defV{final/spot/mistral32_24/original/conflict}{0.35}
\defV{final/spot/mistral32_24/original/control}{0.66}
\defV{final/spot/mistral32_24/original/drop}{0.31}
\defV{final/spot/mistral32_24/original/p}{$2\times10^{-7}$}
\defV{final/spot/mistral32_24/original/unanswered}{0}
\defV{final/spot/mistral32_24/value/conflict}{0.46}
\defV{final/spot/mistral32_24/value/control}{0.51}
\defV{final/spot/mistral32_24/value/drop}{0.05}
\defV{final/spot/mistral32_24/value/near}{0.69}
\defV{final/spot/mistral32_24/value/p}{0.14}
\defV{final/spot/mistral32_24/value/unanswered}{0}
\defV{final/spot/molmo2o_7/named/conflict}{0.40}
\defV{final/spot/molmo2o_7/named/control}{0.51}
\defV{final/spot/molmo2o_7/named/drop}{0.11}
\defV{final/spot/molmo2o_7/named/p}{0.02}
\defV{final/spot/molmo2o_7/named/unanswered}{0}
\defV{final/spot/molmo2o_7/original/conflict}{0.49}
\defV{final/spot/molmo2o_7/original/control}{0.54}
\defV{final/spot/molmo2o_7/original/drop}{0.06}
\defV{final/spot/molmo2o_7/original/p}{0.13}
\defV{final/spot/molmo2o_7/original/unanswered}{0}
\defV{final/spot/molmo2o_7/value/conflict}{0.42}
\defV{final/spot/molmo2o_7/value/control}{0.44}
\defV{final/spot/molmo2o_7/value/drop}{0.02}
\defV{final/spot/molmo2o_7/value/near}{0.58}
\defV{final/spot/molmo2o_7/value/p}{0.79}
\defV{final/spot/molmo2o_7/value/unanswered}{2}
\defV{final/spot/new/k}{2}
\defV{final/spot/new/names}{Gemma-4 12B and Molmo2-O 7B}
\defV{final/spot/pairs}{140}
\defV{final/spot/persists}{7}
\defV{final/spot/qwen35_4/named/conflict}{0.41}
\defV{final/spot/qwen35_4/named/control}{0.64}
\defV{final/spot/qwen35_4/named/drop}{0.23}
\defV{final/spot/qwen35_4/named/p}{$4\times10^{-5}$}
\defV{final/spot/qwen35_4/named/unanswered}{0}
\defV{final/spot/qwen35_4/original/conflict}{0.38}
\defV{final/spot/qwen35_4/original/control}{0.55}
\defV{final/spot/qwen35_4/original/drop}{0.17}
\defV{final/spot/qwen35_4/original/p}{0.001}
\defV{final/spot/qwen35_4/original/unanswered}{0}
\defV{final/spot/qwen35_4/value/conflict}{0.41}
\defV{final/spot/qwen35_4/value/control}{0.60}
\defV{final/spot/qwen35_4/value/drop}{0.19}
\defV{final/spot/qwen35_4/value/near}{0.60}
\defV{final/spot/qwen35_4/value/p}{$2\times10^{-4}$}
\defV{final/spot/qwen35_4/value/unanswered}{9}
\defV{final/spot/qwen35_9/named/conflict}{0.34}
\defV{final/spot/qwen35_9/named/control}{0.73}
\defV{final/spot/qwen35_9/named/drop}{0.39}
\defV{final/spot/qwen35_9/named/p}{$4\times10^{-12}$}
\defV{final/spot/qwen35_9/named/unanswered}{0}
\defV{final/spot/qwen35_9/original/conflict}{0.31}
\defV{final/spot/qwen35_9/original/control}{0.67}
\defV{final/spot/qwen35_9/original/drop}{0.36}
\defV{final/spot/qwen35_9/original/p}{$2\times10^{-12}$}
\defV{final/spot/qwen35_9/original/unanswered}{0}
\defV{final/spot/qwen35_9/value/conflict}{0.31}
\defV{final/spot/qwen35_9/value/control}{0.74}
\defV{final/spot/qwen35_9/value/drop}{0.43}
\defV{final/spot/qwen35_9/value/near}{0.59}
\defV{final/spot/qwen35_9/value/p}{$7\times10^{-14}$}
\defV{final/spot/qwen35_9/value/unanswered}{18}
\defV{final/spot/qwen38_27/named/conflict}{0.28}
\defV{final/spot/qwen38_27/named/control}{0.74}
\defV{final/spot/qwen38_27/named/drop}{0.46}
\defV{final/spot/qwen38_27/named/p}{$3\times10^{-14}$}
\defV{final/spot/qwen38_27/named/unanswered}{0}
\defV{final/spot/qwen38_27/original/conflict}{0.21}
\defV{final/spot/qwen38_27/original/control}{0.70}
\defV{final/spot/qwen38_27/original/drop}{0.49}
\defV{final/spot/qwen38_27/original/p}{$8\times10^{-17}$}
\defV{final/spot/qwen38_27/original/unanswered}{0}
\defV{final/spot/qwen38_27/value/conflict}{0.14}
\defV{final/spot/qwen38_27/value/control}{0.40}
\defV{final/spot/qwen38_27/value/drop}{0.26}
\defV{final/spot/qwen38_27/value/near}{0.18}
\defV{final/spot/qwen38_27/value/p}{$1\times10^{-7}$}
\defV{final/spot/qwen38_27/value/unanswered}{147}
\defV{final/spot/value/cut/k}{147}
\defV{final/spot/value/cut/names}{Qwen3.8 27B}
\defV{final/spot/value/lost/k}{4}
\defV{final/spot/value/n}{9}
\defV{final/spot/value/near/max}{0.71}
\defV{final/spot/value/near/min}{0.21}
\defV{final/spot/value/near/minname}{Gemma-4 12B}
\defV{final/spot/value/near/second}{0.49}
\defV{final/strict/changed}{0}
\defV{final/strict/tests}{80}
\defV{final/supp/cues_leader/decidable}{10}
\defV{final/supp/cues_leader/drop}{10}
\defV{final/supp/cues_leader/notdecidable}{0}
\defV{final/supp/cues_region/decidable}{10}
\defV{final/supp/cues_region/drop}{10}
\defV{final/supp/cues_region/notdecidable}{0}
\defV{final/supp/gsi_placement_k1/decidable}{4}
\defV{final/supp/gsi_placement_k1/drop}{2}
\defV{final/supp/gsi_placement_k1/notdecidable}{6}
\defV{final/supp/isobar_text/acc/max}{1.00}
\defV{final/supp/isobar_text/acc/min}{0.46}
\defV{final/supp/isobar_text/decidable}{2}
\defV{final/supp/isobar_text/drop}{0}
\defV{final/supp/isobar_text/notdecidable}{8}
\defV{final/supp/linechart/decidable}{10}
\defV{final/supp/linechart/drop}{7}
\defV{final/supp/linechart/notdecidable}{0}
\defV{final/supp/linechart_colour/decidable}{3}
\defV{final/supp/linechart_colour/drop}{2}
\defV{final/supp/linechart_colour/notdecidable}{7}
\defV{final/supp/linechart_text/acc/max}{1.00}
\defV{final/supp/linechart_text/acc/min}{1.00}
\defV{final/supp/linechart_text/decidable}{0}
\defV{final/supp/linechart_text/drop}{0}
\defV{final/supp/linechart_text/notdecidable}{10}
\defV{final/supp/pointer_linechart/decidable}{8}
\defV{final/supp/pointer_linechart/drop}{0}
\defV{final/supp/pointer_linechart/notdecidable}{2}
\defV{final/supp/pointer_scatter/decidable}{5}
\defV{final/supp/pointer_scatter/drop}{4}
\defV{final/supp/pointer_scatter/notdecidable}{5}
\defV{final/supp/public_data/decidable}{9}
\defV{final/supp/public_data/drop}{7}
\defV{final/supp/public_data/notdecidable}{1}
\defV{final/supp/scatter_text/acc/max}{1.00}
\defV{final/supp/scatter_text/acc/min}{0.85}
\defV{final/supp/scatter_text/decidable}{1}
\defV{final/supp/scatter_text/drop}{0}
\defV{final/supp/scatter_text/notdecidable}{9}
\defV{final/supp/separation_text/acc/max}{1.00}
\defV{final/supp/separation_text/acc/min}{0.52}
\defV{final/supp/separation_text/decidable}{4}
\defV{final/supp/separation_text/drop}{0}
\defV{final/supp/separation_text/notdecidable}{6}
\defV{final/thinking/qwen38_27/answers}{1120}
\defV{final/thinking/qwen38_27/crop/b}{60}
\defV{final/thinking/qwen38_27/crop/c}{5}
\defV{final/thinking/qwen38_27/crop/ci}{[$-$4.23, $-$2.37]}
\defV{final/thinking/qwen38_27/crop/conflict/acc}{0.51}
\defV{final/thinking/qwen38_27/crop/conflict/c}{$-$0.35}
\defV{final/thinking/qwen38_27/crop/conflict/ci}{[$-$0.16, 0.83]}
\defV{final/thinking/qwen38_27/crop/conflict/d}{0.32}
\defV{final/thinking/qwen38_27/crop/conflict/n}{140}
\defV{final/thinking/qwen38_27/crop/conflict/unparsed}{11}
\defV{final/thinking/qwen38_27/crop/control/acc}{0.91}
\defV{final/thinking/qwen38_27/crop/control/c}{$-$0.28}
\defV{final/thinking/qwen38_27/crop/control/d}{3.50}
\defV{final/thinking/qwen38_27/crop/control/n}{140}
\defV{final/thinking/qwen38_27/crop/control/unparsed}{8}
\defV{final/thinking/qwen38_27/crop/diff}{$-$3.17}
\defV{final/thinking/qwen38_27/crop/p}{$5\times10^{-13}$}
\defV{final/thinking/qwen38_27/crop/pairs}{70}
\defV{final/thinking/qwen38_27/crop/pholm}{$5\times10^{-13}$}
\defV{final/thinking/qwen38_27/crop/status}{yes}
\defV{final/thinking/qwen38_27/direct/b}{92}
\defV{final/thinking/qwen38_27/direct/c}{2}
\defV{final/thinking/qwen38_27/direct/ci}{[$-$4.81, $-$3.24]}
\defV{final/thinking/qwen38_27/direct/conflict/acc}{0.18}
\defV{final/thinking/qwen38_27/direct/conflict/c}{$-$0.48}
\defV{final/thinking/qwen38_27/direct/conflict/ci}{[$-$2.43, $-$1.42]}
\defV{final/thinking/qwen38_27/direct/conflict/d}{$-$1.85}
\defV{final/thinking/qwen38_27/direct/conflict/n}{140}
\defV{final/thinking/qwen38_27/direct/conflict/unparsed}{11}
\defV{final/thinking/qwen38_27/direct/control/acc}{0.82}
\defV{final/thinking/qwen38_27/direct/control/c}{$-$0.22}
\defV{final/thinking/qwen38_27/direct/control/d}{2.04}
\defV{final/thinking/qwen38_27/direct/control/n}{140}
\defV{final/thinking/qwen38_27/direct/control/unparsed}{4}
\defV{final/thinking/qwen38_27/direct/diff}{$-$3.88}
\defV{final/thinking/qwen38_27/direct/p}{$5\times10^{-25}$}
\defV{final/thinking/qwen38_27/direct/pairs}{70}
\defV{final/thinking/qwen38_27/direct/pholm}{$5\times10^{-25}$}
\defV{final/thinking/qwen38_27/direct/status}{yes}
\defV{final/thinking/qwen38_27/failed}{0}
\defV{final/thinking/qwen38_27/gate}{--}
\defV{final/thinking/qwen38_27/matched/b}{90}
\defV{final/thinking/qwen38_27/matched/c}{6}
\defV{final/thinking/qwen38_27/matched/ci}{[$-$5.02, $-$3.35]}
\defV{final/thinking/qwen38_27/matched/conflict/acc}{0.24}
\defV{final/thinking/qwen38_27/matched/conflict/c}{$-$0.67}
\defV{final/thinking/qwen38_27/matched/conflict/ci}{[$-$2.35, $-$1.13]}
\defV{final/thinking/qwen38_27/matched/conflict/d}{$-$1.64}
\defV{final/thinking/qwen38_27/matched/conflict/n}{140}
\defV{final/thinking/qwen38_27/matched/conflict/unparsed}{9}
\defV{final/thinking/qwen38_27/matched/control/acc}{0.84}
\defV{final/thinking/qwen38_27/matched/control/c}{$-$0.20}
\defV{final/thinking/qwen38_27/matched/control/d}{2.40}
\defV{final/thinking/qwen38_27/matched/control/n}{140}
\defV{final/thinking/qwen38_27/matched/control/unparsed}{8}
\defV{final/thinking/qwen38_27/matched/diff}{$-$4.04}
\defV{final/thinking/qwen38_27/matched/p}{$3\times10^{-20}$}
\defV{final/thinking/qwen38_27/matched/pairs}{70}
\defV{final/thinking/qwen38_27/matched/pholm}{$3\times10^{-20}$}
\defV{final/thinking/qwen38_27/matched/status}{yes}
\defV{final/thinking/qwen38_27/truncated}{64}
\defV{final/thinking/qwen38_27/truncated/label_placement_boxes}{64}
\defV{final/thinking/qwen38_27/unparsed}{67}
\defV{final/thinking/qwen38_27/unparsed/pct}{6.0}
\defV{final/thinking/qwen38_27/white/b}{82}
\defV{final/thinking/qwen38_27/white/c}{5}
\defV{final/thinking/qwen38_27/white/ci}{[$-$4.28, $-$2.69]}
\defV{final/thinking/qwen38_27/white/conflict/acc}{0.24}
\defV{final/thinking/qwen38_27/white/conflict/c}{$-$0.63}
\defV{final/thinking/qwen38_27/white/conflict/ci}{[$-$2.14, $-$1.06]}
\defV{final/thinking/qwen38_27/white/conflict/d}{$-$1.54}
\defV{final/thinking/qwen38_27/white/conflict/n}{140}
\defV{final/thinking/qwen38_27/white/conflict/unparsed}{10}
\defV{final/thinking/qwen38_27/white/control/acc}{0.79}
\defV{final/thinking/qwen38_27/white/control/c}{$-$0.25}
\defV{final/thinking/qwen38_27/white/control/d}{1.84}
\defV{final/thinking/qwen38_27/white/control/n}{140}
\defV{final/thinking/qwen38_27/white/control/unparsed}{6}
\defV{final/thinking/qwen38_27/white/diff}{$-$3.38}
\defV{final/thinking/qwen38_27/white/p}{$5\times10^{-19}$}
\defV{final/thinking/qwen38_27/white/pairs}{70}
\defV{final/thinking/qwen38_27/white/pholm}{$5\times10^{-19}$}
\defV{final/thinking/qwen38_27/white/reversal}{yes}
\defV{final/thinking/qwen38_27/white/status}{yes}
\defV{final/unparsed/second/pct}{6.2}
\defV{probe/name/gemma4_31}{Gemma-4 31B}
\defV{probe/name/gemma4_12}{Gemma-4 12B}
\defV{probe/name/qwen38_27}{Qwen3.8 27B}
\defV{probe/name/qwen35_9}{Qwen3.5 9B}
\defV{probe/name/qwen35_4}{Qwen3.5 4B}
\defV{probe/name/internvl35_38}{InternVL3.5 38B}
\defV{probe/name/mistral32_24}{Mistral-Small-3.2 24B}
\defV{probe/name/glm46v_9}{GLM-4.6V-Flash}
\defV{probe/name/llavaov2_8}{LLaVA-OneVision-2 8B}
\defV{probe/name/molmo2o_7}{Molmo2-O 7B}
\defV{probe/name/gpt6_luna}{GPT-6 Luna}
\defV{probe/name/gpt6_sol}{GPT-6.1 Sol}
\defV{probe/remedy/models}{10}
\defV{probe/remedy/models/word}{ten}
\defV{probe/remedy/sigmodels}{7}
\defV{probe/remedy/sigmodels/word}{seven}
\defV{probe/remedy/plain/gemma4_31/control}{0.73}
\defV{probe/remedy/plain/gemma4_31/conflict}{0.29}
\defV{probe/remedy/plain/gemma4_31/drop}{0.44}
\defV{probe/remedy/plain/gemma4_31/dropph}{$1\times10^{-16}$}
\defV{probe/remedy/plain/qwen38_27/control}{0.70}
\defV{probe/remedy/plain/qwen38_27/conflict}{0.21}
\defV{probe/remedy/plain/qwen38_27/drop}{0.49}
\defV{probe/remedy/plain/qwen38_27/dropph}{$7\times10^{-16}$}
\defV{probe/remedy/plain/qwen35_9/control}{0.67}
\defV{probe/remedy/plain/qwen35_9/conflict}{0.31}
\defV{probe/remedy/plain/qwen35_9/drop}{0.36}
\defV{probe/remedy/plain/qwen35_9/dropph}{$1\times10^{-11}$}
\defV{probe/remedy/plain/internvl35_38/control}{0.59}
\defV{probe/remedy/plain/internvl35_38/conflict}{0.42}
\defV{probe/remedy/plain/internvl35_38/drop}{0.16}
\defV{probe/remedy/plain/internvl35_38/dropph}{0.01}
\defV{probe/remedy/plain/mistral32_24/control}{0.66}
\defV{probe/remedy/plain/mistral32_24/conflict}{0.35}
\defV{probe/remedy/plain/mistral32_24/drop}{0.31}
\defV{probe/remedy/plain/mistral32_24/dropph}{$1\times10^{-6}$}
\defV{probe/remedy/plain/glm46v_9/control}{0.56}
\defV{probe/remedy/plain/glm46v_9/conflict}{0.34}
\defV{probe/remedy/plain/glm46v_9/drop}{0.23}
\defV{probe/remedy/plain/glm46v_9/dropph}{$3\times10^{-4}$}
\defV{probe/remedy/plain/llavaov2_8/control}{0.44}
\defV{probe/remedy/plain/llavaov2_8/conflict}{0.24}
\defV{probe/remedy/plain/llavaov2_8/drop}{0.20}
\defV{probe/remedy/plain/llavaov2_8/dropph}{0.002}
\defV{probe/remedy/plain/molmo2o_7/control}{0.54}
\defV{probe/remedy/plain/molmo2o_7/conflict}{0.49}
\defV{probe/remedy/plain/molmo2o_7/drop}{0.06}
\defV{probe/remedy/plain/molmo2o_7/dropph}{0.27}
\defV{probe/remedy/plain/qwen35_4/control}{0.55}
\defV{probe/remedy/plain/qwen35_4/conflict}{0.38}
\defV{probe/remedy/plain/qwen35_4/drop}{0.17}
\defV{probe/remedy/plain/qwen35_4/dropph}{0.005}
\defV{probe/remedy/plain/gemma4_12/control}{0.51}
\defV{probe/remedy/plain/gemma4_12/conflict}{0.49}
\defV{probe/remedy/plain/gemma4_12/drop}{0.03}
\defV{probe/remedy/plain/gemma4_12/dropph}{0.50}
\defV{probe/remedy/upscale/gemma4_31/control}{0.74}
\defV{probe/remedy/upscale/gemma4_31/conflict}{0.34}
\defV{probe/remedy/upscale/gemma4_31/drop}{0.40}
\defV{probe/remedy/upscale/gemma4_31/dropph}{$3\times10^{-15}$}
\defV{probe/remedy/upscale/gemma4_31/gain}{+0.04}
\defV{probe/remedy/upscale/gemma4_31/gainph}{1.00}
\defV{probe/remedy/upscale/gemma4_31/ctrlchange}{+0.01}
\defV{probe/remedy/upscale/gemma4_31/ctrlph}{1.00}
\defV{probe/remedy/upscale/qwen38_27/control}{0.78}
\defV{probe/remedy/upscale/qwen38_27/conflict}{0.19}
\defV{probe/remedy/upscale/qwen38_27/drop}{0.59}
\defV{probe/remedy/upscale/qwen38_27/dropph}{$9\times10^{-23}$}
\defV{probe/remedy/upscale/qwen38_27/gain}{$-$0.01}
\defV{probe/remedy/upscale/qwen38_27/gainph}{1.00}
\defV{probe/remedy/upscale/qwen38_27/ctrlchange}{+0.08}
\defV{probe/remedy/upscale/qwen38_27/ctrlph}{0.72}
\defV{probe/remedy/upscale/qwen35_9/control}{0.78}
\defV{probe/remedy/upscale/qwen35_9/conflict}{0.26}
\defV{probe/remedy/upscale/qwen35_9/drop}{0.52}
\defV{probe/remedy/upscale/qwen35_9/dropph}{$4\times10^{-19}$}
\defV{probe/remedy/upscale/qwen35_9/gain}{$-$0.05}
\defV{probe/remedy/upscale/qwen35_9/gainph}{1.00}
\defV{probe/remedy/upscale/qwen35_9/ctrlchange}{+0.11}
\defV{probe/remedy/upscale/qwen35_9/ctrlph}{0.25}
\defV{probe/remedy/upscale/internvl35_38/control}{0.65}
\defV{probe/remedy/upscale/internvl35_38/conflict}{0.41}
\defV{probe/remedy/upscale/internvl35_38/drop}{0.24}
\defV{probe/remedy/upscale/internvl35_38/dropph}{$4\times10^{-5}$}
\defV{probe/remedy/upscale/internvl35_38/gain}{$-$0.01}
\defV{probe/remedy/upscale/internvl35_38/gainph}{1.00}
\defV{probe/remedy/upscale/internvl35_38/ctrlchange}{+0.06}
\defV{probe/remedy/upscale/internvl35_38/ctrlph}{1.00}
\defV{probe/remedy/upscale/mistral32_24/control}{0.69}
\defV{probe/remedy/upscale/mistral32_24/conflict}{0.33}
\defV{probe/remedy/upscale/mistral32_24/drop}{0.36}
\defV{probe/remedy/upscale/mistral32_24/dropph}{$1\times10^{-9}$}
\defV{probe/remedy/upscale/mistral32_24/gain}{$-$0.02}
\defV{probe/remedy/upscale/mistral32_24/gainph}{1.00}
\defV{probe/remedy/upscale/mistral32_24/ctrlchange}{+0.03}
\defV{probe/remedy/upscale/mistral32_24/ctrlph}{1.00}
\defV{probe/remedy/upscale/glm46v_9/control}{0.60}
\defV{probe/remedy/upscale/glm46v_9/conflict}{0.41}
\defV{probe/remedy/upscale/glm46v_9/drop}{0.19}
\defV{probe/remedy/upscale/glm46v_9/dropph}{$4\times10^{-4}$}
\defV{probe/remedy/upscale/glm46v_9/gain}{+0.07}
\defV{probe/remedy/upscale/glm46v_9/gainph}{1.00}
\defV{probe/remedy/upscale/glm46v_9/ctrlchange}{+0.04}
\defV{probe/remedy/upscale/glm46v_9/ctrlph}{1.00}
\defV{probe/remedy/upscale/llavaov2_8/control}{0.45}
\defV{probe/remedy/upscale/llavaov2_8/conflict}{0.22}
\defV{probe/remedy/upscale/llavaov2_8/drop}{0.23}
\defV{probe/remedy/upscale/llavaov2_8/dropph}{$3\times10^{-4}$}
\defV{probe/remedy/upscale/llavaov2_8/gain}{$-$0.02}
\defV{probe/remedy/upscale/llavaov2_8/gainph}{1.00}
\defV{probe/remedy/upscale/llavaov2_8/ctrlchange}{+0.01}
\defV{probe/remedy/upscale/llavaov2_8/ctrlph}{1.00}
\defV{probe/remedy/upscale/molmo2o_7/control}{0.49}
\defV{probe/remedy/upscale/molmo2o_7/conflict}{0.49}
\defV{probe/remedy/upscale/molmo2o_7/drop}{0.01}
\defV{probe/remedy/upscale/molmo2o_7/dropph}{1.00}
\defV{probe/remedy/upscale/molmo2o_7/gain}{+0.00}
\defV{probe/remedy/upscale/molmo2o_7/gainph}{1.00}
\defV{probe/remedy/upscale/molmo2o_7/ctrlchange}{$-$0.05}
\defV{probe/remedy/upscale/molmo2o_7/ctrlph}{0.72}
\defV{probe/remedy/upscale/qwen35_4/control}{0.71}
\defV{probe/remedy/upscale/qwen35_4/conflict}{0.36}
\defV{probe/remedy/upscale/qwen35_4/drop}{0.36}
\defV{probe/remedy/upscale/qwen35_4/dropph}{$2\times10^{-13}$}
\defV{probe/remedy/upscale/qwen35_4/gain}{$-$0.02}
\defV{probe/remedy/upscale/qwen35_4/gainph}{1.00}
\defV{probe/remedy/upscale/qwen35_4/ctrlchange}{+0.16}
\defV{probe/remedy/upscale/qwen35_4/ctrlph}{0.003}
\defV{probe/remedy/upscale/gemma4_12/control}{0.52}
\defV{probe/remedy/upscale/gemma4_12/conflict}{0.48}
\defV{probe/remedy/upscale/gemma4_12/drop}{0.04}
\defV{probe/remedy/upscale/gemma4_12/dropph}{0.61}
\defV{probe/remedy/upscale/gemma4_12/gain}{$-$0.01}
\defV{probe/remedy/upscale/gemma4_12/gainph}{1.00}
\defV{probe/remedy/upscale/gemma4_12/ctrlchange}{+0.01}
\defV{probe/remedy/upscale/gemma4_12/ctrlph}{1.00}
\defV{probe/remedy/upscale/nremoved}{0}
\defV{probe/remedy/upscale/nremoved/word}{zero}
\defV{probe/remedy/upscale/nreduced}{0}
\defV{probe/remedy/upscale/nreduced/word}{zero}
\defV{probe/remedy/upscale/nkept}{7}
\defV{probe/remedy/upscale/nkept/word}{seven}
\defV{probe/remedy/upscale/nundecided}{0}
\defV{probe/remedy/upscale/nundecided/word}{zero}
\defV{probe/remedy/upscale/ncostly}{0}
\defV{probe/remedy/upscale/ncostly/word}{zero}
\defV{probe/remedy/upscale/notremoved}{7}
\defV{probe/remedy/upscale/notremoved/word}{seven}
\defV{probe/remedy/upscale/dropgrew}{6}
\defV{probe/remedy/upscale/dropgrew/word}{six}
\defV{probe/remedy/instruction/gemma4_31/control}{0.69}
\defV{probe/remedy/instruction/gemma4_31/conflict}{0.35}
\defV{probe/remedy/instruction/gemma4_31/drop}{0.34}
\defV{probe/remedy/instruction/gemma4_31/dropph}{$5\times10^{-11}$}
\defV{probe/remedy/instruction/gemma4_31/gain}{+0.06}
\defV{probe/remedy/instruction/gemma4_31/gainph}{1.00}
\defV{probe/remedy/instruction/gemma4_31/ctrlchange}{$-$0.04}
\defV{probe/remedy/instruction/gemma4_31/ctrlph}{1.00}
\defV{probe/remedy/instruction/qwen38_27/control}{0.67}
\defV{probe/remedy/instruction/qwen38_27/conflict}{0.17}
\defV{probe/remedy/instruction/qwen38_27/drop}{0.50}
\defV{probe/remedy/instruction/qwen38_27/dropph}{$2\times10^{-17}$}
\defV{probe/remedy/instruction/qwen38_27/gain}{$-$0.04}
\defV{probe/remedy/instruction/qwen38_27/gainph}{1.00}
\defV{probe/remedy/instruction/qwen38_27/ctrlchange}{$-$0.03}
\defV{probe/remedy/instruction/qwen38_27/ctrlph}{1.00}
\defV{probe/remedy/instruction/qwen35_9/control}{0.66}
\defV{probe/remedy/instruction/qwen35_9/conflict}{0.27}
\defV{probe/remedy/instruction/qwen35_9/drop}{0.39}
\defV{probe/remedy/instruction/qwen35_9/dropph}{$4\times10^{-10}$}
\defV{probe/remedy/instruction/qwen35_9/gain}{$-$0.04}
\defV{probe/remedy/instruction/qwen35_9/gainph}{1.00}
\defV{probe/remedy/instruction/qwen35_9/ctrlchange}{$-$0.01}
\defV{probe/remedy/instruction/qwen35_9/ctrlph}{1.00}
\defV{probe/remedy/instruction/internvl35_38/control}{0.59}
\defV{probe/remedy/instruction/internvl35_38/conflict}{0.42}
\defV{probe/remedy/instruction/internvl35_38/drop}{0.17}
\defV{probe/remedy/instruction/internvl35_38/dropph}{0.004}
\defV{probe/remedy/instruction/internvl35_38/gain}{+0.00}
\defV{probe/remedy/instruction/internvl35_38/gainph}{1.00}
\defV{probe/remedy/instruction/internvl35_38/ctrlchange}{+0.01}
\defV{probe/remedy/instruction/internvl35_38/ctrlph}{1.00}
\defV{probe/remedy/instruction/mistral32_24/control}{0.69}
\defV{probe/remedy/instruction/mistral32_24/conflict}{0.31}
\defV{probe/remedy/instruction/mistral32_24/drop}{0.39}
\defV{probe/remedy/instruction/mistral32_24/dropph}{$4\times10^{-9}$}
\defV{probe/remedy/instruction/mistral32_24/gain}{$-$0.04}
\defV{probe/remedy/instruction/mistral32_24/gainph}{1.00}
\defV{probe/remedy/instruction/mistral32_24/ctrlchange}{+0.04}
\defV{probe/remedy/instruction/mistral32_24/ctrlph}{1.00}
\defV{probe/remedy/instruction/glm46v_9/control}{0.56}
\defV{probe/remedy/instruction/glm46v_9/conflict}{0.38}
\defV{probe/remedy/instruction/glm46v_9/drop}{0.19}
\defV{probe/remedy/instruction/glm46v_9/dropph}{0.002}
\defV{probe/remedy/instruction/glm46v_9/gain}{+0.04}
\defV{probe/remedy/instruction/glm46v_9/gainph}{1.00}
\defV{probe/remedy/instruction/glm46v_9/ctrlchange}{+0.00}
\defV{probe/remedy/instruction/glm46v_9/ctrlph}{1.00}
\defV{probe/remedy/instruction/llavaov2_8/control}{0.52}
\defV{probe/remedy/instruction/llavaov2_8/conflict}{0.37}
\defV{probe/remedy/instruction/llavaov2_8/drop}{0.15}
\defV{probe/remedy/instruction/llavaov2_8/dropph}{0.02}
\defV{probe/remedy/instruction/llavaov2_8/gain}{+0.13}
\defV{probe/remedy/instruction/llavaov2_8/gainph}{0.13}
\defV{probe/remedy/instruction/llavaov2_8/ctrlchange}{+0.08}
\defV{probe/remedy/instruction/llavaov2_8/ctrlph}{1.00}
\defV{probe/remedy/instruction/molmo2o_7/control}{0.50}
\defV{probe/remedy/instruction/molmo2o_7/conflict}{0.46}
\defV{probe/remedy/instruction/molmo2o_7/drop}{0.04}
\defV{probe/remedy/instruction/molmo2o_7/dropph}{0.69}
\defV{probe/remedy/instruction/molmo2o_7/gain}{$-$0.03}
\defV{probe/remedy/instruction/molmo2o_7/gainph}{1.00}
\defV{probe/remedy/instruction/molmo2o_7/ctrlchange}{$-$0.04}
\defV{probe/remedy/instruction/molmo2o_7/ctrlph}{1.00}
\defV{probe/remedy/instruction/qwen35_4/control}{0.54}
\defV{probe/remedy/instruction/qwen35_4/conflict}{0.36}
\defV{probe/remedy/instruction/qwen35_4/drop}{0.18}
\defV{probe/remedy/instruction/qwen35_4/dropph}{0.004}
\defV{probe/remedy/instruction/qwen35_4/gain}{$-$0.02}
\defV{probe/remedy/instruction/qwen35_4/gainph}{1.00}
\defV{probe/remedy/instruction/qwen35_4/ctrlchange}{$-$0.01}
\defV{probe/remedy/instruction/qwen35_4/ctrlph}{1.00}
\defV{probe/remedy/instruction/gemma4_12/control}{0.52}
\defV{probe/remedy/instruction/gemma4_12/conflict}{0.49}
\defV{probe/remedy/instruction/gemma4_12/drop}{0.04}
\defV{probe/remedy/instruction/gemma4_12/dropph}{0.69}
\defV{probe/remedy/instruction/gemma4_12/gain}{+0.00}
\defV{probe/remedy/instruction/gemma4_12/gainph}{1.00}
\defV{probe/remedy/instruction/gemma4_12/ctrlchange}{+0.01}
\defV{probe/remedy/instruction/gemma4_12/ctrlph}{1.00}
\defV{probe/remedy/instruction/nremoved}{0}
\defV{probe/remedy/instruction/nremoved/word}{zero}
\defV{probe/remedy/instruction/nreduced}{0}
\defV{probe/remedy/instruction/nreduced/word}{zero}
\defV{probe/remedy/instruction/nkept}{7}
\defV{probe/remedy/instruction/nkept/word}{seven}
\defV{probe/remedy/instruction/nundecided}{0}
\defV{probe/remedy/instruction/nundecided/word}{zero}
\defV{probe/remedy/instruction/ncostly}{0}
\defV{probe/remedy/instruction/ncostly/word}{zero}
\defV{probe/remedy/instruction/notremoved}{7}
\defV{probe/remedy/instruction/notremoved/word}{seven}
\defV{probe/remedy/instruction/dropgrew}{5}
\defV{probe/remedy/instruction/dropgrew/word}{five}
\defV{probe/remedy/self_structure/gemma4_31/control}{0.75}
\defV{probe/remedy/self_structure/gemma4_31/conflict}{0.26}
\defV{probe/remedy/self_structure/gemma4_31/drop}{0.49}
\defV{probe/remedy/self_structure/gemma4_31/dropph}{$2\times10^{-16}$}
\defV{probe/remedy/self_structure/gemma4_31/gain}{$-$0.04}
\defV{probe/remedy/self_structure/gemma4_31/gainph}{1.00}
\defV{probe/remedy/self_structure/gemma4_31/ctrlchange}{+0.02}
\defV{probe/remedy/self_structure/gemma4_31/ctrlph}{1.00}
\defV{probe/remedy/self_structure/qwen38_27/control}{0.84}
\defV{probe/remedy/self_structure/qwen38_27/conflict}{0.26}
\defV{probe/remedy/self_structure/qwen38_27/drop}{0.58}
\defV{probe/remedy/self_structure/qwen38_27/dropph}{$6\times10^{-18}$}
\defV{probe/remedy/self_structure/qwen38_27/gain}{+0.05}
\defV{probe/remedy/self_structure/qwen38_27/gainph}{1.00}
\defV{probe/remedy/self_structure/qwen38_27/ctrlchange}{+0.14}
\defV{probe/remedy/self_structure/qwen38_27/ctrlph}{0.009}
\defV{probe/remedy/self_structure/qwen35_9/control}{0.69}
\defV{probe/remedy/self_structure/qwen35_9/conflict}{0.36}
\defV{probe/remedy/self_structure/qwen35_9/drop}{0.33}
\defV{probe/remedy/self_structure/qwen35_9/dropph}{$6\times10^{-9}$}
\defV{probe/remedy/self_structure/qwen35_9/gain}{+0.05}
\defV{probe/remedy/self_structure/qwen35_9/gainph}{1.00}
\defV{probe/remedy/self_structure/qwen35_9/ctrlchange}{+0.01}
\defV{probe/remedy/self_structure/qwen35_9/ctrlph}{1.00}
\defV{probe/remedy/self_structure/internvl35_38/control}{0.58}
\defV{probe/remedy/self_structure/internvl35_38/conflict}{0.46}
\defV{probe/remedy/self_structure/internvl35_38/drop}{0.11}
\defV{probe/remedy/self_structure/internvl35_38/dropph}{0.12}
\defV{probe/remedy/self_structure/internvl35_38/gain}{+0.04}
\defV{probe/remedy/self_structure/internvl35_38/gainph}{1.00}
\defV{probe/remedy/self_structure/internvl35_38/ctrlchange}{$-$0.01}
\defV{probe/remedy/self_structure/internvl35_38/ctrlph}{1.00}
\defV{probe/remedy/self_structure/mistral32_24/control}{0.51}
\defV{probe/remedy/self_structure/mistral32_24/conflict}{0.46}
\defV{probe/remedy/self_structure/mistral32_24/drop}{0.05}
\defV{probe/remedy/self_structure/mistral32_24/dropph}{0.38}
\defV{probe/remedy/self_structure/mistral32_24/gain}{+0.11}
\defV{probe/remedy/self_structure/mistral32_24/gainph}{0.16}
\defV{probe/remedy/self_structure/mistral32_24/ctrlchange}{$-$0.14}
\defV{probe/remedy/self_structure/mistral32_24/ctrlph}{0.08}
\defV{probe/remedy/self_structure/glm46v_9/control}{0.57}
\defV{probe/remedy/self_structure/glm46v_9/conflict}{0.44}
\defV{probe/remedy/self_structure/glm46v_9/drop}{0.14}
\defV{probe/remedy/self_structure/glm46v_9/dropph}{0.09}
\defV{probe/remedy/self_structure/glm46v_9/gain}{+0.10}
\defV{probe/remedy/self_structure/glm46v_9/gainph}{0.45}
\defV{probe/remedy/self_structure/glm46v_9/ctrlchange}{+0.01}
\defV{probe/remedy/self_structure/glm46v_9/ctrlph}{1.00}
\defV{probe/remedy/self_structure/llavaov2_8/control}{0.57}
\defV{probe/remedy/self_structure/llavaov2_8/conflict}{0.44}
\defV{probe/remedy/self_structure/llavaov2_8/drop}{0.13}
\defV{probe/remedy/self_structure/llavaov2_8/dropph}{0.12}
\defV{probe/remedy/self_structure/llavaov2_8/gain}{+0.20}
\defV{probe/remedy/self_structure/llavaov2_8/gainph}{0.004}
\defV{probe/remedy/self_structure/llavaov2_8/ctrlchange}{+0.13}
\defV{probe/remedy/self_structure/llavaov2_8/ctrlph}{0.18}
\defV{probe/remedy/self_structure/molmo2o_7/control}{0.51}
\defV{probe/remedy/self_structure/molmo2o_7/conflict}{0.48}
\defV{probe/remedy/self_structure/molmo2o_7/drop}{0.03}
\defV{probe/remedy/self_structure/molmo2o_7/dropph}{0.58}
\defV{probe/remedy/self_structure/molmo2o_7/gain}{$-$0.01}
\defV{probe/remedy/self_structure/molmo2o_7/gainph}{1.00}
\defV{probe/remedy/self_structure/molmo2o_7/ctrlchange}{$-$0.04}
\defV{probe/remedy/self_structure/molmo2o_7/ctrlph}{1.00}
\defV{probe/remedy/self_structure/qwen35_4/control}{0.56}
\defV{probe/remedy/self_structure/qwen35_4/conflict}{0.41}
\defV{probe/remedy/self_structure/qwen35_4/drop}{0.14}
\defV{probe/remedy/self_structure/qwen35_4/dropph}{0.09}
\defV{probe/remedy/self_structure/qwen35_4/gain}{+0.04}
\defV{probe/remedy/self_structure/qwen35_4/gainph}{1.00}
\defV{probe/remedy/self_structure/qwen35_4/ctrlchange}{+0.01}
\defV{probe/remedy/self_structure/qwen35_4/ctrlph}{1.00}
\defV{probe/remedy/self_structure/gemma4_12/control}{0.56}
\defV{probe/remedy/self_structure/gemma4_12/conflict}{0.36}
\defV{probe/remedy/self_structure/gemma4_12/drop}{0.20}
\defV{probe/remedy/self_structure/gemma4_12/dropph}{$1\times10^{-5}$}
\defV{probe/remedy/self_structure/gemma4_12/gain}{$-$0.13}
\defV{probe/remedy/self_structure/gemma4_12/gainph}{0.03}
\defV{probe/remedy/self_structure/gemma4_12/ctrlchange}{+0.04}
\defV{probe/remedy/self_structure/gemma4_12/ctrlph}{1.00}
\defV{probe/remedy/self_structure/nremoved}{0}
\defV{probe/remedy/self_structure/nremoved/word}{zero}
\defV{probe/remedy/self_structure/nreduced}{1}
\defV{probe/remedy/self_structure/nreduced/word}{one}
\defV{probe/remedy/self_structure/nkept}{3}
\defV{probe/remedy/self_structure/nkept/word}{three}
\defV{probe/remedy/self_structure/nundecided}{3}
\defV{probe/remedy/self_structure/nundecided/word}{three}
\defV{probe/remedy/self_structure/ncostly}{0}
\defV{probe/remedy/self_structure/ncostly/word}{zero}
\defV{probe/remedy/self_structure/notremoved}{7}
\defV{probe/remedy/self_structure/notremoved/word}{seven}
\defV{probe/remedy/self_structure/dropgrew}{2}
\defV{probe/remedy/self_structure/dropgrew/word}{two}
\defV{probe/remedy/crop/gemma4_31/control}{0.88}
\defV{probe/remedy/crop/gemma4_31/conflict}{0.28}
\defV{probe/remedy/crop/gemma4_31/drop}{0.60}
\defV{probe/remedy/crop/gemma4_31/dropph}{$9\times10^{-25}$}
\defV{probe/remedy/crop/gemma4_31/gain}{$-$0.01}
\defV{probe/remedy/crop/gemma4_31/gainph}{1.00}
\defV{probe/remedy/crop/gemma4_31/ctrlchange}{+0.15}
\defV{probe/remedy/crop/gemma4_31/ctrlph}{0.005}
\defV{probe/remedy/crop/qwen38_27/control}{0.89}
\defV{probe/remedy/crop/qwen38_27/conflict}{0.21}
\defV{probe/remedy/crop/qwen38_27/drop}{0.67}
\defV{probe/remedy/crop/qwen38_27/dropph}{$8\times10^{-23}$}
\defV{probe/remedy/crop/qwen38_27/gain}{+0.01}
\defV{probe/remedy/crop/qwen38_27/gainph}{1.00}
\defV{probe/remedy/crop/qwen38_27/ctrlchange}{+0.19}
\defV{probe/remedy/crop/qwen38_27/ctrlph}{$4\times10^{-4}$}
\defV{probe/remedy/crop/qwen35_9/control}{0.89}
\defV{probe/remedy/crop/qwen35_9/conflict}{0.23}
\defV{probe/remedy/crop/qwen35_9/drop}{0.66}
\defV{probe/remedy/crop/qwen35_9/dropph}{$4\times10^{-27}$}
\defV{probe/remedy/crop/qwen35_9/gain}{$-$0.08}
\defV{probe/remedy/crop/qwen35_9/gainph}{0.81}
\defV{probe/remedy/crop/qwen35_9/ctrlchange}{+0.21}
\defV{probe/remedy/crop/qwen35_9/ctrlph}{$5\times10^{-6}$}
\defV{probe/remedy/crop/internvl35_38/control}{0.68}
\defV{probe/remedy/crop/internvl35_38/conflict}{0.39}
\defV{probe/remedy/crop/internvl35_38/drop}{0.29}
\defV{probe/remedy/crop/internvl35_38/dropph}{$2\times10^{-9}$}
\defV{probe/remedy/crop/internvl35_38/gain}{$-$0.03}
\defV{probe/remedy/crop/internvl35_38/gainph}{1.00}
\defV{probe/remedy/crop/internvl35_38/ctrlchange}{+0.09}
\defV{probe/remedy/crop/internvl35_38/ctrlph}{0.37}
\defV{probe/remedy/crop/mistral32_24/control}{0.74}
\defV{probe/remedy/crop/mistral32_24/conflict}{0.34}
\defV{probe/remedy/crop/mistral32_24/drop}{0.41}
\defV{probe/remedy/crop/mistral32_24/dropph}{$8\times10^{-11}$}
\defV{probe/remedy/crop/mistral32_24/gain}{$-$0.01}
\defV{probe/remedy/crop/mistral32_24/gainph}{1.00}
\defV{probe/remedy/crop/mistral32_24/ctrlchange}{+0.09}
\defV{probe/remedy/crop/mistral32_24/ctrlph}{0.37}
\defV{probe/remedy/crop/glm46v_9/control}{0.52}
\defV{probe/remedy/crop/glm46v_9/conflict}{0.50}
\defV{probe/remedy/crop/glm46v_9/drop}{0.02}
\defV{probe/remedy/crop/glm46v_9/dropph}{0.45}
\defV{probe/remedy/crop/glm46v_9/gain}{+0.16}
\defV{probe/remedy/crop/glm46v_9/gainph}{0.001}
\defV{probe/remedy/crop/glm46v_9/ctrlchange}{$-$0.04}
\defV{probe/remedy/crop/glm46v_9/ctrlph}{0.84}
\defV{probe/remedy/crop/llavaov2_8/control}{0.67}
\defV{probe/remedy/crop/llavaov2_8/conflict}{0.34}
\defV{probe/remedy/crop/llavaov2_8/drop}{0.34}
\defV{probe/remedy/crop/llavaov2_8/dropph}{$3\times10^{-7}$}
\defV{probe/remedy/crop/llavaov2_8/gain}{+0.09}
\defV{probe/remedy/crop/llavaov2_8/gainph}{0.64}
\defV{probe/remedy/crop/llavaov2_8/ctrlchange}{+0.23}
\defV{probe/remedy/crop/llavaov2_8/ctrlph}{$2\times10^{-4}$}
\defV{probe/remedy/crop/molmo2o_7/control}{0.51}
\defV{probe/remedy/crop/molmo2o_7/conflict}{0.44}
\defV{probe/remedy/crop/molmo2o_7/drop}{0.07}
\defV{probe/remedy/crop/molmo2o_7/dropph}{0.22}
\defV{probe/remedy/crop/molmo2o_7/gain}{$-$0.05}
\defV{probe/remedy/crop/molmo2o_7/gainph}{1.00}
\defV{probe/remedy/crop/molmo2o_7/ctrlchange}{$-$0.04}
\defV{probe/remedy/crop/molmo2o_7/ctrlph}{0.84}
\defV{probe/remedy/crop/qwen35_4/control}{0.79}
\defV{probe/remedy/crop/qwen35_4/conflict}{0.27}
\defV{probe/remedy/crop/qwen35_4/drop}{0.51}
\defV{probe/remedy/crop/qwen35_4/dropph}{$5\times10^{-19}$}
\defV{probe/remedy/crop/qwen35_4/gain}{$-$0.11}
\defV{probe/remedy/crop/qwen35_4/gainph}{0.22}
\defV{probe/remedy/crop/qwen35_4/ctrlchange}{+0.24}
\defV{probe/remedy/crop/qwen35_4/ctrlph}{$2\times10^{-5}$}
\defV{probe/remedy/crop/gemma4_12/control}{0.81}
\defV{probe/remedy/crop/gemma4_12/conflict}{0.36}
\defV{probe/remedy/crop/gemma4_12/drop}{0.44}
\defV{probe/remedy/crop/gemma4_12/dropph}{$4\times10^{-17}$}
\defV{probe/remedy/crop/gemma4_12/gain}{$-$0.12}
\defV{probe/remedy/crop/gemma4_12/gainph}{0.01}
\defV{probe/remedy/crop/gemma4_12/ctrlchange}{+0.29}
\defV{probe/remedy/crop/gemma4_12/ctrlph}{$2\times10^{-8}$}
\defV{probe/remedy/crop/nremoved}{0}
\defV{probe/remedy/crop/nremoved/word}{zero}
\defV{probe/remedy/crop/nreduced}{0}
\defV{probe/remedy/crop/nreduced/word}{zero}
\defV{probe/remedy/crop/nkept}{7}
\defV{probe/remedy/crop/nkept/word}{seven}
\defV{probe/remedy/crop/nundecided}{0}
\defV{probe/remedy/crop/nundecided/word}{zero}
\defV{probe/remedy/crop/ncostly}{0}
\defV{probe/remedy/crop/ncostly/word}{zero}
\defV{probe/remedy/crop/notremoved}{7}
\defV{probe/remedy/crop/notremoved/word}{seven}
\defV{probe/remedy/crop/dropgrew}{7}
\defV{probe/remedy/crop/dropgrew/word}{seven}
\defV{probe/remedy/nconds}{4}
\defV{probe/remedy/nconds/word}{four}
\defV{probe/remedy/removedany}{0}
\defV{probe/remedy/removedany/word}{zero}
\defV{probe/prompt/models}{7}
\defV{probe/prompt/models/word}{seven}
\defV{probe/prompt/sigmodels}{7}
\defV{probe/prompt/sigmodels/word}{seven}
\defV{probe/prompt/plain/gemma4_31/control}{0.73}
\defV{probe/prompt/plain/gemma4_31/conflict}{0.29}
\defV{probe/prompt/plain/gemma4_31/drop}{0.44}
\defV{probe/prompt/plain/gemma4_31/dropph}{$1\times10^{-16}$}
\defV{probe/prompt/plain/qwen38_27/control}{0.70}
\defV{probe/prompt/plain/qwen38_27/conflict}{0.21}
\defV{probe/prompt/plain/qwen38_27/drop}{0.49}
\defV{probe/prompt/plain/qwen38_27/dropph}{$5\times10^{-16}$}
\defV{probe/prompt/plain/qwen35_9/control}{0.67}
\defV{probe/prompt/plain/qwen35_9/conflict}{0.31}
\defV{probe/prompt/plain/qwen35_9/drop}{0.36}
\defV{probe/prompt/plain/qwen35_9/dropph}{$8\times10^{-12}$}
\defV{probe/prompt/plain/internvl35_38/control}{0.59}
\defV{probe/prompt/plain/internvl35_38/conflict}{0.42}
\defV{probe/prompt/plain/internvl35_38/drop}{0.16}
\defV{probe/prompt/plain/internvl35_38/dropph}{0.004}
\defV{probe/prompt/plain/mistral32_24/control}{0.66}
\defV{probe/prompt/plain/mistral32_24/conflict}{0.35}
\defV{probe/prompt/plain/mistral32_24/drop}{0.31}
\defV{probe/prompt/plain/mistral32_24/dropph}{$7\times10^{-7}$}
\defV{probe/prompt/plain/llavaov2_8/control}{0.44}
\defV{probe/prompt/plain/llavaov2_8/conflict}{0.24}
\defV{probe/prompt/plain/llavaov2_8/drop}{0.20}
\defV{probe/prompt/plain/llavaov2_8/dropph}{0.001}
\defV{probe/prompt/plain/qwen35_4/control}{0.55}
\defV{probe/prompt/plain/qwen35_4/conflict}{0.38}
\defV{probe/prompt/plain/qwen35_4/drop}{0.17}
\defV{probe/prompt/plain/qwen35_4/dropph}{0.002}
\defV{probe/prompt/fewshot/gemma4_31/control}{0.90}
\defV{probe/prompt/fewshot/gemma4_31/conflict}{0.25}
\defV{probe/prompt/fewshot/gemma4_31/drop}{0.65}
\defV{probe/prompt/fewshot/gemma4_31/dropph}{$9\times10^{-23}$}
\defV{probe/prompt/fewshot/gemma4_31/gain}{$-$0.04}
\defV{probe/prompt/fewshot/gemma4_31/gainph}{1.00}
\defV{probe/prompt/fewshot/gemma4_31/ctrlchange}{+0.17}
\defV{probe/prompt/fewshot/gemma4_31/ctrlph}{$6\times10^{-6}$}
\defV{probe/prompt/fewshot/qwen38_27/control}{0.81}
\defV{probe/prompt/fewshot/qwen38_27/conflict}{0.17}
\defV{probe/prompt/fewshot/qwen38_27/drop}{0.64}
\defV{probe/prompt/fewshot/qwen38_27/dropph}{$8\times10^{-22}$}
\defV{probe/prompt/fewshot/qwen38_27/gain}{$-$0.04}
\defV{probe/prompt/fewshot/qwen38_27/gainph}{1.00}
\defV{probe/prompt/fewshot/qwen38_27/ctrlchange}{+0.11}
\defV{probe/prompt/fewshot/qwen38_27/ctrlph}{0.05}
\defV{probe/prompt/fewshot/qwen35_9/control}{0.59}
\defV{probe/prompt/fewshot/qwen35_9/conflict}{0.44}
\defV{probe/prompt/fewshot/qwen35_9/drop}{0.16}
\defV{probe/prompt/fewshot/qwen35_9/dropph}{0.001}
\defV{probe/prompt/fewshot/qwen35_9/gain}{+0.13}
\defV{probe/prompt/fewshot/qwen35_9/gainph}{0.09}
\defV{probe/prompt/fewshot/qwen35_9/ctrlchange}{$-$0.08}
\defV{probe/prompt/fewshot/qwen35_9/ctrlph}{0.43}
\defV{probe/prompt/fewshot/internvl35_38/control}{0.81}
\defV{probe/prompt/fewshot/internvl35_38/conflict}{0.24}
\defV{probe/prompt/fewshot/internvl35_38/drop}{0.56}
\defV{probe/prompt/fewshot/internvl35_38/dropph}{$1\times10^{-17}$}
\defV{probe/prompt/fewshot/internvl35_38/gain}{$-$0.18}
\defV{probe/prompt/fewshot/internvl35_38/gainph}{0.02}
\defV{probe/prompt/fewshot/internvl35_38/ctrlchange}{+0.22}
\defV{probe/prompt/fewshot/internvl35_38/ctrlph}{$5\times10^{-4}$}
\defV{probe/prompt/fewshot/mistral32_24/control}{0.66}
\defV{probe/prompt/fewshot/mistral32_24/conflict}{0.31}
\defV{probe/prompt/fewshot/mistral32_24/drop}{0.35}
\defV{probe/prompt/fewshot/mistral32_24/dropph}{$1\times10^{-8}$}
\defV{probe/prompt/fewshot/mistral32_24/gain}{$-$0.04}
\defV{probe/prompt/fewshot/mistral32_24/gainph}{1.00}
\defV{probe/prompt/fewshot/mistral32_24/ctrlchange}{+0.00}
\defV{probe/prompt/fewshot/mistral32_24/ctrlph}{1.00}
\defV{probe/prompt/fewshot/llavaov2_8/control}{0.41}
\defV{probe/prompt/fewshot/llavaov2_8/conflict}{0.13}
\defV{probe/prompt/fewshot/llavaov2_8/drop}{0.29}
\defV{probe/prompt/fewshot/llavaov2_8/dropph}{$3\times10^{-7}$}
\defV{probe/prompt/fewshot/llavaov2_8/gain}{$-$0.11}
\defV{probe/prompt/fewshot/llavaov2_8/gainph}{0.07}
\defV{probe/prompt/fewshot/llavaov2_8/ctrlchange}{$-$0.03}
\defV{probe/prompt/fewshot/llavaov2_8/ctrlph}{1.00}
\defV{probe/prompt/fewshot/qwen35_4/control}{0.60}
\defV{probe/prompt/fewshot/qwen35_4/conflict}{0.37}
\defV{probe/prompt/fewshot/qwen35_4/drop}{0.23}
\defV{probe/prompt/fewshot/qwen35_4/dropph}{$4\times10^{-6}$}
\defV{probe/prompt/fewshot/qwen35_4/gain}{$-$0.01}
\defV{probe/prompt/fewshot/qwen35_4/gainph}{1.00}
\defV{probe/prompt/fewshot/qwen35_4/ctrlchange}{+0.05}
\defV{probe/prompt/fewshot/qwen35_4/ctrlph}{0.93}
\defV{probe/prompt/fewshot/nremoved}{0}
\defV{probe/prompt/fewshot/nremoved/word}{zero}
\defV{probe/prompt/fewshot/nreduced}{0}
\defV{probe/prompt/fewshot/nreduced/word}{zero}
\defV{probe/prompt/fewshot/nkept}{7}
\defV{probe/prompt/fewshot/nkept/word}{seven}
\defV{probe/prompt/fewshot/nundecided}{0}
\defV{probe/prompt/fewshot/nundecided/word}{zero}
\defV{probe/prompt/fewshot/ncostly}{0}
\defV{probe/prompt/fewshot/ncostly/word}{zero}
\defV{probe/prompt/fewshot/notremoved}{7}
\defV{probe/prompt/fewshot/notremoved/word}{seven}
\defV{probe/prompt/fewshot/dropgrew}{6}
\defV{probe/prompt/fewshot/dropgrew/word}{six}
\defV{probe/prompt/procedure/gemma4_31/control}{0.89}
\defV{probe/prompt/procedure/gemma4_31/conflict}{0.16}
\defV{probe/prompt/procedure/gemma4_31/drop}{0.72}
\defV{probe/prompt/procedure/gemma4_31/dropph}{$1\times10^{-28}$}
\defV{probe/prompt/procedure/gemma4_31/gain}{$-$0.13}
\defV{probe/prompt/procedure/gemma4_31/gainph}{0.01}
\defV{probe/prompt/procedure/gemma4_31/ctrlchange}{+0.16}
\defV{probe/prompt/procedure/gemma4_31/ctrlph}{$8\times10^{-4}$}
\defV{probe/prompt/procedure/qwen38_27/control}{0.81}
\defV{probe/prompt/procedure/qwen38_27/conflict}{0.26}
\defV{probe/prompt/procedure/qwen38_27/drop}{0.55}
\defV{probe/prompt/procedure/qwen38_27/dropph}{$3\times10^{-16}$}
\defV{probe/prompt/procedure/qwen38_27/gain}{+0.05}
\defV{probe/prompt/procedure/qwen38_27/gainph}{0.61}
\defV{probe/prompt/procedure/qwen38_27/ctrlchange}{+0.11}
\defV{probe/prompt/procedure/qwen38_27/ctrlph}{0.02}
\defV{probe/prompt/procedure/qwen35_9/control}{0.81}
\defV{probe/prompt/procedure/qwen35_9/conflict}{0.24}
\defV{probe/prompt/procedure/qwen35_9/drop}{0.58}
\defV{probe/prompt/procedure/qwen35_9/dropph}{$4\times10^{-17}$}
\defV{probe/prompt/procedure/qwen35_9/gain}{$-$0.07}
\defV{probe/prompt/procedure/qwen35_9/gainph}{0.61}
\defV{probe/prompt/procedure/qwen35_9/ctrlchange}{+0.14}
\defV{probe/prompt/procedure/qwen35_9/ctrlph}{0.01}
\defV{probe/prompt/procedure/internvl35_38/control}{0.69}
\defV{probe/prompt/procedure/internvl35_38/conflict}{0.36}
\defV{probe/prompt/procedure/internvl35_38/drop}{0.33}
\defV{probe/prompt/procedure/internvl35_38/dropph}{$1\times10^{-6}$}
\defV{probe/prompt/procedure/internvl35_38/gain}{$-$0.06}
\defV{probe/prompt/procedure/internvl35_38/gainph}{0.61}
\defV{probe/prompt/procedure/internvl35_38/ctrlchange}{+0.10}
\defV{probe/prompt/procedure/internvl35_38/ctrlph}{0.18}
\defV{probe/prompt/procedure/mistral32_24/control}{0.46}
\defV{probe/prompt/procedure/mistral32_24/conflict}{0.48}
\defV{probe/prompt/procedure/mistral32_24/drop}{$-$0.01}
\defV{probe/prompt/procedure/mistral32_24/dropph}{0.89}
\defV{probe/prompt/procedure/mistral32_24/gain}{+0.13}
\defV{probe/prompt/procedure/mistral32_24/gainph}{0.09}
\defV{probe/prompt/procedure/mistral32_24/ctrlchange}{$-$0.19}
\defV{probe/prompt/procedure/mistral32_24/ctrlph}{0.004}
\defV{probe/prompt/procedure/llavaov2_8/control}{0.49}
\defV{probe/prompt/procedure/llavaov2_8/conflict}{0.39}
\defV{probe/prompt/procedure/llavaov2_8/drop}{0.09}
\defV{probe/prompt/procedure/llavaov2_8/dropph}{0.06}
\defV{probe/prompt/procedure/llavaov2_8/gain}{+0.15}
\defV{probe/prompt/procedure/llavaov2_8/gainph}{0.08}
\defV{probe/prompt/procedure/llavaov2_8/ctrlchange}{+0.04}
\defV{probe/prompt/procedure/llavaov2_8/ctrlph}{0.52}
\defV{probe/prompt/procedure/qwen35_4/control}{0.71}
\defV{probe/prompt/procedure/qwen35_4/conflict}{0.28}
\defV{probe/prompt/procedure/qwen35_4/drop}{0.43}
\defV{probe/prompt/procedure/qwen35_4/dropph}{$6\times10^{-11}$}
\defV{probe/prompt/procedure/qwen35_4/gain}{$-$0.10}
\defV{probe/prompt/procedure/qwen35_4/gainph}{0.35}
\defV{probe/prompt/procedure/qwen35_4/ctrlchange}{+0.16}
\defV{probe/prompt/procedure/qwen35_4/ctrlph}{0.01}
\defV{probe/prompt/procedure/nremoved}{0}
\defV{probe/prompt/procedure/nremoved/word}{zero}
\defV{probe/prompt/procedure/nreduced}{0}
\defV{probe/prompt/procedure/nreduced/word}{zero}
\defV{probe/prompt/procedure/nkept}{5}
\defV{probe/prompt/procedure/nkept/word}{five}
\defV{probe/prompt/procedure/nundecided}{2}
\defV{probe/prompt/procedure/nundecided/word}{two}
\defV{probe/prompt/procedure/ncostly}{1}
\defV{probe/prompt/procedure/ncostly/word}{one}
\defV{probe/prompt/procedure/notremoved}{7}
\defV{probe/prompt/procedure/notremoved/word}{seven}
\defV{probe/prompt/procedure/dropgrew}{5}
\defV{probe/prompt/procedure/dropgrew/word}{five}
\defV{probe/prompt/ordering_check/gemma4_31/control}{0.87}
\defV{probe/prompt/ordering_check/gemma4_31/conflict}{0.19}
\defV{probe/prompt/ordering_check/gemma4_31/drop}{0.68}
\defV{probe/prompt/ordering_check/gemma4_31/dropph}{$4\times10^{-28}$}
\defV{probe/prompt/ordering_check/gemma4_31/gain}{$-$0.10}
\defV{probe/prompt/ordering_check/gemma4_31/gainph}{0.20}
\defV{probe/prompt/ordering_check/gemma4_31/ctrlchange}{+0.14}
\defV{probe/prompt/ordering_check/gemma4_31/ctrlph}{0.001}
\defV{probe/prompt/ordering_check/qwen38_27/control}{0.63}
\defV{probe/prompt/ordering_check/qwen38_27/conflict}{0.17}
\defV{probe/prompt/ordering_check/qwen38_27/drop}{0.46}
\defV{probe/prompt/ordering_check/qwen38_27/dropph}{$1\times10^{-13}$}
\defV{probe/prompt/ordering_check/qwen38_27/gain}{$-$0.04}
\defV{probe/prompt/ordering_check/qwen38_27/gainph}{1.00}
\defV{probe/prompt/ordering_check/qwen38_27/ctrlchange}{$-$0.07}
\defV{probe/prompt/ordering_check/qwen38_27/ctrlph}{0.52}
\defV{probe/prompt/ordering_check/qwen35_9/control}{0.76}
\defV{probe/prompt/ordering_check/qwen35_9/conflict}{0.29}
\defV{probe/prompt/ordering_check/qwen35_9/drop}{0.47}
\defV{probe/prompt/ordering_check/qwen35_9/dropph}{$1\times10^{-13}$}
\defV{probe/prompt/ordering_check/qwen35_9/gain}{$-$0.02}
\defV{probe/prompt/ordering_check/qwen35_9/gainph}{1.00}
\defV{probe/prompt/ordering_check/qwen35_9/ctrlchange}{+0.09}
\defV{probe/prompt/ordering_check/qwen35_9/ctrlph}{0.52}
\defV{probe/prompt/ordering_check/internvl35_38/control}{0.69}
\defV{probe/prompt/ordering_check/internvl35_38/conflict}{0.35}
\defV{probe/prompt/ordering_check/internvl35_38/drop}{0.34}
\defV{probe/prompt/ordering_check/internvl35_38/dropph}{$3\times10^{-9}$}
\defV{probe/prompt/ordering_check/internvl35_38/gain}{$-$0.07}
\defV{probe/prompt/ordering_check/internvl35_38/gainph}{1.00}
\defV{probe/prompt/ordering_check/internvl35_38/ctrlchange}{+0.10}
\defV{probe/prompt/ordering_check/internvl35_38/ctrlph}{0.46}
\defV{probe/prompt/ordering_check/mistral32_24/control}{0.74}
\defV{probe/prompt/ordering_check/mistral32_24/conflict}{0.32}
\defV{probe/prompt/ordering_check/mistral32_24/drop}{0.41}
\defV{probe/prompt/ordering_check/mistral32_24/dropph}{$3\times10^{-12}$}
\defV{probe/prompt/ordering_check/mistral32_24/gain}{$-$0.03}
\defV{probe/prompt/ordering_check/mistral32_24/gainph}{1.00}
\defV{probe/prompt/ordering_check/mistral32_24/ctrlchange}{+0.08}
\defV{probe/prompt/ordering_check/mistral32_24/ctrlph}{0.52}
\defV{probe/prompt/ordering_check/llavaov2_8/control}{0.36}
\defV{probe/prompt/ordering_check/llavaov2_8/conflict}{0.26}
\defV{probe/prompt/ordering_check/llavaov2_8/drop}{0.09}
\defV{probe/prompt/ordering_check/llavaov2_8/dropph}{0.05}
\defV{probe/prompt/ordering_check/llavaov2_8/gain}{+0.02}
\defV{probe/prompt/ordering_check/llavaov2_8/gainph}{1.00}
\defV{probe/prompt/ordering_check/llavaov2_8/ctrlchange}{$-$0.09}
\defV{probe/prompt/ordering_check/llavaov2_8/ctrlph}{0.52}
\defV{probe/prompt/ordering_check/qwen35_4/control}{0.53}
\defV{probe/prompt/ordering_check/qwen35_4/conflict}{0.27}
\defV{probe/prompt/ordering_check/qwen35_4/drop}{0.26}
\defV{probe/prompt/ordering_check/qwen35_4/dropph}{$4\times10^{-5}$}
\defV{probe/prompt/ordering_check/qwen35_4/gain}{$-$0.11}
\defV{probe/prompt/ordering_check/qwen35_4/gainph}{0.35}
\defV{probe/prompt/ordering_check/qwen35_4/ctrlchange}{$-$0.02}
\defV{probe/prompt/ordering_check/qwen35_4/ctrlph}{0.78}
\defV{probe/prompt/ordering_check/nremoved}{0}
\defV{probe/prompt/ordering_check/nremoved/word}{zero}
\defV{probe/prompt/ordering_check/nreduced}{0}
\defV{probe/prompt/ordering_check/nreduced/word}{zero}
\defV{probe/prompt/ordering_check/nkept}{6}
\defV{probe/prompt/ordering_check/nkept/word}{six}
\defV{probe/prompt/ordering_check/nundecided}{1}
\defV{probe/prompt/ordering_check/nundecided/word}{one}
\defV{probe/prompt/ordering_check/ncostly}{0}
\defV{probe/prompt/ordering_check/ncostly/word}{zero}
\defV{probe/prompt/ordering_check/notremoved}{7}
\defV{probe/prompt/ordering_check/notremoved/word}{seven}
\defV{probe/prompt/ordering_check/dropgrew}{5}
\defV{probe/prompt/ordering_check/dropgrew/word}{five}
\defV{probe/prompt/nconds}{3}
\defV{probe/prompt/nconds/word}{three}
\defV{probe/prompt/removedany}{0}
\defV{probe/prompt/removedany/word}{zero}
\defV{probe/tool/models}{5}
\defV{probe/tool/models/word}{five}
\defV{probe/tool/sigmodels}{5}
\defV{probe/tool/sigmodels/word}{five}
\defV{probe/tool/plain/gemma4_31/control}{0.73}
\defV{probe/tool/plain/gemma4_31/conflict}{0.29}
\defV{probe/tool/plain/gemma4_31/drop}{0.44}
\defV{probe/tool/plain/gemma4_31/dropph}{$7\times10^{-17}$}
\defV{probe/tool/plain/qwen38_27/control}{0.70}
\defV{probe/tool/plain/qwen38_27/conflict}{0.21}
\defV{probe/tool/plain/qwen38_27/drop}{0.49}
\defV{probe/tool/plain/qwen38_27/dropph}{$3\times10^{-16}$}
\defV{probe/tool/plain/qwen35_9/control}{0.67}
\defV{probe/tool/plain/qwen35_9/conflict}{0.31}
\defV{probe/tool/plain/qwen35_9/drop}{0.36}
\defV{probe/tool/plain/qwen35_9/dropph}{$5\times10^{-12}$}
\defV{probe/tool/plain/mistral32_24/control}{0.66}
\defV{probe/tool/plain/mistral32_24/conflict}{0.35}
\defV{probe/tool/plain/mistral32_24/drop}{0.31}
\defV{probe/tool/plain/mistral32_24/dropph}{$3\times10^{-7}$}
\defV{probe/tool/plain/qwen35_4/control}{0.55}
\defV{probe/tool/plain/qwen35_4/conflict}{0.38}
\defV{probe/tool/plain/qwen35_4/drop}{0.17}
\defV{probe/tool/plain/qwen35_4/dropph}{0.001}
\defV{probe/tool/oracle/gemma4_31/control}{0.61}
\defV{probe/tool/oracle/gemma4_31/conflict}{0.33}
\defV{probe/tool/oracle/gemma4_31/drop}{0.29}
\defV{probe/tool/oracle/gemma4_31/dropph}{$2\times10^{-9}$}
\defV{probe/tool/oracle/gemma4_31/gain}{+0.04}
\defV{probe/tool/oracle/gemma4_31/gainph}{0.47}
\defV{probe/tool/oracle/gemma4_31/ctrlchange}{$-$0.11}
\defV{probe/tool/oracle/gemma4_31/ctrlph}{0.04}
\defV{probe/tool/oracle/gemma4_31/tooluse}{1.00}
\defV{probe/tool/oracle/qwen38_27/control}{0.73}
\defV{probe/tool/oracle/qwen38_27/conflict}{0.57}
\defV{probe/tool/oracle/qwen38_27/drop}{0.16}
\defV{probe/tool/oracle/qwen38_27/dropph}{0.008}
\defV{probe/tool/oracle/qwen38_27/gain}{+0.36}
\defV{probe/tool/oracle/qwen38_27/gainph}{$5\times10^{-10}$}
\defV{probe/tool/oracle/qwen38_27/ctrlchange}{+0.03}
\defV{probe/tool/oracle/qwen38_27/ctrlph}{0.67}
\defV{probe/tool/oracle/qwen38_27/tooluse}{1.00}
\defV{probe/tool/oracle/qwen35_9/control}{0.17}
\defV{probe/tool/oracle/qwen35_9/conflict}{0.04}
\defV{probe/tool/oracle/qwen35_9/drop}{0.13}
\defV{probe/tool/oracle/qwen35_9/dropph}{0.001}
\defV{probe/tool/oracle/qwen35_9/gain}{$-$0.26}
\defV{probe/tool/oracle/qwen35_9/gainph}{$1\times10^{-8}$}
\defV{probe/tool/oracle/qwen35_9/ctrlchange}{$-$0.50}
\defV{probe/tool/oracle/qwen35_9/ctrlph}{$1\times10^{-17}$}
\defV{probe/tool/oracle/qwen35_9/tooluse}{0.99}
\defV{probe/tool/oracle/mistral32_24/control}{0.54}
\defV{probe/tool/oracle/mistral32_24/conflict}{0.46}
\defV{probe/tool/oracle/mistral32_24/drop}{0.08}
\defV{probe/tool/oracle/mistral32_24/dropph}{0.23}
\defV{probe/tool/oracle/mistral32_24/gain}{+0.11}
\defV{probe/tool/oracle/mistral32_24/gainph}{0.09}
\defV{probe/tool/oracle/mistral32_24/ctrlchange}{$-$0.12}
\defV{probe/tool/oracle/mistral32_24/ctrlph}{0.07}
\defV{probe/tool/oracle/mistral32_24/tooluse}{0.05}
\defV{probe/tool/oracle/qwen35_4/control}{0.27}
\defV{probe/tool/oracle/qwen35_4/conflict}{0.21}
\defV{probe/tool/oracle/qwen35_4/drop}{0.06}
\defV{probe/tool/oracle/qwen35_4/dropph}{0.23}
\defV{probe/tool/oracle/qwen35_4/gain}{$-$0.17}
\defV{probe/tool/oracle/qwen35_4/gainph}{0.002}
\defV{probe/tool/oracle/qwen35_4/ctrlchange}{$-$0.28}
\defV{probe/tool/oracle/qwen35_4/ctrlph}{$6\times10^{-7}$}
\defV{probe/tool/oracle/qwen35_4/tooluse}{0.56}
\defV{probe/tool/oracle/nremoved}{0}
\defV{probe/tool/oracle/nremoved/word}{zero}
\defV{probe/tool/oracle/nreduced}{1}
\defV{probe/tool/oracle/nreduced/word}{one}
\defV{probe/tool/oracle/nkept}{2}
\defV{probe/tool/oracle/nkept/word}{two}
\defV{probe/tool/oracle/nundecided}{2}
\defV{probe/tool/oracle/nundecided/word}{two}
\defV{probe/tool/oracle/ncostly}{3}
\defV{probe/tool/oracle/ncostly/word}{three}
\defV{probe/tool/oracle/notremoved}{5}
\defV{probe/tool/oracle/notremoved/word}{five}
\defV{probe/tool/oracle/dropgrew}{0}
\defV{probe/tool/oracle/dropgrew/word}{zero}
\defV{probe/tool/ocr/gemma4_31/control}{0.75}
\defV{probe/tool/ocr/gemma4_31/conflict}{0.37}
\defV{probe/tool/ocr/gemma4_31/drop}{0.38}
\defV{probe/tool/ocr/gemma4_31/dropph}{$9\times10^{-16}$}
\defV{probe/tool/ocr/gemma4_31/gain}{+0.08}
\defV{probe/tool/ocr/gemma4_31/gainph}{0.17}
\defV{probe/tool/ocr/gemma4_31/ctrlchange}{+0.02}
\defV{probe/tool/ocr/gemma4_31/ctrlph}{1.00}
\defV{probe/tool/ocr/gemma4_31/tooluse}{1.00}
\defV{probe/tool/ocr/qwen38_27/control}{0.89}
\defV{probe/tool/ocr/qwen38_27/conflict}{0.26}
\defV{probe/tool/ocr/qwen38_27/drop}{0.62}
\defV{probe/tool/ocr/qwen38_27/dropph}{$6\times10^{-26}$}
\defV{probe/tool/ocr/qwen38_27/gain}{+0.06}
\defV{probe/tool/ocr/qwen38_27/gainph}{1.00}
\defV{probe/tool/ocr/qwen38_27/ctrlchange}{+0.19}
\defV{probe/tool/ocr/qwen38_27/ctrlph}{$1\times10^{-4}$}
\defV{probe/tool/ocr/qwen38_27/tooluse}{1.00}
\defV{probe/tool/ocr/qwen35_9/control}{0.33}
\defV{probe/tool/ocr/qwen35_9/conflict}{0.26}
\defV{probe/tool/ocr/qwen35_9/drop}{0.06}
\defV{probe/tool/ocr/qwen35_9/dropph}{0.24}
\defV{probe/tool/ocr/qwen35_9/gain}{$-$0.04}
\defV{probe/tool/ocr/qwen35_9/gainph}{1.00}
\defV{probe/tool/ocr/qwen35_9/ctrlchange}{$-$0.34}
\defV{probe/tool/ocr/qwen35_9/ctrlph}{$4\times10^{-8}$}
\defV{probe/tool/ocr/qwen35_9/tooluse}{0.98}
\defV{probe/tool/ocr/mistral32_24/control}{0.66}
\defV{probe/tool/ocr/mistral32_24/conflict}{0.36}
\defV{probe/tool/ocr/mistral32_24/drop}{0.30}
\defV{probe/tool/ocr/mistral32_24/dropph}{$5\times10^{-8}$}
\defV{probe/tool/ocr/mistral32_24/gain}{+0.01}
\defV{probe/tool/ocr/mistral32_24/gainph}{1.00}
\defV{probe/tool/ocr/mistral32_24/ctrlchange}{+0.00}
\defV{probe/tool/ocr/mistral32_24/ctrlph}{1.00}
\defV{probe/tool/ocr/mistral32_24/tooluse}{0.01}
\defV{probe/tool/ocr/qwen35_4/control}{0.49}
\defV{probe/tool/ocr/qwen35_4/conflict}{0.36}
\defV{probe/tool/ocr/qwen35_4/drop}{0.12}
\defV{probe/tool/ocr/qwen35_4/dropph}{0.04}
\defV{probe/tool/ocr/qwen35_4/gain}{$-$0.01}
\defV{probe/tool/ocr/qwen35_4/gainph}{1.00}
\defV{probe/tool/ocr/qwen35_4/ctrlchange}{$-$0.06}
\defV{probe/tool/ocr/qwen35_4/ctrlph}{0.82}
\defV{probe/tool/ocr/qwen35_4/tooluse}{1.00}
\defV{probe/tool/ocr/nremoved}{0}
\defV{probe/tool/ocr/nremoved/word}{zero}
\defV{probe/tool/ocr/nreduced}{0}
\defV{probe/tool/ocr/nreduced/word}{zero}
\defV{probe/tool/ocr/nkept}{4}
\defV{probe/tool/ocr/nkept/word}{four}
\defV{probe/tool/ocr/nundecided}{1}
\defV{probe/tool/ocr/nundecided/word}{one}
\defV{probe/tool/ocr/ncostly}{1}
\defV{probe/tool/ocr/ncostly/word}{one}
\defV{probe/tool/ocr/notremoved}{5}
\defV{probe/tool/ocr/notremoved/word}{five}
\defV{probe/tool/ocr/dropgrew}{1}
\defV{probe/tool/ocr/dropgrew/word}{one}
\defV{probe/tool/zoom/gemma4_31/control}{0.76}
\defV{probe/tool/zoom/gemma4_31/conflict}{0.35}
\defV{probe/tool/zoom/gemma4_31/drop}{0.41}
\defV{probe/tool/zoom/gemma4_31/dropph}{$7\times10^{-17}$}
\defV{probe/tool/zoom/gemma4_31/gain}{+0.06}
\defV{probe/tool/zoom/gemma4_31/gainph}{0.67}
\defV{probe/tool/zoom/gemma4_31/ctrlchange}{+0.03}
\defV{probe/tool/zoom/gemma4_31/ctrlph}{1.00}
\defV{probe/tool/zoom/gemma4_31/tooluse}{1.00}
\defV{probe/tool/zoom/qwen38_27/control}{0.14}
\defV{probe/tool/zoom/qwen38_27/conflict}{0.03}
\defV{probe/tool/zoom/qwen38_27/drop}{0.11}
\defV{probe/tool/zoom/qwen38_27/dropph}{0.005}
\defV{probe/tool/zoom/qwen38_27/gain}{$-$0.18}
\defV{probe/tool/zoom/qwen38_27/gainph}{$2\times10^{-5}$}
\defV{probe/tool/zoom/qwen38_27/ctrlchange}{$-$0.56}
\defV{probe/tool/zoom/qwen38_27/ctrlph}{$2\times10^{-16}$}
\defV{probe/tool/zoom/qwen38_27/tooluse}{1.00}
\defV{probe/tool/zoom/qwen35_9/control}{0.47}
\defV{probe/tool/zoom/qwen35_9/conflict}{0.31}
\defV{probe/tool/zoom/qwen35_9/drop}{0.16}
\defV{probe/tool/zoom/qwen35_9/dropph}{0.009}
\defV{probe/tool/zoom/qwen35_9/gain}{+0.01}
\defV{probe/tool/zoom/qwen35_9/gainph}{1.00}
\defV{probe/tool/zoom/qwen35_9/ctrlchange}{$-$0.20}
\defV{probe/tool/zoom/qwen35_9/ctrlph}{0.002}
\defV{probe/tool/zoom/qwen35_9/tooluse}{0.47}
\defV{probe/tool/zoom/mistral32_24/control}{0.67}
\defV{probe/tool/zoom/mistral32_24/conflict}{0.30}
\defV{probe/tool/zoom/mistral32_24/drop}{0.37}
\defV{probe/tool/zoom/mistral32_24/dropph}{$4\times10^{-11}$}
\defV{probe/tool/zoom/mistral32_24/gain}{$-$0.05}
\defV{probe/tool/zoom/mistral32_24/gainph}{1.00}
\defV{probe/tool/zoom/mistral32_24/ctrlchange}{+0.01}
\defV{probe/tool/zoom/mistral32_24/ctrlph}{1.00}
\defV{probe/tool/zoom/mistral32_24/tooluse}{0.00}
\defV{probe/tool/zoom/qwen35_4/control}{0.40}
\defV{probe/tool/zoom/qwen35_4/conflict}{0.34}
\defV{probe/tool/zoom/qwen35_4/drop}{0.06}
\defV{probe/tool/zoom/qwen35_4/dropph}{0.31}
\defV{probe/tool/zoom/qwen35_4/gain}{$-$0.04}
\defV{probe/tool/zoom/qwen35_4/gainph}{1.00}
\defV{probe/tool/zoom/qwen35_4/ctrlchange}{$-$0.15}
\defV{probe/tool/zoom/qwen35_4/ctrlph}{0.02}
\defV{probe/tool/zoom/qwen35_4/tooluse}{0.05}
\defV{probe/tool/zoom/nremoved}{0}
\defV{probe/tool/zoom/nremoved/word}{zero}
\defV{probe/tool/zoom/nreduced}{0}
\defV{probe/tool/zoom/nreduced/word}{zero}
\defV{probe/tool/zoom/nkept}{4}
\defV{probe/tool/zoom/nkept/word}{four}
\defV{probe/tool/zoom/nundecided}{1}
\defV{probe/tool/zoom/nundecided/word}{one}
\defV{probe/tool/zoom/ncostly}{3}
\defV{probe/tool/zoom/ncostly/word}{three}
\defV{probe/tool/zoom/notremoved}{5}
\defV{probe/tool/zoom/notremoved/word}{five}
\defV{probe/tool/zoom/dropgrew}{1}
\defV{probe/tool/zoom/dropgrew/word}{one}
\defV{probe/tool/code/gemma4_31/control}{0.74}
\defV{probe/tool/code/gemma4_31/conflict}{0.34}
\defV{probe/tool/code/gemma4_31/drop}{0.39}
\defV{probe/tool/code/gemma4_31/dropph}{$4\times10^{-15}$}
\defV{probe/tool/code/gemma4_31/gain}{+0.05}
\defV{probe/tool/code/gemma4_31/gainph}{0.38}
\defV{probe/tool/code/gemma4_31/ctrlchange}{+0.01}
\defV{probe/tool/code/gemma4_31/ctrlph}{1.00}
\defV{probe/tool/code/gemma4_31/tooluse}{1.00}
\defV{probe/tool/code/qwen38_27/control}{0.00}
\defV{probe/tool/code/qwen38_27/conflict}{0.00}
\defV{probe/tool/code/qwen38_27/drop}{0.00}
\defV{probe/tool/code/qwen38_27/dropph}{1.00}
\defV{probe/tool/code/qwen38_27/gain}{$-$0.21}
\defV{probe/tool/code/qwen38_27/gainph}{$1\times10^{-8}$}
\defV{probe/tool/code/qwen38_27/ctrlchange}{$-$0.70}
\defV{probe/tool/code/qwen38_27/ctrlph}{$3\times10^{-29}$}
\defV{probe/tool/code/qwen38_27/tooluse}{1.00}
\defV{probe/tool/code/qwen35_9/control}{0.04}
\defV{probe/tool/code/qwen35_9/conflict}{0.05}
\defV{probe/tool/code/qwen35_9/drop}{$-$0.01}
\defV{probe/tool/code/qwen35_9/dropph}{1.00}
\defV{probe/tool/code/qwen35_9/gain}{$-$0.26}
\defV{probe/tool/code/qwen35_9/gainph}{$1\times10^{-7}$}
\defV{probe/tool/code/qwen35_9/ctrlchange}{$-$0.63}
\defV{probe/tool/code/qwen35_9/ctrlph}{$6\times10^{-25}$}
\defV{probe/tool/code/qwen35_9/tooluse}{1.00}
\defV{probe/tool/code/mistral32_24/control}{0.61}
\defV{probe/tool/code/mistral32_24/conflict}{0.36}
\defV{probe/tool/code/mistral32_24/drop}{0.24}
\defV{probe/tool/code/mistral32_24/dropph}{$5\times10^{-5}$}
\defV{probe/tool/code/mistral32_24/gain}{+0.01}
\defV{probe/tool/code/mistral32_24/gainph}{0.88}
\defV{probe/tool/code/mistral32_24/ctrlchange}{$-$0.05}
\defV{probe/tool/code/mistral32_24/ctrlph}{0.87}
\defV{probe/tool/code/mistral32_24/tooluse}{0.09}
\defV{probe/tool/code/qwen35_4/control}{0.05}
\defV{probe/tool/code/qwen35_4/conflict}{0.07}
\defV{probe/tool/code/qwen35_4/drop}{$-$0.02}
\defV{probe/tool/code/qwen35_4/dropph}{1.00}
\defV{probe/tool/code/qwen35_4/gain}{$-$0.31}
\defV{probe/tool/code/qwen35_4/gainph}{$9\times10^{-9}$}
\defV{probe/tool/code/qwen35_4/ctrlchange}{$-$0.50}
\defV{probe/tool/code/qwen35_4/ctrlph}{$9\times10^{-20}$}
\defV{probe/tool/code/qwen35_4/tooluse}{0.90}
\defV{probe/tool/code/nremoved}{0}
\defV{probe/tool/code/nremoved/word}{zero}
\defV{probe/tool/code/nreduced}{0}
\defV{probe/tool/code/nreduced/word}{zero}
\defV{probe/tool/code/nkept}{2}
\defV{probe/tool/code/nkept/word}{two}
\defV{probe/tool/code/nundecided}{3}
\defV{probe/tool/code/nundecided/word}{three}
\defV{probe/tool/code/ncostly}{3}
\defV{probe/tool/code/ncostly/word}{three}
\defV{probe/tool/code/notremoved}{5}
\defV{probe/tool/code/notremoved/word}{five}
\defV{probe/tool/code/dropgrew}{0}
\defV{probe/tool/code/dropgrew/word}{zero}
\defV{probe/tool/nconds}{4}
\defV{probe/tool/nconds/word}{four}
\defV{probe/tool/removedany}{0}
\defV{probe/tool/removedany/word}{zero}
\defV{probe/remedy/condlist}{upscale, instruction, self structure, crop}
\defV{probe/prompt/condlist}{fewshot, procedure, ordering check}
\defV{probe/tool/condlist}{oracle, ocr, zoom, code}
\defV{probe/all/nconds}{11}
\defV{probe/all/nconds/word}{eleven}
\defV{probe/all/reducedconds}{2}
\defV{probe/all/reducedconds/word}{two}
\defV{probe/all/reducedconds/Word}{Two}
\defV{probe/all/reducedmax}{1}
\defV{probe/all/reducedmax/word}{one}
\defV{probe/mech/qwen35_9/agree}{1.00}
\defV{probe/mech/qwen35_9/control}{0.76}
\defV{probe/mech/qwen35_9/conflict}{0.35}
\defV{probe/mech/qwen35_9/drop}{0.41}
\defV{probe/mech/qwen35_9/vmax}{0.71}
\defV{probe/mech/qwen35_9/vlow}{0.647}
\defV{probe/mech/qwen35_9/tmin}{0.40}
\defV{probe/mech/qwen35_9/tmax}{0.48}
\defV{probe/mech/qwen35_9/tup}{0.603}
\defV{probe/mech/qwen35_9/vbefore}{0.533}
\defV{probe/mech/qwen35_9/vafter}{0.564}
\defV{probe/mech/qwen35_9/vdiff}{+0.031}
\defV{probe/mech/qwen35_9/vdifflo}{$-$0.018}
\defV{probe/mech/qwen35_9/vdiffhi}{0.079}
\defV{probe/mech/qwen35_9/detect}{0.89}
\defV{probe/mech/qwen35_9/decision}{undecided}
\defV{probe/mech/qwen35_9/commit}{no}
\defV{probe/mech/qwen38_27/agree}{1.00}
\defV{probe/mech/qwen38_27/control}{0.87}
\defV{probe/mech/qwen38_27/conflict}{0.27}
\defV{probe/mech/qwen38_27/drop}{0.60}
\defV{probe/mech/qwen38_27/vmax}{0.78}
\defV{probe/mech/qwen38_27/vlow}{0.698}
\defV{probe/mech/qwen38_27/tmin}{0.32}
\defV{probe/mech/qwen38_27/tmax}{0.45}
\defV{probe/mech/qwen38_27/tup}{0.582}
\defV{probe/mech/qwen38_27/vbefore}{0.490}
\defV{probe/mech/qwen38_27/vafter}{0.563}
\defV{probe/mech/qwen38_27/vdiff}{+0.073}
\defV{probe/mech/qwen38_27/vdifflo}{0.021}
\defV{probe/mech/qwen38_27/vdiffhi}{0.117}
\defV{probe/mech/qwen38_27/detect}{0.89}
\defV{probe/mech/qwen38_27/decision}{never formed}
\defV{probe/mech/qwen38_27/commit}{yes}
\defV{probe/mech/gemma4_31/agree}{1.00}
\defV{probe/mech/gemma4_31/control}{0.73}
\defV{probe/mech/gemma4_31/conflict}{0.29}
\defV{probe/mech/gemma4_31/drop}{0.44}
\defV{probe/mech/gemma4_31/vmax}{0.77}
\defV{probe/mech/gemma4_31/vlow}{0.711}
\defV{probe/mech/gemma4_31/tmin}{0.49}
\defV{probe/mech/gemma4_31/tmax}{0.55}
\defV{probe/mech/gemma4_31/tup}{0.657}
\defV{probe/mech/gemma4_31/vbefore}{0.643}
\defV{probe/mech/gemma4_31/vafter}{0.747}
\defV{probe/mech/gemma4_31/vdiff}{+0.104}
\defV{probe/mech/gemma4_31/vdifflo}{0.070}
\defV{probe/mech/gemma4_31/vdiffhi}{0.136}
\defV{probe/mech/gemma4_31/detect}{0.99}
\defV{probe/mech/gemma4_31/decision}{undecided}
\defV{probe/mech/gemma4_31/commit}{yes}
\defV{probe/mech/models}{3}
\defV{probe/mech/models/word}{three}
\defV{probe/mech/nneverformed}{1}
\defV{probe/mech/nneverformed/word}{one}
\defV{probe/mech/nundecided}{2}
\defV{probe/mech/nundecided/word}{two}
\defV{probe/mech/nbindswrongly}{0}
\defV{probe/mech/nbindswrongly/word}{zero}
\defV{probe/mech/nformed}{0}
\defV{probe/mech/nformed/word}{zero}
\defV{probe/mech/ncommit}{2}
\defV{probe/mech/ncommit/word}{two}
\defV{probe/mech/vmaxlo}{0.71}
\defV{probe/mech/vmaxhi}{0.78}
\defV{probe/mech/vconfmaxlo}{0.68}
\defV{probe/mech/vconfmaxhi}{0.76}
\defV{probe/mech/threshold}{0.65}
\defV{probe/mech/undecidedgap}{0.01}
\defV{probe/frontier/gpt6_luna/white/control}{0.821}
\defV{probe/frontier/gpt6_luna/white/conflict}{0.207}
\defV{probe/frontier/gpt6_luna/white/drop}{0.614}
\defV{probe/frontier/gpt6_luna/matched/control}{0.729}
\defV{probe/frontier/gpt6_luna/matched/conflict}{0.314}
\defV{probe/frontier/gpt6_luna/matched/drop}{0.414}
\defV{probe/frontier/gpt6_luna/crop/control}{0.807}
\defV{probe/frontier/gpt6_luna/crop/conflict}{0.357}
\defV{probe/frontier/gpt6_luna/crop/drop}{0.450}
\defV{probe/frontier/gpt6_luna/direct/control}{0.743}
\defV{probe/frontier/gpt6_luna/direct/conflict}{0.214}
\defV{probe/frontier/gpt6_luna/direct/drop}{0.529}
\defV{probe/frontier/gpt6_sol/white/control}{0.986}
\defV{probe/frontier/gpt6_sol/white/conflict}{0.586}
\defV{probe/frontier/gpt6_sol/white/drop}{0.400}
\defV{probe/frontier/gpt6_sol/matched/control}{0.971}
\defV{probe/frontier/gpt6_sol/matched/conflict}{0.550}
\defV{probe/frontier/gpt6_sol/matched/drop}{0.421}
\defV{probe/frontier/gpt6_sol/crop/control}{1.000}
\defV{probe/frontier/gpt6_sol/crop/conflict}{0.993}
\defV{probe/frontier/gpt6_sol/crop/drop}{0.007}
\defV{probe/frontier/gpt6_sol/direct/control}{0.971}
\defV{probe/frontier/gpt6_sol/direct/conflict}{0.593}
\defV{probe/frontier/gpt6_sol/direct/drop}{0.379}
\defV{probe/scale/llavaov2_8/control}{0.44}
\defV{probe/scale/llavaov2_8/conflict}{0.24}
\defV{probe/scale/gemma4_12/control}{0.51}
\defV{probe/scale/gemma4_12/conflict}{0.49}
\defV{probe/scale/molmo2o_7/control}{0.54}
\defV{probe/scale/molmo2o_7/conflict}{0.49}
\defV{probe/scale/qwen35_4/control}{0.55}
\defV{probe/scale/qwen35_4/conflict}{0.38}
\defV{probe/scale/glm46v_9/control}{0.56}
\defV{probe/scale/glm46v_9/conflict}{0.34}
\defV{probe/scale/internvl35_38/control}{0.59}
\defV{probe/scale/internvl35_38/conflict}{0.42}
\defV{probe/scale/mistral32_24/control}{0.66}
\defV{probe/scale/mistral32_24/conflict}{0.35}
\defV{probe/scale/qwen35_9/control}{0.67}
\defV{probe/scale/qwen35_9/conflict}{0.31}
\defV{probe/scale/qwen38_27/control}{0.70}
\defV{probe/scale/qwen38_27/conflict}{0.21}
\defV{probe/scale/gemma4_31/control}{0.73}
\defV{probe/scale/gemma4_31/conflict}{0.29}
\defV{probe/scale/gpt6_luna/control}{0.82}
\defV{probe/scale/gpt6_luna/conflict}{0.21}
\defV{probe/scale/gpt6_sol/control}{0.99}
\defV{probe/scale/gpt6_sol/conflict}{0.59}
\defV{probe/scale/rho}{$-$0.23}
\defV{probe/scale/p}{0.47}
\defV{probe/scale/rhoopen}{$-$0.47}
\defV{probe/scale/popen}{0.17}
\defV{probe/scale/n}{12}
\defV{probe/scale/n/word}{twelve}
\defV{probe/scale/nopen}{10}
\defV{probe/scale/nopen/word}{ten}
\defV{probe/scale/below}{9}
\defV{probe/scale/below/word}{nine}
\defV{probe/posthoc/gemma4_31/items}{280}
\defV{probe/posthoc/gemma4_31/calls}{280}
\defV{probe/posthoc/gemma4_31/atA}{41}
\defV{probe/posthoc/gemma4_31/ninnercontrol}{72}
\defV{probe/posthoc/gemma4_31/correctinnercontrol}{51}
\defV{probe/posthoc/gemma4_31/innercontrol}{0.71}
\defV{probe/posthoc/gemma4_31/ninnerconflict}{64}
\defV{probe/posthoc/gemma4_31/correctinnerconflict}{31}
\defV{probe/posthoc/gemma4_31/innerconflict}{0.48}
\defV{probe/posthoc/qwen38_27/items}{280}
\defV{probe/posthoc/qwen38_27/calls}{280}
\defV{probe/posthoc/qwen38_27/atA}{270}
\defV{probe/posthoc/qwen38_27/ninnercontrol}{126}
\defV{probe/posthoc/qwen38_27/correctinnercontrol}{97}
\defV{probe/posthoc/qwen38_27/innercontrol}{0.77}
\defV{probe/posthoc/qwen38_27/ninnerconflict}{126}
\defV{probe/posthoc/qwen38_27/correctinnerconflict}{79}
\defV{probe/posthoc/qwen38_27/innerconflict}{0.63}
\defV{probe/posthoc/qwen35_9/items}{280}
\defV{probe/posthoc/qwen35_9/calls}{276}
\defV{probe/posthoc/qwen35_9/atA}{4}
\defV{probe/posthoc/qwen35_9/ninnercontrol}{21}
\defV{probe/posthoc/qwen35_9/correctinnercontrol}{10}
\defV{probe/posthoc/qwen35_9/innercontrol}{0.48}
\defV{probe/posthoc/qwen35_9/ninnerconflict}{15}
\defV{probe/posthoc/qwen35_9/correctinnerconflict}{2}
\defV{probe/posthoc/qwen35_9/innerconflict}{0.13}
\defV{probe/posthoc/mistral32_24/items}{280}
\defV{probe/posthoc/mistral32_24/calls}{15}
\defV{probe/posthoc/mistral32_24/atA}{0}
\defV{probe/posthoc/mistral32_24/ninnercontrol}{0}
\defV{probe/posthoc/mistral32_24/correctinnercontrol}{0}
\defV{probe/posthoc/mistral32_24/innercontrol}{--}
\defV{probe/posthoc/mistral32_24/ninnerconflict}{0}
\defV{probe/posthoc/mistral32_24/correctinnerconflict}{0}
\defV{probe/posthoc/mistral32_24/innerconflict}{--}
\defV{probe/posthoc/qwen35_4/items}{280}
\defV{probe/posthoc/qwen35_4/calls}{156}
\defV{probe/posthoc/qwen35_4/atA}{0}
\defV{probe/posthoc/qwen35_4/ninnercontrol}{7}
\defV{probe/posthoc/qwen35_4/correctinnercontrol}{2}
\defV{probe/posthoc/qwen35_4/innercontrol}{0.29}
\defV{probe/posthoc/qwen35_4/ninnerconflict}{4}
\defV{probe/posthoc/qwen35_4/correctinnerconflict}{3}
\defV{probe/posthoc/qwen35_4/innerconflict}{0.75}
\defV{probe/posthoc/radius}{15}
\defV{probe/value/gemma4_31/control}{0.91}
\defV{probe/value/gemma4_31/conflict}{0.48}
\defV{probe/value/gemma4_31/drop}{0.42}
\defV{probe/value/gemma4_31/twodrop}{0.36}
\defV{probe/value/gemma4_31/near}{0.52}
\defV{probe/value/gemma4_31/p}{$2\times10^{-20}$}
\defV{probe/value/qwen38_27/control}{0.88}
\defV{probe/value/qwen38_27/conflict}{0.68}
\defV{probe/value/qwen38_27/drop}{0.20}
\defV{probe/value/qwen38_27/twodrop}{0.11}
\defV{probe/value/qwen38_27/near}{0.29}
\defV{probe/value/qwen38_27/p}{$4\times10^{-7}$}
\defV{probe/value/qwen35_9/control}{0.61}
\defV{probe/value/qwen35_9/conflict}{0.47}
\defV{probe/value/qwen35_9/drop}{0.14}
\defV{probe/value/qwen35_9/twodrop}{0.08}
\defV{probe/value/qwen35_9/near}{0.51}
\defV{probe/value/qwen35_9/p}{0.002}
\defV{probe/value/internvl35_38/control}{0.80}
\defV{probe/value/internvl35_38/conflict}{0.60}
\defV{probe/value/internvl35_38/drop}{0.20}
\defV{probe/value/internvl35_38/twodrop}{0.18}
\defV{probe/value/internvl35_38/near}{0.40}
\defV{probe/value/internvl35_38/p}{$2\times10^{-5}$}
\defV{probe/value/mistral32_24/control}{0.78}
\defV{probe/value/mistral32_24/conflict}{0.50}
\defV{probe/value/mistral32_24/drop}{0.28}
\defV{probe/value/mistral32_24/twodrop}{0.32}
\defV{probe/value/mistral32_24/near}{0.50}
\defV{probe/value/mistral32_24/p}{$2\times10^{-10}$}
\defV{probe/value/glm46v_9/control}{0.79}
\defV{probe/value/glm46v_9/conflict}{0.56}
\defV{probe/value/glm46v_9/drop}{0.22}
\defV{probe/value/glm46v_9/twodrop}{0.10}
\defV{probe/value/glm46v_9/near}{0.43}
\defV{probe/value/glm46v_9/p}{$8\times10^{-7}$}
\defV{probe/value/llavaov2_8/control}{0.57}
\defV{probe/value/llavaov2_8/conflict}{0.43}
\defV{probe/value/llavaov2_8/drop}{0.14}
\defV{probe/value/llavaov2_8/twodrop}{0.09}
\defV{probe/value/llavaov2_8/near}{0.56}
\defV{probe/value/llavaov2_8/p}{0.002}
\defV{probe/value/molmo2o_7/control}{0.54}
\defV{probe/value/molmo2o_7/conflict}{0.48}
\defV{probe/value/molmo2o_7/drop}{0.06}
\defV{probe/value/molmo2o_7/twodrop}{0.07}
\defV{probe/value/molmo2o_7/near}{0.52}
\defV{probe/value/molmo2o_7/p}{0.25}
\defV{probe/value/qwen35_4/control}{0.57}
\defV{probe/value/qwen35_4/conflict}{0.54}
\defV{probe/value/qwen35_4/drop}{0.04}
\defV{probe/value/qwen35_4/twodrop}{0.07}
\defV{probe/value/qwen35_4/near}{0.39}
\defV{probe/value/qwen35_4/p}{0.25}
\defV{probe/value/gemma4_12/control}{0.83}
\defV{probe/value/gemma4_12/conflict}{0.62}
\defV{probe/value/gemma4_12/drop}{0.21}
\defV{probe/value/gemma4_12/twodrop}{0.24}
\defV{probe/value/gemma4_12/near}{0.38}
\defV{probe/value/gemma4_12/p}{$6\times10^{-7}$}
\defV{probe/value/n}{10}
\defV{probe/value/n/word}{ten}
\defV{probe/value/nsupported}{8}
\defV{probe/value/nsupported/word}{eight}
\defV{probe/value/controllo}{0.57}
\defV{probe/value/controlhi}{0.91}
\defV{probe/value/conflictlo}{0.43}
\defV{probe/value/conflicthi}{0.68}
\defV{probe/round2/n}{7}
\defV{probe/round2/n/word}{seven}
\defV{probe/ft/qwen35_9/base/contour/control}{0.67}
\defV{probe/ft/qwen35_9/base/contour/conflict}{0.31}
\defV{probe/ft/qwen35_9/base/contour/drop}{0.36}
\defV{probe/ft/qwen35_9/base/contour/p}{$8\times10^{-13}$}
\defV{probe/ft/qwen35_9/base/isobar/control}{0.59}
\defV{probe/ft/qwen35_9/base/isobar/conflict}{0.52}
\defV{probe/ft/qwen35_9/base/isobar/drop}{0.08}
\defV{probe/ft/qwen35_9/base/isobar/p}{0.04}
\defV{probe/ft/qwen35_9/base/scatter/control}{0.96}
\defV{probe/ft/qwen35_9/base/scatter/conflict}{0.56}
\defV{probe/ft/qwen35_9/base/scatter/drop}{0.40}
\defV{probe/ft/qwen35_9/base/scatter/p}{$3\times10^{-21}$}
\defV{probe/ft/qwen35_9/base/gsi/control}{0.58}
\defV{probe/ft/qwen35_9/base/gsi/conflict}{0.39}
\defV{probe/ft/qwen35_9/base/gsi/drop}{0.19}
\defV{probe/ft/qwen35_9/base/gsi/p}{$1\times10^{-5}$}
\defV{probe/ft/qwen35_9/binding/contour/control}{0.99}
\defV{probe/ft/qwen35_9/binding/contour/conflict}{0.94}
\defV{probe/ft/qwen35_9/binding/contour/drop}{0.05}
\defV{probe/ft/qwen35_9/binding/contour/p}{0.02}
\defV{probe/ft/qwen35_9/binding/isobar/control}{0.98}
\defV{probe/ft/qwen35_9/binding/isobar/conflict}{0.98}
\defV{probe/ft/qwen35_9/binding/isobar/drop}{0.01}
\defV{probe/ft/qwen35_9/binding/isobar/p}{0.50}
\defV{probe/ft/qwen35_9/binding/scatter/control}{0.90}
\defV{probe/ft/qwen35_9/binding/scatter/conflict}{0.47}
\defV{probe/ft/qwen35_9/binding/scatter/drop}{0.42}
\defV{probe/ft/qwen35_9/binding/scatter/p}{$6\times10^{-22}$}
\defV{probe/ft/qwen35_9/binding/gsi/control}{0.84}
\defV{probe/ft/qwen35_9/binding/gsi/conflict}{0.53}
\defV{probe/ft/qwen35_9/binding/gsi/drop}{0.30}
\defV{probe/ft/qwen35_9/binding/gsi/p}{$8\times10^{-8}$}
\defV{probe/ft/qwen35_9/answer/contour/control}{0.95}
\defV{probe/ft/qwen35_9/answer/contour/conflict}{0.95}
\defV{probe/ft/qwen35_9/answer/contour/drop}{0.00}
\defV{probe/ft/qwen35_9/answer/contour/p}{0.60}
\defV{probe/ft/qwen35_9/answer/isobar/control}{0.88}
\defV{probe/ft/qwen35_9/answer/isobar/conflict}{0.90}
\defV{probe/ft/qwen35_9/answer/isobar/drop}{$-$0.01}
\defV{probe/ft/qwen35_9/answer/isobar/p}{0.76}
\defV{probe/ft/qwen35_9/answer/scatter/control}{1.00}
\defV{probe/ft/qwen35_9/answer/scatter/conflict}{0.62}
\defV{probe/ft/qwen35_9/answer/scatter/drop}{0.38}
\defV{probe/ft/qwen35_9/answer/scatter/p}{$3\times10^{-23}$}
\defV{probe/ft/qwen35_9/answer/gsi/control}{0.83}
\defV{probe/ft/qwen35_9/answer/gsi/conflict}{0.52}
\defV{probe/ft/qwen35_9/answer/gsi/drop}{0.30}
\defV{probe/ft/qwen35_9/answer/gsi/p}{$2\times10^{-7}$}
\defV{probe/ft/qwen35_9/binding/contour/smaller}{51}
\defV{probe/ft/qwen35_9/binding/contour/larger}{7}
\defV{probe/ft/qwen35_9/binding/contour/predp}{$1\times10^{-9}$}
\defV{probe/ft/qwen35_9/binding/isobar/smaller}{49}
\defV{probe/ft/qwen35_9/binding/isobar/larger}{34}
\defV{probe/ft/qwen35_9/binding/isobar/predp}{0.19}
\defV{probe/ft/qwen35_9/binding/scatter/smaller}{32}
\defV{probe/ft/qwen35_9/binding/scatter/larger}{36}
\defV{probe/ft/qwen35_9/binding/scatter/predp}{1.00}
\defV{probe/ft/qwen35_9/binding/gsi/smaller}{29}
\defV{probe/ft/qwen35_9/binding/gsi/larger}{44}
\defV{probe/ft/qwen35_9/binding/gsi/predp}{1.00}
\defV{probe/ft/qwen35_9/binding/contour/controlchange}{+0.32}
\defV{probe/ft/qwen35_9/binding/contour/controlp}{$6\times10^{-14}$}
\defV{probe/ft/qwen35_9/binding/isobar/controlchange}{+0.39}
\defV{probe/ft/qwen35_9/binding/isobar/controlp}{$1\times10^{-22}$}
\defV{probe/ft/qwen35_9/binding/scatter/controlchange}{$-$0.07}
\defV{probe/ft/qwen35_9/binding/scatter/controlp}{0.02}
\defV{probe/ft/qwen35_9/binding/gsi/controlchange}{+0.26}
\defV{probe/ft/qwen35_9/binding/gsi/controlp}{$1\times10^{-5}$}
\defV{probe/ft/qwen35_9/answer/contour/smaller}{54}
\defV{probe/ft/qwen35_9/answer/contour/larger}{6}
\defV{probe/ft/qwen35_9/answer/contour/predp}{$5\times10^{-11}$}
\defV{probe/ft/qwen35_9/answer/isobar/smaller}{51}
\defV{probe/ft/qwen35_9/answer/isobar/larger}{38}
\defV{probe/ft/qwen35_9/answer/isobar/predp}{0.30}
\defV{probe/ft/qwen35_9/answer/scatter/smaller}{24}
\defV{probe/ft/qwen35_9/answer/scatter/larger}{19}
\defV{probe/ft/qwen35_9/answer/scatter/predp}{0.54}
\defV{probe/ft/qwen35_9/answer/gsi/smaller}{25}
\defV{probe/ft/qwen35_9/answer/gsi/larger}{40}
\defV{probe/ft/qwen35_9/answer/gsi/predp}{0.98}
\defV{probe/ft/qwen35_9/answer/contour/controlchange}{+0.28}
\defV{probe/ft/qwen35_9/answer/contour/controlp}{$3\times10^{-9}$}
\defV{probe/ft/qwen35_9/answer/isobar/controlchange}{+0.28}
\defV{probe/ft/qwen35_9/answer/isobar/controlp}{$1\times10^{-9}$}
\defV{probe/ft/qwen35_9/answer/scatter/controlchange}{+0.04}
\defV{probe/ft/qwen35_9/answer/scatter/controlp}{0.008}
\defV{probe/ft/qwen35_9/answer/gsi/controlchange}{+0.25}
\defV{probe/ft/qwen35_9/answer/gsi/controlp}{$6\times10^{-6}$}
\defV{probe/ft/qwen35_9/bva/binding}{63}
\defV{probe/ft/qwen35_9/bva/answer}{77}
\defV{probe/ft/glm46v_9/base/contour/control}{0.56}
\defV{probe/ft/glm46v_9/base/contour/conflict}{0.34}
\defV{probe/ft/glm46v_9/base/contour/drop}{0.23}
\defV{probe/ft/glm46v_9/base/contour/p}{$2\times10^{-5}$}
\defV{probe/ft/glm46v_9/base/isobar/control}{0.72}
\defV{probe/ft/glm46v_9/base/isobar/conflict}{0.61}
\defV{probe/ft/glm46v_9/base/isobar/drop}{0.10}
\defV{probe/ft/glm46v_9/base/isobar/p}{0.01}
\defV{probe/ft/glm46v_9/base/scatter/control}{0.99}
\defV{probe/ft/glm46v_9/base/scatter/conflict}{0.65}
\defV{probe/ft/glm46v_9/base/scatter/drop}{0.34}
\defV{probe/ft/glm46v_9/base/scatter/p}{$6\times10^{-20}$}
\defV{probe/ft/glm46v_9/base/gsi/control}{0.63}
\defV{probe/ft/glm46v_9/base/gsi/conflict}{0.43}
\defV{probe/ft/glm46v_9/base/gsi/drop}{0.20}
\defV{probe/ft/glm46v_9/base/gsi/p}{$1\times10^{-7}$}
\defV{probe/ft/glm46v_9/binding/contour/control}{0.97}
\defV{probe/ft/glm46v_9/binding/contour/conflict}{0.98}
\defV{probe/ft/glm46v_9/binding/contour/drop}{$-$0.01}
\defV{probe/ft/glm46v_9/binding/contour/p}{0.77}
\defV{probe/ft/glm46v_9/binding/isobar/control}{1.00}
\defV{probe/ft/glm46v_9/binding/isobar/conflict}{0.93}
\defV{probe/ft/glm46v_9/binding/isobar/drop}{0.07}
\defV{probe/ft/glm46v_9/binding/isobar/p}{$6\times10^{-5}$}
\defV{probe/ft/glm46v_9/binding/scatter/control}{0.98}
\defV{probe/ft/glm46v_9/binding/scatter/conflict}{0.60}
\defV{probe/ft/glm46v_9/binding/scatter/drop}{0.38}
\defV{probe/ft/glm46v_9/binding/scatter/p}{$3\times10^{-21}$}
\defV{probe/ft/glm46v_9/binding/gsi/control}{0.76}
\defV{probe/ft/glm46v_9/binding/gsi/conflict}{0.62}
\defV{probe/ft/glm46v_9/binding/gsi/drop}{0.13}
\defV{probe/ft/glm46v_9/binding/gsi/p}{0.02}
\defV{probe/ft/glm46v_9/answer/contour/control}{0.95}
\defV{probe/ft/glm46v_9/answer/contour/conflict}{0.99}
\defV{probe/ft/glm46v_9/answer/contour/drop}{$-$0.04}
\defV{probe/ft/glm46v_9/answer/contour/p}{0.99}
\defV{probe/ft/glm46v_9/answer/isobar/control}{0.99}
\defV{probe/ft/glm46v_9/answer/isobar/conflict}{0.95}
\defV{probe/ft/glm46v_9/answer/isobar/drop}{0.04}
\defV{probe/ft/glm46v_9/answer/isobar/p}{0.006}
\defV{probe/ft/glm46v_9/answer/scatter/control}{0.97}
\defV{probe/ft/glm46v_9/answer/scatter/conflict}{0.81}
\defV{probe/ft/glm46v_9/answer/scatter/drop}{0.16}
\defV{probe/ft/glm46v_9/answer/scatter/p}{$1\times10^{-8}$}
\defV{probe/ft/glm46v_9/answer/gsi/control}{0.80}
\defV{probe/ft/glm46v_9/answer/gsi/conflict}{0.69}
\defV{probe/ft/glm46v_9/answer/gsi/drop}{0.11}
\defV{probe/ft/glm46v_9/answer/gsi/p}{0.03}
\defV{probe/ft/glm46v_9/binding/contour/smaller}{48}
\defV{probe/ft/glm46v_9/binding/contour/larger}{16}
\defV{probe/ft/glm46v_9/binding/contour/predp}{$4\times10^{-5}$}
\defV{probe/ft/glm46v_9/binding/isobar/smaller}{43}
\defV{probe/ft/glm46v_9/binding/isobar/larger}{33}
\defV{probe/ft/glm46v_9/binding/isobar/predp}{0.45}
\defV{probe/ft/glm46v_9/binding/scatter/smaller}{15}
\defV{probe/ft/glm46v_9/binding/scatter/larger}{24}
\defV{probe/ft/glm46v_9/binding/scatter/predp}{0.95}
\defV{probe/ft/glm46v_9/binding/gsi/smaller}{36}
\defV{probe/ft/glm46v_9/binding/gsi/larger}{31}
\defV{probe/ft/glm46v_9/binding/gsi/predp}{0.63}
\defV{probe/ft/glm46v_9/binding/contour/controlchange}{+0.41}
\defV{probe/ft/glm46v_9/binding/contour/controlp}{$9\times10^{-15}$}
\defV{probe/ft/glm46v_9/binding/isobar/controlchange}{+0.28}
\defV{probe/ft/glm46v_9/binding/isobar/controlp}{$3\times10^{-17}$}
\defV{probe/ft/glm46v_9/binding/scatter/controlchange}{$-$0.01}
\defV{probe/ft/glm46v_9/binding/scatter/controlp}{1.00}
\defV{probe/ft/glm46v_9/binding/gsi/controlchange}{+0.12}
\defV{probe/ft/glm46v_9/binding/gsi/controlp}{0.04}
\defV{probe/ft/glm46v_9/answer/contour/smaller}{50}
\defV{probe/ft/glm46v_9/answer/contour/larger}{14}
\defV{probe/ft/glm46v_9/answer/contour/predp}{$4\times10^{-6}$}
\defV{probe/ft/glm46v_9/answer/isobar/smaller}{43}
\defV{probe/ft/glm46v_9/answer/isobar/larger}{29}
\defV{probe/ft/glm46v_9/answer/isobar/predp}{0.12}
\defV{probe/ft/glm46v_9/answer/scatter/smaller}{44}
\defV{probe/ft/glm46v_9/answer/scatter/larger}{8}
\defV{probe/ft/glm46v_9/answer/scatter/predp}{$6\times10^{-7}$}
\defV{probe/ft/glm46v_9/answer/gsi/smaller}{36}
\defV{probe/ft/glm46v_9/answer/gsi/larger}{27}
\defV{probe/ft/glm46v_9/answer/gsi/predp}{0.16}
\defV{probe/ft/glm46v_9/answer/contour/controlchange}{+0.39}
\defV{probe/ft/glm46v_9/answer/contour/controlp}{$6\times10^{-14}$}
\defV{probe/ft/glm46v_9/answer/isobar/controlchange}{+0.28}
\defV{probe/ft/glm46v_9/answer/isobar/controlp}{$8\times10^{-16}$}
\defV{probe/ft/glm46v_9/answer/scatter/controlchange}{$-$0.01}
\defV{probe/ft/glm46v_9/answer/scatter/controlp}{0.45}
\defV{probe/ft/glm46v_9/answer/gsi/controlchange}{+0.16}
\defV{probe/ft/glm46v_9/answer/gsi/controlp}{0.004}
\defV{probe/ft/glm46v_9/bva/binding}{30}
\defV{probe/ft/glm46v_9/bva/answer}{82}
\defV{probe/ft/nremoval}{3}
\defV{probe/ft/nremoval/word}{three}
\defV{probe/ft/ntransfer}{0}
\defV{probe/ft/ntransfer/word}{zero}
\defV{probe/ft/nruns}{4}
\defV{probe/ft/nruns/word}{four}
\defV{probe/route/qwen35_9/control}{0.71}
\defV{probe/route/qwen35_9/conflict}{0.45}
\defV{probe/route/qwen35_9/drop}{0.25}
\defV{probe/route/qwen35_9/n10}{54}
\defV{probe/route/qwen35_9/n01}{16}
\defV{probe/route/qwen35_9/p}{$1\times10^{-5}$}
\defV{probe/route/glm46v_9/control}{0.71}
\defV{probe/route/glm46v_9/conflict}{0.38}
\defV{probe/route/glm46v_9/drop}{0.33}
\defV{probe/route/glm46v_9/n10}{52}
\defV{probe/route/glm46v_9/n01}{3}
\defV{probe/route/glm46v_9/p}{$7\times10^{-12}$}
\defV{probe/route/llavaov2_8/control}{0.48}
\defV{probe/route/llavaov2_8/conflict}{0.49}
\defV{probe/route/llavaov2_8/drop}{$-$0.01}
\defV{probe/route/llavaov2_8/n10}{11}
\defV{probe/route/llavaov2_8/n01}{13}
\defV{probe/route/llavaov2_8/p}{0.73}
\defV{probe/route/molmo2o_7/control}{0.49}
\defV{probe/route/molmo2o_7/conflict}{0.46}
\defV{probe/route/molmo2o_7/drop}{0.03}
\defV{probe/route/molmo2o_7/n10}{6}
\defV{probe/route/molmo2o_7/n01}{1}
\defV{probe/route/molmo2o_7/p}{0.12}
\defV{probe/route/qwen35_4/control}{0.73}
\defV{probe/route/qwen35_4/conflict}{0.44}
\defV{probe/route/qwen35_4/drop}{0.29}
\defV{probe/route/qwen35_4/n10}{60}
\defV{probe/route/qwen35_4/n01}{16}
\defV{probe/route/qwen35_4/p}{$1\times10^{-6}$}
\defV{probe/route/gemma4_12/control}{0.71}
\defV{probe/route/gemma4_12/conflict}{0.36}
\defV{probe/route/gemma4_12/drop}{0.35}
\defV{probe/route/gemma4_12/n10}{64}
\defV{probe/route/gemma4_12/n01}{11}
\defV{probe/route/gemma4_12/p}{$1\times10^{-9}$}
\defV{probe/route/gemma4_31/control}{0.83}
\defV{probe/route/gemma4_31/conflict}{0.37}
\defV{probe/route/gemma4_31/drop}{0.46}
\defV{probe/route/gemma4_31/n10}{72}
\defV{probe/route/gemma4_31/n01}{3}
\defV{probe/route/gemma4_31/p}{$2\times10^{-17}$}
\defV{probe/route/qwen38_27/control}{0.71}
\defV{probe/route/qwen38_27/conflict}{0.39}
\defV{probe/route/qwen38_27/drop}{0.32}
\defV{probe/route/qwen38_27/n10}{62}
\defV{probe/route/qwen38_27/n01}{14}
\defV{probe/route/qwen38_27/p}{$8\times10^{-8}$}
\defV{probe/route/internvl35_38/control}{0.57}
\defV{probe/route/internvl35_38/conflict}{0.46}
\defV{probe/route/internvl35_38/drop}{0.11}
\defV{probe/route/internvl35_38/n10}{24}
\defV{probe/route/internvl35_38/n01}{8}
\defV{probe/route/internvl35_38/p}{0.01}
\defV{probe/route/mistral32_24/control}{0.51}
\defV{probe/route/mistral32_24/conflict}{0.41}
\defV{probe/route/mistral32_24/drop}{0.11}
\defV{probe/route/mistral32_24/n10}{42}
\defV{probe/route/mistral32_24/n01}{26}
\defV{probe/route/mistral32_24/p}{0.10}
\defV{probe/route/n}{10}
\defV{probe/route/n/word}{ten}
\defV{probe/route/neffect}{7}
\defV{probe/route/neffect/word}{seven}
\defV{probe/route/pairs}{150}
\defV{probe/route/pairs/word}{150}
\defV{probe/route/droplo}{0.11}
\defV{probe/route/drophi}{0.46}
\defV{probe/pair/ntests}{50}
\defV{probe/pair/ntests/word}{50}
\defV{probe/pair/nsame}{49}
\defV{probe/pair/nsame/word}{49}
\defV{probe/pair/gemma4_31/white/worse}{47}
\defV{probe/pair/gemma4_31/white/better}{0}
\defV{probe/pair/gemma4_31/white/p}{$1\times10^{-13}$}
\defV{probe/pair/gemma4_31/white/itemp}{$1\times10^{-16}$}
\defV{probe/pair/gemma4_31/matched/worse}{44}
\defV{probe/pair/gemma4_31/matched/better}{1}
\defV{probe/pair/gemma4_31/matched/p}{$2\times10^{-11}$}
\defV{probe/pair/gemma4_31/matched/itemp}{$3\times10^{-15}$}
\defV{probe/pair/gemma4_31/crop/worse}{56}
\defV{probe/pair/gemma4_31/crop/better}{0}
\defV{probe/pair/gemma4_31/crop/p}{$1\times10^{-16}$}
\defV{probe/pair/gemma4_31/crop/itemp}{$4\times10^{-25}$}
\defV{probe/pair/gemma4_31/direct/worse}{52}
\defV{probe/pair/gemma4_31/direct/better}{0}
\defV{probe/pair/gemma4_31/direct/p}{$2\times10^{-15}$}
\defV{probe/pair/gemma4_31/direct/itemp}{$7\times10^{-20}$}
\defV{probe/pair/gemma4_31/separation/worse}{45}
\defV{probe/pair/gemma4_31/separation/better}{3}
\defV{probe/pair/gemma4_31/separation/p}{$2\times10^{-9}$}
\defV{probe/pair/gemma4_31/separation/itemp}{$3\times10^{-10}$}
\defV{probe/pair/gemma4_31/isobar/worse}{56}
\defV{probe/pair/gemma4_31/isobar/better}{7}
\defV{probe/pair/gemma4_31/isobar/p}{$2\times10^{-9}$}
\defV{probe/pair/gemma4_31/isobar/itemp}{$3\times10^{-15}$}
\defV{probe/pair/gemma4_31/scatter/worse}{25}
\defV{probe/pair/gemma4_31/scatter/better}{1}
\defV{probe/pair/gemma4_31/scatter/p}{$1\times10^{-5}$}
\defV{probe/pair/gemma4_31/scatter/itemp}{$2\times10^{-8}$}
\defV{probe/pair/gemma4_31/gsi_placement/worse}{31}
\defV{probe/pair/gemma4_31/gsi_placement/better}{2}
\defV{probe/pair/gemma4_31/gsi_placement/p}{$2\times10^{-6}$}
\defV{probe/pair/gemma4_31/gsi_placement/itemp}{$2\times10^{-8}$}
\defV{probe/pair/gemma4_12/white/worse}{10}
\defV{probe/pair/gemma4_12/white/better}{7}
\defV{probe/pair/gemma4_12/white/p}{0.63}
\defV{probe/pair/gemma4_12/white/itemp}{0.50}
\defV{probe/pair/qwen38_27/white/worse}{53}
\defV{probe/pair/qwen38_27/white/better}{3}
\defV{probe/pair/qwen38_27/white/p}{$7\times10^{-12}$}
\defV{probe/pair/qwen38_27/white/itemp}{$7\times10^{-16}$}
\defV{probe/pair/qwen38_27/matched/worse}{51}
\defV{probe/pair/qwen38_27/matched/better}{3}
\defV{probe/pair/qwen38_27/matched/p}{$2\times10^{-11}$}
\defV{probe/pair/qwen38_27/matched/itemp}{$2\times10^{-15}$}
\defV{probe/pair/qwen38_27/crop/worse}{57}
\defV{probe/pair/qwen38_27/crop/better}{3}
\defV{probe/pair/qwen38_27/crop/p}{$2\times10^{-13}$}
\defV{probe/pair/qwen38_27/crop/itemp}{$3\times10^{-23}$}
\defV{probe/pair/qwen38_27/direct/worse}{6}
\defV{probe/pair/qwen38_27/direct/better}{1}
\defV{probe/pair/qwen38_27/direct/p}{0.38}
\defV{probe/pair/qwen38_27/direct/itemp}{0.80}
\defV{probe/pair/qwen38_27/separation/worse}{32}
\defV{probe/pair/qwen38_27/separation/better}{6}
\defV{probe/pair/qwen38_27/separation/p}{$2\times10^{-4}$}
\defV{probe/pair/qwen38_27/separation/itemp}{$6\times10^{-5}$}
\defV{probe/pair/qwen38_27/isobar/worse}{33}
\defV{probe/pair/qwen38_27/isobar/better}{16}
\defV{probe/pair/qwen38_27/isobar/p}{0.06}
\defV{probe/pair/qwen38_27/isobar/itemp}{0.01}
\defV{probe/pair/qwen38_27/scatter/worse}{31}
\defV{probe/pair/qwen38_27/scatter/better}{0}
\defV{probe/pair/qwen38_27/scatter/p}{$2\times10^{-8}$}
\defV{probe/pair/qwen38_27/scatter/itemp}{$3\times10^{-13}$}
\defV{probe/pair/qwen38_27/gsi_placement/worse}{22}
\defV{probe/pair/qwen38_27/gsi_placement/better}{6}
\defV{probe/pair/qwen38_27/gsi_placement/p}{0.01}
\defV{probe/pair/qwen38_27/gsi_placement/itemp}{0.005}
\defV{probe/pair/qwen35_9/white/worse}{43}
\defV{probe/pair/qwen35_9/white/better}{3}
\defV{probe/pair/qwen35_9/white/p}{$4\times10^{-9}$}
\defV{probe/pair/qwen35_9/white/itemp}{$1\times10^{-11}$}
\defV{probe/pair/qwen35_9/matched/worse}{34}
\defV{probe/pair/qwen35_9/matched/better}{7}
\defV{probe/pair/qwen35_9/matched/p}{$1\times10^{-4}$}
\defV{probe/pair/qwen35_9/matched/itemp}{$6\times10^{-6}$}
\defV{probe/pair/qwen35_9/crop/worse}{65}
\defV{probe/pair/qwen35_9/crop/better}{0}
\defV{probe/pair/qwen35_9/crop/p}{$3\times10^{-19}$}
\defV{probe/pair/qwen35_9/crop/itemp}{$2\times10^{-27}$}
\defV{probe/pair/qwen35_9/direct/worse}{4}
\defV{probe/pair/qwen35_9/direct/better}{1}
\defV{probe/pair/qwen35_9/direct/p}{0.75}
\defV{probe/pair/qwen35_9/direct/itemp}{0.91}
\defV{probe/pair/qwen35_9/separation/worse}{34}
\defV{probe/pair/qwen35_9/separation/better}{6}
\defV{probe/pair/qwen35_9/separation/p}{$1\times10^{-4}$}
\defV{probe/pair/qwen35_9/separation/itemp}{$4\times10^{-5}$}
\defV{probe/pair/qwen35_9/scatter/worse}{50}
\defV{probe/pair/qwen35_9/scatter/better}{3}
\defV{probe/pair/qwen35_9/scatter/p}{$1\times10^{-10}$}
\defV{probe/pair/qwen35_9/scatter/itemp}{$1\times10^{-19}$}
\defV{probe/pair/qwen35_9/gsi_placement/worse}{22}
\defV{probe/pair/qwen35_9/gsi_placement/better}{2}
\defV{probe/pair/qwen35_9/gsi_placement/p}{$3\times10^{-4}$}
\defV{probe/pair/qwen35_9/gsi_placement/itemp}{$2\times10^{-4}$}
\defV{probe/pair/qwen35_4/white/worse}{30}
\defV{probe/pair/qwen35_4/white/better}{10}
\defV{probe/pair/qwen35_4/white/p}{0.009}
\defV{probe/pair/qwen35_4/white/itemp}{0.005}
\defV{probe/pair/qwen35_4/matched/worse}{25}
\defV{probe/pair/qwen35_4/matched/better}{8}
\defV{probe/pair/qwen35_4/matched/p}{0.005}
\defV{probe/pair/qwen35_4/matched/itemp}{0.003}
\defV{probe/pair/qwen35_4/separation/worse}{29}
\defV{probe/pair/qwen35_4/separation/better}{13}
\defV{probe/pair/qwen35_4/separation/p}{0.06}
\defV{probe/pair/qwen35_4/separation/itemp}{0.04}
\defV{probe/pair/qwen35_4/isobar/worse}{33}
\defV{probe/pair/qwen35_4/isobar/better}{22}
\defV{probe/pair/qwen35_4/isobar/p}{0.18}
\defV{probe/pair/qwen35_4/isobar/itemp}{0.14}
\defV{probe/pair/qwen35_4/scatter/worse}{44}
\defV{probe/pair/qwen35_4/scatter/better}{2}
\defV{probe/pair/qwen35_4/scatter/p}{$6\times10^{-10}$}
\defV{probe/pair/qwen35_4/scatter/itemp}{$3\times10^{-18}$}
\defV{probe/pair/qwen35_4/gsi_placement/worse}{34}
\defV{probe/pair/qwen35_4/gsi_placement/better}{9}
\defV{probe/pair/qwen35_4/gsi_placement/p}{0.001}
\defV{probe/pair/qwen35_4/gsi_placement/itemp}{0.002}
\defV{probe/pair/internvl35_38/white/worse}{28}
\defV{probe/pair/internvl35_38/white/better}{12}
\defV{probe/pair/internvl35_38/white/p}{0.05}
\defV{probe/pair/internvl35_38/white/itemp}{0.01}
\defV{probe/pair/internvl35_38/matched/worse}{33}
\defV{probe/pair/internvl35_38/matched/better}{6}
\defV{probe/pair/internvl35_38/matched/p}{$7\times10^{-5}$}
\defV{probe/pair/internvl35_38/matched/itemp}{$2\times10^{-4}$}
\defV{probe/pair/internvl35_38/crop/worse}{33}
\defV{probe/pair/internvl35_38/crop/better}{2}
\defV{probe/pair/internvl35_38/crop/p}{$7\times10^{-8}$}
\defV{probe/pair/internvl35_38/crop/itemp}{$9\times10^{-10}$}
\defV{probe/pair/internvl35_38/direct/worse}{18}
\defV{probe/pair/internvl35_38/direct/better}{8}
\defV{probe/pair/internvl35_38/direct/p}{0.30}
\defV{probe/pair/internvl35_38/direct/itemp}{0.29}
\defV{probe/pair/internvl35_38/isobar/worse}{45}
\defV{probe/pair/internvl35_38/isobar/better}{18}
\defV{probe/pair/internvl35_38/isobar/p}{0.005}
\defV{probe/pair/internvl35_38/isobar/itemp}{0.001}
\defV{probe/pair/internvl35_38/scatter/worse}{64}
\defV{probe/pair/internvl35_38/scatter/better}{2}
\defV{probe/pair/internvl35_38/scatter/p}{$1\times10^{-15}$}
\defV{probe/pair/internvl35_38/scatter/itemp}{$3\times10^{-21}$}
\defV{probe/pair/internvl35_38/gsi_placement/worse}{22}
\defV{probe/pair/internvl35_38/gsi_placement/better}{3}
\defV{probe/pair/internvl35_38/gsi_placement/p}{0.001}
\defV{probe/pair/internvl35_38/gsi_placement/itemp}{0.005}
\defV{probe/pair/mistral32_24/white/worse}{38}
\defV{probe/pair/mistral32_24/white/better}{9}
\defV{probe/pair/mistral32_24/white/p}{$2\times10^{-4}$}
\defV{probe/pair/mistral32_24/white/itemp}{$1\times10^{-6}$}
\defV{probe/pair/mistral32_24/matched/worse}{36}
\defV{probe/pair/mistral32_24/matched/better}{10}
\defV{probe/pair/mistral32_24/matched/p}{$5\times10^{-4}$}
\defV{probe/pair/mistral32_24/matched/itemp}{$2\times10^{-4}$}
\defV{probe/pair/mistral32_24/separation/worse}{21}
\defV{probe/pair/mistral32_24/separation/better}{4}
\defV{probe/pair/mistral32_24/separation/p}{0.005}
\defV{probe/pair/mistral32_24/separation/itemp}{0.009}
\defV{probe/pair/mistral32_24/isobar/worse}{60}
\defV{probe/pair/mistral32_24/isobar/better}{13}
\defV{probe/pair/mistral32_24/isobar/p}{$3\times10^{-7}$}
\defV{probe/pair/mistral32_24/isobar/itemp}{$5\times10^{-10}$}
\defV{probe/pair/mistral32_24/scatter/worse}{53}
\defV{probe/pair/mistral32_24/scatter/better}{4}
\defV{probe/pair/mistral32_24/scatter/p}{$1\times10^{-10}$}
\defV{probe/pair/mistral32_24/scatter/itemp}{$6\times10^{-17}$}
\defV{probe/pair/glm46v_9/white/worse}{33}
\defV{probe/pair/glm46v_9/white/better}{9}
\defV{probe/pair/glm46v_9/white/p}{0.001}
\defV{probe/pair/glm46v_9/white/itemp}{$3\times10^{-4}$}
\defV{probe/pair/llavaov2_8/white/worse}{34}
\defV{probe/pair/llavaov2_8/white/better}{9}
\defV{probe/pair/llavaov2_8/white/p}{0.001}
\defV{probe/pair/llavaov2_8/white/itemp}{0.002}
\defV{probe/pair/llavaov2_8/matched/worse}{38}
\defV{probe/pair/llavaov2_8/matched/better}{14}
\defV{probe/pair/llavaov2_8/matched/p}{0.002}
\defV{probe/pair/llavaov2_8/matched/itemp}{0.001}
\defV{probe/pair/llavaov2_8/crop/worse}{42}
\defV{probe/pair/llavaov2_8/crop/better}{9}
\defV{probe/pair/llavaov2_8/crop/p}{$3\times10^{-6}$}
\defV{probe/pair/llavaov2_8/crop/itemp}{$9\times10^{-8}$}
\defV{probe/pair/llavaov2_8/direct/worse}{2}
\defV{probe/pair/llavaov2_8/direct/better}{2}
\defV{probe/pair/llavaov2_8/direct/p}{1.00}
\defV{probe/pair/llavaov2_8/direct/itemp}{1.00}
\defV{probe/pair/llavaov2_8/separation/worse}{31}
\defV{probe/pair/llavaov2_8/separation/better}{5}
\defV{probe/pair/llavaov2_8/separation/p}{$1\times10^{-4}$}
\defV{probe/pair/llavaov2_8/separation/itemp}{$4\times10^{-5}$}
\defV{probe/pair/llavaov2_8/scatter/worse}{46}
\defV{probe/pair/llavaov2_8/scatter/better}{6}
\defV{probe/pair/llavaov2_8/scatter/p}{$2\times10^{-7}$}
\defV{probe/pair/llavaov2_8/scatter/itemp}{$2\times10^{-10}$}
\defV{probe/pair/molmo2o_7/white/worse}{12}
\defV{probe/pair/molmo2o_7/white/better}{6}
\defV{probe/pair/molmo2o_7/white/p}{0.48}
\defV{probe/pair/molmo2o_7/white/itemp}{0.27}
\defV{probe/mech3/tmax}{0.59}
\defV{probe/mech3/gemma4_31/vA}{0.86}
\defV{probe/mech3/qwen35_9/vA}{0.56}
\defV{probe/mech3/qwen38_27/vA}{0.62}
\defV{probe/mech3/qwenvmax}{0.62}
\defV{probe/prox/nmodels}{6}
\defV{probe/prox/nmodels/word}{six}
\defV{probe/prox/ncombos}{12}
\defV{probe/prox/ncombos/word}{twelve}
\defV{probe/prox/npositive}{11}
\defV{probe/prox/npositive/word}{eleven}
\defV{probe/prox/next/pairs}{46}
\defV{probe/prox/next/pairs/word}{46}
\defV{probe/prox/next/slopelo}{+0.09}
\defV{probe/prox/next/slopehi}{+0.34}
\defV{probe/prox/after/pairs}{28}
\defV{probe/prox/after/pairs/word}{28}
\defV{probe/prox/after/slopelo}{+0.13}
\defV{probe/prox/after/slopehi}{+0.32}
\defV{probe/prox/acc15max}{0.64}
\defV{probe/prox/gemma4_31/next/slope}{+0.17}
\defV{probe/prox/gemma4_31/next/slopeci}{[+0.08, +0.26]}
\defV{probe/prox/gemma4_31/next/acc/0.5}{0.38}
\defV{probe/prox/gemma4_31/next/acc/0.75}{0.50}
\defV{probe/prox/gemma4_31/next/acc/1}{0.47}
\defV{probe/prox/gemma4_31/next/acc/1.25}{0.54}
\defV{probe/prox/gemma4_31/next/acc/1.5}{0.60}
\defV{probe/prox/gemma4_31/after/slope}{+0.18}
\defV{probe/prox/gemma4_31/after/slopeci}{[+0.11, +0.25]}
\defV{probe/prox/gemma4_31/after/acc/0.5}{0.43}
\defV{probe/prox/gemma4_31/after/acc/0.75}{0.54}
\defV{probe/prox/gemma4_31/after/acc/1}{0.57}
\defV{probe/prox/gemma4_31/after/acc/1.25}{0.61}
\defV{probe/prox/gemma4_31/after/acc/1.5}{0.62}
\defV{probe/prox/qwen38_27/next/slope}{+0.34}
\defV{probe/prox/qwen38_27/next/slopeci}{[+0.24, +0.44]}
\defV{probe/prox/qwen38_27/next/acc/0.5}{0.20}
\defV{probe/prox/qwen38_27/next/acc/0.75}{0.37}
\defV{probe/prox/qwen38_27/next/acc/1}{0.37}
\defV{probe/prox/qwen38_27/next/acc/1.25}{0.46}
\defV{probe/prox/qwen38_27/next/acc/1.5}{0.62}
\defV{probe/prox/qwen38_27/after/slope}{+0.32}
\defV{probe/prox/qwen38_27/after/slopeci}{[+0.19, +0.44]}
\defV{probe/prox/qwen38_27/after/acc/0.5}{0.29}
\defV{probe/prox/qwen38_27/after/acc/0.75}{0.48}
\defV{probe/prox/qwen38_27/after/acc/1}{0.50}
\defV{probe/prox/qwen38_27/after/acc/1.25}{0.57}
\defV{probe/prox/qwen38_27/after/acc/1.5}{0.64}
\defV{probe/prox/qwen35_9/next/slope}{+0.14}
\defV{probe/prox/qwen35_9/next/slopeci}{[+0.03, +0.24]}
\defV{probe/prox/qwen35_9/next/acc/0.5}{0.38}
\defV{probe/prox/qwen35_9/next/acc/0.75}{0.48}
\defV{probe/prox/qwen35_9/next/acc/1}{0.49}
\defV{probe/prox/qwen35_9/next/acc/1.25}{0.49}
\defV{probe/prox/qwen35_9/next/acc/1.5}{0.57}
\defV{probe/prox/qwen35_9/after/slope}{+0.15}
\defV{probe/prox/qwen35_9/after/slopeci}{[+0.03, +0.26]}
\defV{probe/prox/qwen35_9/after/acc/0.5}{0.43}
\defV{probe/prox/qwen35_9/after/acc/0.75}{0.48}
\defV{probe/prox/qwen35_9/after/acc/1}{0.52}
\defV{probe/prox/qwen35_9/after/acc/1.25}{0.55}
\defV{probe/prox/qwen35_9/after/acc/1.5}{0.59}
\defV{probe/prox/glm46v_9/next/slope}{+0.17}
\defV{probe/prox/glm46v_9/next/slopeci}{[+0.09, +0.27]}
\defV{probe/prox/glm46v_9/next/acc/0.5}{0.38}
\defV{probe/prox/glm46v_9/next/acc/0.75}{0.49}
\defV{probe/prox/glm46v_9/next/acc/1}{0.49}
\defV{probe/prox/glm46v_9/next/acc/1.25}{0.60}
\defV{probe/prox/glm46v_9/next/acc/1.5}{0.55}
\defV{probe/prox/glm46v_9/after/slope}{+0.13}
\defV{probe/prox/glm46v_9/after/slopeci}{[+0.02, +0.23]}
\defV{probe/prox/glm46v_9/after/acc/0.5}{0.39}
\defV{probe/prox/glm46v_9/after/acc/0.75}{0.55}
\defV{probe/prox/glm46v_9/after/acc/1}{0.55}
\defV{probe/prox/glm46v_9/after/acc/1.25}{0.50}
\defV{probe/prox/glm46v_9/after/acc/1.5}{0.57}
\defV{probe/prox/mistral32_24/next/slope}{+0.22}
\defV{probe/prox/mistral32_24/next/slopeci}{[+0.12, +0.32]}
\defV{probe/prox/mistral32_24/next/acc/0.5}{0.35}
\defV{probe/prox/mistral32_24/next/acc/0.75}{0.52}
\defV{probe/prox/mistral32_24/next/acc/1}{0.50}
\defV{probe/prox/mistral32_24/next/acc/1.25}{0.55}
\defV{probe/prox/mistral32_24/next/acc/1.5}{0.62}
\defV{probe/prox/mistral32_24/after/slope}{+0.13}
\defV{probe/prox/mistral32_24/after/slopeci}{[+0.03, +0.22]}
\defV{probe/prox/mistral32_24/after/acc/0.5}{0.43}
\defV{probe/prox/mistral32_24/after/acc/0.75}{0.54}
\defV{probe/prox/mistral32_24/after/acc/1}{0.55}
\defV{probe/prox/mistral32_24/after/acc/1.25}{0.57}
\defV{probe/prox/mistral32_24/after/acc/1.5}{0.57}
\defV{probe/prox/internvl35_38/next/slope}{+0.09}
\defV{probe/prox/internvl35_38/next/slopeci}{[$-$0.00, +0.18]}
\defV{probe/prox/internvl35_38/next/acc/0.5}{0.43}
\defV{probe/prox/internvl35_38/next/acc/0.75}{0.47}
\defV{probe/prox/internvl35_38/next/acc/1}{0.55}
\defV{probe/prox/internvl35_38/next/acc/1.25}{0.52}
\defV{probe/prox/internvl35_38/next/acc/1.5}{0.52}
\defV{probe/prox/internvl35_38/after/slope}{+0.15}
\defV{probe/prox/internvl35_38/after/slopeci}{[+0.04, +0.26]}
\defV{probe/prox/internvl35_38/after/acc/0.5}{0.46}
\defV{probe/prox/internvl35_38/after/acc/0.75}{0.48}
\defV{probe/prox/internvl35_38/after/acc/1}{0.54}
\defV{probe/prox/internvl35_38/after/acc/1.25}{0.55}
\defV{probe/prox/internvl35_38/after/acc/1.5}{0.64}
\defV{probe/ftmore/nruns}{4}
\defV{probe/ftmore/nruns/word}{four}
\defV{probe/ftmore/linechart/nlarger}{4}
\defV{probe/ftmore/linechart/nlarger/word}{four}
\defV{probe/ftmore/linechart/nreduced}{0}
\defV{probe/ftmore/linechart/nreduced/word}{zero}
\defV{probe/ftmore/qwen35_9/binding/linechart/basedrop}{0.09}
\defV{probe/ftmore/qwen35_9/binding/linechart/drop}{0.11}
\defV{probe/ftmore/qwen35_9/binding/linechart/basecontrol}{0.83}
\defV{probe/ftmore/qwen35_9/binding/linechart/control}{0.78}
\defV{probe/ftmore/qwen35_9/binding/separation/basedrop}{0.24}
\defV{probe/ftmore/qwen35_9/binding/separation/drop}{0.01}
\defV{probe/ftmore/qwen35_9/binding/separation/basecontrol}{0.51}
\defV{probe/ftmore/qwen35_9/binding/separation/control}{0.19}
\defV{probe/ftmore/qwen35_9/answer/linechart/basedrop}{0.09}
\defV{probe/ftmore/qwen35_9/answer/linechart/drop}{0.23}
\defV{probe/ftmore/qwen35_9/answer/linechart/basecontrol}{0.83}
\defV{probe/ftmore/qwen35_9/answer/linechart/control}{0.84}
\defV{probe/ftmore/qwen35_9/answer/separation/basedrop}{0.24}
\defV{probe/ftmore/qwen35_9/answer/separation/drop}{0.06}
\defV{probe/ftmore/qwen35_9/answer/separation/basecontrol}{0.51}
\defV{probe/ftmore/qwen35_9/answer/separation/control}{0.34}
\defV{probe/ftmore/glm46v_9/binding/linechart/basedrop}{0.29}
\defV{probe/ftmore/glm46v_9/binding/linechart/drop}{0.42}
\defV{probe/ftmore/glm46v_9/binding/linechart/basecontrol}{0.85}
\defV{probe/ftmore/glm46v_9/binding/linechart/control}{0.68}
\defV{probe/ftmore/glm46v_9/binding/separation/basedrop}{0.11}
\defV{probe/ftmore/glm46v_9/binding/separation/drop}{$-$0.01}
\defV{probe/ftmore/glm46v_9/binding/separation/basecontrol}{0.56}
\defV{probe/ftmore/glm46v_9/binding/separation/control}{0.24}
\defV{probe/ftmore/glm46v_9/answer/linechart/basedrop}{0.29}
\defV{probe/ftmore/glm46v_9/answer/linechart/drop}{0.31}
\defV{probe/ftmore/glm46v_9/answer/linechart/basecontrol}{0.85}
\defV{probe/ftmore/glm46v_9/answer/linechart/control}{0.81}
\defV{probe/ftmore/glm46v_9/answer/separation/basedrop}{0.11}
\defV{probe/ftmore/glm46v_9/answer/separation/drop}{0.01}
\defV{probe/ftmore/glm46v_9/answer/separation/basecontrol}{0.56}
\defV{probe/ftmore/glm46v_9/answer/separation/control}{0.35}
\defV{probe/attn/half/control}{0.39}
\defV{probe/attn/half/conflict}{0.44}
\defV{probe/attn/half/pdrop}{0.42}
\defV{probe/attn/half/pcontrol}{$1\times10^{-4}$}
\defV{probe/attn/threequarter/control}{0.35}
\defV{probe/attn/threequarter/conflict}{0.29}
\defV{probe/attn/threequarter/pdrop}{0.31}
\defV{probe/attn/threequarter/pcontrol}{$1\times10^{-5}$}
\defV{probe/attn/one/control}{0.62}
\defV{probe/attn/one/conflict}{0.31}
\defV{probe/attn/one/pdrop}{$2\times10^{-8}$}
\defV{probe/attn/one/pcontrol}{1.00}
\defV{probe/attn/oneandhalf/control}{0.39}
\defV{probe/attn/oneandhalf/conflict}{0.34}
\defV{probe/attn/oneandhalf/pdrop}{0.25}
\defV{probe/attn/oneandhalf/pcontrol}{$6\times10^{-6}$}
\defV{probe/solup/control}{0.98}
\defV{probe/solup/conflict}{0.84}
\defV{probe/solup/p}{$1\times10^{-5}$}
\defV{probe/solup/n10}{21}
\defV{probe/solup/n10/word}{21}
\defV{probe/solup/n01}{1}
\defV{probe/solup/n01/word}{one}
\defV{probe/solupthree/control}{0.97}
\defV{probe/solupthree/conflict}{0.81}
\defV{probe/solupthree/p}{$3\times10^{-5}$}
\defV{probe/solupthree/n10}{27}
\defV{probe/solupthree/n10/word}{27}
\defV{probe/solupthree/n01}{4}
\defV{probe/solupthree/n01/word}{four}
\defV{probe/claude/fable/white/control}{0.99}
\defV{probe/claude/fable/white/conflict}{0.75}
\defV{probe/claude/fable/white/control/d}{+4.49}
\defV{probe/claude/fable/white/conflict/d}{+1.35}
\defV{probe/claude/fable/white/drop}{0.24}
\defV{probe/claude/fable/white/p}{$1\times10^{-9}$}
\defV{probe/claude/fable/white/pholm}{$1\times10^{-9}$}
\defV{probe/claude/fable/white/n10}{35}
\defV{probe/claude/fable/white/n10/word}{35}
\defV{probe/claude/fable/white/n01}{1}
\defV{probe/claude/fable/white/n01/word}{one}
\defV{probe/claude/fable/white/pairs}{140}
\defV{probe/claude/fable/white/pairs/word}{140}
\defV{probe/claude/fable/white/items}{280}
\defV{probe/claude/fable/white/items/word}{280}
\defV{probe/claude/fable/white/status}{yes}
\defV{probe/claude/fable/matched/control}{0.96}
\defV{probe/claude/fable/matched/conflict}{0.64}
\defV{probe/claude/fable/matched/control/d}{+3.45}
\defV{probe/claude/fable/matched/conflict/d}{+0.73}
\defV{probe/claude/fable/matched/drop}{0.31}
\defV{probe/claude/fable/matched/p}{$4\times10^{-11}$}
\defV{probe/claude/fable/matched/pholm}{$4\times10^{-11}$}
\defV{probe/claude/fable/matched/n10}{47}
\defV{probe/claude/fable/matched/n10/word}{47}
\defV{probe/claude/fable/matched/n01}{3}
\defV{probe/claude/fable/matched/n01/word}{three}
\defV{probe/claude/fable/matched/pairs}{140}
\defV{probe/claude/fable/matched/pairs/word}{140}
\defV{probe/claude/fable/matched/items}{280}
\defV{probe/claude/fable/matched/items/word}{280}
\defV{probe/claude/fable/matched/status}{yes}
\defV{probe/claude/fable/crop/control}{1.00}
\defV{probe/claude/fable/crop/conflict}{0.93}
\defV{probe/claude/fable/crop/control/d}{+4.91}
\defV{probe/claude/fable/crop/conflict/d}{+2.90}
\defV{probe/claude/fable/crop/drop}{0.07}
\defV{probe/claude/fable/crop/p}{0.002}
\defV{probe/claude/fable/crop/pholm}{0.002}
\defV{probe/claude/fable/crop/n10}{10}
\defV{probe/claude/fable/crop/n10/word}{ten}
\defV{probe/claude/fable/crop/n01}{0}
\defV{probe/claude/fable/crop/n01/word}{zero}
\defV{probe/claude/fable/crop/pairs}{140}
\defV{probe/claude/fable/crop/pairs/word}{140}
\defV{probe/claude/fable/crop/items}{280}
\defV{probe/claude/fable/crop/items/word}{280}
\defV{probe/claude/fable/crop/status}{--}
\defV{probe/claude/fable/direct/control}{0.97}
\defV{probe/claude/fable/direct/conflict}{0.78}
\defV{probe/claude/fable/direct/control/d}{+3.60}
\defV{probe/claude/fable/direct/conflict/d}{+1.53}
\defV{probe/claude/fable/direct/drop}{0.19}
\defV{probe/claude/fable/direct/p}{$3\times10^{-6}$}
\defV{probe/claude/fable/direct/pholm}{$3\times10^{-6}$}
\defV{probe/claude/fable/direct/n10}{26}
\defV{probe/claude/fable/direct/n10/word}{26}
\defV{probe/claude/fable/direct/n01}{2}
\defV{probe/claude/fable/direct/n01/word}{two}
\defV{probe/claude/fable/direct/pairs}{128}
\defV{probe/claude/fable/direct/pairs/word}{128}
\defV{probe/claude/fable/direct/items}{268}
\defV{probe/claude/fable/direct/items/word}{268}
\defV{probe/claude/fable/direct/status}{--}
\defV{probe/claude/fable/up/control}{0.99}
\defV{probe/claude/fable/up/conflict}{0.89}
\defV{probe/claude/fable/up/p}{$3\times10^{-4}$}
\defV{probe/claude/fable/up/n10}{16}
\defV{probe/claude/fable/up/n10/word}{16}
\defV{probe/claude/fable/up/n01}{1}
\defV{probe/claude/fable/up/n01/word}{one}
\defV{probe/claude/fable/up/input}{3176}
\defV{probe/claude/fable/upthree/control}{0.99}
\defV{probe/claude/fable/upthree/conflict}{0.91}
\defV{probe/claude/fable/upthree/p}{0.002}
\defV{probe/claude/fable/upthree/n10}{13}
\defV{probe/claude/fable/upthree/n10/word}{13}
\defV{probe/claude/fable/upthree/n01}{1}
\defV{probe/claude/fable/upthree/n01/word}{one}
\defV{probe/claude/fable/upthree/input}{4912}
\defV{probe/claude/fable/white/input}{935}
\defV{probe/claude/opus/white/control}{0.96}
\defV{probe/claude/opus/white/conflict}{0.48}
\defV{probe/claude/opus/white/control/d}{+3.56}
\defV{probe/claude/opus/white/conflict/d}{$-$0.13}
\defV{probe/claude/opus/white/drop}{0.49}
\defV{probe/claude/opus/white/p}{$1\times10^{-18}$}
\defV{probe/claude/opus/white/pholm}{$2\times10^{-18}$}
\defV{probe/claude/opus/white/n10}{70}
\defV{probe/claude/opus/white/n10/word}{70}
\defV{probe/claude/opus/white/n01}{2}
\defV{probe/claude/opus/white/n01/word}{two}
\defV{probe/claude/opus/white/pairs}{140}
\defV{probe/claude/opus/white/pairs/word}{140}
\defV{probe/claude/opus/white/items}{280}
\defV{probe/claude/opus/white/items/word}{280}
\defV{probe/claude/opus/white/status}{yes}
\defV{probe/claude/opus/matched/control}{0.96}
\defV{probe/claude/opus/matched/conflict}{0.37}
\defV{probe/claude/opus/matched/control/d}{+3.56}
\defV{probe/claude/opus/matched/conflict/d}{$-$0.82}
\defV{probe/claude/opus/matched/drop}{0.59}
\defV{probe/claude/opus/matched/p}{$4\times10^{-24}$}
\defV{probe/claude/opus/matched/pholm}{$9\times10^{-24}$}
\defV{probe/claude/opus/matched/n10}{84}
\defV{probe/claude/opus/matched/n10/word}{84}
\defV{probe/claude/opus/matched/n01}{1}
\defV{probe/claude/opus/matched/n01/word}{one}
\defV{probe/claude/opus/matched/pairs}{140}
\defV{probe/claude/opus/matched/pairs/word}{140}
\defV{probe/claude/opus/matched/items}{280}
\defV{probe/claude/opus/matched/items/word}{280}
\defV{probe/claude/opus/matched/status}{yes}
\defV{probe/claude/opus/crop/control}{1.00}
\defV{probe/claude/opus/crop/conflict}{0.81}
\defV{probe/claude/opus/crop/control/d}{+4.91}
\defV{probe/claude/opus/crop/conflict/d}{+1.92}
\defV{probe/claude/opus/crop/drop}{0.19}
\defV{probe/claude/opus/crop/p}{$3\times10^{-8}$}
\defV{probe/claude/opus/crop/pholm}{$6\times10^{-8}$}
\defV{probe/claude/opus/crop/n10}{26}
\defV{probe/claude/opus/crop/n10/word}{26}
\defV{probe/claude/opus/crop/n01}{0}
\defV{probe/claude/opus/crop/n01/word}{zero}
\defV{probe/claude/opus/crop/pairs}{140}
\defV{probe/claude/opus/crop/pairs/word}{140}
\defV{probe/claude/opus/crop/items}{280}
\defV{probe/claude/opus/crop/items/word}{280}
\defV{probe/claude/opus/crop/status}{--}
\defV{probe/claude/opus/direct/control}{0.97}
\defV{probe/claude/opus/direct/conflict}{0.51}
\defV{probe/claude/opus/direct/control/d}{+3.98}
\defV{probe/claude/opus/direct/conflict/d}{+0.08}
\defV{probe/claude/opus/direct/drop}{0.46}
\defV{probe/claude/opus/direct/p}{$2\times10^{-18}$}
\defV{probe/claude/opus/direct/pholm}{$4\times10^{-18}$}
\defV{probe/claude/opus/direct/n10}{65}
\defV{probe/claude/opus/direct/n10/word}{65}
\defV{probe/claude/opus/direct/n01}{1}
\defV{probe/claude/opus/direct/n01/word}{one}
\defV{probe/claude/opus/direct/pairs}{140}
\defV{probe/claude/opus/direct/pairs/word}{140}
\defV{probe/claude/opus/direct/items}{280}
\defV{probe/claude/opus/direct/items/word}{280}
\defV{probe/claude/opus/direct/status}{--}
\defV{probe/claude/opus/up/control}{1.00}
\defV{probe/claude/opus/up/conflict}{0.63}
\defV{probe/claude/opus/up/p}{$4\times10^{-16}$}
\defV{probe/claude/opus/up/n10}{52}
\defV{probe/claude/opus/up/n10/word}{52}
\defV{probe/claude/opus/up/n01}{0}
\defV{probe/claude/opus/up/n01/word}{zero}
\defV{probe/claude/opus/up/input}{3176}
\defV{probe/claude/opus/upthree/control}{0.99}
\defV{probe/claude/opus/upthree/conflict}{0.67}
\defV{probe/claude/opus/upthree/p}{$1\times10^{-13}$}
\defV{probe/claude/opus/upthree/n10}{44}
\defV{probe/claude/opus/upthree/n10/word}{44}
\defV{probe/claude/opus/upthree/n01}{0}
\defV{probe/claude/opus/upthree/n01/word}{zero}
\defV{probe/claude/opus/upthree/input}{4912}
\defV{probe/claude/opus/white/input}{935}
\defV{probe/human/correct}{107}
\defV{probe/human/correct/word}{107}
\defV{probe/human/n}{108}
\defV{probe/human/n/word}{108}
\defV{probe/human/errors}{1}
\defV{probe/human/errors/word}{one}
\defV{probe/human/cells}{9}
\defV{probe/human/cells/word}{nine}
\defV{probe/human/control/correct}{53}
\defV{probe/human/control/correct/word}{53}
\defV{probe/human/control/n}{54}
\defV{probe/human/control/n/word}{54}
\defV{probe/human/conflict/correct}{54}
\defV{probe/human/conflict/correct/word}{54}
\defV{probe/human/conflict/n}{54}
\defV{probe/human/conflict/n/word}{54}
\defV{probe/human/common/n}{18}
\defV{probe/human/common/n/word}{18}
\defV{probe/human/common/same}{18}
\defV{probe/human/common/same/word}{18}
\defV{probe/human/common/kappa}{1.00}
\defV{probe/human/distinct}{90}
\defV{probe/human/distinct/word}{90}
\newcommand{\VT}[1]{\ifcsname tav@#1\endcsname\V{#1}\else\textcolor{orange}{TODO}\fi}
\newcommand{\supp}[1]{Supp.~#1}

\definecolor{cvprblue}{rgb}{0.21,0.49,0.74}
\usepackage[pagebackref,breaklinks,colorlinks,allcolors=cvprblue]{hyperref}
\hypersetup{pdfauthor={},pdftitle={},pdfsubject={},pdfkeywords={},pdfcreator={},pdfproducer={}}

\title{TerraArt: Reading without Binding in Map and Chart Understanding}
\author{Takaya Kawakatsu\\Preferred Networks, Inc.\\{\tt\small kat.nii.ac.jp@gmail.com}\\{\small\href{https://researchmap.jp/t.kat}{\tt researchmap.jp/t.kat}}
\and
Nam Tuan Ly\\Tokyo Metropolitan University\\{\tt\small namly@tmu.ac.jp}\\{\small\href{https://researchmap.jp/namly}{\tt researchmap.jp/namly}}
\and
Atsuhiro Takasu\\National Institute of Informatics\\{\tt\small takasu@nii.ac.jp}\\{\small\href{https://researchmap.jp/ats.takasu}{\tt researchmap.jp/ats.takasu}}}

\begin{document}
\twocolumn[{%
\renewcommand\twocolumn[1][]{#1}%
\maketitle
\begin{center}
  \centering
  \includegraphics[width=\textwidth]{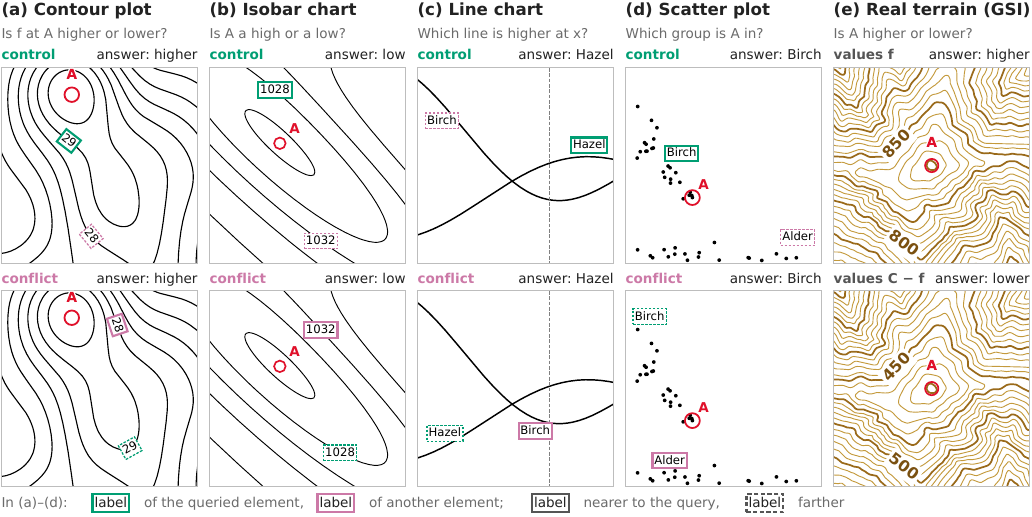}
  \captionof{figure}{Label-placement probes (a)--(e); on (a), GPT-6.1 Sol answers \V{probe/frontier/gpt6_sol/white/control} of control items but \V{probe/frontier/gpt6_sol/white/conflict} of conflict items correctly.}
  \label{fig:teaser}
\end{center}
\vspace{1ex}
}]

\begin{abstract}
We ask whether foundation models, here vision-language models (VLMs), can understand maps and charts.
We take understanding to mean binding each printed number to the geometric element it labels, a line, a point or a region, and reasoning from that binding; reading the number alone is not enough.
Our label-placement probes test this binding with contour plots in which every printed value stays on its own line and only which value lies nearest a query point changes.
GPT-6.1 Sol answered \V{probe/frontier/gpt6_sol/white/control} of the control items correctly, where the nearest label belongs to the innermost line around the query point, but only \V{probe/frontier/gpt6_sol/white/conflict} of the conflict items, where it belongs to the next line outward.
In post hoc runs, Claude Opus 5.5 fell from \V{probe/claude/opus/white/control} to \V{probe/claude/opus/white/conflict} and Claude Fable 5.2 from \V{probe/claude/fable/white/control} to \V{probe/claude/fable/white/conflict}.
In a post hoc analysis, conflict accuracy was below chance (one-sided, uncorrected) in \V{probe/scale/below} of the \V{probe/scale/n} open and GPT models, so most models follow the nearest label rather than guess.
Worked examples raised the control accuracy of the larger models but not their conflict accuracy, so the drop grew in \V{probe/prompt/fewshot/dropgrew} of \V{probe/prompt/sigmodels} open models.
In a blind check, two of the authors gave the true answer on \V{probe/human/correct} of \V{probe/human/n} answers, conflict items included.
In most models the drop is not a misreading, since it remained on items whose two values the model had transcribed correctly and when the question named the two lines.
The same placement also lowered accuracy on isobar charts, scatter plots with group labels and contour maps drawn from real terrain.
Not one of \V{probe/all/nconds} training-free interventions and tools removed the drop in any open model, and only GPT-6.1 Sol lost it in a magnified crop, where both Claude models kept it, and, post hoc, Sol kept a drop when the whole image was upscaled by the same factor as the crop.
Upscaling the whole image by two left the drop in place in all \V{probe/remedy/upscale/nkept/word} open models with a significant drop, so resolution alone does not explain the error.
LoRA fine-tuning on contour plots removed the drop there in \V{probe/ft/nremoval/word} of \V{probe/ft/nruns/word} runs but not on figure types that were not in its training data.
On terrain maps, the same placement changed the answer to whether a trail crosses a given contour line in \V{probe/route/neffect/word} of the \V{probe/route/n/word} open models that we tested.
Current VLMs thus read the labels of maps and charts but tend to bind them to the nearest element.
The generators, questions and answers will be released.
\end{abstract}

\section{Introduction}
\label{sec:intro}

We ask whether foundation models understand maps and charts, and we study vision-language models (VLMs), the foundation models that answer questions about images.
We call a model's reading of a figure \emph{understanding} if it binds each printed number to the geometric element it labels, be it a line, a point or a region, and reasons from that binding; reading the digits correctly is not sufficient.
Benchmarks for chart and map question answering score only the final answer~\cite{masry2022chartqa,xing2025can,ung2025cartomapqa,wang2024charxiv,tang2025chartmuseum}, and to our knowledge none of them varies where a label is printed independently of the element that it belongs to (\cref{sec:related}).
When the nearest label belongs to the element that is queried, a model that takes the nearest label is right, so these scores cannot separate binding from proximity.

We find that current VLMs read the labels but tend to attach each one to the nearest element instead of its own line.
On contour plots that differ only in where a label is placed, GPT-6.1 Sol was correct on \V{probe/frontier/gpt6_sol/white/control} of the items whose nearest label is the true label but on only \V{probe/frontier/gpt6_sol/white/conflict} of those whose nearest label belongs to another line.
Post hoc runs showed that Claude Fable 5.2 fell from \V{probe/claude/fable/white/control} to \V{probe/claude/fable/white/conflict} and Claude Opus 5.5 from \V{probe/claude/opus/white/control} to \V{probe/claude/opus/white/conflict}.
In a post hoc analysis of \V{probe/scale/n} models, conflict accuracy was below chance by an uncorrected one-sided test in \V{probe/scale/below}, among them GPT-6 Luna (\V{probe/frontier/gpt6_luna/white/conflict}), so these models do not guess but follow the nearest label.
Worked examples raised the control accuracy of the larger models but not their conflict accuracy, so the drop grew in \V{probe/prompt/fewshot/dropgrew} of \V{probe/prompt/sigmodels} open models.
One result runs against this pattern: GPT-6.1 Sol lost its drop in a magnified crop (control \V{probe/frontier/gpt6_sol/crop/control}, conflict \V{probe/frontier/gpt6_sol/crop/conflict}), whereas the drop stayed there in all open models with a significant drop, in GPT-6 Luna and in both Claude models.
In contrast, upscaling the whole image by two left the drop in place in all \V{probe/remedy/upscale/nkept/word} open models with a significant drop and, in a post hoc test, a smaller but still significant drop in Sol (control \V{probe/solup/control}, conflict \V{probe/solup/conflict}, p \V{probe/solup/p}), so resolution alone does not account for the error.

A label of another element placed nearest the query also reduced the accuracy of the models on isobar charts of air pressure, on scatter plots whose groups carry printed names and on contour maps of real terrain from the Geospatial Information Authority of Japan (GSI)~\cite{gsi2026dem} (\cref{sec:nearest}).
We also built line charts that name their series; in these charts the competing name is also one of the answer options, and a control without this option removed the drop, so we exclude line charts from the evidence.
Natural maps offer many occasions for this error: in \V{final/prev/windows} random windows of Japanese terrain drawn with the default label placement of a widely used plotting library, the nearest label belonged to a line other than the innermost labelled ring around the point at \V{final/prev/share} of the points with such a ring (\cref{sec:prevalence}).

We isolate this attachment using contour lines, which join points of equal value; a \emph{ring} is a closed contour line around a point, and the lines alone do not tell whether a ring encloses a summit or a pit.
A \emph{label-placement probe} is a contour plot with a query point A in which only two lines have printed values: the innermost labelled ring around the point A (\emph{A's own ring}) and the next ring outward (the \emph{neighbouring ring}).
Each value stays on its own line, and only the choice of which value is printed near A changes (\cref{fig:teaser}a).
In the \emph{control} condition the near label is the \emph{true label} on A's own ring; in the \emph{conflict} condition the near label is the \emph{competing label} on the neighbouring ring, at a matched distance.
The \emph{drop} is the control-minus-conflict difference in accuracy, or in the sensitivity $d'$ from signal-detection theory~\cite{macmillan2004detection}, where $d'=0$ means that a model gives the same answer to both members of a pair and $d'<0$ means that its hit rate lies below its false-alarm rate.
We call a significant drop of this kind \emph{nearest-label binding}; the term names a behavioural outcome, not a mechanism.

We make three contributions.
First, label-placement probes that fix the line each label belongs to and vary only its distance to the query, with a drop that persists in most models with identical line gaps, on correctly transcribed items and on three other figure types (\cref{sec:nearest,sec:alternatives}).
Second, a test of \V{probe/all/nconds} training-free interventions and tools, from instructions and worked examples up to an oracle tool that returns the correct value, none of which removed the drop in any of the open models, together with LoRA fine-tunes on contour plots that removed the drop there but not on figure types never trained (\cref{sec:remedies,sec:teach}).
Third, a probe that moves the competing label in steps, in which accuracy grows with its distance whether it lies on the next line or on the line after it, a hidden-state analysis of three open models in which the correct answer is not decodable above chance at the answer position on conflict items, a count of how often natural contour maps put the nearest label on another line and a route question on terrain maps whose answer the same placement changes (\cref{sec:where,sec:prevalence}).

\section{Related work}
\label{sec:related}

\paragraph{Chart and map question answering.}
Benchmarks of chart question answering test value reading and reasoning over plots~\cite{masry2022chartqa,wang2024charxiv,tang2025chartmuseum}, while map benchmarks~\cite{xing2025can,ung2025cartomapqa,dihan2025mapeval,mukhopadhyay2025mapwise,chang2022mapqa} test route finding, spatial questions and map reading.
In PlotQA~\cite{methani2020plotqa}, ChartQA~\cite{masry2022chartqa} and FigureQA~\cite{kahou2018figureqa}, series are mostly named by a legend rather than by labels printed next to their elements; to our knowledge, no benchmark keeps each label on its own line and reverses the answer by moving only where the labels are printed.

\paragraph{Labels, legends and binding.}
In map analysis, attaching text labels to the lines, points and regions of a map is a long-standing problem~\cite{chiang2014survey}.
In LegendBench, one model followed the order of the legend where it contradicted the plot~\cite{zhang2026legendbench}; its swap variants move names between series, but no variant keeps each name on its own element and prints the name of another element closer to the query than the element's own name.
ChartAB tests the localization of chart elements and their attributes~\cite{bansal2025chartab}, and MeasureBench finds that models misread measuring instruments and improve with reinforcement fine-tuning on synthetic gauges~\cite{lin2025measurebench}; neither moves a correct label relative to its element.
Hong \etal~\cite{hong2026trace} show that VLMs which trace a path drift to a nearby look-alike path, and Niu \etal~\cite{niu2026inside} use activation patching to find where chart values and legend binding are represented; ChartLens attributes the answers on charts to visual marks~\cite{suri2025chartlens}.
Tartaglini \etal~\cite{tartaglini2025diagnosing} decode correct point coordinates from the vision encoder of models that answer wrongly, whereas in our probes the correct answer is barely decodable from the image tokens (\cref{sec:where}).
VLMs understand diagram relations poorly~\cite{hou2025visual} and bind attributes to objects weakly~\cite{yuksekgonul2023bags,campbell2024binding}, and sharpening attention on the relevant region improves spatial relations~\cite{chen2025adaptvis}; the binding we test attaches a printed value to a drawn element rather than a feature to an object, and our probes fix the attachment by construction and test whether it is restored by remedies that target perception, instruction or tool use.
In these studies the near or look-alike element differs from the correct one in more than its position.
Our pairs keep every line and every value fixed and move only the place of one label, so the answer reverses with the place of that label alone, and moving the competing label to the line after the next separates proximity from drift to an adjacent line (\cref{sec:nearest}).

\paragraph{Controlled diagnostics and grouping.}
Diagnostics with minimal pairs expose VLM failures in perception~\cite{tong2024eyes,rahmanzadehgervi2024vision} and show that VLMs follow language priors~\cite{guan2024hallusionbench} and text that conflicts with the image~\cite{deng2025words}.
Humans group elements in a common region or connected by a line even against proximity~\cite{palmer1992common,palmer1994rethinking}; our leader lines and shaded boxes are weaker cues than these, and we collected no human data.

\section{Probes and evaluation}
\label{sec:bench}

\begin{figure}[tb]
  \centering
  \includegraphics[width=\linewidth]{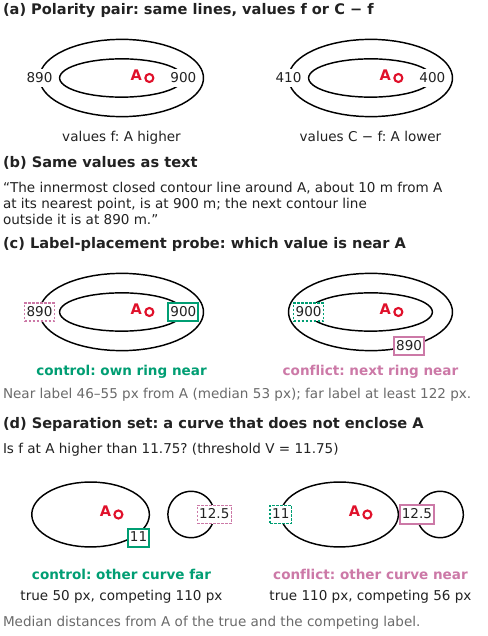}
  \caption{Design of the stimuli (schematic).}
  \label{fig:design}
\end{figure}

\subsection{Label-placement probes}
\label{sec:probes}

The probes use contour plots of smooth sums of Gaussian bumps, drawn as \emph{polarity-controlled pairs} $f$ and $C-f$ for a constant $C$, with A at a local maximum of $f$, so that every line and label position stays in place and only the printed values and the answer change (\cref{fig:design}a).
In \cref{fig:design}c, green boxes mark the true label and magenta boxes the competing label, and the solid box is the one nearer to A.
In the control condition the value of A's own ring is printed near A (\V{final/geom/placement/near/lo}--\V{final/geom/placement/near/hi} pixels) and the value of the neighbouring ring far from A (at least \V{final/geom/placement/far/lo} pixels); in the conflict condition the two positions are exchanged, with distances matched pair by pair.
Each value determines the answer when it is read from its own line; taking the near value as that of A's own ring reverses the answer only in the conflict condition.
Both conditions of a pair are drawn from the same field, so that the lines are identical apart from anti-aliasing and the gaps under the labels.

Each of the four versions of the probe has \V{final/gemma4_31/white/pairs} pairs.
In the default \emph{white-box} version each printed value sits in a white box that covers the line beneath it.
The \emph{matched} version adds an identical blank box at each label place of the other condition, so that the line gaps of the two conditions are identical.
The \emph{crop} version magnifies the area around A and the \emph{direct} version asks for the answer without reasoning.
A typical question ends with ``Is f at A (the centre of the small red ring) higher or lower than most of the values of f within a distance of 0.14 of A (in the units of the axes)? Answer with one word: higher or lower.'', followed by ``Think step by step, then end with `Answer: \textless higher or lower\textgreater'.''.
The same change was applied to other figure types (\cref{fig:teaser}b--e; \supp{Sec.~S6}).

\subsection{Models}
\label{sec:models}

The open VLMs (\V{final/models/n}, \cref{tab:final-headline}) are the Gemma-4 models 31B and 12B~\cite{gemmateam2026gemma}, three Qwen models (Qwen3.8 27B, Qwen3.5 9B and 4B), InternVL3.5 38B~\cite{wang2025internvl35}, Mistral-Small-3.2 24B and GLM-4.6V-Flash, as well as LLaVA-OneVision-2 8B and Molmo2-O 7B (revisions listed in \supp{Sec.~S1}), all served by vLLM~\cite{kwon2023vllm} \V{final/package/vllm} with greedy decoding, reasoning modes off and at most 4{,}000 output tokens.
Every model answered every question once (\V{final/answers/per_model} questions for each model), and every request returned a reply.
A reply without a final answer counts as wrong in accuracy and sign tests and is left out of $d'$, so accuracy and $d'$ can disagree for a model with many such replies; such replies were at most \V{final/unparsed/second/pct}\% of the answers of each model except Qwen3.8 27B (\V{final/qwen38_27/unparsed/pct}\%, all stopped at the token limit; \supp{Tab.~S2}).
GPT-6 Luna and GPT-6.1 Sol answered the four probe versions through their APIs at the default temperature.
Post hoc, Claude Fable 5.2 (medium effort) and Claude Opus 5.5 answered the same items and the upscaled ones through their API at the default temperature; Fable returned \V{probe/claude/fable/direct/items} of the \V{probe/claude/opus/direct/items} direct items.
In a blind check, two authors gave \V{probe/human/n} answers on \V{probe/human/distinct} distinct items from \V{probe/human/cells/word} cells (one of them a route cell) and gave the true answer on \V{probe/human/control/correct} of \V{probe/human/control/n} control and \V{probe/human/conflict/correct} of \V{probe/human/conflict/n} conflict answers, with the same answers on all \V{probe/human/common/n} items that both answered; the one error was a route item whose trail barely touches the asked line.

\subsection{Measures and tests}
\label{sec:measures}

For each condition, let $H$ be the rate of one answer (``higher'' or ``yes'') on items where it is true and $F$ its rate where it is false, with the correction $(k+0.5)/(n+1)$ for $k$ such answers among $n$ items; then $d'=z(H)-z(F)$, with $z$ the inverse standard normal distribution function~\cite{macmillan2004detection}.
All intervals are 95\% percentile intervals from 2{,}000 bootstrap draws over pairs~\cite{efron1994bootstrap}.
Each contrast is decided by an exact two-sided sign test on accuracy over items paired by pair and member; $n_{10}$ and $n_{01}$ count the items correct only in the control and only in the conflict condition, and $\Delta d'$ is descriptive.
Because the two members of a pair share every line, we also ran each test with pairs as units, counting the pairs with more correct answers in the control or the conflict condition (post hoc); \V{probe/pair/nsame/word} of the \V{probe/pair/ntests/word} decisions were unchanged and the one that changed is the contrast of Qwen3.8 27B on isobar charts (Holm-corrected $p$ of \V{probe/pair/qwen38_27/isobar/p} instead of \V{probe/pair/qwen38_27/isobar/itemp}).
A contrast is \emph{decidable} when the control $d'$ is at least 0.5 and $n_{10}+n_{01}\ge10$; for GSI maps, the control $d'$ on separate readability pairs must also be at least 0.5.

The \emph{primary test} is the Holm-corrected white-box contrast over models at $\alpha=0.05$~\cite{holm1979simple}; the \emph{reversal} holds when it is passed and the upper end of the interval of the conflict $d'$ is below zero.
For the models passing the primary test, the matched, crop and direct versions are tested in this order at $\alpha=0.025$ with a stop at the first failure, Holm-corrected over models at each step, and the decidable contrasts on the four other figure types form a single Holm family across models and figure types at $\alpha=0.025$.
We count the reversal over the decidable models, a choice made after the results; Gemma-4 31B had already answered every set while we checked the generators, so we use Mistral-Small-3.2 24B as the example.
Analyses added after the results are marked \emph{post hoc}.
Every probe had its items, tests and analysis plan stored under version control before its first answer.

\section{Nearest-label binding}
\label{sec:results}

\begin{table*}[tb]
  \centering
  \footnotesize
  \setlength{\tabcolsep}{2.4pt}
  \begin{tabular}{@{}lccccccccc@{}}
\toprule
Model & Acc.\ control & Acc.\ conflict & $d'$ control / conflict & $\Delta d'$ [95\% interval] & $n_{10}{:}n_{01}$ & $p$ & Drop & Reversal & Read near / far \\
\midrule
Gemma-4 31B$^{*}$ & \V{final/gemma4_31/white/control/acc} & \V{final/gemma4_31/white/conflict/acc} & \V{final/gemma4_31/white/control/d} / \V{final/gemma4_31/white/conflict/d} & \V{final/gemma4_31/white/diff} \V{final/gemma4_31/white/ci} & \V{final/gemma4_31/white/b}:\V{final/gemma4_31/white/c} & \V{final/gemma4_31/white/pholm} & \V{final/gemma4_31/white/status} & \V{final/gemma4_31/white/reversal} & \V{final/probe/gemma4_31/near} / \V{final/probe/gemma4_31/far} \\
Gemma-4 12B & \V{final/gemma4_12/white/control/acc} & \V{final/gemma4_12/white/conflict/acc} & \V{final/gemma4_12/white/control/d} / \V{final/gemma4_12/white/conflict/d} & \V{final/gemma4_12/white/diff} \V{final/gemma4_12/white/ci} & \V{final/gemma4_12/white/b}:\V{final/gemma4_12/white/c} & \V{final/gemma4_12/white/pholm} & \V{final/gemma4_12/white/status} & \V{final/gemma4_12/white/reversal} & \V{final/probe/gemma4_12/near} / \V{final/probe/gemma4_12/far} \\
Qwen3.8 27B & \V{final/qwen38_27/white/control/acc} & \V{final/qwen38_27/white/conflict/acc} & \V{final/qwen38_27/white/control/d} / \V{final/qwen38_27/white/conflict/d} & \V{final/qwen38_27/white/diff} \V{final/qwen38_27/white/ci} & \V{final/qwen38_27/white/b}:\V{final/qwen38_27/white/c} & \V{final/qwen38_27/white/pholm} & \V{final/qwen38_27/white/status} & \V{final/qwen38_27/white/reversal} & \V{final/probe/qwen38_27/near} / \V{final/probe/qwen38_27/far} \\
Qwen3.5 9B & \V{final/qwen35_9/white/control/acc} & \V{final/qwen35_9/white/conflict/acc} & \V{final/qwen35_9/white/control/d} / \V{final/qwen35_9/white/conflict/d} & \V{final/qwen35_9/white/diff} \V{final/qwen35_9/white/ci} & \V{final/qwen35_9/white/b}:\V{final/qwen35_9/white/c} & \V{final/qwen35_9/white/pholm} & \V{final/qwen35_9/white/status} & \V{final/qwen35_9/white/reversal} & \V{final/probe/qwen35_9/near} / \V{final/probe/qwen35_9/far} \\
Qwen3.5 4B & \V{final/qwen35_4/white/control/acc} & \V{final/qwen35_4/white/conflict/acc} & \V{final/qwen35_4/white/control/d} / \V{final/qwen35_4/white/conflict/d} & \V{final/qwen35_4/white/diff} \V{final/qwen35_4/white/ci} & \V{final/qwen35_4/white/b}:\V{final/qwen35_4/white/c} & \V{final/qwen35_4/white/pholm} & \V{final/qwen35_4/white/status} & \V{final/qwen35_4/white/reversal} & \V{final/probe/qwen35_4/near} / \V{final/probe/qwen35_4/far} \\
InternVL3.5 38B & \V{final/internvl35_38/white/control/acc} & \V{final/internvl35_38/white/conflict/acc} & \V{final/internvl35_38/white/control/d} / \V{final/internvl35_38/white/conflict/d} & \V{final/internvl35_38/white/diff} \V{final/internvl35_38/white/ci} & \V{final/internvl35_38/white/b}:\V{final/internvl35_38/white/c} & \V{final/internvl35_38/white/pholm} & \V{final/internvl35_38/white/status} & \V{final/internvl35_38/white/reversal} & \V{final/probe/internvl35_38/near} / \V{final/probe/internvl35_38/far} \\
Mistral-Small-3.2 24B & \V{final/mistral32_24/white/control/acc} & \V{final/mistral32_24/white/conflict/acc} & \V{final/mistral32_24/white/control/d} / \V{final/mistral32_24/white/conflict/d} & \V{final/mistral32_24/white/diff} \V{final/mistral32_24/white/ci} & \V{final/mistral32_24/white/b}:\V{final/mistral32_24/white/c} & \V{final/mistral32_24/white/pholm} & \V{final/mistral32_24/white/status} & \V{final/mistral32_24/white/reversal} & \V{final/probe/mistral32_24/near} / \V{final/probe/mistral32_24/far} \\
GLM-4.6V-Flash & \V{final/glm46v_9/white/control/acc} & \V{final/glm46v_9/white/conflict/acc} & \V{final/glm46v_9/white/control/d} / \V{final/glm46v_9/white/conflict/d} & \V{final/glm46v_9/white/diff} \V{final/glm46v_9/white/ci} & \V{final/glm46v_9/white/b}:\V{final/glm46v_9/white/c} & \V{final/glm46v_9/white/pholm} & \V{final/glm46v_9/white/status} & \V{final/glm46v_9/white/reversal} & \V{final/probe/glm46v_9/near} / \V{final/probe/glm46v_9/far} \\
LLaVA-OneVision-2 8B & \V{final/llavaov2_8/white/control/acc} & \V{final/llavaov2_8/white/conflict/acc} & \V{final/llavaov2_8/white/control/d} / \V{final/llavaov2_8/white/conflict/d} & \V{final/llavaov2_8/white/diff} \V{final/llavaov2_8/white/ci} & \V{final/llavaov2_8/white/b}:\V{final/llavaov2_8/white/c} & \V{final/llavaov2_8/white/pholm} & \V{final/llavaov2_8/white/status} & \V{final/llavaov2_8/white/reversal} & \V{final/probe/llavaov2_8/near} / \V{final/probe/llavaov2_8/far} \\
Molmo2-O 7B & \V{final/molmo2o_7/white/control/acc} & \V{final/molmo2o_7/white/conflict/acc} & \V{final/molmo2o_7/white/control/d} / \V{final/molmo2o_7/white/conflict/d} & \V{final/molmo2o_7/white/diff} \V{final/molmo2o_7/white/ci} & \V{final/molmo2o_7/white/b}:\V{final/molmo2o_7/white/c} & \V{final/molmo2o_7/white/pholm} & \V{final/molmo2o_7/white/status} & \V{final/molmo2o_7/white/reversal} & \V{final/probe/molmo2o_7/near} / \V{final/probe/molmo2o_7/far} \\
\midrule
GPT-6 Luna & \V{final/frontier/gpt6_luna/white/control/acc} & \V{final/frontier/gpt6_luna/white/conflict/acc} & \V{final/frontier/gpt6_luna/white/control/d} / \V{final/frontier/gpt6_luna/white/conflict/d} & -- & \V{final/frontier/gpt6_luna/white/b}:\V{final/frontier/gpt6_luna/white/c} & \V{final/frontier/gpt6_luna/white/pholm} & \V{final/frontier/gpt6_luna/white/status} & -- & -- \\
GPT-6.1 Sol & \V{final/frontier/gpt6_sol/white/control/acc} & \V{final/frontier/gpt6_sol/white/conflict/acc} & \V{final/frontier/gpt6_sol/white/control/d} / \V{final/frontier/gpt6_sol/white/conflict/d} & -- & \V{final/frontier/gpt6_sol/white/b}:\V{final/frontier/gpt6_sol/white/c} & \V{final/frontier/gpt6_sol/white/pholm} & \V{final/frontier/gpt6_sol/white/status} & -- & -- \\
Claude Fable 5.2$^{\dagger}$ & \V{probe/claude/fable/white/control} & \V{probe/claude/fable/white/conflict} & \V{probe/claude/fable/white/control/d} / \V{probe/claude/fable/white/conflict/d} & -- & \V{probe/claude/fable/white/n10}:\V{probe/claude/fable/white/n01} & \V{probe/claude/fable/white/pholm} & \V{probe/claude/fable/white/status} & -- & -- \\
Claude Opus 5.5$^{\dagger}$ & \V{probe/claude/opus/white/control} & \V{probe/claude/opus/white/conflict} & \V{probe/claude/opus/white/control/d} / \V{probe/claude/opus/white/conflict/d} & -- & \V{probe/claude/opus/white/n10}:\V{probe/claude/opus/white/n01} & \V{probe/claude/opus/white/pholm} & \V{probe/claude/opus/white/status} & -- & -- \\
\bottomrule
\end{tabular}
  \caption{Label-placement probe with white boxes, the same \V{final/gemma4_31/white/pairs} pairs of fields for every model; $^{\dagger}$post hoc.}
  \label{tab:final-headline}
\end{table*}

\subsection{The drop and its robustness}
\label{sec:nearest}

The primary test was passed by \V{final/dec/white/k} of the \V{final/models/n} models (\cref{tab:final-headline}; $p$ is Holm-corrected, n.d.\ denotes a contrast that cannot be decided, the asterisk marks the model that had already seen the sets, and Read gives the shares of near and far labels transcribed correctly as in \cref{sec:alternatives}).
For Mistral-Small-3.2 24B the accuracy fell from \V{final/mistral32_24/white/control/acc} to \V{final/mistral32_24/white/conflict/acc} and $d'$ from \V{final/mistral32_24/white/control/d} to \V{final/mistral32_24/white/conflict/d} ($n_{10}{:}n_{01}={}$\V{final/mistral32_24/white/b}:\V{final/mistral32_24/white/c}), while the lines and the two printed values were the same.
In \V{final/dec/reversal/k} of the \V{final/dec/reversal/decidable} decidable models the conflict $d'$ was below zero with its whole interval.
For Gemma-4 12B and GLM-4.6V-Flash the contrast was not decidable, as their control $d'$ fell below 0.5, and the drop of Molmo2-O 7B was not significant with only \V{final/molmo2o_7/white/b}:\V{final/molmo2o_7/white/c} discordant items.

\begin{table*}[tb]
  \centering
  \footnotesize
  \setlength{\tabcolsep}{3pt}
  \begin{tabular}{@{}lccccccc@{}}
\toprule
Model & Matched & Crop & Direct & Separ. & Isobar & Scatter & GSI \\
\midrule
Gemma-4 31B & \V{final/gemma4_31/matched/diff} \V{final/gemma4_31/matched/status} & \V{final/gemma4_31/crop/diff} \V{final/gemma4_31/crop/status} & \V{final/gemma4_31/direct/diff} \V{final/gemma4_31/direct/status} & \V{final/gemma4_31/separation/diff} \V{final/gemma4_31/separation/status} & \V{final/gemma4_31/isobar/diff} \V{final/gemma4_31/isobar/status} & \V{final/gemma4_31/scatter/diff} \V{final/gemma4_31/scatter/status} & \V{final/gemma4_31/gsi_placement/diff} \V{final/gemma4_31/gsi_placement/status} \\
Gemma-4 12B & \V{final/gemma4_12/matched/diff} \V{final/gemma4_12/matched/status} & \V{final/gemma4_12/crop/diff} \V{final/gemma4_12/crop/status} & \V{final/gemma4_12/direct/diff} \V{final/gemma4_12/direct/status} & \V{final/gemma4_12/separation/diff} \V{final/gemma4_12/separation/status} & \V{final/gemma4_12/isobar/diff} \V{final/gemma4_12/isobar/status} & \V{final/gemma4_12/scatter/diff} \V{final/gemma4_12/scatter/status} & \V{final/gemma4_12/gsi_placement/diff} \V{final/gemma4_12/gsi_placement/status} \\
Qwen3.8 27B & \V{final/qwen38_27/matched/diff} \V{final/qwen38_27/matched/status} & \V{final/qwen38_27/crop/diff} \V{final/qwen38_27/crop/status} & \V{final/qwen38_27/direct/diff} \V{final/qwen38_27/direct/status} & \V{final/qwen38_27/separation/diff} \V{final/qwen38_27/separation/status} & \V{final/qwen38_27/isobar/diff} \V{final/qwen38_27/isobar/status} & \V{final/qwen38_27/scatter/diff} \V{final/qwen38_27/scatter/status} & \V{final/qwen38_27/gsi_placement/diff} \V{final/qwen38_27/gsi_placement/status} \\
Qwen3.5 9B & \V{final/qwen35_9/matched/diff} \V{final/qwen35_9/matched/status} & \V{final/qwen35_9/crop/diff} \V{final/qwen35_9/crop/status} & \V{final/qwen35_9/direct/diff} \V{final/qwen35_9/direct/status} & \V{final/qwen35_9/separation/diff} \V{final/qwen35_9/separation/status} & \V{final/qwen35_9/isobar/diff} \V{final/qwen35_9/isobar/status} & \V{final/qwen35_9/scatter/diff} \V{final/qwen35_9/scatter/status} & \V{final/qwen35_9/gsi_placement/diff} \V{final/qwen35_9/gsi_placement/status} \\
Qwen3.5 4B & \V{final/qwen35_4/matched/diff} \V{final/qwen35_4/matched/status} & \V{final/qwen35_4/crop/diff} \V{final/qwen35_4/crop/status} & \V{final/qwen35_4/direct/diff} \V{final/qwen35_4/direct/status} & \V{final/qwen35_4/separation/diff} \V{final/qwen35_4/separation/status} & \V{final/qwen35_4/isobar/diff} \V{final/qwen35_4/isobar/status} & \V{final/qwen35_4/scatter/diff} \V{final/qwen35_4/scatter/status} & \V{final/qwen35_4/gsi_placement/diff} \V{final/qwen35_4/gsi_placement/status} \\
InternVL3.5 38B & \V{final/internvl35_38/matched/diff} \V{final/internvl35_38/matched/status} & \V{final/internvl35_38/crop/diff} \V{final/internvl35_38/crop/status} & \V{final/internvl35_38/direct/diff} \V{final/internvl35_38/direct/status} & \V{final/internvl35_38/separation/diff} \V{final/internvl35_38/separation/status} & \V{final/internvl35_38/isobar/diff} \V{final/internvl35_38/isobar/status} & \V{final/internvl35_38/scatter/diff} \V{final/internvl35_38/scatter/status} & \V{final/internvl35_38/gsi_placement/diff} \V{final/internvl35_38/gsi_placement/status} \\
Mistral-Small-3.2 24B & \V{final/mistral32_24/matched/diff} \V{final/mistral32_24/matched/status} & \V{final/mistral32_24/crop/diff} \V{final/mistral32_24/crop/status} & \V{final/mistral32_24/direct/diff} \V{final/mistral32_24/direct/status} & \V{final/mistral32_24/separation/diff} \V{final/mistral32_24/separation/status} & \V{final/mistral32_24/isobar/diff} \V{final/mistral32_24/isobar/status} & \V{final/mistral32_24/scatter/diff} \V{final/mistral32_24/scatter/status} & \V{final/mistral32_24/gsi_placement/diff} \V{final/mistral32_24/gsi_placement/status} \\
GLM-4.6V-Flash & \V{final/glm46v_9/matched/diff} \V{final/glm46v_9/matched/status} & \V{final/glm46v_9/crop/diff} \V{final/glm46v_9/crop/status} & \V{final/glm46v_9/direct/diff} \V{final/glm46v_9/direct/status} & \V{final/glm46v_9/separation/diff} \V{final/glm46v_9/separation/status} & \V{final/glm46v_9/isobar/diff} \V{final/glm46v_9/isobar/status} & \V{final/glm46v_9/scatter/diff} \V{final/glm46v_9/scatter/status} & \V{final/glm46v_9/gsi_placement/diff} \V{final/glm46v_9/gsi_placement/status} \\
LLaVA-OneVision-2 8B & \V{final/llavaov2_8/matched/diff} \V{final/llavaov2_8/matched/status} & \V{final/llavaov2_8/crop/diff} \V{final/llavaov2_8/crop/status} & \V{final/llavaov2_8/direct/diff} \V{final/llavaov2_8/direct/status} & \V{final/llavaov2_8/separation/diff} \V{final/llavaov2_8/separation/status} & \V{final/llavaov2_8/isobar/diff} \V{final/llavaov2_8/isobar/status} & \V{final/llavaov2_8/scatter/diff} \V{final/llavaov2_8/scatter/status} & \V{final/llavaov2_8/gsi_placement/diff} \V{final/llavaov2_8/gsi_placement/status} \\
Molmo2-O 7B & \V{final/molmo2o_7/matched/diff} \V{final/molmo2o_7/matched/status} & \V{final/molmo2o_7/crop/diff} \V{final/molmo2o_7/crop/status} & \V{final/molmo2o_7/direct/diff} \V{final/molmo2o_7/direct/status} & \V{final/molmo2o_7/separation/diff} \V{final/molmo2o_7/separation/status} & \V{final/molmo2o_7/isobar/diff} \V{final/molmo2o_7/isobar/status} & \V{final/molmo2o_7/scatter/diff} \V{final/molmo2o_7/scatter/status} & \V{final/molmo2o_7/gsi_placement/diff} \V{final/molmo2o_7/gsi_placement/status} \\
\midrule
Holds / tested & \V{final/dec/matched/k}/\V{final/dec/matched/tested} & \V{final/dec/crop/k}/\V{final/dec/crop/tested} & \V{final/dec/direct/k}/\V{final/dec/direct/tested} & \V{final/dec/separation/k}/\V{final/dec/separation/tested} & \V{final/dec/isobar/k}/\V{final/dec/isobar/tested} & \V{final/dec/scatter/k}/\V{final/dec/scatter/tested} & \V{final/dec/gsi_placement/k}/\V{final/dec/gsi_placement/tested} \\
\bottomrule
\end{tabular}
  \caption{Main tests: $\Delta d'$ and outcome per model.}
  \label{tab:final-tests}
\end{table*}

With identical line gaps the drop held in \V{final/dec/matched/k} of the \V{final/dec/matched/tested} models tested, and in a magnified crop in all \V{final/dec/crop/tested} (\cref{tab:final-tests}; n.d.\ marks a contrast that the items did not decide and a dash a test that the stopping rule skipped).
With the direct prompt the contrast was decidable only for Gemma-4 31B, where it held.
With reasoning on and up to 24{,}000 tokens, Qwen3.8 27B kept the drop in each of the four versions (white box $\Delta d'={}$\V{final/thinking/qwen38_27/white/diff}).
For GPT-6 Luna, accuracy fell from \V{final/frontier/gpt6_luna/white/control/acc} to \V{final/frontier/gpt6_luna/white/conflict/acc} and for GPT-6.1 Sol from \V{final/frontier/gpt6_sol/white/control/acc} to \V{final/frontier/gpt6_sol/white/conflict/acc}, both significant after Holm correction over the two, and the drop remained when the line gaps were identical; in the magnified crop it vanished in Sol (control \V{final/frontier/gpt6_sol/crop/control/acc}, conflict \V{final/frontier/gpt6_sol/crop/conflict/acc}) but not in Luna, in either Claude model or in any open model with a significant drop of its own.
In the open models, upscaling the whole image by two left the drop in place in all \V{probe/remedy/upscale/nkept/word} models that had a significant drop, so resolution alone does not explain their error.
In a post hoc test, GPT-6.1 Sol kept a drop on images that were upscaled by two (control \V{probe/solup/control}, conflict \V{probe/solup/conflict}, p \V{probe/solup/p}) and by three (control \V{probe/solupthree/control}, conflict \V{probe/solupthree/conflict}, p \V{probe/solupthree/p}), which matches the magnification of the crop.
The crop keeps both of the labels but removes the lines far from A, so what it leaves out, rather than magnification, may have removed the error of Sol.
Post hoc, Claude Fable 5.2 and Claude Opus 5.5 both kept a significant drop in the magnified crop (conflict \V{probe/claude/fable/crop/conflict} and \V{probe/claude/opus/crop/conflict}, $n_{10}{:}n_{01}$ \V{probe/claude/fable/crop/n10}:\V{probe/claude/fable/crop/n01} and \V{probe/claude/opus/crop/n10}:\V{probe/claude/opus/crop/n01}) and on the images upscaled by two (conflict \V{probe/claude/fable/up/conflict} and \V{probe/claude/opus/up/conflict}) and by three (\V{probe/claude/fable/upthree/conflict} and \V{probe/claude/opus/upthree/conflict}).
Post hoc, with the direct prompt most open models gave the same answer to nearly every item, whatever was drawn.
Across the \V{probe/scale/n} models, conflict accuracy was below chance in \V{probe/scale/below}, by a one-sided binomial test without correction, and the rank correlation between control and conflict accuracy across these models remained undecided ($\rho={}$\V{probe/scale/rho}, $p={}$\V{probe/scale/p}; post hoc).

The same placement of a competing label also lowered accuracy on a separation set, where the competing label lies on a closed curve that does not enclose A (\V{final/dec/separation/k} of \V{final/dec/separation/tested} models), on isobar charts (\V{final/dec/isobar/k} of \V{final/dec/isobar/tested}), on scatter plots with group labels (\V{final/dec/scatter/k} of \V{final/dec/scatter/tested}) and on contour maps of real terrain from the Geospatial Information Authority of Japan (GSI)~\cite{gsi2026dem} (\V{final/dec/gsi_placement/k} of \V{final/dec/gsi_placement/tested}); except for the reversal counted over all models, all decisions across models held (\supp{Sec.~S6}).
On polarity-controlled pairs of real terrain maps, stating the values and the lines around A in one extra sentence raised $d'$ in every model, and short leader lines or light shaded boxes kept the drop on scatter plots in every model with either cue; these cues show which group a label names, not which group holds the query point (\supp{Sec.~S6}).
On isobar charts, a version that asks which of the two printed values belongs to the smallest labelled isobar around A showed a significant drop in \V{probe/value/nsupported/word} of \V{probe/value/n/word} models, whose control accuracy was \V{probe/value/controllo}--\V{probe/value/controlhi} (Gemma-4 31B: control \V{probe/value/gemma4_31/control}, conflict \V{probe/value/gemma4_31/conflict}).

\begin{figure}[tb]
  \centering
  \includegraphics[width=\linewidth]{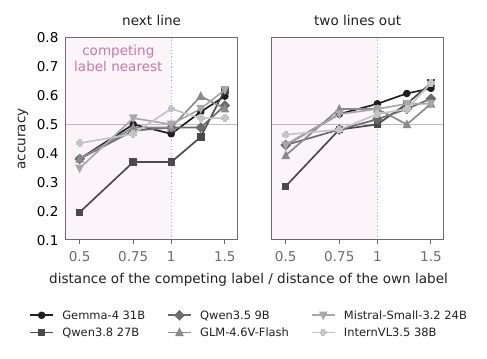}
  \caption{Accuracy against the distance of the competing label.}
  \label{fig:proximity}
\end{figure}

\paragraph{Distance of the competing label.}
To ask whether the error is a drift to the adjacent line while tracing~\cite{hong2026trace} or a pull of the nearer label, we redrew \V{probe/prox/next/pairs} white-box pairs with the label of A's own ring about 120 pixels away from A and the competing label at 0.5 to 1.5 times that distance on the next line outward, and \V{probe/prox/after/pairs} pairs with the competing label on the line after that (post hoc; \cref{fig:proximity}).
Accuracy rose with the distance of the competing label in \V{probe/prox/npositive/word} of the \V{probe/prox/ncombos/word} combinations of \V{probe/prox/nmodels/word} open models and two lines, with slopes per unit of log ratio of \V{probe/prox/next/slopelo} to \V{probe/prox/next/slopehi} on the next line and \V{probe/prox/after/slopelo} to \V{probe/prox/after/slopehi} on the line after; for Qwen3.8 27B accuracy rose from \V{probe/prox/qwen38_27/next/acc/0.5} to \V{probe/prox/qwen38_27/next/acc/1.5}.
The slopes were in a similar range when the competing label lay two lines away, so the error follows the distance of the label and not whether its line is adjacent to the ring of A.
Accuracy changed gradually and not abruptly at the ratio of one where the nearest label changes, and it stayed at most \V{probe/prox/acc15max} when the label of A's ring was the nearer one, so the distance of the own label from A also matters.

\subsection{Misreading and question reading}
\label{sec:alternatives}

In a transcription probe, every model read out the two printed values of the same white-box pictures without the question about A.
On the conflict items the far label was transcribed correctly in \V{final/probe/far/min} to \V{final/probe/far/max} of the items, and the near label was read no better than the far label (mean difference \V{final/probe/gap}).
Post hoc, on the items whose two values a model had transcribed correctly in both the control and the conflict condition (\V{final/probe/item/kept/min} to \V{final/probe/item/kept/max} items per model), the drop stayed significant after Holm correction in \V{final/probe/item/kholm} of the \V{final/dec/reversal/decidable} decidable models (Mistral-Small-3.2 24B from \V{final/probe/mistral32_24/item/control} to \V{final/probe/mistral32_24/item/conflict}); this does not estimate the drop on all items, as the kept items differ across models.
A model may also take the question to ask for the number printed nearest A, as for a spot height.
When the question named the lines (``Find the innermost closed contour line around A and the next contour line outward.''), the drop persisted in \V{final/spot/persists} of the \V{final/spot/decided} models whose drop was significant in this run; for Mistral-Small-3.2 24B it fell from \V{final/spot/mistral32_24/original/drop} to \V{final/spot/mistral32_24/named/drop}, so part of its drop may stem from the question.

\section{Robustness to training-free interventions}
\label{sec:remedies}

\begin{table}[tb]
  \centering
  \footnotesize
  \setlength{\tabcolsep}{2pt}
  \begin{tabular}{@{}llccccc@{}}
    \toprule
    Kind & Condition & Models & Removed & Reduced & Kept & Undecided \\
    \midrule
    Image & Upscale & \V{probe/remedy/sigmodels} & \V{probe/remedy/upscale/nremoved} & \V{probe/remedy/upscale/nreduced} & \V{probe/remedy/upscale/nkept} & \V{probe/remedy/upscale/nundecided} \\
    & Crop & \V{probe/remedy/sigmodels} & \V{probe/remedy/crop/nremoved} & \V{probe/remedy/crop/nreduced} & \V{probe/remedy/crop/nkept} & \V{probe/remedy/crop/nundecided} \\
    Prompt & Instruction & \V{probe/remedy/sigmodels} & \V{probe/remedy/instruction/nremoved} & \V{probe/remedy/instruction/nreduced} & \V{probe/remedy/instruction/nkept} & \V{probe/remedy/instruction/nundecided} \\
    & Describe first & \V{probe/remedy/sigmodels} & \V{probe/remedy/self_structure/nremoved} & \V{probe/remedy/self_structure/nreduced} & \V{probe/remedy/self_structure/nkept} & \V{probe/remedy/self_structure/nundecided} \\
    & Worked examples & \V{probe/prompt/sigmodels} & \V{probe/prompt/fewshot/nremoved} & \V{probe/prompt/fewshot/nreduced} & \V{probe/prompt/fewshot/nkept} & \V{probe/prompt/fewshot/nundecided} \\
    & Procedure & \V{probe/prompt/sigmodels} & \V{probe/prompt/procedure/nremoved} & \V{probe/prompt/procedure/nreduced} & \V{probe/prompt/procedure/nkept} & \V{probe/prompt/procedure/nundecided} \\
    & Ordering check & \V{probe/prompt/sigmodels} & \V{probe/prompt/ordering_check/nremoved} & \V{probe/prompt/ordering_check/nreduced} & \V{probe/prompt/ordering_check/nkept} & \V{probe/prompt/ordering_check/nundecided} \\
    Tool & Zoom & \V{probe/tool/sigmodels} & \V{probe/tool/zoom/nremoved} & \V{probe/tool/zoom/nreduced} & \V{probe/tool/zoom/nkept} & \V{probe/tool/zoom/nundecided} \\
    & OCR & \V{probe/tool/sigmodels} & \V{probe/tool/ocr/nremoved} & \V{probe/tool/ocr/nreduced} & \V{probe/tool/ocr/nkept} & \V{probe/tool/ocr/nundecided} \\
    & Code & \V{probe/tool/sigmodels} & \V{probe/tool/code/nremoved} & \V{probe/tool/code/nreduced} & \V{probe/tool/code/nkept} & \V{probe/tool/code/nundecided} \\
    & Ring oracle & \V{probe/tool/sigmodels} & \V{probe/tool/oracle/nremoved} & \V{probe/tool/oracle/nreduced} & \V{probe/tool/oracle/nkept} & \V{probe/tool/oracle/nundecided} \\
    \bottomrule
  \end{tabular}
  \caption{Training-free interventions and tools: models by decision.}
  \label{tab:remedies}
\end{table}

\subsection{Conditions and outcomes}

We call each training-free intervention or tool a \emph{remedy}, and each remedy was posed on the \V{final/gemma4_31/white/pairs} white-box pairs to the open models whose white-box drop was significant, or to five of these for the tools (column Models in \cref{tab:remedies}).
The image remedies upscale the picture or use the magnified crop.
The prompt remedies add one sentence stating that a printed number gives the value of the line on which it sits (\emph{instruction}), ask the model in a first turn to list each label near A with its line and whether that line encloses A and then answer on the basis of this description (\emph{describe first}), give three worked examples on GSI contour maps that share no field with any test item, each with a reasoning written by a fixed template (\emph{worked examples}), prescribe the steps of the reading (\emph{procedure}) or ask the model to check the ordering of the two values (\emph{ordering check}).
The tool remedies let a model call, up to eight times, a zoom tool, a label tool that returns the printed values with their positions (OCR in \cref{tab:remedies}), a code interpreter, or a \emph{ring oracle} that returns the value of the innermost labelled ring around a queried point; the oracle returns the true value at A on all but a few items and thus gives an upper bound on what perception could contribute.
A remedy \emph{removes} the drop in a model when the drop becomes less than half of the drop without the remedy and not significant while conflict accuracy rises significantly (Holm over models), \emph{reduces} it when conflict accuracy rises significantly without removing the drop, \emph{keeps} it when neither condition holds and the drop remains significant, and is \emph{undecided} otherwise.

\subsection{No remedy removed the drop}

No condition removed the drop in any model (\cref{tab:remedies}).
\V{probe/all/reducedconds/Word} conditions reduced it in \V{probe/all/reducedmax/word} model each: describe first did so in LLaVA-OneVision-2 8B, largely through fewer replies without any answer, and the ring oracle in Qwen3.8 27B (conflict accuracy from \V{probe/tool/plain/qwen38_27/conflict} to \V{probe/tool/oracle/qwen38_27/conflict}, with a drop of \V{probe/tool/oracle/qwen38_27/drop} that stayed significant).
Most of the undecided cases are costly ones in which a remedy also lowered control accuracy, often because tool conversations used all calls and ended without an answer.

The remedies that help perception or instruction raised control accuracy but not conflict accuracy.
With worked examples, Gemma-4 31B rose from \V{probe/prompt/plain/gemma4_31/control} to \V{probe/prompt/fewshot/gemma4_31/control} on control items while its conflict accuracy went from \V{probe/prompt/plain/gemma4_31/conflict} to \V{probe/prompt/fewshot/gemma4_31/conflict}, and the drop grew in \V{probe/prompt/fewshot/dropgrew} of \V{probe/prompt/sigmodels} models; the magnified crop widened it in \V{probe/remedy/crop/dropgrew} of \V{probe/remedy/sigmodels}.
Low control accuracy therefore does not limit conflict accuracy, since models that solve nearly all control items still follow the near label.

Post hoc, we asked where the models queried the ring oracle and what they did with its answer.
Only Qwen3.8 27B placed its first call near A (within \V{probe/posthoc/radius} pixels on \V{probe/posthoc/qwen38_27/atA} of \V{probe/posthoc/qwen38_27/items} items; Gemma-4 31B on \V{probe/posthoc/gemma4_31/atA}).
On the items where the oracle had returned the true value of A's own ring, accuracy was still lower in the conflict condition, \V{probe/posthoc/gemma4_31/innercontrol} against \V{probe/posthoc/gemma4_31/innerconflict} for Gemma-4 31B and \V{probe/posthoc/qwen38_27/innercontrol} against \V{probe/posthoc/qwen38_27/innerconflict} for Qwen3.8 27B (without a significance test).
Holding the correct value in the conversation did not make these models use it.

Post hoc, we scaled the attention logits of the image tokens in Qwen3.5 9B by a fixed factor, a fixed-factor form of the attention adjustment of AdaptVis~\cite{chen2025adaptvis}.
Softer attention (factor 0.5 or 0.75) and sharper attention (1.5) all made the drop non-significant, but only because control accuracy fell from \V{probe/attn/one/control} to between \V{probe/attn/threequarter/control} and \V{probe/attn/half/control}; conflict accuracy rose only at 0.5, to \V{probe/attn/half/conflict}, and a factor of 2 stopped the model from giving answers.

\section{Fine-tuning on contour plots}
\label{sec:teach}

\begin{table}[tb]
  \centering
  \footnotesize
  \setlength{\tabcolsep}{4pt}
  \begin{tabular}{@{}llcccc@{}}
    \toprule
    Model & Training & Contour & Isobar & Scatter & GSI \\
    \midrule
    Qwen3.5 9B & None & \V{probe/ft/qwen35_9/base/contour/drop} & \V{probe/ft/qwen35_9/base/isobar/drop} & \V{probe/ft/qwen35_9/base/scatter/drop} & \V{probe/ft/qwen35_9/base/gsi/drop} \\
    & Answer only & \V{probe/ft/qwen35_9/answer/contour/drop} & \V{probe/ft/qwen35_9/answer/isobar/drop} & \V{probe/ft/qwen35_9/answer/scatter/drop} & \V{probe/ft/qwen35_9/answer/gsi/drop} \\
    & Binding & \V{probe/ft/qwen35_9/binding/contour/drop} & \V{probe/ft/qwen35_9/binding/isobar/drop} & \V{probe/ft/qwen35_9/binding/scatter/drop} & \V{probe/ft/qwen35_9/binding/gsi/drop} \\
    \midrule
    GLM-4.6V-Flash & None & \V{probe/ft/glm46v_9/base/contour/drop} & \V{probe/ft/glm46v_9/base/isobar/drop} & \V{probe/ft/glm46v_9/base/scatter/drop} & \V{probe/ft/glm46v_9/base/gsi/drop} \\
    & Answer only & \V{probe/ft/glm46v_9/answer/contour/drop} & \V{probe/ft/glm46v_9/answer/isobar/drop} & \V{probe/ft/glm46v_9/answer/scatter/drop} & \V{probe/ft/glm46v_9/answer/gsi/drop} \\
    & Binding & \V{probe/ft/glm46v_9/binding/contour/drop} & \V{probe/ft/glm46v_9/binding/isobar/drop} & \V{probe/ft/glm46v_9/binding/scatter/drop} & \V{probe/ft/glm46v_9/binding/gsi/drop} \\
    \bottomrule
  \end{tabular}
  \caption{Drop in accuracy after LoRA fine-tuning on contour plots only.}
  \label{tab:teach}
\end{table}

Since no training-free remedy removed the drop, we asked whether training removes it and whether the removal carries over to figure types never seen in training.
We fine-tuned Qwen3.5 9B and GLM-4.6V-Flash with LoRA~\cite{hu2022lora} of rank 16 on the language model, for one epoch on 20{,}000 white-box contour items whose fields, query points and images are disjoint from every test item.
In the \emph{binding} fine-tune each target names the innermost labelled ring around A, the next ring outward and the ring that carries the near label before it compares the two values and answers; the \emph{answer-only} fine-tune uses the same items with the answer alone.
Both were evaluated on the held-out white-box pairs and on the isobar, scatter and GSI terrain sets of \cref{sec:nearest} without further training.

On contour plots the drop fell below half of its base value and was no longer significant in \V{probe/ft/nremoval/word} of the \V{probe/ft/nruns/word} fine-tunes of \cref{tab:teach}.
Control accuracy also rose, for GLM-4.6V-Flash with binding targets from \V{probe/ft/glm46v_9/base/contour/control} to \V{probe/ft/glm46v_9/binding/contour/control}; for Qwen3.5 9B with binding targets the drop fell from \V{probe/ft/qwen35_9/base/contour/drop} to \V{probe/ft/qwen35_9/binding/contour/drop} but stayed significant ($p={}$\V{probe/ft/qwen35_9/binding/contour/p}).
A significant reduction (Holm over sets) on at least two of the three unseen sets occurred in \V{probe/ft/ntransfer/word} of the \V{probe/ft/nruns/word} fine-tunes.
On scatter plots and GSI maps the drop stayed significant in every fine-tune; the only significant reduction on any unseen set was that of the answer-only fine-tune of GLM-4.6V-Flash on scatter plots with \V{probe/ft/glm46v_9/answer/scatter/smaller}:\V{probe/ft/glm46v_9/answer/scatter/larger} items whose control-minus-conflict difference fell and rose and \mbox{$p={}$\V{probe/ft/glm46v_9/answer/scatter/predp}}.
Binding targets did not transfer better than answers alone (items with the smaller control-minus-conflict difference under binding and under answer-only targets \V{probe/ft/qwen35_9/bva/binding}:\V{probe/ft/qwen35_9/bva/answer} for Qwen3.5 9B and \V{probe/ft/glm46v_9/bva/binding}:\V{probe/ft/glm46v_9/bva/answer} for GLM-4.6V-Flash), and the binding fine-tune of Qwen3.5 9B lowered its control accuracy on scatter plots by \V{probe/ft/qwen35_9/binding/scatter/controlchange} ($p={}$\V{probe/ft/qwen35_9/binding/scatter/controlp}).
On line charts and on the separation set, which were neither trained nor used in the evaluation above (post hoc), no fine-tune reduced the drop on line charts and \V{probe/ftmore/linechart/nlarger/word} of the \V{probe/ftmore/nruns/word} enlarged it, for GLM-4.6V-Flash with binding targets from \V{probe/ftmore/glm46v_9/binding/linechart/basedrop} to \V{probe/ftmore/glm46v_9/binding/linechart/drop}; on the separation set the drop shrank only because control accuracy fell, for Qwen3.5 9B with binding targets from \V{probe/ftmore/qwen35_9/binding/separation/basecontrol} to \V{probe/ftmore/qwen35_9/binding/separation/control}.
Training on one figure type thus removed the nearest-label answer there but did not teach a binding that carried over to the other figure types, so each figure type needs its own test.

\section{Where the nearest-label answer forms}
\label{sec:where}

For Gemma-4 31B, Qwen3.5 9B and Qwen3.8 27B, we fed each model its own recorded white-box reasoning (teacher forcing) and trained linear probes on the hidden states of several layers to decode the correct answer or the answer implied by the near label; a first version of this analysis failed a validity check and was replaced by this one.
Each probe is a ridge classifier trained on control and conflict items together and scored in ten folds on fields held out from training; the two answers agree on control items and are opposite on conflict items, so on conflict items a probe for the correct answer that had learned the near-label answer would fall below chance, and a probe at chance carries neither of the two answers.
On conflict items the correct answer was not decodable above chance at the answer position at any layer, whereas the nearest-label answer was decodable with accuracies of up to \V{probe/mech/vconfmaxlo} to \V{probe/mech/vconfmaxhi}.
By a threshold rule on decoding accuracy, the correct answer was \emph{never formed} in \V{probe/mech/nneverformed/word} model, and the other \V{probe/mech/nundecided/word} remained undecided within \V{probe/mech/undecidedgap} of the thresholds.
In \V{probe/mech/ncommit/word} models the answer given by the nearest label became more decodable over the first sentence of the reasoning that ties the near value to the query, so the binding is committed during reasoning; this locates the error in time but does not establish its cause.
In a post hoc probe of the image tokens before any text, the correct answer on conflict items was decodable at most at \V{probe/mech3/tmax}, about as well as on control items and already in the first layer, whereas the nearest-label answer, decoded over all items, control items included, reached up to \V{probe/mech3/gemma4_31/vA} in Gemma-4 31B and at most \V{probe/mech3/qwenvmax} in the two Qwen models.

\section{Nearest labels in natural contour maps}
\label{sec:prevalence}

We drew \V{final/prev/windows} random windows of 768 by 768 pixels over the land of Japan from the public elevation tiles of GSI at 5~m resolution, or at 10~m where those are missing, with contours every 10~m and labels on every 50~m line at the default placement of matplotlib~\cite{hunter2007matplotlib}.
For the \V{final/prev/own} of \V{final/prev/points} random points that lie inside a labelled ring, the nearest label was on a line other than the innermost labelled ring at \V{final/prev/share} \V{final/prev/share/ci} of the points, and to the next labelled line outward (the case of our probes) at \V{final/prev/neighbour} \V{final/prev/neighbour/ci}; this measures how often the nearest label is wrong, not how often a model errs on such maps.

To ask whether the placement changes a decision on real terrain, we drew \V{probe/route/pairs/word} pairs of GSI terrain windows drawn in the style of the GSI terrain maps of \cref{sec:nearest}, each carrying a synthetic trail, and asked whether this trail crosses the \emph{index line} (the thick, labelled contour line every 50~m) of a named elevation, which decides whether a walk passes that elevation.
Both the drawn lines and the elevation model along the trail gave the same answer; the two members of a pair differ only in whether the label near the trail is that of the crossed line or that of the adjacent index line.
Accuracy fell significantly in \V{probe/route/neffect/word} of the \V{probe/route/n/word} open models (Holm over models), by \V{probe/route/droplo} to \V{probe/route/drophi}, and for Gemma-4 31B from \V{probe/route/gemma4_31/control} to \V{probe/route/gemma4_31/conflict}.
LLaVA-OneVision-2 8B and Molmo2-O 7B, which answered yes to nearly every item, and Mistral-Small-3.2 24B showed no significant drop.

\section{Limitations}
\label{sec:discussion}

Open models reach only moderate control accuracy (\V{final/headline/control/min} to \V{final/headline/control/max}); remedies that raised it did not raise the conflict accuracy (\cref{sec:remedies}), but the results still describe models that solve the task in part.
The drop does not separate a model that sometimes takes the near value from one that always does; a post hoc conditional analysis shows a spread across models (\supp{Tab.~S4}).
On \V{final/nat/items} natural conflict points of the random maps and as many other points, the \V{final/nat/models} open models answered with accuracies from \V{final/nat/acc/min} to \V{final/nat/acc/max}, and only \V{final/nat/harmful} was significantly less accurate on the conflict points, so the cost of natural label placement is not measured; the route task measures the cost of a placed conflict on synthetic trails.

The matched version equalises the line gaps but not all local differences between label places, and the crop that removed the drop of GPT-6.1 Sol differs from the image upscaled by three not only in the lines far from A but also in how much of the field it shows.
The hidden-state analysis of three models uses linear probes and finds the sentence that ties the near value to the query by its wording, so it locates the point at which the nearest-label answer becomes decodable but not its cause.
The tool remedies depend on how often a model calls a tool, which was rare in some models, and we analysed the oracle calls post hoc, without a significance test.

The synthetic figures use mostly black lines, one font and few elements, and each item was read once with a single English chain-of-thought prompt.
The human check covers \V{probe/human/distinct} items answered by two authors, not a sample of map readers.

The fine-tunes cover two models of about 9B parameters with one rank and one training set, and the route task was not posed to the commercial models.
We tested open models of up to 38B parameters and four commercial models from two providers, the latter on the label-placement probes only and the two Claude models post hoc; agreement across models that read the same items does not constitute an independent replication.
The work uses no personal data; its main risk is overreliance on VLM map reading in civilian mountain navigation, which the probes are meant to expose.

\section{Conclusion}
\label{sec:conclusion}

Label-placement probes that keep each label on its own line and move only its distance to the query lowered accuracy in \V{final/dec/white/k} of \V{final/models/n} open VLMs and in the four commercial models, two of which were tested post hoc.
The drop survived correct transcription, a question that names the lines, magnification in the open models, instructions, procedures, worked examples and tools, among them an oracle that directly returns the correct value, and in three open models the answer decodable at the answer position is the answer given by the nearest label.
Fine-tuning removed it on contour plots but not on figure types never trained, and the same placement changed route decisions on terrain maps in \V{probe/route/neffect/word} of \V{probe/route/n/word} models.
Since natural contour maps put the nearest label on another line at most points inside a labelled ring, attaching each label to its own line is a reading skill that benchmarks should test directly.
We will release the generators, the questions and answers and the scoring code under CC BY 4.0 for data and MIT for code, with the GSI-derived images under the terms of GSI, and with Croissant metadata and a dataset card.

\section*{LLM Usage Statement}
LLM-based assistants were used in this work.
Claude was used to write and revise the code that renders the maps and charts, computes their ground truth, queries the open and GPT models and computes all statistics.
Claude was also used for surveys of related work, for drafting and editing the text and for simulated reviews of drafts.
ChatGPT was used for ideation, surveys of related work and simulated reviews of drafts.
The released code computed every reported result from the replies of the open VLMs on local servers and of the four commercial models queried through their API; the Claude replies were collected by a co-author with a separate script and imported and scored by the released code.
No LLM other than the evaluated models produced, edited or selected any reading, score or reported number.
The authors have checked all code, analyses and text.

\clearpage
\setcounter{section}{0}\setcounter{table}{0}\setcounter{figure}{0}\setcounter{footnote}{0}
\renewcommand{\theHsection}{S\arabic{section}}\renewcommand{\theHtable}{S\arabic{table}}\renewcommand{\theHfigure}{S\arabic{figure}}
\onecolumn
\begin{center}{\Large\bfseries TerraArt Supplementary Material}\end{center}\vspace{1em}
\renewcommand{\TODO}[1]{\textcolor{orange}{[TODO: #1]}}
\providecommand{\VT}{}
\renewcommand{\VT}[1]{\ifcsname tav@#1\endcsname\V{#1}\else\textcolor{orange}{TODO}\fi}
\renewcommand{\thesection}{S\arabic{section}}
\renewcommand{\thetable}{S\arabic{table}}
\renewcommand{\thefigure}{S\arabic{figure}}
\providecommand{\q}{}
\renewcommand{\q}[1]{``\textit{#1}''}

\small

This supplement gives the models, replies and supplementary contrasts of the final run (\cref{S-sec:final}), the sizes of the sets (\cref{S-sec:sets}), the questions (\cref{S-sec:prompts}), the generation of stimuli and the removal of shortcuts (\cref{S-sec:gen}), the data sources and licences (\cref{S-sec:data}), and the other figure types, map pairs and supplementary contrasts summarised in the main paper (\cref{S-sec:moved}).
Numbers with the prefix S refer to this supplement; all others refer to the main paper.
Every number is generated from the stored answers and the answer keys of the released sets and of the released probe records.
No 6.1 release of GPT-6 Luna was available to us, so the two GPT models differ in version.
Before the analysed GPT-6.1 Sol run, 101 replies of a stopped run and 20 pilot replies of GPT-6 Sol were set aside unread and unscored.
A co-author collected the Claude replies through the Anthropic API, with at most 16,000 output tokens for Fable 5.2 and 4,000 for Opus 5.5.

\section{Final run}
\label{S-sec:final}

The \V{final/models/n} models of the final run, their pinned revisions (\cref{S-tab:final-models}), the replies without an answer (\cref{S-tab:final-parse}), the supplementary contrasts (\cref{S-tab:final-supp}) and the post hoc conditional accuracy (\cref{S-tab:final-conditional}) are given here.
The models were served with vLLM \V{final/package/vllm}.
\Cref{S-tab:final-parse} gives per model the answers, failed requests, replies without an answer and replies cut at the limit of 4{,}000 output tokens.
\Cref{S-tab:final-supp} gives the conflict-minus-control $\Delta d'$ of each supplementary contrast; a drop is counted in the main text when the contrast is decidable, $n_{10}>n_{01}$ and the uncorrected sign test gives $p<0.05$.
The text versions reached an accuracy of \V{final/supp/separation_text/acc/min} to \V{final/supp/separation_text/acc/max} on the separation texts, \V{final/supp/isobar_text/acc/min} to \V{final/supp/isobar_text/acc/max} on isobar texts, \V{final/supp/scatter_text/acc/min} to \V{final/supp/scatter_text/acc/max} on scatter texts across all models and conditions, and \V{final/supp/linechart_text/acc/max} on line-chart texts in all models and conditions.
For the white-box probe, \cref{S-tab:final-conditional} gives the accuracy per condition, the conflict replies without an answer (counted as wrong), and conflict accuracy over the items whose control item of the same pair and member was answered correctly.

\begin{table}[h]
  \centering
  \small
  \begin{tabular}{@{}lll@{}}
\toprule
Model & Hugging Face name & Revision \\
\midrule
Gemma-4 31B & \texttt{\V{final/gemma4_31/repo}} & \texttt{\V{final/gemma4_31/revision}} \\
Gemma-4 12B & \texttt{\V{final/gemma4_12/repo}} & \texttt{\V{final/gemma4_12/revision}} \\
Qwen3.8 27B & \texttt{\V{final/qwen38_27/repo}} & \texttt{\V{final/qwen38_27/revision}} \\
Qwen3.5 9B & \texttt{\V{final/qwen35_9/repo}} & \texttt{\V{final/qwen35_9/revision}} \\
Qwen3.5 4B & \texttt{\V{final/qwen35_4/repo}} & \texttt{\V{final/qwen35_4/revision}} \\
InternVL3.5 38B & \texttt{\V{final/internvl35_38/repo}} & \texttt{\V{final/internvl35_38/revision}} \\
Mistral-Small-3.2 24B & \texttt{\V{final/mistral32_24/repo}} & \texttt{\V{final/mistral32_24/revision}} \\
GLM-4.6V-Flash & \texttt{\V{final/glm46v_9/repo}} & \texttt{\V{final/glm46v_9/revision}} \\
LLaVA-OneVision-2 8B & \texttt{\V{final/llavaov2_8/repo}} & \texttt{\V{final/llavaov2_8/revision}} \\
Molmo2-O 7B & \texttt{\V{final/molmo2o_7/repo}} & \texttt{\V{final/molmo2o_7/revision}} \\
\bottomrule
\end{tabular}
  \caption{Models of the final run.}
  \label{S-tab:final-models}
\end{table}

\begin{table}[h]
  \centering
  \small
  \begin{tabular}{@{}lcccc@{}}
\toprule
Model & Answers & Failed & Without an answer & Cut at the token limit \\
\midrule
Gemma-4 31B & \V{final/gemma4_31/answers} & \V{final/gemma4_31/failed} & \V{final/gemma4_31/unparsed} & \V{final/gemma4_31/truncated} \\
Gemma-4 12B & \V{final/gemma4_12/answers} & \V{final/gemma4_12/failed} & \V{final/gemma4_12/unparsed} & \V{final/gemma4_12/truncated} \\
Qwen3.8 27B & \V{final/qwen38_27/answers} & \V{final/qwen38_27/failed} & \V{final/qwen38_27/unparsed} & \V{final/qwen38_27/truncated} \\
Qwen3.5 9B & \V{final/qwen35_9/answers} & \V{final/qwen35_9/failed} & \V{final/qwen35_9/unparsed} & \V{final/qwen35_9/truncated} \\
Qwen3.5 4B & \V{final/qwen35_4/answers} & \V{final/qwen35_4/failed} & \V{final/qwen35_4/unparsed} & \V{final/qwen35_4/truncated} \\
InternVL3.5 38B & \V{final/internvl35_38/answers} & \V{final/internvl35_38/failed} & \V{final/internvl35_38/unparsed} & \V{final/internvl35_38/truncated} \\
Mistral-Small-3.2 24B & \V{final/mistral32_24/answers} & \V{final/mistral32_24/failed} & \V{final/mistral32_24/unparsed} & \V{final/mistral32_24/truncated} \\
GLM-4.6V-Flash & \V{final/glm46v_9/answers} & \V{final/glm46v_9/failed} & \V{final/glm46v_9/unparsed} & \V{final/glm46v_9/truncated} \\
LLaVA-OneVision-2 8B & \V{final/llavaov2_8/answers} & \V{final/llavaov2_8/failed} & \V{final/llavaov2_8/unparsed} & \V{final/llavaov2_8/truncated} \\
Molmo2-O 7B & \V{final/molmo2o_7/answers} & \V{final/molmo2o_7/failed} & \V{final/molmo2o_7/unparsed} & \V{final/molmo2o_7/truncated} \\
\bottomrule
\end{tabular}
  \caption{Replies per model.}
  \label{S-tab:final-parse}
\end{table}

\begin{table}[h]
  \centering
  \small
  \begin{tabular}{@{}lccccc@{}}
\toprule
Contrast & Gemma-4 31B & Gemma-4 12B & Qwen3.8 27B & Qwen3.5 9B & Qwen3.5 4B \\
\midrule
GSI, next line but one & \V{final/gemma4_31/gsi_placement_k1/diff} & \V{final/gemma4_12/gsi_placement_k1/diff} & \V{final/qwen38_27/gsi_placement_k1/diff} & \V{final/qwen35_9/gsi_placement_k1/diff} & \V{final/qwen35_4/gsi_placement_k1/diff} \\
Separation, text & \V{final/gemma4_31/separation_text/diff} & \V{final/gemma4_12/separation_text/diff} & \V{final/qwen38_27/separation_text/diff} & \V{final/qwen35_9/separation_text/diff} & \V{final/qwen35_4/separation_text/diff} \\
Isobar, text & \V{final/gemma4_31/isobar_text/diff} & \V{final/gemma4_12/isobar_text/diff} & \V{final/qwen38_27/isobar_text/diff} & \V{final/qwen35_9/isobar_text/diff} & \V{final/qwen35_4/isobar_text/diff} \\
Scatter, text & \V{final/gemma4_31/scatter_text/diff} & \V{final/gemma4_12/scatter_text/diff} & \V{final/qwen38_27/scatter_text/diff} & \V{final/qwen35_9/scatter_text/diff} & \V{final/qwen35_4/scatter_text/diff} \\
Line chart, black & \V{final/gemma4_31/linechart/diff} & \V{final/gemma4_12/linechart/diff} & \V{final/qwen38_27/linechart/diff} & \V{final/qwen35_9/linechart/diff} & \V{final/qwen35_4/linechart/diff} \\
Line chart, colour & \V{final/gemma4_31/linechart_colour/diff} & \V{final/gemma4_12/linechart_colour/diff} & \V{final/qwen38_27/linechart_colour/diff} & \V{final/qwen35_9/linechart_colour/diff} & \V{final/qwen35_4/linechart_colour/diff} \\
Line chart, text & \V{final/gemma4_31/linechart_text/diff} & \V{final/gemma4_12/linechart_text/diff} & \V{final/qwen38_27/linechart_text/diff} & \V{final/qwen35_9/linechart_text/diff} & \V{final/qwen35_4/linechart_text/diff} \\
Scatter, leader line & \V{final/gemma4_31/cues_leader/diff} & \V{final/gemma4_12/cues_leader/diff} & \V{final/qwen38_27/cues_leader/diff} & \V{final/qwen35_9/cues_leader/diff} & \V{final/qwen35_4/cues_leader/diff} \\
Scatter, common region & \V{final/gemma4_31/cues_region/diff} & \V{final/gemma4_12/cues_region/diff} & \V{final/qwen38_27/cues_region/diff} & \V{final/qwen35_9/cues_region/diff} & \V{final/qwen35_4/cues_region/diff} \\
Scatter, code label & \V{final/gemma4_31/pointer_scatter/diff} & \V{final/gemma4_12/pointer_scatter/diff} & \V{final/qwen38_27/pointer_scatter/diff} & \V{final/qwen35_9/pointer_scatter/diff} & \V{final/qwen35_4/pointer_scatter/diff} \\
Line chart, code label & \V{final/gemma4_31/pointer_linechart/diff} & \V{final/gemma4_12/pointer_linechart/diff} & \V{final/qwen38_27/pointer_linechart/diff} & \V{final/qwen35_9/pointer_linechart/diff} & \V{final/qwen35_4/pointer_linechart/diff} \\
NCEP/NCAR isobars & \V{final/gemma4_31/public_data/diff} & \V{final/gemma4_12/public_data/diff} & \V{final/qwen38_27/public_data/diff} & \V{final/qwen35_9/public_data/diff} & \V{final/qwen35_4/public_data/diff} \\
\midrule
Contrast & InternVL3.5 38B & Mistral-Small-3.2 24B & GLM-4.6V-Flash & LLaVA-OneVision-2 8B & Molmo2-O 7B \\
\midrule
GSI, next line but one & \V{final/internvl35_38/gsi_placement_k1/diff} & \V{final/mistral32_24/gsi_placement_k1/diff} & \V{final/glm46v_9/gsi_placement_k1/diff} & \V{final/llavaov2_8/gsi_placement_k1/diff} & \V{final/molmo2o_7/gsi_placement_k1/diff} \\
Separation, text & \V{final/internvl35_38/separation_text/diff} & \V{final/mistral32_24/separation_text/diff} & \V{final/glm46v_9/separation_text/diff} & \V{final/llavaov2_8/separation_text/diff} & \V{final/molmo2o_7/separation_text/diff} \\
Isobar, text & \V{final/internvl35_38/isobar_text/diff} & \V{final/mistral32_24/isobar_text/diff} & \V{final/glm46v_9/isobar_text/diff} & \V{final/llavaov2_8/isobar_text/diff} & \V{final/molmo2o_7/isobar_text/diff} \\
Scatter, text & \V{final/internvl35_38/scatter_text/diff} & \V{final/mistral32_24/scatter_text/diff} & \V{final/glm46v_9/scatter_text/diff} & \V{final/llavaov2_8/scatter_text/diff} & \V{final/molmo2o_7/scatter_text/diff} \\
Line chart, black & \V{final/internvl35_38/linechart/diff} & \V{final/mistral32_24/linechart/diff} & \V{final/glm46v_9/linechart/diff} & \V{final/llavaov2_8/linechart/diff} & \V{final/molmo2o_7/linechart/diff} \\
Line chart, colour & \V{final/internvl35_38/linechart_colour/diff} & \V{final/mistral32_24/linechart_colour/diff} & \V{final/glm46v_9/linechart_colour/diff} & \V{final/llavaov2_8/linechart_colour/diff} & \V{final/molmo2o_7/linechart_colour/diff} \\
Line chart, text & \V{final/internvl35_38/linechart_text/diff} & \V{final/mistral32_24/linechart_text/diff} & \V{final/glm46v_9/linechart_text/diff} & \V{final/llavaov2_8/linechart_text/diff} & \V{final/molmo2o_7/linechart_text/diff} \\
Scatter, leader line & \V{final/internvl35_38/cues_leader/diff} & \V{final/mistral32_24/cues_leader/diff} & \V{final/glm46v_9/cues_leader/diff} & \V{final/llavaov2_8/cues_leader/diff} & \V{final/molmo2o_7/cues_leader/diff} \\
Scatter, common region & \V{final/internvl35_38/cues_region/diff} & \V{final/mistral32_24/cues_region/diff} & \V{final/glm46v_9/cues_region/diff} & \V{final/llavaov2_8/cues_region/diff} & \V{final/molmo2o_7/cues_region/diff} \\
Scatter, code label & \V{final/internvl35_38/pointer_scatter/diff} & \V{final/mistral32_24/pointer_scatter/diff} & \V{final/glm46v_9/pointer_scatter/diff} & \V{final/llavaov2_8/pointer_scatter/diff} & \V{final/molmo2o_7/pointer_scatter/diff} \\
Line chart, code label & \V{final/internvl35_38/pointer_linechart/diff} & \V{final/mistral32_24/pointer_linechart/diff} & \V{final/glm46v_9/pointer_linechart/diff} & \V{final/llavaov2_8/pointer_linechart/diff} & \V{final/molmo2o_7/pointer_linechart/diff} \\
NCEP/NCAR isobars & \V{final/internvl35_38/public_data/diff} & \V{final/mistral32_24/public_data/diff} & \V{final/glm46v_9/public_data/diff} & \V{final/llavaov2_8/public_data/diff} & \V{final/molmo2o_7/public_data/diff} \\
\bottomrule
\end{tabular}
  \caption{Supplementary contrasts: $\Delta d'$ per model.}
  \label{S-tab:final-supp}
\end{table}

\begin{table}[h]
  \centering
  \small
  \begin{tabular}{@{}lcccc@{}}
\toprule
Model & Acc.\ control & Acc.\ conflict & Conflict without an answer & Conflict acc.\ given control correct \\
\midrule
Gemma-4 31B & \V{final/gemma4_31/white/control/acc} & \V{final/gemma4_31/white/conflict/acc} & \V{final/gemma4_31/white/conflict/unparsed} & \V{final/gemma4_31/white/conditional} \\
Gemma-4 12B & \V{final/gemma4_12/white/control/acc} & \V{final/gemma4_12/white/conflict/acc} & \V{final/gemma4_12/white/conflict/unparsed} & \V{final/gemma4_12/white/conditional} \\
Qwen3.8 27B & \V{final/qwen38_27/white/control/acc} & \V{final/qwen38_27/white/conflict/acc} & \V{final/qwen38_27/white/conflict/unparsed} & \V{final/qwen38_27/white/conditional} \\
Qwen3.5 9B & \V{final/qwen35_9/white/control/acc} & \V{final/qwen35_9/white/conflict/acc} & \V{final/qwen35_9/white/conflict/unparsed} & \V{final/qwen35_9/white/conditional} \\
Qwen3.5 4B & \V{final/qwen35_4/white/control/acc} & \V{final/qwen35_4/white/conflict/acc} & \V{final/qwen35_4/white/conflict/unparsed} & \V{final/qwen35_4/white/conditional} \\
InternVL3.5 38B & \V{final/internvl35_38/white/control/acc} & \V{final/internvl35_38/white/conflict/acc} & \V{final/internvl35_38/white/conflict/unparsed} & \V{final/internvl35_38/white/conditional} \\
Mistral-Small-3.2 24B & \V{final/mistral32_24/white/control/acc} & \V{final/mistral32_24/white/conflict/acc} & \V{final/mistral32_24/white/conflict/unparsed} & \V{final/mistral32_24/white/conditional} \\
GLM-4.6V-Flash & \V{final/glm46v_9/white/control/acc} & \V{final/glm46v_9/white/conflict/acc} & \V{final/glm46v_9/white/conflict/unparsed} & \V{final/glm46v_9/white/conditional} \\
LLaVA-OneVision-2 8B & \V{final/llavaov2_8/white/control/acc} & \V{final/llavaov2_8/white/conflict/acc} & \V{final/llavaov2_8/white/conflict/unparsed} & \V{final/llavaov2_8/white/conditional} \\
Molmo2-O 7B & \V{final/molmo2o_7/white/control/acc} & \V{final/molmo2o_7/white/conflict/acc} & \V{final/molmo2o_7/white/conflict/unparsed} & \V{final/molmo2o_7/white/conditional} \\
\bottomrule
\end{tabular}
  \caption{White-box probe: conditional accuracy (post hoc).}
  \label{S-tab:final-conditional}
\end{table}

\section{Sets}
\label{S-sec:sets}

\begin{table}[h]
  \centering
  \small
  \begin{tabular}{@{}lr@{}}
    \toprule
    Set & Questions \\
    \midrule
    Map pairs (four conditions) & \V{final/set/map_pairs/q} \\
    Label placement (four versions) & \V{final/set/label_placement_boxes/q} \\
    GSI placement / readability check & \V{final/set/gsi_placement/q} / \V{final/set/gsi_placement_development/q} \\
    Separation, picture / text & \V{final/set/separation_shortcut_free/q} / \V{final/set/separation_shortcut_free_text/q} \\
    Isobar charts, picture / text & \V{final/set/transfer_isobar/q} / \V{final/set/transfer_isobar_text/q} \\
    Line charts (black and colour) / text & \V{final/set/transfer_linechart/q} / \V{final/set/transfer_linechart_text/q} \\
    Scatter plots, picture / text & \V{final/set/transfer_scatter/q} / \V{final/set/transfer_scatter_text/q} \\
    Pointer controls, scatter / line charts & \V{final/set/pointer_control_scatter/q} / \V{final/set/pointer_control_linechart/q} \\
    NCEP/NCAR isobar charts & \V{final/set/public_data/q} \\
    Grouping cues & \V{final/set/grouping_cues/q} \\
    All sets & \V{final/answers/per_model} \\
    \bottomrule
  \end{tabular}
  \caption{Questions per set.}
  \label{S-tab:sets}
\end{table}

\Cref{S-tab:sets} lists the questions per set of the final run, asked of every model; each pair enters a condition with both members, so a condition holds twice as many items as pairs.

\section{Questions}
\label{S-sec:prompts}

Every question ends with \q{Think step by step, then end with `Answer: \textless X or Y\textgreater'.}, where X and Y are the two options, except the direct version, ending with \q{Reply only with `Answer: \textless X or Y\textgreater'.}, and the questions answered in JSON, which end with \q{Think step by step, then end with the JSON object.}
The texts below are verbatim, with placeholders or one example of the numbers.
The worked examples come from the released development split of the GSI placement set; an earlier run with three newly generated examples, replaced so that every example comes from released data, reduced the drop in LLaVA-OneVision-2 8B and left it in the other six models.

Map pairs: \q{This is a contour map (north is up): brown contour lines every 10 m, thicker index lines every 50 m; the scale bar is below the map. The index lines are labelled with their elevation in metres. Is the ground at A (the centre of the small red ring) higher or lower than most of the ground around it, within about 80 m of A? Answer with one word: higher or lower.}
The text conditions insert one sentence before the question: values only, \q{The two elevation values printed nearest to A are $v_1$ m (label centre about $d_1$ m from A) and $v_2$ m (label centre about $d_2$ m from A).}; structure only, \q{A lies inside a closed contour line about $n$ m from A at its nearest point; the next contour line outward encloses it at about $m$ m from A at its nearest point.}; values + structure, \q{The innermost closed contour line around A, about $d$ m from A at its nearest point, is at $z_0$ m; the next contour line outside it is at $z_1$ m.}

GSI placement: \q{This is a contour map (north is up): brown contour lines every 10 m, thicker index lines every 50 m; the scale bar is below the map. Only two index lines are labelled with their elevation in metres. Is the ground at A (the centre of the small red ring) higher or lower than most of the ground around it, within about 80 m of A? Answer with one word: higher or lower.}

Label placement: \q{This is a contour plot of a scalar field f(x, y): black contour lines at equal steps of 2 in f. Only two contour lines are labelled with their value of f. Is f at A (the centre of the small red ring) higher or lower than most of the values of f within a distance of 0.14 of A (in the units of the axes)? Answer with one word: higher or lower.}
Separation: \q{This is a contour plot of a scalar field f(x, y): black contour lines at equal steps of 10 in f. Only two contour lines are labelled with their value of f. Is f at A (the centre of the small red ring) higher than 165? Answer with one word: yes or no.}, with questions about higher and lower thresholds balanced.

Isobar charts (synthetic and NCEP/NCAR): \q{This is a weather chart: black lines are isobars (lines of equal sea-level pressure) drawn every 4 hPa. Only two isobars are labelled with their pressure in hPa. Is A (the centre of the small red ring) inside a high-pressure system or a low-pressure system? Answer with one word: high or low.}
Line charts: \q{This line chart shows 3 series; each series is labelled once with its name printed next to its line. At x = 7 (dashed vertical line), which is higher: Birch or Maple? Answer with one word: Birch or Maple.}
Scatter plots with group labels: \q{This scatter plot shows 3 groups of points; each group is named once by a label printed next to it. Which group does the point A (inside the small red ring) belong to: Hazel or Rowan? Answer with one word: Hazel or Rowan.}

Text versions state the labels and the elements that they belong to, for example for the separation set \q{A scalar field f(x, y) has contour lines at equal steps of 10. Point A lies directly inside a closed contour line of value 180 (no other contour line lies between A and that line). Another closed contour line, of value 150, lies nearby but does not enclose A. Is f at A higher than 165? Answer with one word: yes or no.}; scatter text versions list the coordinates of A and of the points of the two queried groups, and line-chart text versions give the values of the two queried series at the queried $x$.
Prompts of the crop, value-choice, named and value-first versions and of the remedies are verbatim in the released sets and code.

\section{Stimulus generation and shortcut removal}
\label{S-sec:gen}

\paragraph{Map pairs.}
Areas of the GSI 5\,m elevation model are drawn at 768 pixels with 10\,m contours and 50\,m index lines; index-line labels are placed by matplotlib's contour labelling, kept along their line and turned to read left to right, and reused with the same position and angle on the $C-f$ member, with both values padded to the same width.
No depression ticks are drawn.

\paragraph{Contour-plot probes.}
Fields are smooth sums of Gaussian bumps with smoothed noise.
Label placement keeps fields in which both placements are feasible on the same rings; near distances are matched pair by pair, and label positions are identical in both members.
In the separation set A lies inside its enclosing line, and the threshold lies midway between the two printed values, so that only the label read from the enclosing line decides the answer.

\paragraph{Transfer sets.}
Isobar fields are shifted per pair so that the two printed values are $T\pm2$\,hPa and the partner member draws the field $2T-p$, where $p$ is the pressure field; both members print the same two numbers with their places exchanged, so any rule on the values alone is at chance.
Every other line or group is at least three times as far from a label centre as the label's own line or group.
Near distances are spread over three ranges, and whether the true label is above the competing one alternates over pairs.
Line-chart partners exchange the two queried names on the same pixels; scatter partners exchange the two group names.
For the leader-line cue, each label is moved straight away from its group until its leader line is long enough to be seen; pairs where a moved label would touch a line or change the nearest label were dropped before any answer.
Shaded boxes are light blue-grey rectangles that enclose a queried label and the nearby part of its own group, but neither another group nor the query point.
Every generator asserts per item the true answer, the answer when each label is read from its own element, and the answer when the nearest label is used.

\paragraph{Public data.}
NCEP/NCAR charts show a 40$^\circ\times$40$^\circ$ window with black isobars every 4\,hPa and no coastlines; one member of each pair shows the real values and the other draws $2T-p$ as in the synthetic charts.

\section{Data sources and licences}
\label{S-sec:data}

Elevation: GSI digital elevation model tiles (DEM5A, else DEM10B), used with approval for this project; images derived from them will be released under the GSI terms, and the generators reproduce them from the public tiles.
Sea-level pressure: NCEP/NCAR Reanalysis 1 daily means of 2024, NOAA Physical Sciences Laboratory, public.
Acknowledgement requested by the provider: NCEP Reanalysis data provided by the NOAA PSL, Boulder, Colorado, USA.
Human check: both annotators are authors of this paper and took part by their own consent; no other person took part and no personal data were collected, so no review by an institutional review board was sought.
All other figures are synthetic.
The generators, questions, stored answers and scoring code will be released under CC BY 4.0 (data) and MIT (code).

\section{Other figure types, map pairs and supplementary contrasts}
\label{S-sec:moved}

This section gives the sets and results that the main paper summarises in one sentence each; references to sections and tables without the prefix S point to the main paper.

\subsection{Sets}
The \emph{GSI placement set} applies the same manipulation to labels on index lines (the thick contour lines every 50\,m that carry values) of real GSI terrain: the near label lies on the next index line (\V{final/gemma4_31/gsi_placement/pairs} pairs) or on the line after it (\V{final/gemma4_31/gsi_placement_k1/pairs} pairs).
Further pairs from separate areas test whether a model can read such maps at all; a model passes this \emph{GSI readability check} when its $d'$ on their control items, where the true label lies nearest A, is at least 0.5.

The \emph{separation set} separates the proximity of a label from enclosure.
It asks whether $f$ at A is above a stated threshold $V$, and the competing label lies on a closed curve that does not enclose A, either farther from A than the true label (control; median \V{final/geom/separation/control/true/median} pixels for the true and \V{final/geom/separation/control/competing/median} pixels for the competing label) or nearer to A (conflict; \V{final/geom/separation/conflict/true/median} and \V{final/geom/separation/conflict/competing/median} pixels), with \V{final/gemma4_31/separation/pairs} pairs.
As in the probe, the two label places are exchanged, so in the conflict condition the true label also moves away from A.
The threshold $V$ lies midway between both printed values, so the two values are equally far from $V$ and the answer depends only on which value belongs to the curve around A.

In the conflict version of each transfer item, the true label lies far from the query and a competing label lies near it, in one of three distance ranges balanced over pairs (\V{final/geom/transfer_linechart/bands} pixels for line charts and scatter plots, \V{final/geom/transfer_isobar/bands} pixels for isobar charts); the control version exchanges the two label places.
\emph{Isobar charts} show a synthetic pressure field with two labelled closed isobars around A and ask whether A lies in a high-pressure or in a low-pressure system.
\emph{Line charts} show three or four black series, each named once next to its own line, and ask which of two series is higher at an $x$ marked by a dashed line; a colour version draws each series together with its name in one colour.
\emph{Scatter plots with group labels} show three or four groups of black dots, each group named once near its dots, and ask which group a ringed point belongs to.
On line charts and scatter plots the competing name is one of the two answer options, so a drop there may also come from the salience of this name.
We removed shortcuts while designing the generators and froze the generators before the reported sets were answered: both members of an isobar pair print the same two numbers, line charts and scatter plots balance the vertical order of the labels, and every other line or group lies at least three times as far from a label as the element that the label names.
\emph{Text versions} replace the picture by a text and state the labels and the elements they belong to but not their positions, so their conflict and control texts are identical; we use them only as a reference for the accuracy that can be reached without reading label positions.
A pointer control replaces the competing label by a three-letter code that is not an answer option (\V{final/gemma4_31/pointer_scatter/pairs} pairs per chart type), so that a drop there cannot come from the salience of an option name.

We also redrew public data with the same manipulation: isobar charts of sea-level pressure from the NCEP/NCAR Reanalysis for 2024 (\V{final/gemma4_31/public_data/pairs} pairs).
The grouping-cue sets add to the scatter plots either a leader line from each queried label to its own group or a \emph{shaded box} (a light rectangle around each queried label and the nearby part of its group); every queried label receives the cue.

Each of the \V{final/map/pairs} map pairs shows a square area of the 5\,m elevation model of the Geospatial Information Authority of Japan (GSI), which we drew with matplotlib at 768 pixels in the style of GSI topographic maps.
Thin contour lines are drawn at every 10\,m and a thick \emph{index line} every 50\,m, and only index lines carry printed values.
The second member draws $C$ minus the elevation, where $C$ is a multiple of 50\,m, and it reuses the label positions and angles.
Labels follow their line and read left to right, so the two members differ only in the digits and in sub-pixel anti-aliasing, and no depression ticks (short marks inside closed lines around hollows) are drawn.
The point A is marked by a small red ring, and the question asks whether the ground at A is higher or lower than most of the ground within about 80\,m; in the summit member of a pair every sample around A lies at least one contour interval (10\,m) below A, whereas in the other member A is a pit.
To answer the question, a reader has to find the values of A's own ring and the neighbouring ring and compare them.
Text conditions keep the picture and add one sentence: \emph{values only} gives the two printed values nearest A with their distances but does not say which line encloses A; \emph{structure only} gives the distances of A's own ring and the neighbouring ring without their values; \emph{values + structure} states the value of each of these two lines.

\subsection{Decisions across models}
The tests form five families, each corrected by the Holm method over its models and tests: the white-box contrast, the matched, crop and direct versions, and the four contrasts on the other figure types, and each family controls the familywise error rate at its own level.
Across models, we summarise four outcomes: the reversal in two thirds of the models or more, the matched contrast in at least two thirds of the models passing the primary test, each figure type in at least half of the tested models (decidable and passing the primary test), and at least two models passing the GSI readability check.
All other contrasts are \emph{supplementary} and use uncorrected $p$ and a predicted direction; we count a supplementary drop if the contrast is decidable, $n_{10}>n_{01}$ and $p<0.05$, a counting rule chosen after the answers were seen.
The GSI readability check was passed by \V{final/dec/gate/k} of the \V{final/models/n} models.
Three outcomes (\cref{S-tab:outcomes}) held; the reversal over all models as fixed beforehand was not met (\V{final/dec/reversal/k}/\V{final/dec/reversal/n}), and it held only when counted over the decidable models, a choice made after the results.
\begin{table}[h]
  \centering
  \scriptsize
  \setlength{\tabcolsep}{2pt}
  \begin{tabular}{@{}llcc@{}}
    \toprule
    Decision & Threshold & Result & Met \\
    \midrule
    Reversal, decidable models & $\ge$2/3 & \V{final/dec/reversal/k}/\V{final/dec/reversal/decidable} & \V{final/rule/reversal} \\
    Reversal, all models & $\ge$2/3 & \V{final/dec/reversal/k}/\V{final/dec/reversal/n} & \V{final/rule/reversal/all} \\
    Matched version & $\ge$2/3 of tested & \V{final/dec/matched/k}/\V{final/dec/matched/tested} & \V{final/rule/matched} \\
    Separation & $\ge$1/2 of tested & \V{final/dec/separation/k}/\V{final/dec/separation/tested} & \V{final/rule/separation} \\
    Isobar chart & $\ge$1/2 of tested & \V{final/dec/isobar/k}/\V{final/dec/isobar/tested} & \V{final/rule/isobar} \\
    Scatter plot & $\ge$1/2 of tested & \V{final/dec/scatter/k}/\V{final/dec/scatter/tested} & \V{final/rule/scatter} \\
    GSI placement & $\ge$1/2 of tested & \V{final/dec/gsi_placement/k}/\V{final/dec/gsi_placement/tested} & \V{final/rule/gsi_placement} \\
    GSI readability check & $\ge$2 models & \V{final/dec/gate/k} & \V{final/rule/gate} \\
    \bottomrule
  \end{tabular}
  \caption{Decisions across models and their outcomes.}
  \label{S-tab:outcomes}
\end{table}
\begin{table}[h]
  \centering
  \scriptsize
  \setlength{\tabcolsep}{2pt}
  \begin{tabular}{@{}lcccc@{}}
\toprule
Model & Picture & Values & Structure & Both \\
\midrule
Gemma-4 31B & \V{final/gemma4_31/map/index_labels/acc} (\V{final/gemma4_31/map/index_labels/d}) & \V{final/gemma4_31/map/values_only/acc} (\V{final/gemma4_31/map/values_only/d}) & \V{final/gemma4_31/map/structure_only/acc} (\V{final/gemma4_31/map/structure_only/d}) & \V{final/gemma4_31/map/values_structure/acc} (\V{final/gemma4_31/map/values_structure/d}) \\
Gemma-4 12B & \V{final/gemma4_12/map/index_labels/acc} (\V{final/gemma4_12/map/index_labels/d}) & \V{final/gemma4_12/map/values_only/acc} (\V{final/gemma4_12/map/values_only/d}) & \V{final/gemma4_12/map/structure_only/acc} (\V{final/gemma4_12/map/structure_only/d}) & \V{final/gemma4_12/map/values_structure/acc} (\V{final/gemma4_12/map/values_structure/d}) \\
Qwen3.8 27B & \V{final/qwen38_27/map/index_labels/acc} (\V{final/qwen38_27/map/index_labels/d}) & \V{final/qwen38_27/map/values_only/acc} (\V{final/qwen38_27/map/values_only/d}) & \V{final/qwen38_27/map/structure_only/acc} (\V{final/qwen38_27/map/structure_only/d}) & \V{final/qwen38_27/map/values_structure/acc} (\V{final/qwen38_27/map/values_structure/d}) \\
Qwen3.5 9B & \V{final/qwen35_9/map/index_labels/acc} (\V{final/qwen35_9/map/index_labels/d}) & \V{final/qwen35_9/map/values_only/acc} (\V{final/qwen35_9/map/values_only/d}) & \V{final/qwen35_9/map/structure_only/acc} (\V{final/qwen35_9/map/structure_only/d}) & \V{final/qwen35_9/map/values_structure/acc} (\V{final/qwen35_9/map/values_structure/d}) \\
Qwen3.5 4B & \V{final/qwen35_4/map/index_labels/acc} (\V{final/qwen35_4/map/index_labels/d}) & \V{final/qwen35_4/map/values_only/acc} (\V{final/qwen35_4/map/values_only/d}) & \V{final/qwen35_4/map/structure_only/acc} (\V{final/qwen35_4/map/structure_only/d}) & \V{final/qwen35_4/map/values_structure/acc} (\V{final/qwen35_4/map/values_structure/d}) \\
InternVL3.5 38B & \V{final/internvl35_38/map/index_labels/acc} (\V{final/internvl35_38/map/index_labels/d}) & \V{final/internvl35_38/map/values_only/acc} (\V{final/internvl35_38/map/values_only/d}) & \V{final/internvl35_38/map/structure_only/acc} (\V{final/internvl35_38/map/structure_only/d}) & \V{final/internvl35_38/map/values_structure/acc} (\V{final/internvl35_38/map/values_structure/d}) \\
Mistral-Small-3.2 24B & \V{final/mistral32_24/map/index_labels/acc} (\V{final/mistral32_24/map/index_labels/d}) & \V{final/mistral32_24/map/values_only/acc} (\V{final/mistral32_24/map/values_only/d}) & \V{final/mistral32_24/map/structure_only/acc} (\V{final/mistral32_24/map/structure_only/d}) & \V{final/mistral32_24/map/values_structure/acc} (\V{final/mistral32_24/map/values_structure/d}) \\
GLM-4.6V-Flash & \V{final/glm46v_9/map/index_labels/acc} (\V{final/glm46v_9/map/index_labels/d}) & \V{final/glm46v_9/map/values_only/acc} (\V{final/glm46v_9/map/values_only/d}) & \V{final/glm46v_9/map/structure_only/acc} (\V{final/glm46v_9/map/structure_only/d}) & \V{final/glm46v_9/map/values_structure/acc} (\V{final/glm46v_9/map/values_structure/d}) \\
LLaVA-OneVision-2 8B & \V{final/llavaov2_8/map/index_labels/acc} (\V{final/llavaov2_8/map/index_labels/d}) & \V{final/llavaov2_8/map/values_only/acc} (\V{final/llavaov2_8/map/values_only/d}) & \V{final/llavaov2_8/map/structure_only/acc} (\V{final/llavaov2_8/map/structure_only/d}) & \V{final/llavaov2_8/map/values_structure/acc} (\V{final/llavaov2_8/map/values_structure/d}) \\
Molmo2-O 7B & \V{final/molmo2o_7/map/index_labels/acc} (\V{final/molmo2o_7/map/index_labels/d}) & \V{final/molmo2o_7/map/values_only/acc} (\V{final/molmo2o_7/map/values_only/d}) & \V{final/molmo2o_7/map/structure_only/acc} (\V{final/molmo2o_7/map/structure_only/d}) & \V{final/molmo2o_7/map/values_structure/acc} (\V{final/molmo2o_7/map/values_structure/d}) \\
\bottomrule
\end{tabular}
  \caption{Map pairs: accuracy ($d'$) per text condition.}
  \label{S-tab:final-maps}
\end{table}

\subsection{Supplementary contrasts}
These contrasts are exploratory, because their counting rule was chosen after the answers were seen (Tab.~S3).
By construction, each text version uses identical conflict and control texts, so no drop is expected; no model showed a significant one (accuracies in Sec.~S1).
On black line charts the drop was significant in \V{final/supp/linechart/drop} of \V{final/models/n} models.
Drawing each label in the colour of its line raised accuracy in both conditions to near ceiling in most models, leaving \V{final/supp/linechart_colour/notdecidable} models with too few discordant items; a significant drop remained in \V{final/supp/linechart_colour/drop}.
With the competing name replaced by a code that is not one of the answer options, a significant drop remained on scatter plots in \V{final/supp/pointer_scatter/drop} models and in no model on line charts.
The line charts are therefore not counted as evidence of nearest-label binding.
A leader line or a shaded box left a significant drop on scatter plots in every model with either cue.
The drop was significant on redrawn NCEP/NCAR isobar charts in \V{final/supp/public_data/drop} models, and on GSI maps with the near label one index line further out in \V{final/supp/gsi_placement_k1/drop} of the \V{final/supp/gsi_placement_k1/decidable} decidable models.

\subsection{Text conditions on the map pairs}
From the picture alone, $d'$ on the map pairs (\V{final/map/n} items per condition) ranged from \V{final/map/index_labels/min} to \V{final/map/index_labels/max} (\cref{S-tab:final-maps}).
With both the values and the structure stated in text, $d'$ rose to \V{final/map/values_structure/min}--\V{final/map/values_structure/max}, higher than from the picture alone in every model.
Values only (\V{final/map/values_only/min}--\V{final/map/values_only/max}) and structure only (\V{final/map/structure_only/min}--\V{final/map/structure_only/max}) changed $d'$ in both directions across models.
Because replies without an answer are left out of $d'$, both Qwen3.8 27B and LLaVA-OneVision-2 8B combine, in some text conditions, accuracy below one half with a positive $d'$.
Alone, the values + structure sentence determines the answer, so this condition bounds accuracy from above when the picture need not be read and does not locate the failing stage by itself.

\subsection{Value-choice and magnification probes}

The value-choice probe uses the isobar items unchanged and asks which of the two printed values belongs to the smallest labelled isobar enclosing A; the answer is one of the two values, parsed from a JSON field, and unparsed replies count as wrong.
Its test is an exact sign test on the pairs, Holm-corrected over the \V{probe/value/n/word} models; \cref{S-tab:value} gives the accuracies, the drop and, for comparison, the drop of the two-choice question on the same items.
A second probe magnified one window per isobar pair, shared by its two members, with and without upscaling and against a resampling placebo, in \V{probe/round2/n/word} models; it found no evidence that magnification turns newly read pairs into nearest-label answers.

\begin{table}[h]
  \centering
  \small
  \caption{Value-choice probe on isobar charts.}
  \label{S-tab:value}
  \begin{tabular}{lccccc}
    \toprule
    Model & Control & Conflict & Drop & Two-choice drop & $p$ (Holm) \\
    \midrule
    \V{probe/name/gemma4_31} & \V{probe/value/gemma4_31/control} & \V{probe/value/gemma4_31/conflict} & \V{probe/value/gemma4_31/drop} & \V{probe/value/gemma4_31/twodrop} & \V{probe/value/gemma4_31/p} \\
    \V{probe/name/qwen38_27} & \V{probe/value/qwen38_27/control} & \V{probe/value/qwen38_27/conflict} & \V{probe/value/qwen38_27/drop} & \V{probe/value/qwen38_27/twodrop} & \V{probe/value/qwen38_27/p} \\
    \V{probe/name/gemma4_12} & \V{probe/value/gemma4_12/control} & \V{probe/value/gemma4_12/conflict} & \V{probe/value/gemma4_12/drop} & \V{probe/value/gemma4_12/twodrop} & \V{probe/value/gemma4_12/p} \\
    \V{probe/name/internvl35_38} & \V{probe/value/internvl35_38/control} & \V{probe/value/internvl35_38/conflict} & \V{probe/value/internvl35_38/drop} & \V{probe/value/internvl35_38/twodrop} & \V{probe/value/internvl35_38/p} \\
    \V{probe/name/glm46v_9} & \V{probe/value/glm46v_9/control} & \V{probe/value/glm46v_9/conflict} & \V{probe/value/glm46v_9/drop} & \V{probe/value/glm46v_9/twodrop} & \V{probe/value/glm46v_9/p} \\
    \V{probe/name/mistral32_24} & \V{probe/value/mistral32_24/control} & \V{probe/value/mistral32_24/conflict} & \V{probe/value/mistral32_24/drop} & \V{probe/value/mistral32_24/twodrop} & \V{probe/value/mistral32_24/p} \\
    \V{probe/name/qwen35_9} & \V{probe/value/qwen35_9/control} & \V{probe/value/qwen35_9/conflict} & \V{probe/value/qwen35_9/drop} & \V{probe/value/qwen35_9/twodrop} & \V{probe/value/qwen35_9/p} \\
    \V{probe/name/llavaov2_8} & \V{probe/value/llavaov2_8/control} & \V{probe/value/llavaov2_8/conflict} & \V{probe/value/llavaov2_8/drop} & \V{probe/value/llavaov2_8/twodrop} & \V{probe/value/llavaov2_8/p} \\
    \V{probe/name/molmo2o_7} & \V{probe/value/molmo2o_7/control} & \V{probe/value/molmo2o_7/conflict} & \V{probe/value/molmo2o_7/drop} & \V{probe/value/molmo2o_7/twodrop} & \V{probe/value/molmo2o_7/p} \\
    \V{probe/name/qwen35_4} & \V{probe/value/qwen35_4/control} & \V{probe/value/qwen35_4/conflict} & \V{probe/value/qwen35_4/drop} & \V{probe/value/qwen35_4/twodrop} & \V{probe/value/qwen35_4/p} \\
    \bottomrule
  \end{tabular}
\end{table}

\subsection{Further analyses}
In a post hoc count, the nearest label printed another value than the innermost ring at \V{final/prev/other} of the points and the label of the innermost ring was a median \V{final/prev/ratio} times as far away as the nearest label.
The \emph{named} version names the lines (``Find the innermost closed contour line around A and the next contour line outward.''), and the \emph{value-first} version asks a model to report the value printed on the innermost closed line around A before it answers.
In the value-first version, the value a model reported for the innermost line on conflict items was the near label in \V{final/spot/value/near/min} of the items for \V{final/spot/value/near/minname} and in \V{final/spot/value/near/second} to \V{final/spot/value/near/max} for the other models.
We excluded \V{final/spot/value/cut/names} from these shares because \V{final/spot/value/cut/k} of its \V{final/spot/items} replies hit the token limit.
Post hoc, the value-first version showed no significant drop in \V{final/spot/value/lost/k} models, but in three of them control accuracy also fell, so reporting the value first is not shown to remove the error.
Grouping by common region and connectedness can override proximity for humans, but our cues do not test these principles: the short leader lines and shaded boxes mark which group each label names, not which group holds the query point, and colour may act as a key, not a grouping cue.
\end{document}